\documentclass[10pt,english]{article}
\usepackage[utf8]{inputenc}
\usepackage{cite}
\usepackage[shortlabels]{enumitem}
\usepackage{amsmath,amsthm,amssymb}
\usepackage{longtable}
\usepackage{array}
\usepackage{color}
\usepackage{comment}
\usepackage{xspace}
\usepackage{makecell}
\usepackage[table]{xcolor}
\usepackage{multirow}
\usepackage{placeins}
\usepackage[framemethod=TikZ]{mdframed}
\usetikzlibrary{shapes.symbols}
\usetikzlibrary{arrows.meta}
\usetikzlibrary{calc}
\usepackage{float}
\usepackage[]{graphicx}
\usepackage{alltt}
\usepackage{diagbox}
\usepackage[T1]{fontenc}
\usepackage[utf8]{inputenc}
\usepackage[top=100pt,bottom=100pt,left=68pt,right=66pt]{geometry}
\usepackage{lipsum}
\usepackage[english]{babel}
\usepackage{fancyhdr}
\usepackage{titlesec}
\titleformat{\chapter}
   {\normalfont\LARGE\bfseries}{\thechapter.}{1em}{}
\titlespacing{\chapter}{0pt}{50pt}{2\baselineskip}
\usepackage{float}
\usepackage{soul}
\floatstyle{plaintop}
\restylefloat{table}
\usepackage{hyperref}
\hypersetup{hidelinks,linkcolor = black} % Changes the link color to black and hides the hideous red border that usually is created
\graphicspath{ {images/} }

\newtheorem{defi}{Definition}[section]
\newenvironment{definition}{\begin{mdframed}[roundcorner=6pt,linecolor=blue,backgroundcolor=blue!5]
\begin{defi}}{\end{defi}
\end{mdframed}}

\newtheorem{cor}{Corollary}[section]
\newenvironment{corollary}{\begin{mdframed}[roundcorner=6pt,linecolor=green,backgroundcolor=green!5]
\begin{cor}}{\end{cor}\end{mdframed}}

\newtheorem{thm}{Theorem}[section]
\newenvironment{theorem}{\begin{mdframed}[roundcorner=6pt,linecolor=red,backgroundcolor=red!5]
\begin{thm}}{\end{thm}
\end{mdframed}}

\newtheorem{exampl}{Example}[section]

\newtheorem{mdl}{Model}[section]

\newtheorem{alg}{Algorithm}[section]
\newenvironment{algorithm}{\begin{mdframed}[roundcorner=6pt,linecolor=black,backgroundcolor=black!5]
\begin{alg}}{\end{alg}
\end{mdframed}}

\newtheorem{srcfile}{Source}[section]
\newenvironment{sourcefile}[8]{\begin{mdframed}[linecolor=blue,backgroundcolor=blue!5]
\begin{srcfile}[#1]#2
\begin{center}
\begin{tabular}{p{4.5cm}|p{6.5cm}}
\hline
\multicolumn{2}{c}{\textbf{Contents}}\\
\hline
\multicolumn{2}{p{15cm}}{#3}\\
\hline
\multicolumn{2}{c}{\textbf{Resolution}}\\
\hline
\textbf{Time-Frame} & #4\\
\textbf{Regional-Level (Status)} & #5\\
\textbf{Age-Resolution} & #6\\
\hline
\multicolumn{2}{c}{\textbf{Source Information}}\\
\hline
\textbf{Filename} & #7\\
\textbf{Source (URL)} & #8\\
\textbf{Licence} &\noindent}{\\
\hline
\end{tabular}
\end{center}
\end{srcfile}
\end{mdframed}
}

\newcommand{\tikzsrc}[9]{
\node (#1) at ($(#2) + (0,-1.2)$){};
\path[fill=red!#6] ($ (#1) + (#4) $) rectangle ($ (#1) + (#5) + (0,-0.5) $);
\path[fill=blue!#7] ($ (#1) + (#4) + (0,-0.5)$) rectangle ($ (#1) + (#5) + (0,-1) $);
\path[draw] ($ (#1) + (#4) $) rectangle ($ (#1) + (#5) + (0,-1) $) node[pos=0.5, text=#8] {Src.#3};
}
\newcommand{\tikzusesrc}[3]{
\path[draw=green!50!black, line width=0.8mm] ($ (#1) + (#2) $) rectangle ($ (#1) + (#3) + (0,-1) $);
}

\newcommand{\tikzbbox}[3]{
\draw let  
      \p1 = (T1952.west),
      \p2 = (T2101.east),
      \p3 = (#1),
      \p4 = (#2)
      in ($ (\x1,\y3) + (0.1,0.1) $) rectangle ($ (\x2,\y4) + (-0.1,-1.1) $);
\path
        let 
        \p1 = (#1.north),
        \p2 = (#2.south)  
    in 
    node[rotate=80,align=center] at ($ (-2, \y1)!0.5!(-2,\y2) $) {#3};
}

\newcommand{\tikzcolormap}[1]{
\path[] ($ (#1) + (0,1) $) rectangle ($ (#1) + (9,0) $)  node[pos=0.5]{regional resolution};
\path[fill=red!50] ($ (#1) + (0,0) $) rectangle ($ (#1) + (9,-1) $) node[pos=0.5]{registration districts};
\path[fill=red!40] ($ (#1) + (0,-1) $) rectangle ($ (#1) + (9,-2) $) node[pos=0.5]{municipalities};
\path[fill=red!20] ($ (#1) + (0,-2) $) rectangle ($ (#1) + (9,-3) $) node[pos=0.5]{districts};
\path[fill=red!10] ($ (#1) + (0,-3) $) rectangle ($ (#1) + (9,-4) $) node[pos=0.5]{federal states};
\path[fill=red!0] ($ (#1) + (0,-4) $) rectangle ($ (#1) + (9,-5) $) node[pos=0.5]{none};

\path[] ($ (#1) + (10,1) $) rectangle ($ (#1) + (19,0) $)  node[pos=0.5]{sex \& age resolution};
\path[fill=blue!50] ($ (#1) + (10,0) $) rectangle ($ (#1) + (19,-1) $) node[pos=0.5]{sex \& single age classes};
\path[fill=blue!30] ($ (#1) + (10,-1) $) rectangle ($ (#1) + (19,-2) $) node[pos=0.5]{sex \& fine age classes};
\path[fill=blue!20] ($ (#1) + (10,-2) $) rectangle ($ (#1) + (19,-3) $) node[pos=0.5]{sex \& coarse age classes};
\path[fill=blue!10] ($ (#1) + (10,-3) $) rectangle ($ (#1) + (19,-4) $) node[pos=0.5]{sex};
\path[fill=blue!0] ($ (#1) + (10,-4) $) rectangle ($ (#1) + (19,-5) $) node[pos=0.5]{none};
}

\newtheorem{paramfile}{Parameter Value}[section]
\newenvironment{parameterfile}[2]{\begin{mdframed}[linecolor=green,backgroundcolor=green!5]
\begin{paramfile}[#1]
\begin{center}
\begin{tabular}{p{5.5cm}|p{8.5cm}}}
{\end{tabular}
\end{center}
\end{paramfile}
\end{mdframed}
}

\colorlet{mycolor}{green!30}
\sethlcolor{mycolor}

\begin{document}

%%% Selects the language to be used for the first couple of pages
\selectlanguage{english}

%%%%% Adds the title page
\begin{titlepage}
	\clearpage\thispagestyle{empty}
	\centering
	\includegraphics[width=0.2\linewidth]{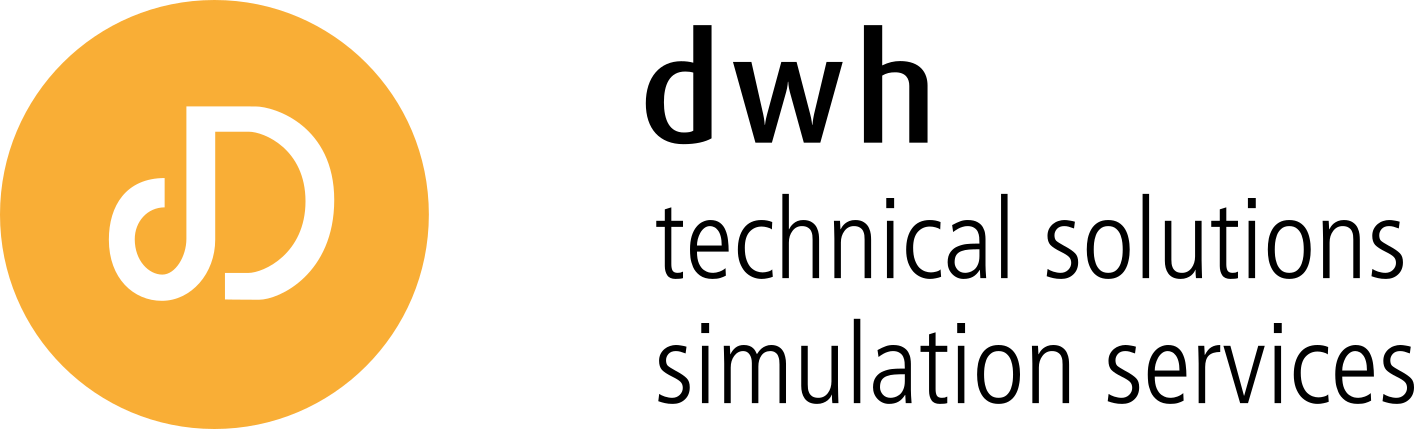}\hspace{1cm}
	\includegraphics[width=0.2\linewidth]{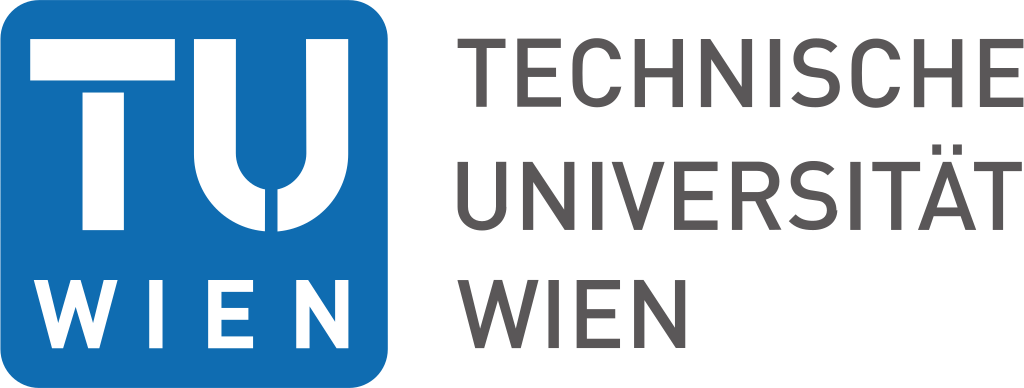}\\
    \vspace{0.5cm}
%	{\large dwh white paper }\\
	\vspace{1cm}
	\rule{12cm}{1.2pt}\\
	\vspace{0.5cm}
	{\Huge \textbf{GEPOC Parameters}} \\
 \vspace{0.1cm}
   {\Large \textbf{Open Source Parametrisation and Validation for Austria}}\\
   {\large \textbf{Version 2.0}}\\
	\vspace{0.5cm}
	\vspace{1cm}
	{\large \today}\\
	\vspace{0.5cm}
    {\large
    Martin Bicher$^{1,3,*}$,
    Maximilian Viehauser$^{2,3}$, Daniele Giannandrea$^{2,3}$, Hannah Kastinger$^{3}$, Dominik Brunmeir$^{1,3}$, Claire Rippinger$^{3}$, Christoph Urach$^{2,3}$, Niki Popper$^{1,2,3}$\\
    }
    \vspace{0.5cm}
    % Affiliation list
    {\small
    $^{1}$ TU Wien, Institute of Information Systems Engineering, Favoritenstra\ss e 9-11, 1040 Vienna, Austria\\
    $^{2}$ TU Wien, Institute of Statistics and Mathematical Methods in Economics, Wiedner Hauptstraße 8-10, 1040 Vienna, Austria\\
    $^{3}$ dwh GmbH, dwh simulation services, Neustiftgasse 57--59, 1070 Vienna, Austria\\
    $^{*}$Correspondence: martin.bicher@dwh.at; martin.bicher@tuwien.ac.at
    }\\[0.5em]
	\rule{12cm}{1.2pt}\\
    \vspace{2cm}
    {\normalsize
    \begin{abstract}
    GEPOC, short for Generic Population Concept, is a collection of models and methods for analysing population-level research questions. For the valid application of the models for a specific country or region, stable and reproducible data processes are necessary, which provide valid and ready-to-use model parameters. This work contains a complete description of the data-processing methods for computation of model parameters for Austria, based exclusively on freely and publicly accessible data. In addition to the description of the source data used, this includes all algorithms used for aggregation, disaggregation, fusion, cleansing or scaling of the data, as well as a description of the resulting parameter files. The document places particular emphasis on the computation of parameters for the most important GEPOC model, GEPOC ABM, a continuous-time agent-based population model. An extensive validation study using this particular model was made and is presented at the end of this work.
    \end{abstract}
    }
	\pagebreak

\end{titlepage}
\tableofcontents
\newpage
\section{Introduction} 
In the present document, we will describe relevant data processes of GEPOC (Generic Population Concept), a collection of models and methods for analysing population-level research questions. We will lay a special focus on the parametrisation of the agent-based model GEPOC ABM, in specific, GEPOC ABM Version 2.2 and the two modules GEPOC ABM Geography, necessary for spatial analysis, and GEPOC ABM IM, necessary for internal migration analysis. We refer to the openly available model documentation\cite{bicher2025gepoc} for details on the model and to \cite{bicher_gepoc_2018} for information about Version 1.0 of the data processes. The model will be parametrised for the country Austria, based exclusively on freely available data. The data source is primarily Statistics Austria and its different publication platforms for open data. 

The document contains
\begin{enumerate}
    \item the different source data sets, their origin, their content and their license,
    \item methods for aggregation, disaggregation, fusion, cleansing or scaling of the data to be suited as model input,
    \item the different derived parameter values, which can be used as model input, and 
    \item a rigorous quantitative validation of the model with the presented parameter values.
\end{enumerate}

The document is structured as follows: First, in Section \ref{sec:model_parameters}, we state the parameters required by GEPOC ABM, most of which are equally relevant to the other GEPOC models (e.g. GEPOC SD\cite{bicher_definition_2015}, GEPOC PDE\cite{bicher_mean-field_2016}). Sections \ref{sec:demographic_relations} and \ref{sec:algorithms} include the methodological background of the parameter calculation. The prior contains specification of important demographic terminology and states the most important demographic relations, balance equations and formulas. The latter contains relevant own-developed generic algorithms, which are not limited to demography in application. Equipped with the relevant terminology, formulas and algorithms, and the source data, in Section \ref{sec:source_data}, the parameter calculation process follows in Section \ref{sec:parameter_calculation}, which we regard as the core part of the document. It displays how the different source data, which all have different levels of aggregation in temporal, spatial, age, and sex resolution, are processed to get one high-quality set of model parameters. Finally, the validation section \ref{sec:validation} provides evidence that the derived parameter values lead to quantitatively valid simulations.

We want to emphasise that in GEPOC, as well as in the census data, \textbf{sex}, i.e. \textit{female}, \textit{male}, is solely interpreted from the biological point of view. In GEPOC, we regard a person as female if it is, according to their biological properties, capable of producing offspring. It is important to mention that this does not necessarily reflect the person's gender (which is not implemented in GEPOC ABM).

\section{Model Parameters}
\label{sec:model_parameters}
In this section we quickly summarise the parameters required for successful parametrisation of GEPOC ABM. Note that these can be found in \cite{bicher2025gepoc}, Sections 3.3.3, 4.3.3, and 5.3.3.

To differentiate the model parameter from the actual demographic quantity (see Section \ref{sec:demographic_relations}), we mark the parameter with a hat symbol: $\hat{(\cdot)}$.

Since we will compute parameters for GEPOC ABM Geography for different regional-levels (see \ref{defi:regional-level}), the parameters for GEPOC ABM directly result for the country-level.

\subsection{GEPOC ABM}
\begin{table}[H]
\begin{center}
\begin{tabular}{p{1.5cm}|p{5.5cm}|c|c|p{4cm}}
\hline
Parameter & Dimensions & Unit & P. Space   & Interpretation\\
\hline
$\hat{\alpha}_m$ & - &  probability & $[0,1]$  & probability for male person-agent at birth\\
\hline
$\hat{a}_{max}$ & - & years & $\mathbb{N}/\{0\}$ & maximum age regarded in the parameters\\
\hline
$\hat{P}(y,s,a)$ & $y\in \{y_0,\dots,y_e\}$, $a\in \{0,\dots, a_{max}\}$, $s\in \{\text{male},\text{female}\}$ & persons & $\mathbb{N}\cup\{0\}$ &  total population per age $a$, sex $s$ at the start of year $y$.\\
\hline
$\hat{I}(y,s,a)$ & $y\in \{y_0,\dots,y_e\}$, $a\in \{0,\dots, a_{max}\}$, $s\in \{\text{male},\text{female}\}$ & persons & $\mathbb{N}\cup\{0\}$ &  total immigrants with age $a$ (at time of immigration), sex $s$ in the course of year $y$.\\
\hline
$\hat{D^p}(y,s,a)$ & $y\in \{y_0,\dots,y_e\}$, $a\in \{0,\dots, a_{max}\}$, $s\in \{\text{male},\text{female}\}$ & probability & $[0,1]$ &  probability of a person with sex $s$, who has had its $a$-th birthday in year $y$, to die before its $a+1$-st birthday.\\
\hline
$\hat{E^p}(y,s,a)$ & $y\in \{y_0,\dots,y_e\}$, $a\in \{0,\dots, a_{max}\}$, $s\in \{\text{male},\text{female}\}$ & probability & $[0,1]$ &  probability of a person with sex $s$, who has had its $a$-th birthday in year $y$, to emigrate before its $a+1$-st birthday.\\
\hline
$\hat{B^p}(y,s,a)$ & $y\in \{y_0,\dots,y_e\}$, $a\in \{0,\dots, a_{max}\}$, $s\in \{\text{male},\text{female}\}$ & probability & $[0,1]$ &  probability of a female person, who has had her $a$-th birthday in year $y$, to give birth to a child before her $a+1$-st birthday. This probability must compensate for multiple-births, which are not depicted in the model.\\
\hline
\end{tabular}
\end{center}
\caption{Parameters of GEPOC ABM. Note that the sex variable $s$ in GEPOC ABM is interpreted from a perspective of reproduction. Agents which are in principle capable of reproduction are called \textit{female}, all other are called \textit{male}.\vspace{2mm}}
\label{tbl:params_gepoc}
\end{table}

\subsection{GEPOC ABM Geography}
\begin{table}[H]
\begin{center}
    \begin{tabular}{p{1.7cm}|p{5.7cm}|c|c|p{4cm}}
\hline
Parameter & Dimensions & Unit & P. Space   & Interpretation\\
\hline
$r_x$ & $x\in\{0,d,e,b,i,min\}$& name & various & regional-levels used for initialisation, death, emigration, birth and immigration processes.\\
\hline
$A_j^{r_x}$ & $x\in\{0,d,e,b,i,min\}, j\in \{1,\dots,q_x\}$& $\{(long,lat)\}$ & $\subset \mathbb{R}^2$ & Specification of the region-families matching to the specified regional-levels with a suitable area-status. Hereby, regional-level $r_x$ has $q_x$ regions.\\
\hline
$\hat{P}(y,i,s,a)$ & $y\in \{y_0,\dots,y_e\}$, $i\in \{1,\dots, q_0\}$, $\quad a\in \{0,\dots, a_{max}\}$, $s\in \{\text{male},\text{female}\}$ & persons & $\mathbb{N}\cup\{0\}$ &  total population per region $A_i^{r_0}$, age $a$, sex $s$ at the start of year $y$.\\
\hline
$\hat{I}(y,i,s,a)$ & $y\in \{y_0,\dots,y_e\}$, $i\in \{1,\dots, q_i\}$, $\quad a\in \{0,\dots, a_{max}\}$, $s\in \{\text{male},\text{female}\}$ & persons & $\mathbb{N}\cup\{0\}$ &  total immigrants to region $A_i^{r_i}$ with age $a$ (at time of immigration), sex $s$ in the course of year $y$.\\
\hline
$\hat{D^p}(y,i,s,a)$ & $y\in \{y_0,\dots,y_e\}$, $i\in \{1,\dots, q_d\}$, $\quad a\in \{0,\dots, a_{max}\}$, $s\in \{\text{male},\text{female}\}$ & probability & $[0,1]$ &  Probability of a person with sex $s$ living in region $A_i^{r_d}$, who has had its $a$-th birthday in year $y$, to die before its $a+1$-st birthday.\\
\hline
$\hat{E^p}(y,i,s,a)$ & $y\in \{y_0,\dots,y_e\}$, $i\in \{1,\dots, q_e\}$, $\quad a\in \{0,\dots, a_{max}\}$, $s\in \{\text{male},\text{female}\}$ & probability & $[0,1]$ &  Probability of a person with sex $s$ living in region $A_i^{r_e}$, who has had its $a$-th birthday in year $y$, to emigrate before its $a+1$-st birthday.\\
\hline
$\hat{B^p}(y,i,s,a)$ & $y\in \{y_0,\dots,y_e\}$, $i\in \{1,\dots, q_b\}$, $\quad a\in \{0,\dots, a_{max}\}$, $s\in \{\text{male},\text{female}\}$ & probability & $[0,1]$ &  Probability of a female person living in region $A_i^{r_b}$, who has had her $a$-th birthday in year $y$, to give birth to a child before her $a+1$-st birthday. This probability must compensate for multiple-births which are not depicted in the model.\\
\hline
\end{tabular}
\end{center}
\caption{Additional parameters of GEPOC ABM Geography extending Table \ref{tbl:params_gepoc}.\vspace{2mm}}
\label{tbl:params_gepoc_geo}
\end{table}

\subsection{GEPOC ABM IM}
\begin{table}[H]
\begin{center}
    \begin{tabular}{p{1.9cm}|p{5.7cm}|c|c|p{4cm}}
\hline
Parameter & Dimensions & Unit & P. Space   & Interpretation\\
\hline
$r_{im}$ & - & name & various & regional-level used for internal migration.\\
\hline
$A_j^{r_{im}}$ & $j\in \{1,\dots,q_{im}\}$& $\{(lat,long)\}$ & $\subset \mathbb{R}^2$ & Specification of the regional set-families for internal migration.\\
\hline
$\hat{IE}(y,i,s,a)$ & $y\in \{y_0,\dots,y_e\}$, $i\in \{1,\dots, q_0\}$, $\quad a\in \{0,\dots, a_{max}\}$, $s\in \{\text{male},\text{female}\}$ & probability & $[0,1]$ &  Probability of a person with sex $s$ living in region $i$, who has had its $a$-th birthday in year $y$, to emigrate internally before its $a+1$-st birthday.\\
\hline
\multicolumn{5}{c}{Interregional model}\\
\hline
$\hat{OD}(y,i,s,j)$ & $y\in \{y_0,\dots,y_e\}$, $i\in \{1,\dots, q_{im}\}$, $s\in \{\text{male},\text{female}\}$, $j\in \{1,\dots, q_{im}\}$ & persons & $\mathbb{N}\cup\{0\}$ &  total migrants from region $i$ to $j$ with sex $s$ in the course of year $y$.\\
\hline
\multicolumn{5}{c}{Biregional model}\\
\hline
$\hat{II}(y,j,s,a)$ & $y\in \{y_0,\dots,y_e\}$, $j\in \{1,\dots, q_{im}\}$, $s\in \{\text{male},\text{female}\}$, $a\in \{0,\dots, a_{max}\}$ & persons & $\mathbb{N}\cup\{0\}$ &  internal immigrants into region $j$ with sex $s$ and age $a$ in the course of year $y$.\\
\hline
\multicolumn{5}{c}{Full Regional model}\\
\hline
$\hat{M}(y,i,s,a,j)$ & $y\in \{y_0,\dots,y_e\}$, $i\in \{1,\dots, q_{im}\}$, $s\in \{\text{male},\text{female}\}$, $a\in \{0,\dots, a_{max}\}$, $j\in \{1,\dots, q_{im}\}$ & persons & $\mathbb{N}\cup\{0\}$ &  internal migrants from region $i$ into $j$ with sex $s$ and age $a$ in the course of year $y$.\\
\hline
\end{tabular}
\caption{Additional parameters of GEPOC ABM IM extending Table \ref{tbl:params_gepoc_geo}.\vspace{2mm}}
\label{tbl:params_gepoc_im}
\end{center}
\end{table}

\newpage
\section{Demographic Terminology and Relations}
\label{sec:demographic_relations}
\subsection{Regional-Level and Identifiers}\label{sec:regional_levels}
The common option to communicate a specific location in a given country is to specify the sub-region of the country in which the point lies in. The smaller the sub-region, the more accurately the statement describes the point. To standardise communication, there are various ways how the country is divided into multiple regions with comparable sizes.
\begin{definition}[regional-level, region-id]\label{defi:regional-level}
    We denote the way how a country is divided into sub-regions as \textbf{regional-level}. Each region the country is divided into for a regional-level is identified by a specific \textbf{region-id}.
\end{definition}
In some situations it might be useful to compare regional-levels w.r. to how specific they describe a location:
\begin{definition}[fine/coarse]\label{defi:fine_coarse}
A regional-level is said to be more \textbf{fine} than the other, if every region of the latter can be split into regions of the prior. In this situation, the latter is also said to be more \textbf{coarse} than the prior.
\end{definition}
In Austria there are various well-known regional-levels used for different purposes, from division of legislative, executive, juristic competences up to sole statistical purposes. The most important ones are surely the nine federal-states. Their region-ids range from \textit{AT-1} to \textit{AT-9}, referring to the names of the states in alphabetical order. Internationally, the federal-states count to the NUTS-2 (Nomenclature des unités territoriales statistiques) regional-level. The Austrian contribution for the NUTS-1 level, which is coarser than NUTS-2, is not so often used. It refers to three combinations of federal-states: (\textit{AT-1+AT-3+AT-9}, \textit{AT-2+AT-6}, and \textit{AT-4+AT-5+AT-7+AT-8}). The federal-state level can be refined into the 35 NUTS-3 regions or into the ~95 political districts of Austria. Note that the district level is not finer than NUTS-3, since some districts are split into two different NUTS-3 regions. The ~2000 Municipalities are finer than NUTS-3 and the district level and count to the internationally used LAU (local administrative unit) regions.

We furthermore explain, in detail, the regional-levels and corresponding region-ids used for the GEPOC parametrisation.
\begin{center}
\begin{tabular}{p{3.5cm}|p{6cm}|p{6cm}}
identifier & meaning & ID (ISO) structure \\
\hline
country & No regional resolution. Data is given for Austria as a total. & \textit{AT}\\\hline
federalstates & Data is given for the nine federal-states of Austria. & \textit{AT-1} to \textit{AT-9}\\\hline
districts & Data is given for the roughly 100 (94 as of 2022) political districts in Austria. Note that Vienna, as a whole, is one of those. & Three digit ISO code. The first digit (i.e. the 100-digit) refers to the federalstate (\textit{AT-1} to \textit{AT-9}), the other two are (more less) ascending integers. E.g. \textit{301} is the ``first'' district (Krems an der Donau (Stadt)) in AT-3 (Lower Austria). Due to created and removed districts, we find occasional gaps in the ascenging order. E.g. district \textit{613} (Mürzzuschlag) was removed and integrated into newly developed district \textit{621} (Bruck-Mürzzuschlag) in 2013. Vienna, as a special case, has ISO code \textit{900}.\\\hline
districts\_districts & Data is given for the 116 (as of 2022) NUTS3 regions in Austria. This is equivalent with the districts, yet Vienna is split into its 23 separate ``Gemeindebezirke''. & Same as \textit{districts}. ISO codes within Vienna are \textit{901-923}.\\\hline
municipalities & Data is given for the 1941 (as of 2022) municipalities in Austria. Note that Vienna, as a whole, is one of those. & Five digit ISO code whereas the first three refer to the political district. The last two digits are more less ascending integers analogous to the \textit{district}-level. E.g. \textit{30101} (Krems an der Donau) is (the only) municipality in district \textit{301} ((Krems an der Donau (Stadt))). Following the scheme, Vienna has ISO code \textit{90001}\\\hline
municipalities\ \_districts & Data is given for the 1940+23=1964 (as of 2022) municipalities in Austria whereas Vienna is split into its 23 ``Gemeindebezirke''. & Same as \textit{municipalities}, yet Viennese ``Gemeindebezirke'' have codes \textit{90101-92301} which also follows the scheme applied to the \textit{districts\_districts} level.\\\hline
municipalities\ \_registrationdistricts & To get a finer resolution within Vienna, data in the ``Gemeindebezirke'' are split into even finer regions - so called ``Zaehlbezirke'' (loosely translated, registration-districts). These are primarily used for statistical reasons. Anyway, this data includes 1940+247=2187 (Gebietsstand 2022) regions. & Outside of Vienna: same as \textit{municipalities}. Registration-districts within Vienna follow a different scheme: seven-digit ISO whereas the first five refer to the district (\textit{municipalities\_districts} level). The last two are ascending integers.
\end{tabular}
\end{center}

The statistical and political regions and their borders in Austria tend to change from time to time - the finer, the more often. E.g., municipalities change almost yearly, while the federal-states changed in 1987 for the last time, when Vienna was announced its own federal-state. As a result, any regional data must be accompanied with information for which time-span the data is valid w.r. to the political regions involved. 
\begin{definition}[area-status]\label{defi:area-status}
The term area-status (loosely translated from German ``Gebietsstand'') refers to the year in which the geographical landscape of the country for a specific region-level is compatible with the given information.
\end{definition}
Note that the area-status of a data set does not necessarily imply that the data is also given only for this year. First, we have no breaches in the time-series if the geographical landscape remains unchanged for several years (e.g. time-series for the federal-states are consistent for over hundred years). Second, data which was collected under an outdated area-status could also be updated to a new one if the collection process allows it. This is currently done by Statistics Austria, which recomputes historic data to an updated area-status every year.

Before diving into the details of parameter calculation for GEPOC, we want to clarify details about the key quantities, our computations will be based on. Note that we still talk about (observing) the real system and not about a model.

\subsection{Census}
\label{sec:quantities}
First of all, we discuss quantities which can be collected from the inhabitants by counting individuals.

\subsubsection{Population \texorpdfstring{$P$}{P}} 
For a given regional-level, $P(y,r,s,a)$ stands for the overall population with sex $s$ (biological sex at birth, male/female) and age $a$ (in years) in region-id $r$ at day $y-01-01$ (start of the year). For Austria, this refers to the state of the central register of residents (ZMR, Zentrales Melderegister) at the given date, which are all persons with an official primary place of residence in Austria. In the absence of better instruments to measure the inhabitants of Austria, and in absence of a better definition of ``who counts as inhabitant of Austria'' this data is denoted as ground-truth for parametrisation and validation of the simulation. 

\subsubsection{Births \texorpdfstring{$B$}{B}, Deaths \texorpdfstring{$D$}{D}, Emigrants \texorpdfstring{$E$}{E}, Immgrants \texorpdfstring{$I$}{I}.} For a given regional-level, $B(y,r,s,a)$, $D(y,r,s,a)$, $I(y,r,s,a)$, and $E(y,r,s,a)$ stand for the total number of new-born, died, immigrated, and emigrated persons with sex $s$ and age $a$ in the course of year $y$ (i.e. between $y-01-01$ and $y-12-31$). Age $a$ always refers to the age at the corresponding event, and $r$ refers to the region the person lived before death and emigration, the person was born in, or the person migrated into according to the ZMR. Note that $B(y,r,s,a)$ is only nonzero for $a=0$.

\subsubsection{Births by Mother \texorpdfstring{$B_m$}{Bm}.} For a given regional-level, $B_m(y,r,s,a)$ stands for the total number of newborns by sex $s$ and age $a$ of the mother in the course of year $y$ (i.e. between $y-01-01$ and $y-12-31$). As before, age $a$ refers to the age at the corresponding event, and $r$ refers to the region the person lived at the time of birth. Note that $B_m(y,r,s,a)$ is only non-zero for $s=f$ (compare with the definition of sex, earlier in the text) and that
\[\sum_{s}B(y,r,s,0)=\sum_{a}B_m(y,r,f,a).\]

\subsubsection{Internal Migrants \texorpdfstring{$M$}{M} and Net Internal Migration \texorpdfstring{$\Delta M$}{DM}.}
For a given regional-level, $M(y,r,s,a,r_2)$ refers to the number of persons with age $a$ and sex $s$ who migrated from region $r$ into region $r_2$ in the course of year $y$. We furthermore define the net internal migration via
\[\Delta M(y,r,s,a)=\sum_{r_2}M(y,r_2,s,a,r)-M(y,r,s,a,r_2).\]

\subsubsection{Aggregation}
For all introduced quantities we use a very native notation for aggregation: $\forall X\in\{P,E,I,D,B,B_m,\Delta M\}$
\begin{align*}
    X(y,r,s)&:=\sum_{a}X(y,r,s,a),\\
    X(y,r,a)&:=\sum_{s}X(y,r,s,a),\\
    X(y,s,a)&:=\sum_{r}X(y,r,s,a),\\
    X(y,r)&:=\sum_{s}\sum_{a}X(y,r,s,a),\\
    X(y,a)&:=\sum_{s}\sum_{r}X(y,r,s,a),\\
    X(y,s)&:=\sum_{r}\sum_{a}X(y,r,s,a),\\
    X(y)&:=\sum_{r}\sum_{s}\sum_{a}X(y,r,s,a).
\end{align*}
Note that this notation might not be mathematically sound, but it is highly useful for communication as long as we care about using the correct variable names. This notation is directly extended to internal-migrants, for which we also introduce new terms:
\begin{align*}
    IE(y,r,s,a)&:=\sum_{r_2}M(y,r,s,a,r_2),\\
    OD(y,r,s,r_2)&:=\sum_{a}M(y,r,s,a,r_2),\\
    II(y,s,a,r_2)&:=\sum_{r}M(y,r,s,a,r_2),\\
\end{align*}
The terms stand for internal emigrants, migrants' origin-destination, and internal immigrants. 

Finally, we also introduce an age cohort $a_{max}^+$ so that $\forall X\in\{P,E,I,D,B,\Delta M\}$
\begin{align*}
    X(y,r,s,a_{max}^+)&:=\sum_{a\geq a_{max}}X(y,r,s,a),\\
    M(y,r,s,a_{max}^+,r_2)&:=\sum_{a\geq a_{max}}M(y,r,s,a,r_2).
\end{align*}
This age class $a_{max}^+$ refers to all individuals with age higher or equal to $a_{max}$. Note that different quantities use different values of $a_{max}$.

\subsection{Event Probabilities and Rates}
\label{sec:event_probabilities}
Talking about total number of births, deaths, etc, we may think of shifting these quantities from the country- to the individual level in terms of rates and probabilities for the corresponding event. 

\subsubsection{Event Probabilities \texorpdfstring{$X^p$}{Xp}}
Defining individual event probabilities is not a straight forward task, since minor details may massively influence the outcome. In our work we define:
\begin{definition}[probability of an event]\label{def:probability}
    For a certain event, the quantity $X^p(y,r,s,a)$ stands for the probability that the event occurs to/for a person with sex $s$, which has its $a$\textsuperscript{th} birthday in the course of year $y$ in region $r$, until the person turns $a+1$.
    We will always indicate probabilities by a superscript ``p''.
\end{definition}
This definition is in-line with the classic definition of the term \textit{death-probability} as it is generally understood by statistics offices (compare with the online glossary of Statistics Finland~ \cite{statistics_finland_probability_nodate}) and it matches the dynamic update concept of the GEPOC model. 

In our studies, $D^p$, $E^p$, $IE^p$, and $M^p$ refer to the probability that the person dies, emigrates, emigrates internally, and internally migrates to a certain destination. Probability $B^p$ refers to the probability that a person gives birth to an offspring (can only be nonzero for $s=$female). Hereby it takes a special role, since the target persons for which the probability is specified, i.e. the female inhabitants, are not the ones who are recorded via $B$, namely the newborn children. Furthermore, the probability-concept does not make sense for immigration processes since the affected person is not part of the observed population before.

\subsubsection{Event Rates \texorpdfstring{$X^r$}{Xr}}
Intuitively, dividing the number of events caused by a group of persons, by the size of the group, fulfils the requirements of a Laplace-space and can accordingly be interpreted as a probability. Therefore, we would expect that the term 
\[\frac{X(y,r,s,a)}{P(y,r,s,a)}\]
provides proper insights into the likelihood of the event $X$ per individual. Unfortunately, it is not that easy, because the group of persons responsible for the events recorded by $X(y,r,s,a)$ is not $P(y,r,s,a)$. A great part of the individuals responsible for the events already had their $a$-th birthday in year $y$ before the event. Therefore, they have been members of the cohort $P(y-1,r,s,a)$ and not $P(y,r,s,a)$, meaning that the actual denominator must have been larger. Nevertheless, the expression is meaningful as it describes an average rate of the event in year $y$, in particular if the average size of the cohort $P_{avg}$ in the course of the year is used instead of the size $P$ at the start of the year:
\begin{definition}[average rate of an event]\label{def:avgrate}
    For a certain event, 
    \begin{equation}
        X^r(y,r,s,a):=\frac{X(y,r,s,a)}{P^{avg}(y,r,s,a)}
    \end{equation}
    defines the average rate of the event in the course of year $y$. Hereby $P^{avg}$ describes the average size of the population over the course of the year.
\end{definition}
In case the census does not provide average population information, it can be approximated by the arithmetic mean of the populations on new-year
\begin{equation}P_{avg}(y,r,s,a)\approx \frac{P(y,r,s,a)+P(y+1,r,s,a)}{2}.\end{equation}
In the following, we will apply this approximation for $P_{avg}$ for all parameter computations.

Although $X^r$ is not directly applicable as a probability, it is nevertheless often applied as demographic quantity. For example, the age-dependent rate of fertility $B_m^{r}(y,r,f,a)$ multiplied by $1000$ is an important demographic indicator stating the average number of newborns per $1000$ women with age $a$. It helps defining meaningful derived indicators:
\begin{definition}[total fertility rate $TFR$ and mean age at childbearing $MAC$]\label{def:fertilityrates}
    With the average fertility rate $B_m^r$, the total fertility rate defines as
    \begin{equation}
        TFR(y,r,f)=\sum_{a}B_m^r(y,r,f,a),
    \end{equation}
     and the average fertility age defines as
    \begin{equation}
        MAC(y,r,f)=\frac{\sum_{a}aB_m^r(y,r,f,a)}{\sum_{a}B_m^r(y,r,f,a)}.
    \end{equation}
    The prior estimates the number of newborns in a woman's lifetime, the latter the average age of a woman giving birth to a child.
\end{definition}

\subsubsection{Probabilities from Rates and Census}

Due to mentioned population-group mismatch due to ageing, precise calculation of actual probabilities from census data alone is usually impossible. However, there are well established methods which give good estimates.

One of the most important ones dates back to life-science pioneer William Farr. In a paper published in 1859\cite{farr_construction_1859}, he presented relevant concepts for the computation of death-tables for Great Britain, including an important method for the estimation of death probabilities. The formula, found on page 848, is based on the idea that the average rate of mortality $D^r$, as defined in the section before, computes the rate of death per lived year of life, but not per individual. Farr assumed that the population of the cohort with age $a$ remains constant and that, therefore, any of the $D$ individuals lost by death must have been replaced by an equivalent new individual in the course of the year to maintain the total cohort size. Therefore, the total number of individuals under observation is larger than the overall lived years of life. Farr found the formula \begin{equation}D^p(y,r,s,a) = 1-\frac{1-\frac{1}{2}D^r(y,r,s,a)}{1+\frac{1}{2}D^r(y,r,s,a)} =  \frac{D^r(y,r,s,a)}{1+\frac{1}{2} D^r(y,r,s,a)}.\end{equation}
to compensate for this bias. The formula can be reasoned by comparing an individual-level ratio with a population-level ratio for ``deaths-per-person-year'':

When an individual turns $a$, its chance for surviving the upcoming year is $(1-D^p_a)$, and  its probability of death is $D^p_a$. Assuming that the time of death is uniformly distributed in the course of the individual's life-year, the average proportion of the year spent by the individual is
\[(1-D^p(y,r,s,a))+\frac{1}{2}D^p(y,r,s,a)=1-\frac{1}{2}D^p(y,r,s,a).\]
Therefore, we observe
\[\frac{D^p(y,r,s,a)}{1-\frac{1}{2}D^p(y,r,s,a)}\]
deaths-per-person-year for the individual. 

On the population scope, the age band of individuals with age $a$ is assumed to be of constant size $P_a$. Therefore, the total years of life spent by persons of age $a$ over the observed year is precisely $P_a$. With $D_a$ recorded deaths of persons with age $a$, there are $\frac{D_a}{P_a}$ deaths per person year on the population-scope. Since the scopes must be identical, we have
\[\frac{D_a}{P_a}=\frac{D^p(y,r,s,a)}{1-\frac{1}{2}D^p(y,r,s,a)}\]
which can be transformed to
\[D^p(y,r,s,a)=\frac{\frac{D_a}{P_a}}{1+\frac{1}{2}\frac{D_a}{P_a}}.\]
Approximating $\frac{D_a}{P_a}\approx D^r(y,r,s,a)$ gives the stated formula.

Clearly, the formula incorporates several inaccuracies, some of which can be made more accurate without major computations. One of the most important ones refers to the uniformity of the time of death in the course of the year. In particular for newborn infants, the rate of mortality in the first months of life is a lot larger than later on. Therefore, the average time an individual that died with age $0$ spent in the cohort is usually a lot smaller than $1/2$. Leaving this average time free for parametrisation leads to the modern-time formulation of the death-rate formula:
\begin{theorem}[Farr's Death Rate Formula (modern version)]\label{thm:farr}
Let $1-\alpha(a)$ stand for the expected year-of-life spent by a person in its age cohort, given that the person is going to die, then
\begin{equation}\frac{D^r(y,r,s,a)}{1+\alpha(a)D^r(y,r,s,a)} = \frac{D(y,r,s,a)}{P_{avg}(y,r,s,a)+\alpha(a)D(y,r,s,a)},\end{equation}
is a good model for the probability of of death $D^p(y,r,s,a)$ as given by Definition \ref{def:probability}. Hereby, $P_{avg}$ stands for the average population over the course of year $y$.
\end{theorem}
As mentioned, it is legitimate to define $\alpha(a)=1/2$ for $a>0$, however, since infants are way more likely to die within the first few weeks of their life than in the rest of their first life-year, typically $\alpha(0)>0.9$ is applied. Moreover, it is worth mentioning that the formula is equally meaningful when applied to the $a_{max}^+$ cohort to get a constant death probability $D^p(y,r,s,a_{max}^+)$ for any individual with age $\geq a_{max}$.

Analogous to the mentioned fertility rates the formula in Theorem \ref{thm:farr} is an internationally recognized concept for the computation of a demographic indicator. Besides being a good approximation, it is also the one used by Statistics Austria to compute mortality- and life-tables\cite{hanika2014}. As a result, it will be used for any parametrisation strategy involving statistical indicators like death probabilities and mortality tables. For model parametrisation from census data, the formula can be improved:

One of the key advantages of the formula is that the probabilities of death can be computed from census information from one single year. This became possible by the assuming
\[\frac{D_a}{P_a}=\frac{D(y,r,s,a)}{P_{avg}(y,r,s,a)} = D^r(y,r,s,a).\]
However, the year over which the deaths-per-life-year are computed on the individual-scope, namely the individual's $a+1$-st year of life, is not equivalent with the one over which the rate of mortality is computed, namely Jan $1^{st}$ to Jan $1^{st}$. In particular, if the probabilities are under subject of high dynamics, this results in a half-year time-lag. In case one is not bound to apply data from a single year, we can compensate the bias by taking the arithmetic mean
\begin{equation}
    \frac{1}{2}\frac{D(y,r,s,a)}{P_{avg}(y,r,s,a)+\alpha(a)D(y,r,s,a)}+\frac{1}{2}\frac{D(y+1,r,s,a)}{P_{avg}(y+1,r,s,a)+\alpha(a)D(y+1,r,s,a)}.
\end{equation}
Since all probabilities in GEPOC ABM are defined in the same manner as the death probability, we may also apply the formula for parametrisation of any probability occurring in the update in a slightly modified form:
\begin{corollary}[Farr Formula (model parametrisation)]
\label{cor:farr_param}
Let $X\in \{B,D,E,IE\}$, and $Q(y,r,s,a):=E(y,r,s,a)+D(y,r,s,a)$ as the total number of individuals leaving the age cohort in the course of year $y$, and let $y_N$ refer to the last available year in the population census. With
\begin{align}
    P_{avg}(y,r,s,a)&:=\frac{1}{2}(P(y,r,s,a)+P(\max(y+1,y_N),r,s,a)),\\
    X^{p,-}(y,r,s,a)&:=\frac{X(y,r,s,a)}{P_{avg}(y,r,s,a)+\frac{1}{2}Q(y,r,s,a)},\text{ and}\\
    X^{p,+}(y,r,s,a)&:=\frac{X(\max(y+1,y_N),r,s,a)}{P_{avg}(\max(y+1,y_N),r,s,a)+\frac{1}{2}Q(\max(y+1,y_N),r,s,a)},
\end{align}
we get
\begin{equation}
    \frac{1}{2}X^{p,-}(y,r,s,a)+\frac{1}{2}X^{p,+}(y,r,s,a)
\end{equation}
as a good approximation for the probability $X^p(y,r,s,a)$ of the event according to Definition \ref{def:probability}.
\end{corollary}
The formula can be reasoned for $X\neq D$ using the same ideas as presented before, namely comparing the individual occurrences of the event $X$ per individual life-year with the corresponding population-scope observation. Since the model does not regard skewed infant-deaths or emigrations, usage of any $\alpha(a)$ different from $1/2$ is not necessary. Finally, the model does, mechanistically, not differentiate between deaths and emigrants - in both cases, agents leave the model and future events are cancelled. As a result, joining $D$ and $E$ as a common census for individuals leaving the model boundaries is useful.

\subsubsection{Life Tables and Life-Expectancy \texorpdfstring{$LE$}{LE}}

Under $LE(y,r,s,a)$ we understand the life expectancy of a person with sex $s$, living in region $r$, at the point of its $a$-th birthday in year $y$. Note that this is usually a rising function when increasing $a$ and $y$ at the same time, i.e. $LE(y,r,s,a)<LE(y+1,r,s,a+1)$, since already surviving for $a$ years reduces the chances for dying young. The function is not necessarily increasing when $y$ is fixed, e.g. due to medical advancements.

Alike the computation of death/event probabilities, creating a mapping between deaths/death rates (probabilities) and the life expectancy is not straightforward either and requires assumptions. The most common way to approach this problem is by applying the concept of life-/death-/mortality-tables from survival analysis.

These tables are solely based on the death probabilities $D^p$ computed from overall deaths with Farr's Death Rate Formula, Theorem \ref{thm:farr}. For given sex $s$, region $r$, and year $y$, the table is filled as follows (Sullivan method \cite{klotz2016}):
\begin{enumerate}
   \item First of all, we define the series of death probabilities 
    \begin{equation}q_i:=\left\lbrace\begin{matrix} D^p(y,r,s,i),\quad 0\leq i\leq a_{max}\\ D^p(y,r,s,a_{max}^+),\quad i>a_{max}\end{matrix}\right.\end{equation}
    \item In the next step, we compute population series $l$ and absolute death series $d$ recursively. We start with an arbitrary fictional start population $l_0$ (typically $l_0=100000$ is chosen):
    \begin{align}
        d_i &:= l_i\cdot q_i\\
        l_{i+1}&:=l_i - d_i.
    \end{align}
    This leads $l_i=l_0\cdot \prod_{j=0}^{i-1}(1-q_j)$ and $d_i=l_0\cdot q_{i}\cdot \prod_{j=0}^{i-1}(1-q_j)$.
    \item Furthermore, the population at-risk $L$ is computed. Again, we find $\alpha$, as defined in Theorem \ref{thm:farr}, in this formula:
    \begin{equation}
        L_i:=l_i-\alpha(i) d_i.
    \end{equation}
    This leads to the explicit formula
    \begin{equation*}
    L_i=l_0\left(1-\alpha(i)q_i\right)\prod_{j=0}^{i-1}(1-q_j).
    \end{equation*}
    \item We furthermore compute the cumulative number of population years at risk $T$:
    \begin{equation}
        T_i:=\sum_{j=i}^{\infty}L_j,
    \end{equation}
    which can be written explicitly as
    \begin{equation*}
        T_i:=l_0\sum_{j=i}^{\infty}\left(1-\alpha(j) q_j\right)\prod_{k=0}^{j-1}(1-q_k).
    \end{equation*}
    Note that the infinite sum can be computed analytically, since $q_{k}=q_{a_{max}}, \alpha(k)=0.5$ for $k\geq a_{max}$. Thus, for $k>a_{max}$, $L_i$ forms a geometric series with
    \begin{multline*}
    \frac{l_i}{l_{i-1}} = (1-q_{a_{max}})\Rightarrow \sum_{i=a_{max}}^{\infty}l_i=l_{a_{max}}\frac{1}{q_{a_{max}}}\\
    L_{i}=l_i-\alpha(i)l_iq_{i}=l_i(1-\alpha(i)q_{i})\Rightarrow \sum_{i=a_{max}}^{\infty}L_i= l_{a_{max}}\frac{1-\alpha(a_{max})q_{a_{max}}}{q_{a_{max}}}=l_{a_{max}}\frac{1-\frac{q_{a_{max}}}{2}}{q_{a_{max}}}.
    \end{multline*}
    \item Finally, we compute the life expectancy vector $e$ by
    \begin{equation}
        e_i:=\frac{T_i}{l_i}.
    \end{equation}
\end{enumerate}
\newpage
Putting it all together, we find the following closed formula
\begin{definition}[Life Expectancy Formula]\label{def:life_expectancy}
The life expectancy of a person with sex $s$, living in region $r$ at the point of its $a$-th birthday in year $y$ can be approximated with
\begin{equation}
    LE(D^p,y,r,s,a)=\frac{\sum_{j=a}^{\infty}\left(1-\alpha(j) D^p(y,r,s,j)\right)\prod_{k=0}^{j-1}(1-D^p(y,r,s,k))}{\prod_{j=0}^{a-1}(1-D^p(y,r,s,j))}.
\end{equation}
\end{definition}
Most importantly, $l_0$ cancels out which renders its choice irrelevant for computing the life expectancy.

\subsection{Demographic Balance Equations}
In the next step, we will state some balance equations, which can be used to calculate relations between the different quantities. The first balance refers to the overall population.
\begin{corollary}[Overall Population Balance]\label{cor:balance_overall}
For all $y$ and $s$,
\begin{equation}
    P(y+1,s)=P(y,s)+B(y,s)+I(y,s)-D(y,s)-E(y,s).
\end{equation}
\end{corollary}
This equation does not require additional explanation, but we want to emphasise that it does not at all hold without full summation over age and region.
\begin{corollary}[Region-Specific Population Balance]\label{cor:balance_regional}
For all $y$, $r$ and $s$,
\begin{equation}
    P(y+1,r,s)=P(y,r,s)+B(y,r,s)+I(y,r,s)-D(y,r,s)-E(y,r,s)+\Delta M(y,r,s).
\end{equation}
\end{corollary}
With introduction of the spatial component, the (net) internal migration becomes relevant.

Development of age-dependent balance equations is possible, however, additional assumptions need to be made (e.g. how demographic events are distributed over the life-year). Moreover, these formulas turned out to be numerically unstable with respect to data errors. As a result, we will not state any of those.

\newpage
\section{Disaggregation Algorithms}
\label{sec:algorithms}
In the last section, several formulas have been introduced, which can be used to compute probability parameters for the model when census information is provided. With these formulas, GEPOC should be configurable for as long a period as possible, extending not only as far back into the past as possible, but also into the future. 

This process, in turn, brings with it other difficulties that have nothing to do with demographics per-se: Since both historical and forecast census information are not available in the same resolution as current data, they must be harmonised. If the coarsest common level of detail was chosen for harmonisation, an enormous loss of information for the current data would result, making the model unnecessarily inaccurate. Accordingly, the opposite approach is chosen: data with lower resolution is \textbf{disaggregated} to the finest level of resolution using assumptions about the distribution.

In the following, we distinguish between two problem statements. In the \textit{one-sided disaggregation} problem, one data-set is strictly finer than the other, and the goal of the procedure is to elevate the resolution of the coarser data-set to that of the finer one. In the \textit{two-sided disaggregation} problem, both data-sets each have resolution deficiencies in different dimensions, and the final result harmonises the data on the joint finest resolution. The latter can be regarded as estimating a distribution given its marginals, and occurs in the computation of internal migration parameters.

\subsection{One-Sided Disaggregation}
\label{sec:one_sided_disaggregation}
In the one-sided problem, one data-set is strictly finer than another one. Let $\Psi_1:=(Y_1,R_1,S_1,A_1)$ be the set of all year-region-sex-age tuples of the data entries $X_1(y,r,s,a),(y,r,s,a)\in \Psi_1$ of the fine-grained data and $\Psi_2:=(Y_2,R_2,S_2,A_2)$ be the indices to the coarse data $X_2$, then there is an aggregation mapping $f: \Psi_1\rightarrow \Psi_2$ so that $X_1$ can be aggregated to the coarse level by summation over all of the inputs of $f$ with the same image. That means, let
\[A(y,r,s,a):=\{(y',r',s',a')\in \Psi_1:f((y',r',s',a'))=(y,r,s,a)\},\]
then
\[X_{1,agg}(y,r,s,a)=\sum_{(y',r',s',a')\in A(y,r,s,a)}X_1(y',r',s',a').\]
While this computation is uniquely defined, the inverse problem, i.e. bringing $X_2$ to the fine level $X_{2,disagg}$, requires additional assumptions. Clearly,
\begin{equation}\label{eq:disagg_cond}
    X_2(y,r,s,a)=\sum_{(y',r',s',a')\in A(y,r,s,a)}X_{2,disagg}(y',r',s',a'),
\end{equation}
must be fulfilled, apart from that, however, the computation of $X_{2,disagg}$ has no further constraints. To solve this problem, we usually define a distribution which the disaggregated data should follow, i.e. for every $(y,r,s,a)\in \Psi_2$:
\begin{multline}\label{eq:disagg_cond_2}
   \vec{X}_{2,disagg}:= \left(X_{2,disagg}(y',r',s',a')\right)_{(y',r',s',a')\in A(y,r,s,a)}\sim\\
\left(p(y',r',s',a')\right)_{(y',r',s',a')\in A(y,r,s,a)}=:\vec{p}(y,r,s,a)\in (\mathbb{R}^+)^{|A(y,r,s,a)|}
\end{multline}
In most cases, this distribution is gained from the fine-resolution data-set.

\subsubsection{Proportional Disaggregation}
\label{sec:proportional_disaggregation}
The most straightforward strategy to disaggregate the value will be called \textit{proportional disaggregation}:
\begin{algorithm}[Proportional Disaggregation]
\label{alg:prop_disagg}
    Let $X_2(y,r,s,a)$ be a value to be disaggregated and $\vec{p}(y,r,s,a)$ a vector which the disaggregated values should follow, then
    \begin{equation}
        \vec{X}_{2,disagg}\leftarrow \vec{p}(y,r,s,a)\frac{X_2(y,r,s,a)}{\sum \vec{p}(y,r,s,a)}.
    \end{equation}
\end{algorithm}
It is clear that the disaggregated vector fulfils conditions (\ref{eq:disagg_cond}) and (\ref{eq:disagg_cond_2}) perfectly.

Unfortunately, there are still cases where a different algorithm is required, namely when it comes to conserve whole numbers. Both the initial population and the immigration process require integer-valued parameters.

\subsubsection{Integer-Valued Disaggregation}
The problem of disaggregating a given natural number of elements $X_2(y,r,s,a)\in \mathbb{N}$ onto a number of parties according to a distribution vector $\vec{p}$ into an integer-valued vector $\vec{X}_{2,disagg}$ is usually called an apportionment problem. This problem has a wide range of applications, most famously in politics, to match the result of demographic elections onto a finite number of seats in the parliament. Interestingly, the problem is well known (proven, \cite{balinski2010fair}) for having no fully fair solution and that corresponding apportionment algorithms for disaggregation will always generate paradoxical and/or biased situations, such as the famous Alabama Paradox: The apportionment strategy used by that time in the US for assigning parliament seats was based on ranking division remainders (Hamilton-method). However, an employee of the census office found out that the method would assign the state of Alabama 8 seats if 299 total seats were available, but only 7 if 300 total seats were available ~\cite{robinson1982alabama}.

For GEPOC, we decided to apply the Huntington-Hill apportionment strategy \cite{balinski_huntington_1977}. The method is known to be considerably fair and is based on iterative disaggregation of a value by drawing indices:
\begin{algorithm}[Huntington-Hill Disaggregation]\label{alg:hh_disaggregation}
 Let $X_2(y,r,s,a)\in \mathbb{N}$ be a value to be disaggregated and $\vec{p}(y,r,s,a)\in (\mathbb{R}^+)^{n}$ a vector which the disaggregated values should follow, then the Huntington-Hill method is given as follows:
 \begin{enumerate}
 \item  Initialise two state vectors 
 $\vec{v}\leftarrow \vec{p}(y,r,s,a)$ and $\vec{w}\leftarrow (0)_{i=1}^{n}$.
 \item Perform the following step $X_2(y,r,s,a)$-times:
 \item Let $j\leftarrow \text{argmax}(\vec{v})$, then $\vec{w}_j\leftarrow \vec{w}_j+1$ and 
 \[\vec{v}_j\leftarrow \frac{\vec{p}_j(y,r,s,a)}{\sqrt{\vec{w}_j(\vec{w}_j+1)}}.\]
 Hereby, highest $\vec{p}(y,r,s,a)$ and lowest $j$ are used as tiebreakers if multiple highest values occur.
 \end{enumerate}
Set $\vec{X}_{2,disagg}\leftarrow \vec{w}$.
\end{algorithm}
It must be noted that $\vec{p}(y,r,s,a)$ is not normed in this process. In general, $\vec{p}(y,r,s,a)\in(\mathbb{N}^0)^n$ can even be helpful for computation. With $N:=\sum \vec{p}(y,r,s,a)$, then, for any $X_2(y,r,s,a)=kN,k\in \mathbb{N}$, the algorithm will return the values $k\vec{p}(y,r,s,a)$. That means, if $X_2(y,r,s,a)>N$, we may compute $k:=\lfloor X_2(y,r,s,a)/N\rfloor$, apply the algorithm only on $X_2(y,r,s,a)-Nk$, and add $k\vec{p}$ at the end.

As an alternative to an apportionment algorithm, the strategy can also be interpreted as a drawing process from a probability distribution. Yet, in contrast to using random numbers for drawing, the resulting sequence is deterministic and creates fully reproducible results without stochastic fluctuations. Quite the opposite is true: convergence properties to the distributions have been thoroughly investigated~\cite{balinski_huntington_1977}.

\subsection{Two-Sided Disaggregation}
\label{sec:two_sided_disaggregation}
Finally, we require algorithms to match two data-sets, both of which are coarse on different dimensions. To be specific, we will regard datasets $X_1$ and $X_2$ so that the corresponding index sets $\Psi_1$ and $\Psi_2$ fulfil the following three properties:
\begin{itemize}
    \item there is a dimension $j$ along which $X_1$ is fully aggregated, i.e. $|\{x_j,x\in \Psi_1\}|=1$,
    \item there is a different dimension $j\neq k$ along which $X_2$ is fully aggregated, i.e. $|\{x_k,x\in \Psi_2\}|=1$,
    \item for all other dimensions, both datasets have the same resolution.
\end{itemize}
Let, without loss of generality, be $j=2$ and $k=4$, then dataset $X_1$ differentiates age but is aggregated overall regions, and dataset $X_2$ differentiates regions but is fully aggregated over all age-classes. For every year $y$ and sex $s$, the problem states as follows: find a data-set $X_{1,2}(y,r,s,a)$ so that
\begin{align}\label{eq:ipf_condition_1}
\sum_{r'}X_{1,2}(y,r',s,a) &= X_1(y,s,a), \text{and}\\\label{eq:ipf_condition_2}
\sum_{a'}X_{1,2}(y,r,s,a') &= X_2(y,r,s),\end{align}
assuming that
\begin{equation}
    \sum_{a'}X_1(y,s,a')=\sum_{r'}X_2(y,r',s)
\end{equation}
holds for all $y$ and $s$. 

The main field of application of this problem, in terms of the parametrisation of GEPOC, is internal migration. Data which includes origin and destination information often does not include any other dimensions such as age. Internal emigration or immigration data, on the other hand, does not include any information about the individual's origin/destination. As a result, there are different data available, each showing one side of the medal, but not all. They can essentially be interpreted as the marginals of an unknown distribution.

\subsubsection{Disaggregation of Marginal Distributions}
For any values of the codimensions, problems (\ref{eq:ipf_condition_1}) and (\ref{eq:ipf_condition_2}) translate to a very fundamental mathematical problem. Find a positive matrix $M\in (\mathbb{R}^+)^{m\times n}$ so that the sum over all columns matches $\vec{a}\in (\mathbb{R}^+)^{m}$ and the sum over all rows matches $\vec{b}\in (\mathbb{R}^+)^{n}$, i.e.
\[\sum_{j=1}^{n}M_{\cdot,j}=\vec{a},\quad \sum_{i=1}^{m}M_{i,\cdot}=\vec{b}.\]
For any $n\geq m>2$, the problem is well underdetermined ($m\times n$ vs. $m+n$ degrees of freedom) and potentially has various solutions. One well-known method to find one of them is \textit{Iterative Proportional Fitting} (IPF), sometimes also called RAS algorithm or biproportional fitting. It iteratively divides a given initial estimate by the target row- and column sums, leading to convergence.
\begin{algorithm}[Iterative Proportional Fitting (IPF)]\label{alg:ipf_2d}
    Let $\vec{a}\in (\mathbb{R}^+)^{m}$ and $\vec{b}\in (\mathbb{R}^+)^{n}$ be row- and column-sums of an unknown matrix $M\in (\mathbb{R}^+)^{m\times n}$, where $\sum_{i=1}^{m}\vec{a} = \sum_{j=1}^{b}\vec{b}$ is given. Furthermore, let $M_0\in (\mathbb{R}/\{0\})^{m\times n}$ be an initial guess for $M$ with positive values. Define $X:=M_0$ as the state, and 
    \begin{equation}
        res(X):=\left\lVert\sum_{i=1}^{m}M_{i,\cdot}-\vec{b}\right\rVert+\left\lVert\sum_{j=1}^{n}M_{\cdot,j}-\vec{a}\right\rVert
    \end{equation}
    as the residual of the current state $X$. Until the residual is smaller than a defined tolerance or does not improve any more, perform the following steps:
    \begin{enumerate}
        \item Update:
        \begin{equation}
            \forall i,j: X_{i,j}\leftarrow X_{i,j}\frac{a_i}{\sum_{j=1}^{n}X_{i,j}}.
        \end{equation}
        \item Update:
        \begin{equation}
            \forall i,j: X_{i,j}\leftarrow X_{i,j}\frac{b_j}{\sum_{i=1}^{m}X_{i,j}}.
        \end{equation}
    \end{enumerate}
\end{algorithm}
The algorithm has been proven to converge as long as a solution exists and has been shown to have minimum distance (maximum likelihood) to the initial guess~\cite{deming1940least}. It is worth mentioning that the same concept can also be used in three dimensions, as it is required to fully entangle internal-emigration data and internal-immigration data, both with high age resolution, with origin-destination data without age information.

\begin{algorithm}[Iterative Proportional Fitting 3D (IPF-3D)]\label{alg:ipf_3d}
    Let $A\in (\mathbb{R}^+)^{m\times n}$, $B\in (\mathbb{R}^+)^{n\times r}$ and $C\in (\mathbb{R}^+)^{m\times r}$ be the marginals of an unknown tensor $M\in (\mathbb{R}^+)^{m\times n\times r}$, i.e.
    \begin{equation}
        \sum_{k=1}^{r}M_{\cdot,\cdot,k} = A_{\cdot,\cdot}\quad ,\quad \sum_{i=1}^{m}M_{i,\cdot,\cdot} = B_{\cdot,\cdot}\quad , \quad \sum_{j=1}^{n}M_{\cdot,j,\cdot} = C_{\cdot,\cdot}
    \end{equation}
    whereas we assume that
    \begin{equation}
        \sum_{i=1}^{m}A_{i,\cdot} =\sum_{j=1}^{r}B_{\cdot,j}\quad ,\quad \sum_{j=1}^{n} A_{\cdot,j} =\sum_{j=1}^{r}C_{\cdot,j}\quad ,\quad \sum_{i=1}^n B_{i,\cdot} =\sum_{i=1}^{m}C_{i,\cdot}
    \end{equation}
    is given. Furthermore, let $M_0\in (\mathbb{R}/\{0\})^{m\times n\times r}$ be an initial guess for $M$ with positive values. Define $X:=M_0$ as the state, and 
    \begin{equation}
        res(X):=\left\lVert
        \sum_{k=1}^{r}M_{\cdot,\cdot,k} - A_{\cdot,\cdot}\right\rVert +  \left\lVert\sum_{i=1}^{m}M_{i,\cdot,\cdot} - B_{\cdot,\cdot}\right\rVert+  \left\lVert\sum_{j=1}^{n}M_{\cdot,j,\cdot} - C_{\cdot,\cdot}\right\rVert
    \end{equation}
    as the residual of the current state $X$ (hereby $\lVert\cdot \rVert$ refers to a suitable matrix-norm). Until the residual is smaller than a defined tolerance or does not improve any more, perform the following steps:
    \begin{enumerate}
        \item Update:
        \begin{equation}
            \forall i,j,k: X_{i,j,k}\leftarrow X_{i,j,k}\frac{A_{i,j}}{\sum_{k=1}^{r}X_{i,j,k}}.
        \end{equation}
        \item Update:
        \begin{equation}
            \forall i,j,k: X_{i,j,k}\leftarrow X_{i,j,k}\frac{B_{j,k}}{\sum_{i=1}^{m}X_{i,j,k}}.
        \end{equation}
        \item Update:
        \begin{equation}
            \forall i,j,k: X_{i,j,k}\leftarrow X_{i,j,k}\frac{C_{i,k}}{\sum_{j=1}^{n}X_{i,j,k}}.
        \end{equation}
    \end{enumerate}
\end{algorithm}

\subsubsection{Disaggregation of Integer-Valued Marginals (Concepts)}
Clearly, the ideas from the last section can also be adapted to sole integer-valued disaggregation. For this purpose, we developed two algorithms ourselves, one for 2D and one for 3D. We will sketch the ideas:

In the first step of the 2D algorithm, marginal $\vec{a}$ is disaggregated to the second dimension using the Huntington-Hill Strategy, Algorithm \ref{alg:hh_disaggregation}, to get an initial estimate $M_0=:X\in \mathbb{N}^{m\times n}$. We may use the IPF algorithm to get a good initial distribution for the Huntington-Hill Disaggregation algorithm. This way, we can be sure that the equations for the first marginal are fulfilled right from the start.

At this point, the actual algorithm starts, which can be motivated by the idea of connecting pins on a board by strings. Every unit-value in $X$ corresponds to an individual (e.g. $X_{i,j}=4$ refers to four individuals) and we imagine the individual to be a string on a wall of pins. The wall has one column of pins for every dimension and one pin for every possible index in this dimension. An individual with the indices $i,j$ (e.g. origin $i$ and destination $j$) connects the pin $i$ in the first column and $j$ in the second column.

If we take a string $(i_0,j_1)$ and change its second pin to $j_2$, then the first marginal equation would remain unchanged. The operation would correspond to $X_{i_0,j_1}-=1$ and $X_{i_0,j_2}+=1$. That means if $j_1$ was chosen so that $\sum_{i=1}^mX_{i,j_1}-\vec{b}_{j_1}>0$ and $j_2$ so that $\sum_{i=1}^mX_{i,j_2}-\vec{b}_{j_2}<0$, the algorithm reduces the marginal. By design, the algorithm will converge if a solution exists.

\begin{figure}
    \centering
    \includegraphics[width=0.8\linewidth]{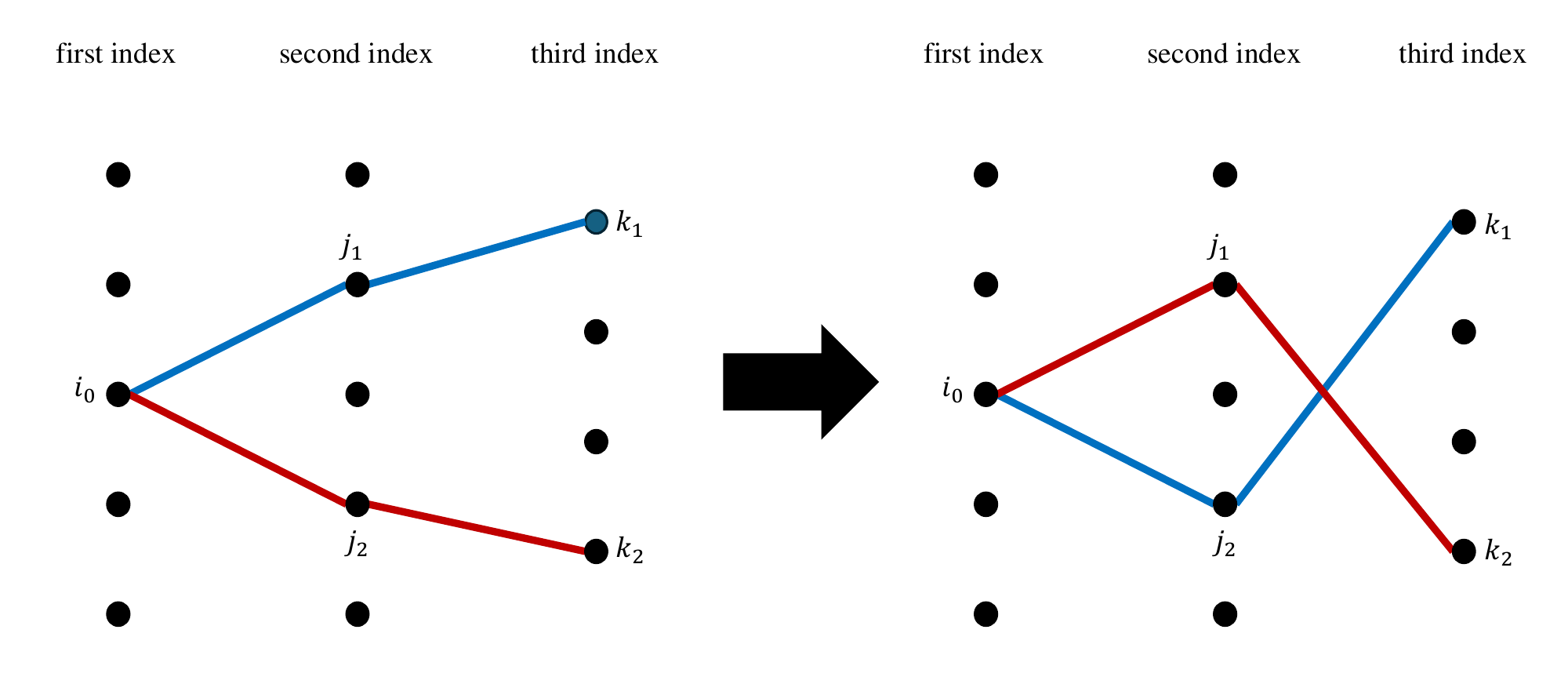}
    \caption{Concept of the Advanced Migration Matcher Algorithm via strings connecting pins on a wall.}
    \label{fig:strings}
\end{figure}

For the three dimensional variant, we use the same strategy, however, with three columns of pins on the wall (see Figure \ref{fig:strings}). In the first step, we use the 2D version of the algorithm to compute an initial guess $M_0=X\in \mathbb{N}^{m\times n\times r}$ which fulfils the $A$ and $C$ marginal equations. 

The main part of the algorithm is similar to the 2D variant: we look for two strings with same first index $(i_0,j_1,k_1)$ and $(i_0,j_2,k_2)$. A swap of the second index $j_1\leftrightarrow j_2$ will conserve all connections between the first and second index. As a result, the $A$ marginal equation will be conserved. Also, the swap will conserve the connections between the first and third dimension, thereby leaving the $C$ marginal untouched. The $B$-marginal, however, changes, since the number of $(j_1,k_1)$ and $(j_1,k_1)$ connections will be diminished by one, whereas $(j_1,k_2)$ and $(j_2,k_1)$ will grow by one. When suitable pairs are chosen, the algorithm makes progress in the right direction. However, convergence is not guaranteed (and usually not reached).

Currently, no integer-valued two-sided disaggregation is applied in the GEPOC parametrisation. So we do not give a reproducible statement of the two mentioned algorithms.

\newpage
\section{Source Data}
\label{sec:source_data}
In this section we describe all source data used for parametrisation of GEPOC as of September 2025. Note that the parameter calculation might change as soon as data changes its format, is not updated anymore, or better data becomes freely available.

\subsection{Sources for GEPOC ABM Parametrisation}
The following tables show the source files for the parametrisation of GEPOC. Most of the files are updated yearly, therefore we will indicate the most recent year for which population information on Jan $1^{st}$ is available as $y_0$. 
\begin{sourcefile}
{Bevölkerungsstand}
{Bevölkerung zu Jahresbeginn (einheitlicher Gebietsstand $y_0$)}
{Multiple files with population status for specific year (1.1.) for single age classes, sex and municipalities.}
{2002-$y_0$}
{municipalities\_districts ($y_0$)\\\textbf{Sex-Resolution} & $\{m,f\}$}
{$[0,100^+]$}
{OGD\_bevstandjbab2002\_....}
{\href{https://data.statistik.gv.at}{https://data.statistik.gv.at}}
%{2025-12-10}
\label{src:popBase}CC Namensnennung 4.0 International
\end{sourcefile}
\newpage
\begin{sourcefile}
{Bevölkerungsprognose 2014 bis 2024 - Zählbezirke (1) Wien}
{VIE-Bevölkerungsprognose für 250 Wiener Zählbezirke nach Altersgruppen und Geschlecht 2014 bis 2024}
{Population (and forecast) for Viennese registration-districts.}
{2014-2024}
{municipalities\_registrationdistricts (2014)\\\textbf{Sex-Resolution} & $\{m,f\}$}
{\{0,3,6,10,15,20,25,30,45,60,75\textsuperscript{+}\}}
{vie303}
{\href{https://www.data.gv.at/}{https://www.data.gv.at/}}
%{2021-12-10}
\label{src:popRegistrationdistricts}CC Namensnennung 4.0 International
\end{sourcefile}

\begin{sourcefile}
{Bevölkerung zum Jahresanfang 1952 bis 2101}
{Prognose zur Bevölkerung zum Jahresanfang 1952 bis 2101 nach Alter in Einzeljahren, Geschlecht und Hauptszenario}
{Population (forecast) for Austria for single age-classes, sex and federalstate}
{1952-2101}
{federalstates (1987)\\\textbf{Sex-Resolution} & $\{m,f\}$}
{$[0,100^+]$}
{OGD\_bevjahresanf\_PR\_BEVJA\_4}
{\href{https://data.statistik.gv.at/}{https://data.statistik.gv.at/}}
%{2025-09-24}
\label{src:popForecast}CC Namensnennung 4.0 International
\end{sourcefile}

\begin{sourcefile}
{Wanderungen mit dem Ausland ab 2002 nach Alter, Geschlecht und Staatsangehörigkeit}
{Außenwanderungen ab 2002 nach Jahr, Alter in Einzeljahren, Geschlecht, Staatsangehörigkeit}
{Immigrants and Emigrants since 2002 with respect to age and sex. No regional resolution!}
{2002-$(y_0-1)$}
{country (1945)\\\textbf{Sex-Resolution} & $\{m,f\}$}
{$[0,100^+]$}
{OGD\_bevwan020\_AUSSENWAND\_100}
{\href{https://data.statistik.gv.at/web/meta.jsp?dataset=OGD_bevwan020_AUSSENWAND_100}{https://data.statistik.gv.at/...}}
%{2025-05-26}
\label{src:migrationCountry}CC Namensnennung 4.0 International
\end{sourcefile}

\begin{sourcefile}
{Wanderungen mit dem Ausland von 2002 bis 2014 nach Altersgruppen, Gemeinde und Staatsangehörigkeit}
{Außenwanderungen 2002-2014 nach Jahr, 5jährige Altersgruppen, Gemeinde, Staatsangehörigkeit (Länderguppen)}
{Immigrants and Emigrants w.r. to 5year age groups and municipality. Additional info w.r. to origin countries. Gebietsstand 2022 not mentioned but used.}
{2002-2014}
{municipalities\_districts ($y_0$)\\\textbf{Sex-Resolution} & none}
{$\{0,5,10,15,20,25,30,35,40,45,50,55,60,65,70,75,80,85,90,95,100^+\}$}
{OGD\_bevwan020\_AUSSENWAND\_201}
{\href{https://data.statistik.gv.at/web/meta.jsp?dataset=OGD_bevwan020_AUSSENWAND_201}{https://data.statistik.gv.at/...}}
%{2022-06-08}
\label{src:migrationMuni_1}CC Namensnennung 4.0 International
\end{sourcefile}

\begin{sourcefile}
{Wanderungen mit dem Ausland ab 2015 nach Altersgruppen, Gemeinde und Staatsangehörigkeit}
{Außenwanderungen ab 2015 nach Jahr, 5jährige Altersgruppen, Gemeinde, Staatsangehörigkeit (Länderguppen)}
{Same as Source \ref{src:migrationMuni_1} but for years from 2015. No obvious reason found, why these two files are separated.}
{2015-$(y_0-1)$}
{municipalities\_districts ($y_0$)\\\textbf{Sex-Resolution} & none}
{$\{0,5,10,15,20,25,30,35,40,45,50,55,60,65,70,75,80,85,90,95,100^+\}$}
{OGD\_bevwan020\_AUSSENWAND\_202}
{\href{https://data.statistik.gv.at/web/meta.jsp?dataset=OGD_bevwan020_AUSSENWAND_202}{https://data.statistik.gv.at/...}}
%{2025-05-26}
\label{src:migrationMuni_2}CC Namensnennung 4.0 International
\end{sourcefile}
\newpage
\begin{sourcefile}
{Bevölkerungsbewegung 1961 bis 2100 nach Bundesland, Bewegungsarten und Szenarien}
{Bevölkerungsbewegung 1961 bis 2100, Bundesland, Bewegungsarten, Hauptszenario, Wachstum, Alterung, Fertilität, Wanderung}
{Forecast of total number of died, newborn, internal and external migrants by federalstate.}
{1961-2100 (births, deaths)\\
&2002-2100 (migrants)}
{federalstates (1987)\\\textbf{Sex-Resolution} & none}
{none}
{OGD\_bevbewegung\_BEV\_BEW\_2}
{\href{https://data.statistik.gv.at/web/meta.jsp?dataset=OGD_bevbewegung_BEV_BEW_2}{https://data.statistik.gv.at/...}}
%{2024-11-27}
\label{src:migrationForeacst}CC Namensnennung 4.0 International
\end{sourcefile}

\begin{sourcefile}
{Tabellensammlung Außenwanderung}
{Wanderungen mit dem Ausland (Außenwanderungen) nach Bundesländern 1996–$y_0$}
{Total number of emigrants and immigrants per federalstate.}
{1996-$(y_0-1)$}
{federalstates (1987)\\\textbf{Sex-Resolution} & none}
{none}
{Tabellensammlung\_Aussenwanderung\_\dots}
{\href{https://www.statistik.at/statistiken/bevoelkerung-und-soziales/bevoelkerung/migration-und-einbuergerung/wanderungen-insgesamt}{https://data.statistik.at/statistiken/...}}
%{2024-11-27}
\label{src:migrationBackcast}CC Namensnennung 4.0 International
\end{sourcefile}

\newpage
\begin{sourcefile}
{Demographische Zeitreihenindikatoren}
{Demographische Indikatoren sind international anerkannte und gebräuchliche Kennzahlen zur Beschreibung der Bevölkerungsstruktur und Bevölkerungsbewegung...}
{Various sources (one for each federalstate) with various indicators related to demography. This includes an age-specific death census, age-specific fertility, population with single age classes, and with reduced resolution internal and external migration counts per age-class.}
{1961-$y_0$ (fertility, death)\\
&2002-$y_0$ (migration)}
{federalstates (1987)\\\textbf{Sex-Resolution} & $\{m,f\}$}
{varies}
{xxxx\_Zeitreihenindikatoren\_1961\_y0.ods}
{\href{https://www.statistik.at/statistiken/bevoelkerung-und-soziales/bevoelkerung/demographische-indikatoren-und-tafeln/demographische-zeitreihenindikatoren}{https://www.statistik.at/statistiken/...}}
%{2025-09-15}
\label{src:indicators}CC Namensnennung 4.0 International
\end{sourcefile}

\begin{sourcefile}
{Demographische Indikatoren 1961 bis 2100}
{ Demografische Indikatoren nach Zeit, Bundesland und Szenarien}
{Forecast for various fertility and death related parameters: Total fertility rate, gross reproduction rate, net reproduction rate, average fertility age, and life expectancy of men and women at birth and at the age of 65. The quality of the dataset is limited by the fact that Statistics Austria rounded every indicator to the nearest integer.
}
{1961-2100}
{federalstates (1987)\\\textbf{Sex-Resolution} & $\{m,f\}$}
{none}
{OGD\_demoind\_DEM\_IND\_2}
{\href{https://data.statistik.gv.at/web/meta.jsp?dataset=OGD_demoind_DEM_IND_2}{https://data.statistik.gv.at/...}}
%{2024-11-27}
\label{src:indicatorsForecast}CC Namensnennung 4.0 International
\end{sourcefile}

\newpage
\subsection{Sources Specific for GEPOC ABM Geography Parametrisation}

\begin{sourcefile}
{GeoJSON/TopoJSON Austria (2016-2017)}
{Geo- and TopoJSON files of municipalities, districts and states in Austria, as of January 2017.}
{GeoJSON file containing all borders of all municipalities in Austria including districts within Vienna as of region-status 2017. Since it was sufficient for GEPOC, a 95\% simplified version was used.}
{2017}
{municipalities\_districts (2017)\\\textbf{Sex-Resolution} & -}
{-}
{gemeinden\_95\_geo.json}
{\href{https://github.com/ginseng666/GeoJSON-TopoJSON-Austria}{https://github.com/ginseng666/GeoJSON-TopoJSON-Austria}}
%{commit 2017-01-29}
\label{src:geojson2017}CC BY 4.0, Flooh Perlot
\end{sourcefile}

\begin{sourcefile}
{Zählbezirksgrenzen Wien}
{Zählbezirke sind statistische Definitionen. Die 23 Bezirke werden in 250 Zählbezirke unterteilt.}
{GeoJSON file containing all borders of the Viennese registration-districts. Although internally consistent, it does not seamlessly fit with Source \ref{src:geojson2017}}
{2019}
{municipalities\_registrationdistricts (2019)\\\textbf{Sex-Resolution} & -}
{-}
{ZAEHLBEZIRKOGD}
{\href{https://www.data.gv.at/katalog/dataset/stadt-wien_zhlbezirksgrenzenwien/resource/d1415d3f-57c9-4258-b515-ebd3ae52dc46}{https://www.data.gv.at/...}}
%{2019-01-01}
\label{src:geojsonVienna}CC Namensnennung 4.0 International
\end{sourcefile}

\newpage
\begin{sourcefile}
{GHS resident population grid}
{-}
{Raster image of Europe. Value per raster estimates the number of inhabitants.}
{2016}
{$100[m]\times100[m]$ raster map\\\textbf{Sex-Resolution} & -}
{-}
{Austria2016\_100m.json}
{\href{https://www.eea.europa.eu/data-and-maps/data/external/ghs-resident-population-grid}{https://www.eea.europa.eu/...}}
%{2016}
\label{src:ghsmap}Creative Commons Attribution 4.0 International
\end{sourcefile}

\subsection{Sources Specific for GEPOC ABM IM Parametrisation}

\begin{sourcefile}
{Wanderungen innerhalb Österreichs}
{Wanderungen innerhalb Österreichs ab 2002 (einheitlicher Gebietsstand $y_0$) }
{Number of internal migrants between (origin-destination) municipalities/districts in Austria w.r. to migration year and sex.}
{2002-$(y_0-1)$}
{municipalities\_districts ($y_0$)\\\textbf{Sex-Resolution} & $\{m,f\}$}
{none}
{OGDEXT\_BINNENWAND\_1}
{\href{https://data.statistik.gv.at/web/meta.jsp?dataset=OGDEXT_BINNENWAND_1}{https://data.statistik.gv.at/...}}
%{2024-05-28}
\label{src:internalMigrants}CC Namensnennung 4.0 International
\end{sourcefile}

\newpage
\begin{sourcefile}
{Binnenwanderungen innerhalb Österreichs ab 2002}
{Jahr und Alter in Einzeljahren nach Wanderungen innerhalb Österreichs und Politischer Bezirk / Wiener Gemeindebezirk - Herkunftsort nach Geschlecht}
{Total number of internal emigrants and immigrants per district collected manually from STATcube. Vienna and remaining Austria are collected in different files due to download limitations.}
{2002-$(y_0-1)$}
{districts\_districts ($y_0$)\\\textbf{Sex-Resolution} & $\{m,f\}$}
{$[0,95^+]$}
{-}
{\href{https://statcube.at/statistik.at/ext/statcube/jsf/tableView/tableView.xhtml}{https://statcube.at/\dots}}
%{2025-05-26}
\label{src:internalMigrantsStatCube}CC Namensnennung 4.0 International as long as the data is downloaded manually
\end{sourcefile}

\begin{sourcefile}
{Tabellensammlung Binnenwanderungen 2024}
{Wanderungen innerhalb Österreichs (Binnenwanderungen) zwischen und innerhalb der Bundesländer 1996–2024}
{Total number of internal emigrants and immigrants per federalstate without age-resolution.}
{1996-2024}
{federalstates (1987) \\\textbf{Sex-Resolution} & none}
{none}
{Tabellensammlung\_Binnenwanderungen\_2024}
{\href{https://www.statistik.at/statistiken/bevoelkerung-und-soziales/bevoelkerung/migration-und-einbuergerung/binnenwanderungen}{https://www.statistik.at/statistiken/\dots}}
%{2025-05-26}
\label{src:internalMigrantsBC}CC Namensnennung 4.0 International
\end{sourcefile}

\newpage
\section{GEPOC Parameter Calculation}
\label{sec:parameter_calculation}
In the following we describe how we calculate the GEPOC parameters as introduced in Section \ref{sec:model_parameters} from the source data introduced in Section \ref{sec:source_data} in a reproducible way. We will refer to the year for which the last population-census data is available as $y_0$. This will also correspond to the used area-status and we assume that the updated data for population change (migration, births, deaths, etc.) is available for this area-status and up to year $(y_0-1)$.

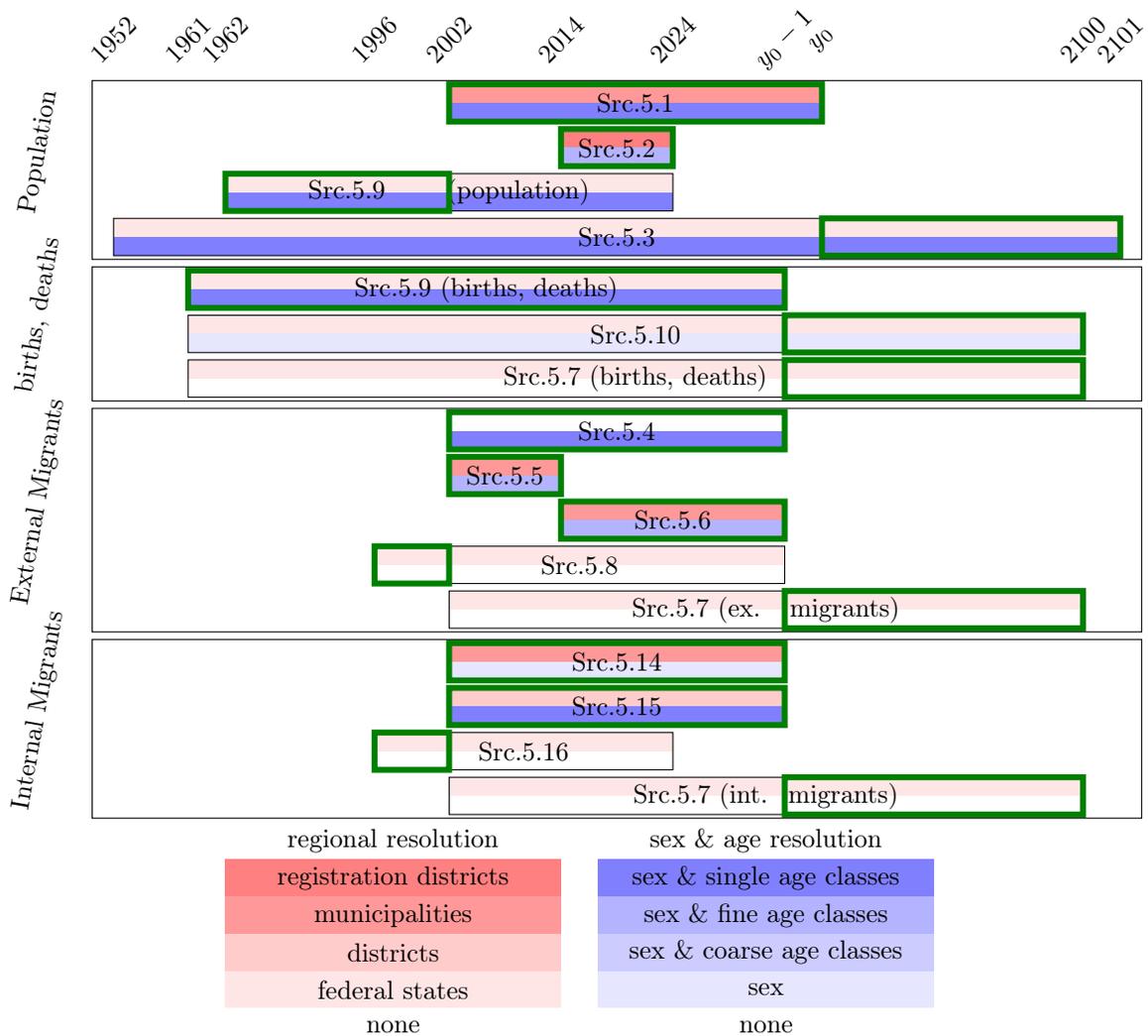
\begin{figure}
    \centering
    \begin{tikzpicture}[scale=0.5]
\node[rotate=45](T1952) at (0,0) {$1952$};
\node[rotate=45](T1961) at (2,0) {$1961$};
\node[rotate=45](T1962) at (3,0) {$1962$};
\node[rotate=45](T1996) at (7,0) {$1996$};
\node[rotate=45](T2002) at (9,0) {$2002$};
\node[rotate=45](T2014) at (12,0) {$2014$};
\node[rotate=45](T2024) at (15,0) {$2024$};
\node[rotate=45](Ty0m1) at (18,0) {$y_0-1$};
\node[rotate=45](Ty0) at (19,0) {$y_0$};
\node[rotate=45](T2100) at (26,0) {$2100$};
\node[rotate=45](T2101) at (27,0) {$2101$};

\tikzsrc{P1}{T1952}{\ref{src:popBase}}{T2002}{Ty0}{40}{50}{black};
\tikzusesrc{P1}{T2002}{Ty0};
\tikzsrc{P2}{P1}{\ref{src:popRegistrationdistricts}}{T2014}{T2024}{50}{30}{black};
\tikzusesrc{P2}{T2014}{T2024};
\tikzsrc{P3}{P2}
{\ref{src:indicators} \qquad (population)}{T1962}{T2024}{10}{50}{black};
\tikzusesrc{P3}{T1962}{T2002};
\tikzsrc{P4}{P3}
{\ref{src:popForecast}}{T1952}{T2101}{10}{50}{black};
\tikzusesrc{P4}{Ty0}{T2101};
\tikzbbox{P1}{P4}{Population};

\node[] (C0) at ($(P4)+(0,-0.2)$){};
\tikzsrc{C1}{C0}
{\ref{src:indicators} (births, deaths)}{T1961}{Ty0m1}{10}{50}{black};
\tikzusesrc{C1}{T1961}{Ty0m1};
\tikzsrc{C2}{C1}
{\ref{src:indicatorsForecast}}{T1961}{T2100}{10}{10}{black};
\tikzusesrc{C2}{Ty0m1}{T2100};
\tikzsrc{C3}{C2}
{\ref{src:migrationForeacst} (births, deaths)}{T1961}{T2100}{10}{0}{black};
\tikzusesrc{C3}{Ty0m1}{T2100};
\tikzbbox{C1}{C3}{births, deaths};

\node[] (M0) at ($(C3)+(0,-0.2)$){};
\tikzsrc{M1}{M0}
{\ref{src:migrationCountry}}{T2002}{Ty0m1}{0}{50}{black};
\tikzusesrc{M1}{T2002}{Ty0m1};
\tikzsrc{M2}{M1}
{\ref{src:migrationMuni_1}}{T2002}{T2014}{40}{30}{black};
\tikzusesrc{M2}{T2002}{T2014};
\tikzsrc{M3}{M2}
{\ref{src:migrationMuni_2}}{T2014}{Ty0m1}{40}{30}{black};
\tikzusesrc{M3}{T2014}{Ty0m1};
\tikzsrc{M4}{M3}
{\ref{src:migrationBackcast}}{T1996}{Ty0m1}{10}{0}{black};
\tikzusesrc{M4}{T1996}{T2002};
\tikzsrc{M6}{M4}
{\ref{src:migrationForeacst} (ex. \ \ migrants)}{T2002}{T2100}{10}{0}{black};
\tikzusesrc{M6}{Ty0m1}{T2100};
\tikzbbox{M1}{M6}{External Migrants};

\node[] (IM0) at ($(M6)+(0,-0.2)$){};
\tikzsrc{IM1}{IM0}
{\ref{src:internalMigrants}}{T2002}{Ty0m1}{40}{10}{black};
\tikzusesrc{IM1}{T2002}{Ty0m1};
\tikzsrc{IM2}{IM1}
{\ref{src:internalMigrantsStatCube}}{T2002}{Ty0m1}{20}{50}{black};
\tikzusesrc{IM2}{T2002}{Ty0m1};
\tikzsrc{IM3}{IM2}
{\ref{src:internalMigrantsBC}}{T1996}{T2024}{10}{0}{black};
\tikzusesrc{IM3}{T1996}{T2002};
\tikzsrc{IM4}{IM3}
{\ref{src:migrationForeacst} (int. \ migrants)}{T2002}{T2100}{10}{0}{black};
\tikzusesrc{IM4}{Ty0m1}{T2100};
\tikzbbox{IM1}{IM4}{Internal Migrants};
\tikzcolormap{3,-22}
\end{tikzpicture}
    \caption{Overview of the time periods covered by the different data-sources. The darker the colour the higher the resolution of the corresponding data, red for spatial resolution, blue for age \& sex resolution. Several data-sources are only used for a part of the time-frame which they cover, indicated by a green rectangle.}
    \label{fig:time_coverage}
\end{figure}

In general, calculation of parameters involves two steps. In the first and most complicated step, we compute harmonized census data (population, births, deaths, \dots) for as long a period as possible. Our goal is to have a continuous time-seres of census numbers ranging from the past into the far future with the same spatial and age-resolution. Browsing over the multiple stated source-files from Section \ref{sec:source_data}, it becomes clear that the age-, sex-, and regional resolutions of the files are inconsistent. In particular, forecasts are usually given on coarser regional and age resolution than historic data. This is visualised in Figure \ref{fig:time_coverage}. As mentioned in Section \ref{sec:algorithms}, we will harmonize all data on the finest level of resolution using one of the presented disaggregation algorithms.

In the second step we will use the harmonized data series to compute probabilities using the adapted Farr formula \ref{cor:farr_param}.

We will henceforth use the following notation for a disaggregation procedure with either of the two one-sided algorithms (Proportional Disaggregation method, Algorithm \ref{alg:prop_disagg}, and Huntington-Hill Disaggregation method, Algorithm \ref{alg:hh_disaggregation}). The expression
\begin{equation}\label{eq:deaggregation}
    X_1:(Y_1,R_1,S_1,A_1)\rightarrow (Y_2,R_2,S_2,A_2) \text{ via }X_2\text{ key } (K_1,K_2,\dots).
\end{equation}
means that dataset $X_1$ is disaggregated from year/region/sex/age resolution $(Y_1,R_1,S_1,A_1)$ to the finer resolution $(Y_2,R_2,S_2,A_2)$ using a distribution from dataset $X_2$. Hereby, $(K_1,K_2,\dots)$ must be a sub-vector of $(Y_1,R_1,S_1,A_1)$ and refers to all dimensions along which an individual distribution is used for disaggregation. To gain a better understanding, we refer to the first application of the notation below, since the concept is best explained on an example.

\subsection{Population \texorpdfstring{$\hat{P}$}{P}}
\label{sec:calculation_population}
Since providing a static micro-census of the population is one of the key objectives of GEPOC, our goal is to get a parameter set on the finest-possible age and regional resolution. Therefore, several data sources have to be disaggregated and joined. Note that in this processing strategy we will apply the disaggregation method of Huntington-Hill, Algorithm \ref{alg:hh_disaggregation}, since we require whole numbers for initialisation of the agent population. 

The following files pose the basis for the population processing.

{\setlength{\extrarowheight}{1mm}
\begin{center}
    \begin{tabular}{ccccc}
    \hline
    \makecell{\textbf{name}\\\textbf{of source}} &
    \makecell{Bevölkerungsstand} &
    \makecell{Bevölkerung zum \\ Jahresanfang\\ 1952 bis 2101} &
    \makecell{Demographische\\ Zeitreihenindikatoren} &
    \makecell{Bevölkerungsprognose\\ 2014 bis 2024\\ Zählbezirke (1) Wien} \\
    \textbf{source} & Source \ref{src:popBase} & Source \ref{src:popForecast} & 
    \makecell{Source \ref{src:indicators} \\ (Table 19) }
    &Source \ref{src:popRegistrationdistricts}\\
    \cline{1-5}
    \textbf{variable} & $S_{p}(y,r,s,a)$ & $S_{pf}(y,r,s,a)$ & $S_{i}^p(y,r,s,a)$ & $S_{pv}(y,r,s,a)$ \\
    $y\in$ & [2002,$y_0$] & [1952,2101] & [1962,2021] & [2014,2024]\\
    $r\in$ & municipalities\_districts & federalstates & federalstates & \makecell{registrationdistricts\\(only Vienna)}\\
    $s\in $ & $\{m,f\}$ & $\{m,f\}$ & $\{m,f\}$ & $\{m,f\}$\\
    $a\in$ & [0,100\textsuperscript{+}] & [0,100\textsuperscript{+}] & [0,95\textsuperscript{+}] & \{0,3,6,\dots, 75\textsuperscript{+}\}\\
    \hline
\end{tabular}
\end{center}
}
Note that we will use Source \ref{src:indicators} between 1962 and 2001 to generate a population census earlier than 2002 and not Source \ref{src:popForecast}, even though the latter would provide data until 1952 (see also, Figure \ref{fig:time_coverage}). The reason for this is that we observed that Source \ref{src:indicators} perfectly aligns with the high-resolution population data from Source \ref{src:popBase} for the overlapping years 2002 to 2021 whereas the other does not. Apparently, Source \ref{src:popForecast} should be interpreted as a forecast over its whole time period and incorporates corresponding uncertainty.
\begin{enumerate}
    \item We start processing with the population forecast (pf) $S_{pf}(y,r,s,a)$ (Source \ref{src:popForecast}) restricted to $y\in \{y_0+1,\dots,2101\}$ (earlier data will be ignored). To be compatible with $S_{p}$ (Source \ref{src:popBase}), we need to disaggregate it w.r. to a municipality distribution. 
    \item We aggregate the last three years of the historic population (p) census $S_{p}$: 
    \[X_1(r,s,a)=\sum_{y=y_0-2}^{y_0} S_{p}(y,r,s,a).\]
    \item We will use $X_1$ for disaggregation of $S_{pf}$ using the Huntington-Hill Disaggregation method:
    \begin{multline*}
        S_{pf}:([y_0+1,2101],\{\text{federalstates}\},\{m,f\},[0,100^+])\\
        \rightarrow ([y_0+1,2101],\{\text{municipalities\_districts}\},\{m,f\},[0,100^+])\\
        \text{ via }X_1\text{ key } (\{\text{federalstates}\},\{m,f\},[0,100^+])
    \end{multline*} 
    Since this is the first time we use the disaggregation notation introduced earlier, we describe the process in detail:
    \begin{enumerate}
        \item Disaggregation will be applied along the regional dimensions. So, for every federal-state id $r\in \{\textit{AT-1},\dots,\textit{AT-9}\}$, we compute the list $\vec{z}(r)$ of all municipality ids which lie within $r$. E.g.
        \[\vec{z}(\textit{AT-1})=(\textit{10101},\textit{10201},\dots,\textit{10932})\]
        \item According to the key-tuple $(K_1,K_2,\dots)=(\{\text{federalstates}\},\{m,f\},[0,100^+])$, we iterate over all federal-state ids $r$, sex $s\in \{m,f\}$, and age classes $a\in [0,100^+]$, and define
        \[\vec{P}(r,s,a)=\left(X_1(x,s,a)\right)_{x\in \vec{z}(r)}\]
        \item We initialise a data-set $X_2(\cdot,\cdot,\cdot,\cdot)$ with zero entries and same spatial resolution as $X_1$ and sex and age resolution as $S_{pf}$.
        \item For every year $y\in [y_0+1,2101]$, we iterate over all federal-states $r$, sex $s\in \{m,f\}$, and age classes $a\in [0,100^+]$. We use $\vec{P}(r,s,a)$ as the distribution vector of Algorithm \ref{alg:hh_disaggregation}, disaggregate $S_{pf}(y,r,s,a)$ into a vector $\vec{q}$, and integrate it into $X_2$ via $X_2\left(y,\vec{z}(r)_i,s,a\right):=\vec{q}_i$.
    \end{enumerate}
    The final dataset $X_2$ is a refined forecast until 2101:
    \[X_2(y,r,s,a): y\in [y_0+1,2101],r\in \text{municipalities\_districts},s\in\{m,f\},a\in [0,100^+]\]
    It has the following properties:
    \begin{itemize}
        \item The municipalities\_districts-distribution per federal-state matches, within the numerical accuracy of the algorithm, the distribution of $X_1$ from $(y_0-2)$ to $y_0$.
        \item It is integer-valued, which is reasonable for a population census and important for agent-based model parametrisation.
        \item Aggregation of $X_2$ from municipalities\_districts to federalstates would perfectly result in $S_{pf}$.
    \end{itemize}
    \item In the next step, we extend the population data-set beyond the lower time-frame bound (2002) from Source \ref{src:popBase}.
    using $S_i^p$ between 1962 and 2001.
    \item Since $S_i^p$ has $a_{max}=95$, whereas $a_{max}=100$ is needed to match with $S_{p}$ and $X_2$, we first need to disaggregate the final age-cohort.
    \item For all federal-state ids $r$, sex $s\in\{m,f\}$ and age $a\geq 95$ define $X_3(r,s,a):=\sum_{r'\in r}\sum_{y=2002}^{2004}S_{p}(y,r',s,a)$. This way we receive an age distribution for persons $\geq 95$. We use it to disaggregate the $95^+$ cohort from $S_i^p$ as follows:
    \begin{multline*}S_i^p:([1962,2001],\{\text{federalstates}\},\{m,f\},95^+)\\
    \rightarrow ([1962,2001],\{\text{federalstates}\},\{m,f\},[95,100^+])\\
    \text{ via }X_3\text{ key } (\{\text{federalstates}\},\{m,f\})\end{multline*}
    The resulting dataset $X_4$ has now the required $a_{max}=100$:
    \[X_4(y,r,s,a): y\in [1962,2001],r\in \{\text{federalstates}\},s\in\{m,f\},a\in [0,100^+]\]
    \item Furthermore we extrapolate from federalstates to municipalities\_districts. Analogous to steps 2 and 3, we define
    \[X_5(r,s,a)=\sum_{y=2002}^{2004} S_{p}(y,r,s,a)\]
    and distribute
    \begin{multline*}
    X_4:([1962,2001],\{\text{federalstates}\},\{m,f\},[0,100^+])\\
    \rightarrow ([1962,2001],\{\text{municipalities\_districts}\},\{m,f\},[0,100^+])\\ 
    \text{ via }X_5\text{ key } (\{\text{federalstates}\},\{m,f\},[0,100^+])
    \end{multline*}
    using the Huntington-Hill Disaggregation method. The resulting data-set $X_6$ has the same resolutions as $S_{p}$ and $X_2$: 
    \[X_6(y,r,s,a): y\in [1962,2001],r\in \{\text{municipalities\_districts}\},s\in\{m,f\},a\in [0,100^+]\]
    Since their time-frames [1962,2001],[2002,$y_0$] and [$y_0+1$,2101] fit seamlessly, we merge them into a combined dataset $X_7$.
    \[X_7(y,r,s,a): y\in [1962,2101],r\in \{\text{municipalities\_districts}\},s\in\{m,f\},a\in [0,100^+]\]
    \item It remains to disaggregate Viennese districts into registration-districts using the corresponding population census for Vienna $S_{pv}$ (Source \ref{src:popRegistrationdistricts}). This is challenging due to (a) the unusual age-classes and (b) the limited time-frame of $S_{pv}$.
    
    We define the auxiliary data-set
    \[X_8(y,r,s,a)=S_{pv}(\min(\max(y,2014),2023),r,s,a).\]
     to extend the time-frame of $S_{pv}$. We use it for the disaggregation process of $X_7$ for the Viennese regions using the Huntington-Hill Disaggregation method:
    \begin{multline*}X_7:([1962,2101],\{\text{districts (Vienna)}\},\{m,f\},[0,100^+])\\
    \rightarrow ([1962,2101],\{\text{registrationdistricts (Vienna)}\},\{m,f\},[0,100^+])\\
    \text{ via }X_8\text{ key }([1962,2101],\{\text{districts (Vienna)}\},\{m,f\},\{0,3,6,\dots,75^+\})\end{multline*}
    The resulting data-set $\hat{P}$ finally has the target resolution
    \[\hat{P}(y,r,s,a): y\in [1962,2101],r\in \{\text{municipalities\_registrationdistricts}\},s\in\{m,f\},a\in [0,100^+]\]
    and is the final result of the population data processing.
    \item To be more flexible, we aggregated the data to coarser regional-levels as well.
\end{enumerate}
\begin{parameterfile}{Population $\hat{P}$}{municipalities\_registrationdistricts}
\hline
\multicolumn{2}{c}{\textbf{Contents}\label{param:population}}\\
\hline
\multicolumn{2}{p{15cm}}{Population of Austria per age, region, sex at the start of the years between 1962 and 2101}\\
\hline
\multicolumn{2}{c}{\textbf{Resolution}}\\
\hline
\textbf{Time-Frame} & $[1962,2101]$\\
\textbf{Regional-Level} & municipalities\_registrationdistricts\\
\textbf{Age-Resolution} & $[0,100^+]$\\
\hline
\textbf{Other Regional-Levels} & 
municipalities\_districts / municipalities / districts\_districts / districts / federalstates / country\\
\hline
\end{parameterfile}

In comparison with the ground-truth provided by Statistics-Austria, we would score the validity of the parameter values (0 invalid, 10 fully valid) as follows:
\begin{center}
    \begin{tabular}{c|c|c|c|c|p{5cm}}
    years & regions & sex & age & score & description\\
    \hline
    $[2002,y_0]$ & actual municipalities & all & all & 10 & raw data\\
    $[2002,y_0]$ & Viennese registrationdistricts & all & all & 9.5 & statistical linkage with population data from Vienna\\
    $[y_0+1,2101]$ & all & all & $[0,100^+]$ & 8 & federalstates extrapolated to municipalities and registrationdistricts using the distribution in years 2022-2024.\\
    $[1990,2001]$ & all & all & $[0,94]$ & 8 & federalstates extrapolated to municipalities and registrationdistricts using the distribution in years 2002-2004.\\
    $[1990,2001]$ & all & all & $[95,100^+]$ & 7 & age cohorts extrapolated using the distribution in years 2002-2004.\\
    \end{tabular}
\end{center}

\subsection{Births \texorpdfstring{$\hat{B}_m,\hat{B}$}{Bm,B} and Probability of a Male Child \texorpdfstring{$\hat{\alpha}_m$}{alpham}}
\label{sec:calculation_birth}
In this section we compute a harmonized census for births (by sex/age of children and by sex/age of corresponding mothers).
\subsubsection{Births by Age of the Mother \texorpdfstring{$\hat{B_m}$}{Bm}}
We start by computing births by the age of the mother. Like the population data, we need to disaggregate and join a series of different data with different resolutions. The situation becomes even more complicated, since forecast information for the age distribution of births is only available in form of fertility rates and mean fertility age (see Source \ref{src:indicators}). Thus, additional assumptions need to be made. 

Moreover, we want to emphasise that computed births are no longer required to be integers, since we will only use them in form of probabilities in the model (see later). Therefore, we will use the Proportional Disaggregation algorithm, Algorithm \ref{alg:prop_disagg}, because it perfectly conserves the distribution.

We will use the following sources for computing births:
{\setlength{\extrarowheight}{1mm}
\begin{center}
    \begin{tabular}{ccccc}
    \hline
    \makecell{\textbf{name}\\\textbf{of source}}&
    \makecell{Demographische\\  Zeitreihen\\-indikatoren} &
    \makecell{Bevölkerungs\\-bewegung\\ 1961 bis 2100\\ nach Bundesland,\\ Bewegungsarten\\ und Szenarien} &
    \makecell{Demographische\\ Indikatoren\\ 1961 bis 2100} &
    \makecell{Population $\hat{P}$} \\
    \textbf{source} & S. \ref{src:indicators} (Table 1)& S. \ref{src:migrationForeacst} (births) & S. \ref{src:indicatorsForecast} (fertility age) & \makecell{Parameter\\Value \ref{param:population}}\\
    \cline{1-5}
    \textbf{variable} & $S_i^b(y,r,s,a)$ & $S_{mf}^b(y,r)$ & $S_{if}^b(y,r)$ & $\hat{P}(y,r,s,a)$ \\
    $y\in$ & [1961,$y_0-1$] & [1961,2100] & [1961,2100] & [1962,2101]\\
    $r\in$ & federalstates & federalstates & federalstates & federalstates\\
    $s\in $ & $\{m,f\}$ & none & - & $\{m,f\}$\\
    $a\in$ & $[15,49]$ & none & - & $[0,100^+]$\\
    \hline
\end{tabular}
\end{center}
}
The population-parameters $\hat{P}$ derived in the previous section are required to compute birth rates from births and vice versa.

\begin{figure}
    \centering
    \includegraphics[width=0.8\linewidth]{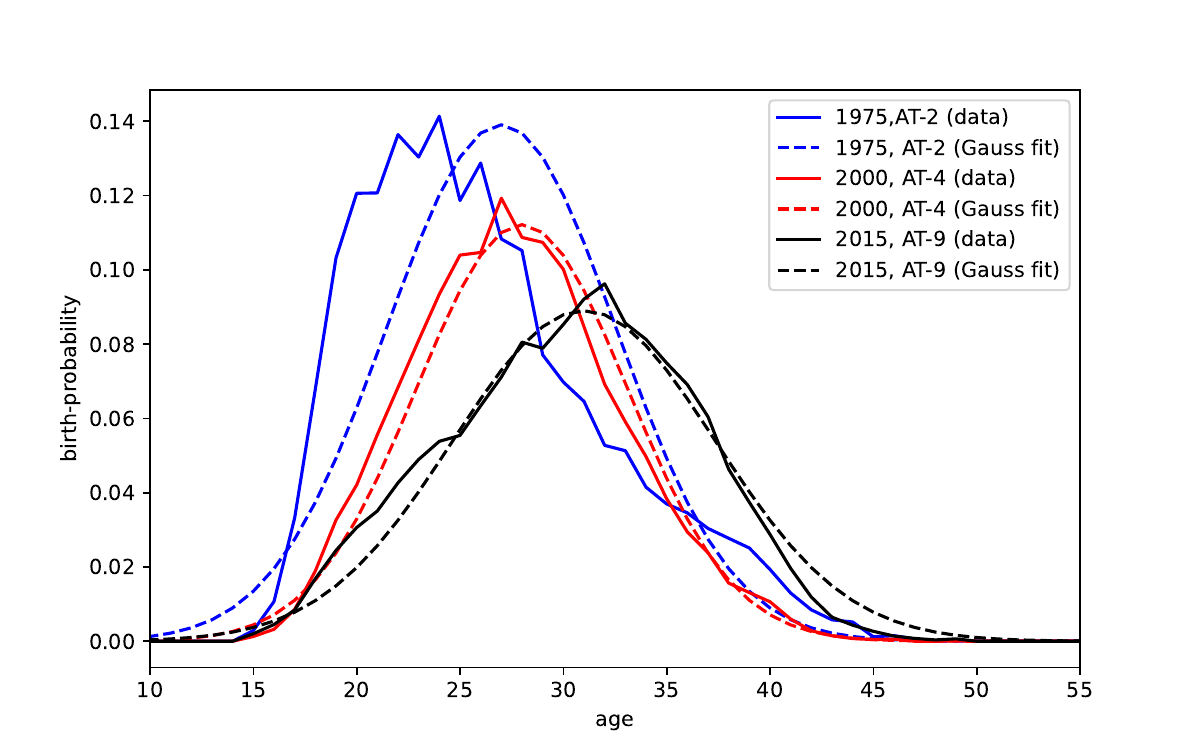}
    \caption{Fit of the Gaussian bell curve through the relative births per female inhabitant ($S_i^b/1000$)}
    \label{fig:gauss_fit}
\end{figure}

\begin{enumerate}
    \item First of all, we use the age specific births from the time-series indicators source (Table 1 ``Lebendgeborene'' from Source \ref{src:indicators}) by age of the corresponding mother. We refer to this data as $S_i^b(y,r,s,a)$.
    \item We will use this data in combination with the female population $\hat{P}(y,r,f,a)$ to compute the age-dependent average fertility rate (see Definition \ref{def:avgrate}) by using the classic approximation for the average population. The data suggests that the rate for women outside of the given age-range $[15,49]$ can be set to zero. So we define:
    \[X_1(y,r,f,a)=\begin{array}{cc}\frac{S_i^b(y,r,f,a)}{0.5(\hat{P}(y,r,f,a)+\hat{P}(y+1,r,f,a))},&a\in [15,49]\\
    0,&a\in [0,14]\vee a\in [50,100^+].
    \end{array}\]
    At this point, $X_1$ has the following resolution:
    \[X_1(y,r,s,a): y\in [1962,y_0-1],r\in \{\text{federalstates}\},s\in\{f\},a\in [0,100^+].\]
    Note that we will drop data for year 1961, since no population data is available.
    \item For computing the forecast, we make use of the migration forecast $S_{mf}^b(y,r)$ (Source \ref{src:migrationForeacst}), which also includes births and deaths, and the fertility related fields from the indicator forecast $S_{if}^b(y,r)$ (Source \ref{src:indicatorsForecast}). The prior contains the \textit{total number of births} for a given year and federalstate without further specifying age and sex, neither of the child nor the mother. The latter contains an estimate about the dynamics of the \textit{mean-fertility age} (MAC, see Definition \ref{def:fertilityrates}) for every federalstate. That means, for each year and federalstate, the forecast of 101 birth-probabilities (one for each age-class) will be based on two scalar variables.
    \item For each year $y\in [y_0,2100]$ and federalstate $r\in \{\text{AT-1},\dots,\text{AT-9}\}$ we define the following optimisation problem:
    
    Let $\vec{b}\in (\mathbb{R}^+)^{101}$ stand for the average age-dependent birth-rate vector for year $y$ and federalstate $r$ then
    \[f_1:(\mathbb{R}^+)^{101}\rightarrow \mathbb{R}^+: \vec{b}\mapsto \sum_{i=0}^{100}\vec{b}_i\frac{\hat{P}(y,r,f,i)+\hat{P}(y+1,r,f,i)}{2}\]
     maps the rates to the \textit{total number of births} and
    \[f_2:(\mathbb{R}^+)^{101}\rightarrow \mathbb{R}^+: \vec{b}\mapsto \frac{\sum_{i=0}^{100}i\vec{b}_{i+1}}{\sum_{i=0}^{100}\vec{b}_{i+1}}\]
    maps them to the MAC. With these functions, we aim to minimise
    \begin{equation}\label{eq:optproblem_births}
        F:(\mathbb{R}^+)^{101}\rightarrow \mathbb{R}^+: \vec{b}\mapsto \frac{|f_1(\vec{b})-S_{mf}^b(y,r)|}{100}+|f_2(\vec{b})-S_{if}^b(y,r)|.
    \end{equation}
    The scaling factor $1/100$ balances that the two objectives are on different scales (MAC ~30, births $>10000$). With $101$ degrees of freedom, this optimisation problem would be heavily underdetermined. Consequently, we reduce the degrees of freedom by
    \begin{equation}\label{eq:gauss}
    G:\mathbb{R}^3\rightarrow (\mathbb{R}^+)^{101}: \vec{\theta}\mapsto \vec{b}=\left(\vec{\theta}_1 \exp\left(-\frac{((i-1)-\vec{\theta}_2)^2}{\vec{\theta}_3^2}\right)\right)_{i=0}^{100}.
    \end{equation}
    Using the Gaussian bell curve fit we achieve the reduced minimisation problem $F\circ G$ which has three-dimensional input. As seen in Figure \ref{fig:gauss_fit}, the Gaussian bell curve $G$ is a proper model for the age-distribution of the birth rates later than 1990, after which the distribution stopped being skewed to the left.
    \item After defining the minimisation problem $F\circ G$ we solve it numerically for every federalstate and year $y\in [y_0,2100]$ using a Broyden-Fletcher-Goldfarb-Shanno Quasi Newton method. For the optimised $\vec{\theta}_{opt}$, we compute $\vec{b}_{opt}=G(\vec{\theta}_{opt})$ and use it to get the entries for the age-dependent birth forecast $X_2$:
    \[X_2(y,r,f,i)=\vec{b}_{opt,i}\frac{\hat{P}(y,r,f,i)+\hat{P}(y+1,r,f,i)}{2}.\]
    \item We combine $X_1$ and $X_2$ into $\hat{B}_m$ which has now the following resolution:
    \[\hat{B}_m(y,r,s,a): y\in [1962,2100],r\in \{\text{federalstates}\},s\in\{f\},a\in [0,100^+].\]
    \item To be more flexible we computed the data for other regional-levels as well. Hereby the data was not only aggregated to coarser levels but also disaggregated to a finer one. Hereby the fertility rates were disaggregated, assuming that the rates are equivalent for each subregion. Recomputing the births using the fine-grained population data gives an estimate for high-resolution birth data.
\end{enumerate}

\begin{parameterfile}{Births per Mother $\hat{B}_m$}{federalstates}
\hline
\multicolumn{2}{c}{\textbf{Contents}\label{param:births_mothers}}\\
\hline
\multicolumn{2}{p{15cm}}{Total number of newborn children of female persons with age $a$ living in region $r$ in the course of year $y$.}\\
\hline
\multicolumn{2}{c}{\textbf{Resolution}}\\
\hline
\textbf{Time-Frame} & $[1962,2100]$\\
\textbf{Regional-Level} & federalstates\\
\textbf{Age-Resolution} & $[0,100^+]$\\
\hline
\textbf{Other Regional-Levels} & districts\_districts / country\\
\hline
\end{parameterfile}
In comparison with the ground-truth provided by Statistics-Austria, we would score the validity of the parameter values (0 invalid, 10 fully valid) as follows:
\begin{center}
    \begin{tabular}{c|c|c|c|c|p{5cm}}
    years & regions & sex & age & score & description\\
    \hline
    $[1962,y_0-1]$ & all & all & all & 10 & raw data\\
    $[y_0,2100]$ & all & all & all & 6.5 & extrapolation method with a Gaussian bell curve to conserve certain sums (total births, average fertility age ...)
    \end{tabular}
\end{center}

\subsubsection{Birth Probability of a Male Child \texorpdfstring{$\hat{\alpha}_m$}{alpham}}
Next, we will compute the probability that a newborn person in Austria is male or female. This is done using Source \ref{src:indicators} for Austria as a whole. Table 1 ``Lebendgeborene'' includes the total number of male and female newborn since 1961 which can be used to compute the ratio. In the regarded time-frame, the ratio of male newborn varies between $0.5082$ in 1982 and $0.5170$ in 2012, yet, in general, no trend can be observed. The average lies at $0.513234$ which we will furthermore use as a static parameter value.

\begin{parameterfile}{Probability of a male newborn $\hat{\alpha}_m$}{-}
\hline
\multicolumn{2}{c}{\textbf{Contents}\label{param:male_prob}}\\
\hline
\multicolumn{2}{p{15cm}}{Probability that a newborn child is male: 
\[P(s=m|s\in \{f,m\})=\hat{\alpha}_m=0.513234\]}\\
\hline
\end{parameterfile}
\subsubsection{Births \texorpdfstring{$\hat{B}$}{B}}
Finally, combining probability $\hat{\alpha}_m$ with the total births per mother's age $\hat{B}_m(y,r,f,a)$ via
\begin{equation}
    \hat{B}(y,r,s,0)=\left\lbrace \begin{array}{ll}
        \hat{\alpha}_m\sum_{a=0}^{100}\hat{B}_m(y,r,f,a), & s=m, \\
         (1-\hat{\alpha}_m)\sum_{a=0}^{100}\hat{B}_m(y,r,f,a), & s=f,
    \end{array} \right.
\end{equation}
we get a proper estimate for the total number of newborns per sex of the newborn $\hat{B}$.
\begin{parameterfile}{Births $\hat{B}$}{federalstates}
\hline
\multicolumn{2}{c}{\textbf{Contents}\label{param:births}}\\
\hline
\multicolumn{2}{p{15cm}}{Total number of newborn children with sex $s$ born in region $r$ in the course of year $y$. Regional-levels can be districts\_districts, country or federalstates}\\
\hline
\multicolumn{2}{c}{\textbf{Resolution}}\\
\hline
\textbf{Time-Frame} & $[1962,2100]$\\
\textbf{Regional-Level} & federalstates\\
\textbf{Age-Resolution} & $\{0\}$\\
\hline
\textbf{Other Regional-Levels} & districts\_districts / country\\
\hline
\end{parameterfile}

\subsection{Deaths \texorpdfstring{$\hat{D}$}{D}}
\label{sec:calculation_death}
In this section we compute a harmonized census for deaths. Like the processing steps for the births, additional assumptions need to be made due to the forecast information. 

Again, we will use the Proportional Disaggregation method, Algorithm \ref{alg:prop_disagg} because it perfectly conserves the distribution.

\begin{figure}
    \centering
    \includegraphics[width=0.8\linewidth]{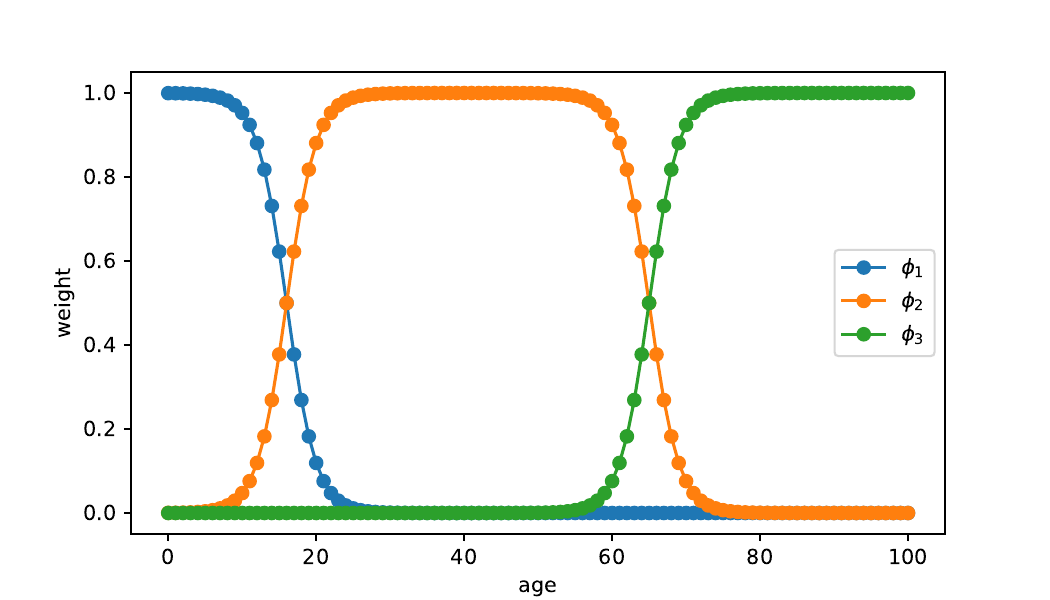}
    \caption{Activation functions $f_1, f_2$ and $f_3$ used to parametrise the distribution.}
    \label{fig:actfun}
\end{figure}

\begin{figure}
    \centering
    \includegraphics[width=0.8\linewidth]{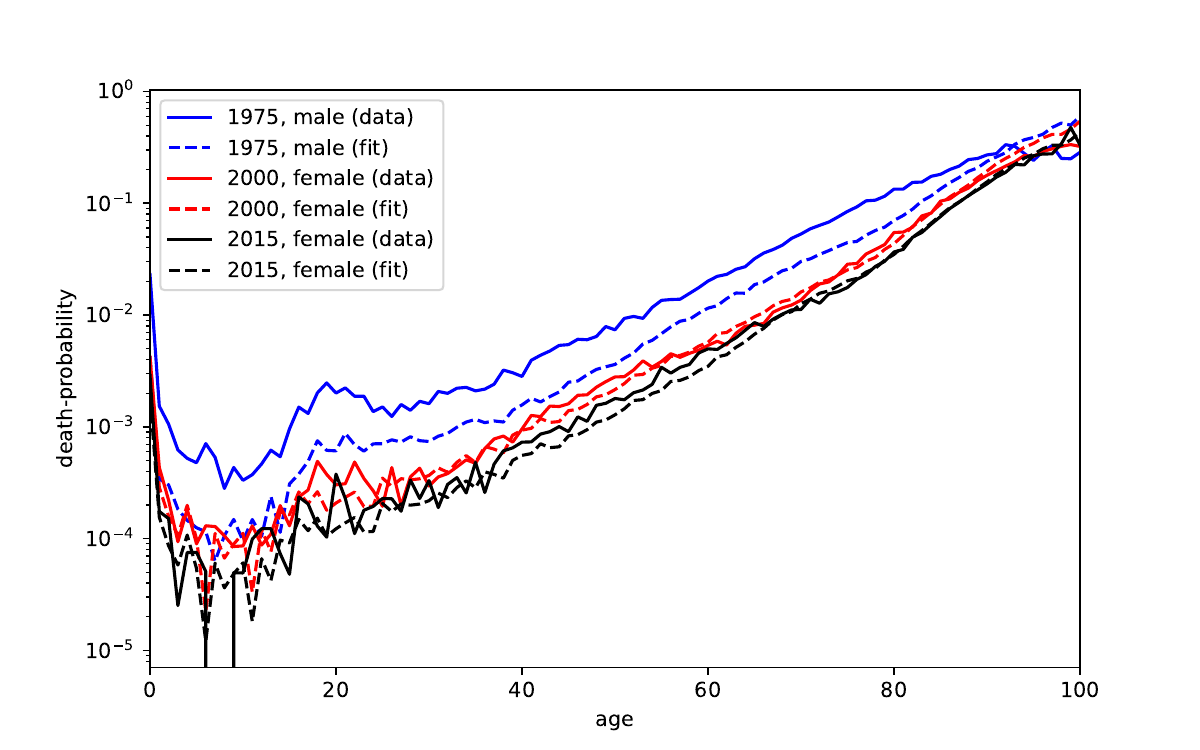}
    \caption{Fit of the parametrised curve through the death probabilities given in the data. The fit is better the more recent the data which justifies its use for the forecast.}
    \label{fig:death_fit}
\end{figure}

We will use the following sources for computing deaths:

{\setlength{\extrarowheight}{1mm}
\begin{center}
    \begin{tabular}{ccccc}
    \hline
    \makecell{\textbf{name}\\\textbf{of source}}&
    \makecell{Demographische\\  Zeitreihen\\-indikatoren} &
    \makecell{Bevölkerungs\\-bewegung\\ 1961 bis 2100\\ nach Bundesland,\\ Bewegungsarten\\ und Szenarien} &
    \makecell{Demographische\\ Indikatoren\\ 1961 bis 2100} &
    \makecell{Population $\hat{P}$} \\
    \textbf{source} & S. \ref{src:indicators} (Table 3)& S. \ref{src:migrationForeacst} (deaths) & \makecell{S. \ref{src:indicatorsForecast}\\(life expectancies)}& Parameter Value \ref{param:population}\\
    \cline{1-5}
    \textbf{variable} & $S_i^d(y,r,s,a)$ & $S_{mf}^d(y,r)$ & $S_{if}^d(y,s,a)$ & $\hat{P}(y,r,s,a)$ \\
    $y\in$ & [1961,$y_0-1$] & [1961,2100] & [1961,2100] & [1962,2101]\\
    $r\in$ & federalstates & federalstates & country & federalstates\\
    $s\in $ & $\{m,f\}$ & none & $\{m,f\}$ & $\{m,f\}$\\
    $a\in$ & $[0,a_{max}^+]$, $a_{max}$ varies & none & $\{0,65\}$ & $[0,100^+]$\\
    \hline
\end{tabular}
\end{center}
}
Due to the lack of age-information for the forecast, calculation of deaths works quite similar to the one for births: we will specify a minimisation problem. 
\begin{enumerate}
    \item We will start with the death table $S_i^d$, which contains the absolute number of deaths per federalstate. 
    \item In the next step we will take care about deficiencies in this data set due to the varying $a_{max}$ class. As with the births, we will drop year 1961 due to missing population data.
    \item Since $a_{max}=95$ for years before 1971, we use the $[95,96,97,98,99+]$ data for 1971-1973 to disaggregate the $95^+$ cohort with the Proportional Disaggregation method, Algorithm \ref{alg:prop_disagg}. Let $X_1(r,s,a)=\sum_{y=1971}^{1973}S_i^d(y,r,s,a)$ and compute
    \begin{multline*}
        S_i^d: ([1962,1970],\{\text{federalstates}\},\{m,f\},95^+)\\
        \rightarrow ([1962,1970],\{\text{federalstates}\},\{m,f\},\{95,96,97,98,99^+\})\\
        \text{ via }X_1\text{ key }(\{\text{federalstates}\},\{m,f\})
    \end{multline*}
    \item Since $a_{max}=99$ for years before 2017, we use the $[99,100^+]$ data for 2017-2019 to disaggregate the $99^+$ cohort. Let $X_2(r,s,a)=\sum_{y=2017}^{2019}S_i^d(y,r,s,a)$ and compute
    \begin{multline*}
        S_i^d: ([1962,2016],\{\text{federalstates}\},\{m,f\},99^+)\\
        \rightarrow ([1962,2016],\{\text{federalstates}\},\{m,f\},\{99,100^+\})\\
        \text{ via }X_2\text{ key }(\{\text{federalstates}\},\{m,f\})
    \end{multline*}
    At this stage we have the total number of deaths by age, region, and sex via
    \[X_2(y,r,s,a): y\in [1962,y_0],r\in \{\text{federalstates}\},s\in\{m,f\},a\in [0,100^+]\]
    \item In the next steps, we will compute death probabilities using the classic Farr Formula from Theorem \ref{thm:farr}. Our goal is to emulate Statistic Austria's death tables to create a forecast which is compatible with the provided indicators
    \[X_3(y,r,s,a)=\frac{X_2(y,r,s,a)}{\frac{1}{2}\left(\hat{P}(y,r,s,a)+\hat{P}(y+1,r,s,a)\right)+\alpha(a)X_2(y,r,s,a)}\]
    We identified that we can reproduce the death tables of Statistic Austria using $\alpha(0)=0.923$ and $\alpha(a)=0.5$ else.
    \item Analogous to the birth computation (see Section \ref{sec:calculation_birth}) we furthermore define a minimisation problem. First we need to specify the reference:
    
    For any given year $y\in [y_0,2100]$ five values are available: 
    The total number of deaths $d=\sum_{r}S_{mf}^d(y,r)$, the life-expectancies of $0$-year old males $c_{0,m}=S_{10}(y,m,0)$ and females $c_{0,f}=S_{10}(y,f,0)$, and the life expectancies of $65$-year old males  $c_{65,m}=S_{10}(y,m,65)$ and females $c_{65,f}=S_{10}(y,f,65)$. These five values will provide the target of the minimisation problem.
    
    The two death-probability vectors for males and females $\vec{q}_m$ and $\vec{q}_f$ pose the degrees of freedom. Furthermore we define the five functions $f_1(\vec{q}_m,\vec{q}_f)$, $f_2(\vec{q}_m)$,$f_3(\vec{q}_f)$, $f_4(\vec{q}_m)$ and $f_5(\vec{q}_f)$ which map the death probabilities onto the five desired outcomes: total deaths, and life-expectancy of 0/65-year old males and females.

    We furthermore define the target function for the optimisation:
    \begin{multline*}F:(\mathbb{R}^{+})^{2\cdot 101}\rightarrow \mathbb{R}^+:\\
    (\vec{q}_m,\vec{q}_f)\mapsto\left\lVert\frac{f_1(\vec{q}_m,\vec{q}_f)-d}{2000}\right\rVert_2+\left\lVert f_2(\vec{q}_m)-c_{0,m}\right\rVert_2+\left\lVert f_3(\vec{q}_f)-c_{0,f}\right\rVert_2+\left\lVert f_4(\vec{q}_m)-c_{65,m}\right\rVert_2+\left\lVert f_5(\vec{q}_f)-c_{65,f}\right\rVert_2\end{multline*}
    The weight of $1/2000$ was determined experimentally to give useful results, since $f_1$ is on a much higher scale than the other four outcomes. 
    \item 
    Before defining the five functions $f_1-f_5$, we reduce the degrees of freedom of the problem (currently $2\cdot 101=202$). Our approach to solve this problem is a parametrised reference distribution. We define two reference distributions $q_{ref,m}$, $q_{ref,f}$ from the last three available data years via
\[q_{ref,m}(a)=\frac{1}{3}(X_3(y_0-2,m,a)+X_3(y_0-1,m,a)+X_3(y_0,m,a)).\]
\[q_{ref,f}(a)=\frac{1}{3}(X_3(y_0-2,f,a)+X_3(y_0-1,f,a)+X_3(y_0,f,a)).\]
Note that for the most recent update, the years 2020,2021 and 2022 were omitted in this computation due to slightly different mortality by the COVID-19 crisis.

Furthermore, we parametrise the reference distributions with six parameters, three for each sex. We define 
\begin{equation}G:\mathbb{R}^6\rightarrow (\mathbb{R}^{+})^{2\cdot 101}:
(\vec{\theta})\mapsto\left( \begin{matrix}
   (\theta_1\phi_1(a)+\theta_2\phi_2(a)+\theta_3\phi_3(a))\cdot q_{ref,m}(a),\quad a\in[0,100]\\
   (\theta_4\phi_1(a)+\theta_5\phi_2(a)+\theta_6\phi_3(a))\cdot q_{ref,f}(a),\quad a\in[0,100]
\end{matrix}\right).
\end{equation}
The three actuation functions $\phi_1,\phi_2,\phi_3$ are chosen to smoothly raise and lower death probabilities in certain age groups. We chose
\begin{align}
\phi_1(a)&=1-\frac{1}{1+\exp(-0.5\cdot(a-16))},\\
\phi_2(a)&=\frac{1}{1+\exp(-0.5\cdot(a-16))}-\frac{1}{1+\exp(-0.5\cdot(a-65))},\\
\phi_3(a)&=\frac{1}{1+\exp(-0.5\cdot(a-65))}.
\end{align}
Figure \ref{fig:actfun} displays the three functions. However, with $F\circ G$ the optimisation problem becomes $6$ dimensional. Figure \ref{fig:death_fit} shows, how well the strategy fits historic data, analogous to Figure \ref{fig:gauss_fit} for the birth parameter processing.
\item We furthermore define $f_1$ to $f_5$. The first one maps the probabilities onto the total number of deaths per federalstate. We compute this using the inverse of the Farr formula and a sum over all ages and sex:
\[f_1(\vec{q}_m,\vec{q}_f)=\sum_{s\in\{m,f\}}\sum_{i=1}^{101}\frac{\hat{P}(y,r,s,i-1)+\hat{P}(y+1,r,s,i-1)}{2}\frac{(\vec{q}_s)_i}{1-\alpha(i-1)(\vec{q}_s)_i}\]

Moreover, $f_2$ to $f_5$ use the life expectancy formula from Definition \ref{def:life_expectancy}:
\[f_2(\vec{q}_m)=LE(\vec{q}_m,y,r,m,0),\ f_3(\vec{q}_f)=LE(\vec{q}_f,y,r,f,0),\]
\[f_4(\vec{q}_m)=LE(\vec{q}_m,y,r,m,65),\ f_5(\vec{q}_f)=LE(\vec{q}_f,y,r,f,65).\]
\item After defining the minimisation problem $F\circ G$ we solve it numerically for every federalstate and year $y\in [y_0,2100]$ using a Broyden-Fletcher-Goldfarb-Shanno (BFGS) Quasi Newton method. The method requires feasible bounds and initial values to converge properly. We furthermore call the optimised vector $\vec{\theta}_{opt}$.  
\item Finally, we compute the target vectors $\vec{q}_{m},\vec{q}_{f}$ for the forecast by $G(\vec{\theta}_{opt})$ and apply the Farr Formula inversely
\[\frac{\hat{P}(y,r,s,i-1)+\hat{P}(y+1,r,s,i-1)}{2}\frac{(\vec{q}_s)_i}{1-\alpha(i-1)(\vec{q}_s)_i}\]
to get the total number of deaths per age. We merge them with $X_3$ to get a complete data-set $\hat{D}$:
\[\hat{D}(y,r,s,a):y\in [1962,2100],r\in \{\text{federalstates}\},s\in \{m,f\},a\in [0,100^+]\]
\item To be more flexible we also computed the death data for other regional-levels as well. For those with finer regional resolution we applied the same strategy as for the birth, i.e. assuming that death rates are equivalent in subregions.
\end{enumerate}

\begin{parameterfile}{Deaths $\hat{D}$}{federalstates}
\hline
\multicolumn{2}{c}{\textbf{Contents}\label{param:deaths}}\\
\hline
\multicolumn{2}{p{15cm}}{Total number of died persons living in region $r$ with sex $s$ and $a$-th birthday in year $y$.}\\
\hline
\multicolumn{2}{c}{\textbf{Resolution}}\\
\hline
\textbf{Time-Frame} & $[1962,2100]$\\
\textbf{Regional-Level} &  federalstates \\
\textbf{Age-Resolution} & $[0,100^+]$\\
\hline
\textbf{Other Regional-Levels} & districts\_districts / country \\
\hline
\end{parameterfile}
In comparison with the ground-truth provided by Statistics-Austria, we would score the validity of the parameter values (0 invalid, 10 fully valid) as follows:
\begin{center}
    \begin{tabular}{c|c|c|c|c|p{5cm}}
    years & regions & sex & age & score & description\\
    \hline
    $[1962,y_0-1]$ & all & all & $[0,94]$ & 10 & raw data\\
    $[1971,y_0-1]$ & all & all & $[0,98]$ & 10 & raw data\\
    $[2017,y_0-1]$ & all & all & $[0,100^+]$ & 10 & raw data\\
    $[1961,1970]$ & all & all & $[95,100^+]$ & 9 & distributed deaths from 95+ according to distribution for 95-99 as of 1971-1973\\
    $[1971,2016]$ & all & all & $[99,100^+]$ & 9 & distributed deaths from 95+ according to distribution 99-100+ as of 2017-2019\\
    $[y_0,2100]$ & all & all & $[99,100^+]$ & 6 & extrapolation method with the probability distribution from 2017-2019 to conserve certain sums (total deaths, life expectancies,...)\\
    \end{tabular}
\end{center}

\subsection{Emigrants \texorpdfstring{$\hat{E}$}{E}}
\label{sec:calculation_emigration}
In this section we compute a harmonized census for emigrants.

Again, we will use the Proportional Disaggregation method, Algorithm \ref{alg:prop_disagg} because it perfectly conserves the distribution.

We will use the following sources:
{\setlength{\extrarowheight}{1mm}
\begin{center}
    \begin{tabular}{cccc}
    \hline
    \makecell{\textbf{name}\\\textbf{of source}}&
    \makecell{Wanderungen mit \\dem Ausland von \\2002 bis 2014\\ nach Altersgruppen,\\ Gemeinde und\\Staatsangehörigkeit} &
    \makecell{Wanderungen mit \\dem Ausland ab 2015 \\nach Altersgruppen, \\Gemeinde und \\Staatsangehörigkeit} &
    \makecell{Wanderungen mit \\dem Ausland ab 2002 \\nach Alter, \\Geschlecht und\\ Staatsangehörigkeit} \\
    \textbf{source} & S. \ref{src:migrationMuni_1} (emigration) & S. \ref{src:migrationMuni_2} (emigration) & S. \ref{src:migrationCountry} (emigration) \\
    \hline
    \textbf{variable} & $S_{m1}^e(y,r,s,a)$ & $S_{m2}^e(y,r,s,a)$ & $S_{mc}^e(y,s,a)$  \\
    $y\in$ & [2002,2014] & [2015,$y_0-1$] & [2002,$y_0-1$] \\
    $r\in$ & municipalities\_districts & municipalities\_districts & country \\
    $s\in $ & none & none & $\{m,f\}$ \\
    $a\in$ & $\{0,5,\dots,100^+\}$ & $\{0,5,\dots,100^+\}$ & $[0,100^+]$ \\
    \hline
    \hline
    \makecell{\textbf{name}\\\textbf{of source}}&
    \makecell{Bevölkerungs\\-bewegung \\1961 bis 2100 \\nach Bundesland,\\ Bewegungsarten\\ und Szenarien} &
    \makecell{Tabellensammlung \\Außenwanderung} &
    \makecell{Population $\hat{P}$} \\
    \textbf{source} &  S. \ref{src:migrationForeacst} (emigrants) & S. \ref{src:migrationBackcast} (emigrants) & Parameter Value \ref{param:population} \\
    \hline
    \textbf{variable} & $S_{mf}^e(y,r)$ & $S_{mb}^e(y,r)$ & $\hat{P}(y,r,s,a)$\\
    $y\in$ & [2002,2100] & [1996,$y_0$] &[1962,2101]\\
    $r\in$  & federalstates & federalstates & districts\_districts\\
    $s\in $ & none & none & $\{m,f\}$\\
    $a\in$  & none & none & $[0,100^+]$\\
    \hline
\end{tabular}
\end{center}
}
\begin{enumerate}
    \item In the first step we merge the two datasets $S^e_{m1}$ and $S^e_{m2}$ from the two sources with external migration data (Sources \ref{src:migrationMuni_1} and Source \ref{src:migrationMuni_2}). They contain total number of emigrants per year, municipality\_district resolution and five-year age-classes and can be merged seamlessly into a data set $X_1$:
    \[X_1(y,r,a):y\in [2002,y_0-1],r\in \{\text{municipalities\_districts}\},a\in \{0,5,\dots,100^+\}.\]
    \item To get a sex-resolution and a finer age-resolution, we use $S_{mc}^e$ from Source \ref{src:migrationCountry} which contains single-age resolved migration data on country level. Since there is a straightforward mapping between the 5-year and single-year age-classes, we will use $S_{mc}^e$ to fine-tune data from $X_1$ with the Proportional Disaggregation strategy:
    \begin{multline*}
        X_1: ([2002,y_0-1],\{\text{municipalities\_districts}\},\{0,5,\dots,100^+\})\\
        \rightarrow ([2002,y_0-1],\{\text{municipalities\_districts}\},\{m,f\},[0,100^+])\\
        \text{ via }S_{mc}^e\text{ key }(\{0,5,\dots,100^+\}).
    \end{multline*}
    The resulting $X_2$ has a sex resolution and single-age classes $[0,100^+]$.
    \item For the parameter values beyond $[2002,y_0-1]$ we use a similar yet much simpler strategy as for births and deaths. Since the only quantity available for computing the forecast is a scalar total number of emigrants per federalstate from $S_{mf}^e$, we can only assume that the age- and regional distribution of emigrants remains the same. We define
    \[X_3(r,s,a)=\sum_{y=y_0-3}^{y_0-1}X_1(y,r,s,a)\]
    and disaggregate $S_{mf}^e$:
    \begin{multline*}
        S_{mf}^e: ([y_0,2100],\{\text{federalstates}\},none,none)\\
        \rightarrow ([y_0,2100],\{\text{municipalities\_districts}\},\{m,f\},[0,100^+])\\
        \text{ via }X_3\text{ key }(\{\text{federalstates}\}).
    \end{multline*}
    We merge the result with $X_2$ to get
    \[X_4(y,r,s,a):y\in [2002,2100],r\in \{\text{municipalities\_districts}\},s\in \{m,f\},a\in [0,100^+].\]
    \item Since the historic data $S^e_{mb}$ has precisely the same resolution as the forecast data, we use the same strategy to extend the data into the past until $1996$ using the distribution for $2002,2003,2004$ as reference for disaggregation. 
    \item Having the data with the given regional resolution (municipalities\_districts) is not only unnecessarily fine for the simulation model, but also much too noisy to compute stable probabilities. As a result, we decided to aggregate the data to the districts\_districts level and use this as the main output of the algorithm. Thus
\[\hat{E}(y,r,s,a):y\in [1996,2100],r\in \{\text{districts\_districts}\},s\in \{m,f\},a\in [0,100^+].\]
\item Finally, to be more flexible, we decided to keep the high-resolution data and aggregate it to various coarser regional-levels as well.
\end{enumerate}

\begin{parameterfile}{Emigrants $\hat{E}$}{districts\_districs}
\hline
\multicolumn{2}{c}{\textbf{Contents}\label{param:emigrants}}\\
\hline
\multicolumn{2}{p{15cm}}{Total number of emigrated persons living in region $r$ with sex $s$ and $a$-th birthday in year $y$.}\\
\hline
\multicolumn{2}{c}{\textbf{Resolution}}\\
\hline
\textbf{Time-Frame} & $[1996,2100]$\\
\textbf{Regional-Level} & districts\_districts \\
\textbf{Age-Resolution} & $[0,100^+]$\\
\hline
\textbf{Other Regional-Levels} & country / federalstates / municipalities\_districts\\
\hline
\end{parameterfile}
In comparison with the ground-truth provided by Statistics-Austria, we would score the validity of the parameter values (0 invalid, 10 fully valid) as follows:
\begin{center}
    \begin{tabular}{c|c|c|c|c|p{5cm}}
    years & regions & sex & age & score & description\\
    \hline
    $[2002,y_0]$ & all & all & all & 8.5 & Statistical linkage between single-age data and age-class data.\\
    $[y_0+1,2100]$ & all & all & all & 4 & Used district, age and sex distribution of the last three available years to extrapolate until 2100 given only the total expected number of emigrants per federalstate and sex\\
    $[1996,2001]$ & all & all & all & 4 & Used district, age and sex distribution of the first three available years to extrapolate before 2002 given only the total number of emigrants per federalstate and sex\\
    \end{tabular}
\end{center}

\subsection{Immigrants \texorpdfstring{$\hat{I}$}{I}}
\label{sec:calculation_immigration}
In this section we compute the parameter values for the immigration processes. That means, we need to identify the total number of immigrants per year, so that the interface-agent can sample them into the model. Since the model needs to sample new coordinates for the involved agents, we aim to get the same resolution as for $\hat{P}$.

Note that the model requires the numbers to be integers. Therefore, we will finally apply the Huntington-Hill Disaggregation method, Algorithm \ref{alg:hh_disaggregation}, to get a well structured parameter file.

Essentially we will use the same sources for immigration as for emigration, however we will make use of the information we already have from the other harmonised census data $\hat{B},\hat{E},\hat{D}$:
{\setlength{\extrarowheight}{1mm}
\begin{center}
    \begin{tabular}{cccc}
    \hline
    \makecell{\textbf{name}\\\textbf{of source}}&
    \makecell{Wanderungen mit \\dem Ausland von \\2002 bis 2014\\ nach Altersgruppen,\\ Gemeinde und\\Staatsangehörigkeit} &
    \makecell{Wanderungen mit \\dem Ausland ab 2015 \\nach Altersgruppen, \\Gemeinde und \\Staatsangehörigkeit} &
    \makecell{Wanderungen mit \\dem Ausland ab 2002 \\nach Alter, \\Geschlecht und\\ Staatsangehörigkeit} \\
    \textbf{source} & S. \ref{src:migrationMuni_1} (immigration) & S. \ref{src:migrationMuni_2} (immigration) & S. \ref{src:migrationCountry} (immigration) \\
    \hline
    \textbf{variable} & $S_{m1}^i(y,r,s,a)$ & $S_{m2}^i(y,r,s,a)$ & $S_{mc}^i(y,s,a)$  \\
    $y\in$ & [2002,2014] & [2015,$y_0-1$] & [2002,$y_0-1$] \\
    $r\in$ & municipalities\_districts & municipalities\_districts & country \\
    $s\in $ & none & none & $\{m,f\}$ \\
    $a\in$ & $\{0,5,\dots,100^+\}$ & $\{0,5,\dots,100^+\}$ & $[0,100^+]$\\
    \hline
    \hline
    \makecell{\textbf{name}\\\textbf{of source}}&
    \makecell{Bevölkerungs\\-bewegung \\1961 bis 2100 \\nach Bundesland,\\ Bewegungsarten\\ und Szenarien} &
    \makecell{Tabellensammlung \\Außenwanderung} &
    \makecell{Population $\hat{P}$} \\
    \textbf{source} &  S. \ref{src:migrationForeacst} (immigrants) & S. \ref{src:migrationBackcast} (immigrants) & Parameter Value \ref{param:population} \\
    \hline
    \textbf{variable} & $S_{mf}^i(y,r)$ & $S_{mb}^i(y,r)$ & $\hat{P}(y,r,s,a)$ \\
    $y\in$ & [2002,2100] & [1996,$y_0$]& [1962,2101]\\
    $r\in$  & federalstates & federalstates & municipalities\_registrationdistricts \\
    $s\in $ & $\{m,f\}$ & $\{m,f\}$& $\{m,f\}$\\
    $a\in$  & none & none & $[0,100^+]$\\
    \hline
    \hline
    \makecell{\textbf{name}\\\textbf{of source}}&
    \makecell{Births $\hat{B}$} &
    \makecell{Deaths $\hat{D}$} &
    \makecell{Emigrants $\hat{E}$} \\
    \textbf{source} &  Parameter Value \ref{param:births} & Parameter Value 
    \ref{param:deaths} & Parameter Value \ref{param:emigrants}\\
    \hline
    \textbf{variable} & $\hat{B}(y,s,a)$ & $\hat{D}(y,s,a)$ & $\hat{E}(y,s,a)$\\
    $y\in$ & [1962,2100] & [1962,2100]&  [1996,2100]\\
    $r\in$  & country & country & country\\
    $s\in $ & $\{m,f\}$ & $\{m,f\}$& $\{m,f\}$\\
    $a\in$  & $\{0\}$ & $[0,100^+]$ & $[0,100^+]$\\
    \hline
\end{tabular}
\end{center}
}
\begin{enumerate}
    \item The first steps of immigration processing will be analogous to the one from calculating the emigration parameters: We merge $S_{m1}^i$ and $S_{m2}^i$ into one dataset, and use $S_{mc}^i$ to get a sex resolution and to refine the age-classes into single-age groups. Let
    \[X_1(y,r,s,a):y\in [2002,y_0-1],r\in \{\text{municipalities\_districts}\},s\in \{m,f\},a\in [0,100^+],\]
    stand for the resulting file containing absolute number of immigrants per region $r$, sex $s$, age $a$ and year $y$.
    \item Also for the forecast we use a similar strategy as for the emigration. However, since $S_{mf}^i$ only contains total number of immigrants without sex-resolution, and immigrants (in particular refugees) are heavily biased here, we use a different reference though: We use balance equation
    from Corollary \ref{cor:balance_overall} for $y\in [y_0,2100]$ and get
    \[X_2(y,s)=\left[\sum_{a=0}^{101}\hat{P}(y+1,s,a)-\hat{P}(y,s,a)-\hat{B}(y,s,a)+\hat{E}(y,s,a)+\hat{D}(y,s,a)\right].\]
    Since the immigration data must contain integers only, $[\cdot]$ refers to rounding to the nearest integer. With $X_2$ we find the total number of immigrants per sex and year
    \[X_2(y,s):y\in [y_0,2100],s\in \{m,f\},\]
    so that they perfectly (apart from rounding) align with the other parameter-files in the model processing. It is clear that any of the four (immigrants, emigrants, births, deaths) can be computed knowing the other three. We chose immigrants in our processing steps in favour of the others since it is the most fluctuating quantity over the years.
    \item For the first disaggregation step we use the federal-state distribution from $S_{mf}^i$ using the Huntington-Hill Disaggregation strategy:
    \begin{multline*}
        X_2: ([y_0,2100],none,\{m,f\},none)\rightarrow ([y_0,2100],\{\text{federalstates}\},\{m,f\},none)\\
        \text{ via }S_{mf}^i\text{ key }([y_0,2100]).
    \end{multline*}
    \item From this point we continue analogous to the emigration parameter calculation again. We use  
    \[X_3(r,s,a)=\sum_{y=y_0-3}^{y_0-1}X_2(y,r,s,a)\]
    as distribution and disaggregate $X_2$:
    \begin{multline*}
        X_2: ([y_0,2100],\{\text{federalstates}\},\{m,f\},none)\\
        \rightarrow ([y_0,2100],\{\text{municipalities\_districts}\},\{m,f\},[0,100^+])\\
        \text{ via }X_3\text{ key }(\{\text{federalstates}\},\{m,f\}).
    \end{multline*}
    We summarise all data into
    \[X_4(y,r,s,a):y\in [2002,2100],r\in \{\text{municipalities\_districts}\},s\in\{m,f\},a\in [0,100^+]. \]
    \item We proceed analogously with the historic data from 1996 and get
    \[X_5(y,r,s,a):y\in [1996,2100],r\in \{\text{municipalities\_districts}\},s\in\{m,f\},a\in [0,100^+]. \]
    \item To match the spatial resolution of the population data $\hat{P}$, we disaggregate $X_5$ further into Viennese registration-districts. With the assumption that immigrants distribute analogously to the population, we disaggregate with the Huntington-Hill Disaggregation method:
    \begin{multline*}
        X_5: ([1996,2100],\{\text{municipalities\_districts}\},\{m,f\},[0,100^+])\\
        \rightarrow ([1996,2100],\{\text{municipalities\_registrationdistricts}\},\{m,f\},[0,100^+])\\
        \text{ via }\hat{P}\text{ key }([1996,2100],\{\text{municipalities\_districts}\},\{m,f\},[0,100^+]).
    \end{multline*}
    The resulting data-set $\hat{I}$ can be regarded as the result of this parameter calculation process.
    \item To be more flexible, we also aggregated the data to coarser regional-levels.
\end{enumerate}

\begin{parameterfile}{Immigrants $\hat{I}$}{regional-level municipalities\_registrationdistricts}
\hline
\multicolumn{2}{c}{\textbf{Contents}\label{param:immigrants}}\\
\hline
\multicolumn{2}{p{15cm}}{Total number of immigrated persons into region $r$ with sex $s$ and $a$-th birthday in year $y$.}\\
\hline
\multicolumn{2}{c}{\textbf{Resolution}}\\
\hline
\textbf{Time-Frame} & $[1996,2100]$\\
\textbf{Regional-Level} & municipalities\_registrationdistricts\\
\textbf{Age-Resolution} & $[0,100^+]$\\
\hline
\textbf{Other Regional-Levels} & country / federalstates / districts\_districts / municipalities\_districts \\
\hline
\end{parameterfile}
In comparison with the ground-truth provided by Statistics-Austria, we would score the validity of the parameter values (0 invalid, 10 fully valid) as follows:
\begin{center}
    \begin{tabular}{c|c|c|c|c|p{5cm}}
    years & regions & sex & age & score & description\\
    \hline
    $[2002,y_0-1]$ & real municipalities & all & all & 9.5 & Statistical linkage between single-age data and age-class data.\\
    $[2002,y_0-1]$ & Viennese registration districts & all & all & 8.5 & Statistical linkage with population data from Vienna.\\
    $[y_0,2100]$ & all & all & all & 4 & Total numbers computed by balance equation, age/sex/municipality distribution from the last three available years\\
    $[1996,2001]$ & all & all & all & 4 & Total numbers computed by balance equation, age/sex/municipality distribution from the first three available years\\
    \end{tabular}
\end{center}

\subsection{Birth-, Death-, and Emigration probabilities \texorpdfstring{$\hat{B}^p,\hat{D}^p,\hat{E}^p$}{Bp,Dp,Ep}}
\label{sec:calculation_probabilities}
In the final step, we compute probabilities from the computed harmonised / augmented census information developed in the previous sections:
{\setlength{\extrarowheight}{1mm}
\begin{center}
    \begin{tabular}{ccccc}
    \hline
    \makecell{\textbf{name}\\\textbf{of source}}&
    \makecell{Population $\hat{P}$} &
    \makecell{Deaths $\hat{D}$} &
    \makecell{Emigrants $\hat{E}$} &
    \makecell{Births per Mother $\hat{B}_m$} \\
    \textbf{source} & Parameter Value \ref{param:population} & Parameter Value \ref{param:deaths}  & Parameter Value \ref{param:emigrants}  & Parameter Value \ref{param:births_mothers}\\
    \hline
    \textbf{variable} & $\hat{P}(y,r,s,a)$ & $\hat{D}(y,r,s,a)$ & $\hat{E}(y,r,s,a)$ & $\hat{B}_m(y,r,s,a)$ \\
    $y\in$ & [1962,2101] & [1996,2100] & [1996,2100] & [1962,2100]\\
    $r\in$ & \multicolumn{4}{c}{country / federalstates / districts\_districts}\\
    $s\in $ & $\{m,f\}$ & $\{m,f\}$ & $\{m,f\}$ & $\{f\}$\\
    $a\in$ & $[0,100]$ & $[0,100]$ & $[0,100]$ & $[0,100]$\\
    \hline
\end{tabular}
\end{center}
}

With
\begin{equation*}
    Q(y,r,s,a) = \hat{D}(y,r,s,a)+\hat{E}(y,r,s,a)
\end{equation*}
as the total number of individuals leaving the cohort, we apply the enhanced Farr Formula, Corollary \ref{cor:farr_param}, for computation of probabilities which are compatible with definition \ref{def:probability}. Note that we can use only the overlapping time period $[1996,2100]$ for this computation. This results in the following data sets:

\begin{parameterfile}{Birth-Probabilities $\hat{B}^p$}{federalstates}
\hline
\multicolumn{2}{c}{\textbf{Contents}\label{param:birth_prob}}\\
\hline
\multicolumn{2}{p{15cm}}{Probability that a female agent living in region $r$ who has its $a$-th birthday in year $y$ gives birth to a child until its next birthday.}\\
\hline
\multicolumn{2}{c}{\textbf{Resolution}}\\
\hline
\textbf{Time-Frame} & $[1996,2100]$\\
\textbf{Regional-Level} & federalstates\\
\textbf{Age-Resolution} & $[0,100^+]$\\
\hline
\textbf{Other Regional-Levels} & country / districts\_districts\\
\hline
\end{parameterfile}

\begin{parameterfile}{Death-Probabilities $\hat{D}^p$}{federalstates}
\hline
\multicolumn{2}{c}{\textbf{Contents}\label{param:death_prob}}\\
\hline
\multicolumn{2}{p{15cm}}{Probability that an agent with sex $s$ living in region $r$ who has its $a$-th birthday in year $y$ dies until its next birthday.}\\
\hline
\multicolumn{2}{c}{\textbf{Resolution}}\\
\hline
\textbf{Time-Frame} & $[1996,2100]$\\
\textbf{Regional-Level} & federalstates\\
\textbf{Age-Resolution} & $[0,100^+]$\\
\hline
\textbf{Other Regional-Levels} & country / districts\_districts\\
\hline
\end{parameterfile}

\begin{parameterfile}{Emigration-Probabilities $\hat{E}^p$}{districts\_districts}
\hline
\multicolumn{2}{c}{\textbf{Contents}\label{param:emigration_prob}}\\
\hline
\multicolumn{2}{p{15cm}}{Probability that an agent with sex $s$ living in region $r$ who has its $a$-th birthday in year $y$ emigrates until its next birthday.}\\
\hline
\multicolumn{2}{c}{\textbf{Resolution}}\\
\hline
\textbf{Time-Frame} & $[1996,2100]$\\
\textbf{Regional-Level} & districts\_districts\\
\textbf{Age-Resolution} & $[0,100^+]$\\
\hline
\textbf{Other Regional-Levels} & country / federalstates\\
\hline
\end{parameterfile}

Note that we chose federalstates as the default regional-level for death and birth probabilities and districts\_districts as the one for emigration. This was chosen as a compromise between having regional features well represented, which motivates a fine regional-level, and outlier-robust probability computation, which motivates larger population counts and therefore a coarse regional-level.

\subsection{Region-Families \texorpdfstring{$(A^{r_x}_j)$}{Arj}}
\label{sec:calculation_regional}
For all regional-levels $r$ specified in Section \ref{sec:regional_levels} we specify the regions for the region families $A_j^{r_x}$ in form of so called multi-polygons with internal rings (MPIR). That means, every region $A_j^{r_x}$ is defined as a tuple
\begin{multline*}
    A_j^{r_x}=\left((P_1,\dots,P_{J_j}),(PI_1,\dots,PI_{K_j})\right):\\
    P_{i}=((lat,long)_{1},(lat,long)_2,\dots,(lat,long)_{L_{ji}}),\\
    PI_{i}=((lat,long)_{1},(lat,long)_2,\dots,(lat,long)_{M_{ji}})
\end{multline*}
We can make use of this structure in various ways. First of all, we can easily find in which region a given point lies in using an appropriate point-in-polygon algorithm (see \cite{huang1997complexity} for examples). A point lies in a region $A_j^{r_x}$ if it lies in one of the polygons $P_i$ but not in any of the interior rings $PI$. Moreover, we may triangulate the regions using a Constrained Delaunay triangulation \cite{chew1987constrained}. This allows for efficient sampling of uniformly distributed random coordinates inside each MPIR (see \cite{bicher_gepoc_2018}).

For $r_{min}$, which is needed to fine-tune the coordinate sampling process (we refer to the documentation of GEPOC ABM Geography in \cite{bicher2025gepoc} for details), we use a raster-image format. I.e. the regions are defined as rectangles:
\[A_{i,j}^{r_{min}}=\{(x,y): i\Delta lat \leq x-lat_0 \leq (i+1)\Delta lat, j\Delta long \leq y-long_0 \leq (j+1)\Delta long \}.\]
This makes a lot of computations easier and faster, e.g. finding if a given coordinate lies inside a certain region or not.

\subsubsection{Multi-Polygon Processing}
To get compatible model input files, we require high quality MPIR files for whole Austria which are consistent across different regional-levels. The only way to guarantee consistency is by aggregating data from finer to coarser regional-levels. Suppose we have regions $A_j$ for the municipalities\_registrationdistricts level, we can aggregate:
\begin{enumerate}
    \item municipalities\_registrationdistricts $\rightarrow$ municipalities\_districts, by merging all those regions in Vienna which share the same first three digits in their id (leading to 23 districts in Vienna),
    \item municipalities\_districts $\rightarrow$ municipalities, by merging all regions in Vienna into one region,
    \item municipalities\_districts $\rightarrow$ districts\_districts, by merging all regions which share the same first three digits in their id,
    \item districts\_districts $\rightarrow$ districts, by merging all regions in Vienna,
    \item districts $\rightarrow$ federalstates, by merging all regions which share the same first number in their id,
    \item federalstates $\rightarrow$ country, by merging all regions.
\end{enumerate}
It remains to get proper data for the municipalities\_registrationdistricts level. For this purpose, we use Source \ref{src:geojson2017} and \ref{src:geojsonVienna}. The former GEOJSON file contains the MPIR borders of all municipalities in Austria, the latter for all registrationdistricts in Vienna for area-status 2017 and 2019 respectively. While the data seems to be well suited for GEPOC parametrisation, there were a couple of preprocessing steps involved before we could use it. Since many steps were done manually, it is not possible to describe them in a reproducible fashion:
\begin{enumerate}
    \item First of all, both data files were internally inconsistent. That means that the included geometries had (a) (self-)intersections and (b) gaps in between bordering regions. Problem (a) was solved by cleaning the geometries automatically, intersecting all geometries with each other, and cutting out duplicate areas, problem (b) was solved manually afterwards. Using \href{https://www.google.com/url?sa=t&rct=j&q=&esrc=s&source=web&cd=&cad=rja&uact=8&ved=2ahUKEwjSi92d0reEAxXEQ_EDHUWoC10QFnoECAYQAQ&url=https%3A%2F%2Fwww.qgis.org%2Fde%2Fsite%2F&usg=AOvVaw2xSxrH0LmwlqL0siokWUq-&opi=89978449}{QGIS} and \href{https://mapshaper.org/}{MapShaper}, left-out areas within Austria were identified and manually assigned to suitable regions.
    \item In a second step, both datasets have been joined to get a map for the municipalities in Austria including registration districts in Vienna. For the Viennese borders, the same problems (a) and (b) could be identified which were solved analogously.
    \item Since the registrationdistricts have not changed for a while, the resulting data has area-status 2017. It remains to update the area-status to a given year $y_0$ which involves various merges and splits of municipalities. Until 2025, for example, these are:
    \paragraph{2017$\rightarrow$2018}
    \begin{itemize}
        \item 40803$+$40819 $\rightarrow$ 40835 (Peuerbach)
        \item 41625$+$41340 $\rightarrow$ 41628 (Vorderweißenbach)
    \end{itemize}
    \paragraph{2018$\rightarrow$2019}
    \begin{itemize}
        \item 41310$+$41302$\rightarrow$41345 (Helfenberg)
        \item 41335$+$41301$\rightarrow$41346 (St. Stefan-Afiesl)
    \end{itemize}
    \paragraph{2019$\rightarrow$2020}
    \begin{itemize}
        \item 61056$+$61058$+$62347 $\rightarrow$61060$+$61061 (Sankt Veit in der Südsteiermark, Straß in Steiermark)
    \end{itemize}
    \paragraph{2021$\rightarrow$2022}
    \begin{itemize}
        \item 70327$+$70330$+$70341$\rightarrow$70370 (Matrei am Brenner)
    \end{itemize}
    \paragraph{2024$\rightarrow$2025}
    \begin{itemize}
        \item 62252$+$62267$\rightarrow$62280 (Fürstenfeld)
    \end{itemize}
    With exception of the 3 to 2 merge in 2020, which was manually executed using the geo-software QGIS, all merges could be done automatically using standard shape tools.
\end{enumerate}
\begin{parameterfile}{Region-Families $A^{r_x}$}{area-status $y_0$}
\hline
\multicolumn{2}{c}{\textbf{Contents}\label{param:regions}}\\
\hline
\multicolumn{2}{p{15cm}}{Borders of regions with different regional-levels within Austria valid for $y_0$-01-01.}\\
\hline
\multicolumn{2}{c}{\textbf{Resolution}}\\
\hline
\textbf{Regional-Level} & all levels specified in Section \ref{sec:regional_levels}\\
\hline
\end{parameterfile}

\subsubsection{Raster Image for \texorpdfstring{$A^{r_{min}}$}{Ar}}
For $A^{r_{min}}$ we use the Global Human Settlement layer of Europe, Source \ref{src:ghsmap}. To make usage more efficient we use a tight cutout around the borders of Austria.
\begin{parameterfile}{$A^{r_{min}}$}{-}
\hline
\multicolumn{2}{c}{\textbf{Contents}\label{param:regions_min}}\\
\hline
\multicolumn{2}{p{15cm}}{Estimated number of inhabitants per grid-cell.}\\
\hline
\multicolumn{2}{c}{\textbf{Resolution}}\\
\hline
\textbf{Regional-Level} & 100$\times$100m raster\\
\textbf{Age Resolution} & no age resolution\\
\hline
\end{parameterfile}

\subsection{Internal Migrants \texorpdfstring{$\hat{OD}$,$\hat{IE}$, $\hat{II}$,$\hat{M}$}{OD,IE,II,M}}
In the following we compute parameters for the internal migration models in GEPOC, in specific the interregional, biregional and fully regional model of GEPOC ABM IM.

\subsubsection{Interregional Flows \texorpdfstring{$\hat{OD}$}{OD}}
\label{sec:calculation_interregional}
The first parameter values which will be computed for parametrisation of internal migrations will be the ones for the interregional model (see GEPOC IM, \cite{bicher2025gepoc}), i.e. the model which computes the destination region (solely) based on the origin region and sex. 

For one-sided disaggregation we will apply the Proportional Disaggregation method, Algorithm \ref{alg:prop_disagg}, since this data will only be used as probabilities in the model. For two-sided disaggregation we will use the 2D Iterative Proportional Fitting (IPF 2D) method, Algorithm \ref{alg:ipf_2d}.

We will use the following sources:
{\setlength{\extrarowheight}{1mm}
\begin{center}
    \begin{tabular}{cccc}
    \hline
    \makecell{\textbf{name}\\\textbf{of source}}& 
    \makecell{Wanderungen\\ innerhalb \\Österreichs} &
    \makecell{Bevölkerungsbewegung \\1961 bis 2100 \\nach Bundesland,\\ Bewegungsarten\\und Szenarien} &
    \makecell{Tabellensammlung \\Binnenwanderungen 2024} \\
    \textbf{source} & S. \ref{src:internalMigrants} & 
    \makecell{S. \ref{src:migrationForeacst} \\(i. emigration and immigration)}
    & \makecell{S. \ref{src:internalMigrantsBC} \\(i. emigration and immigration)}\\
    \hline
    \textbf{variable} &  $S_{od}^e(y,r,s,r_2)$ & $S_{mf}^{ie}(y,r),S_{mf}^{ii}(y,r)$ & $S_{mb}^{ie}(y,r),S_{mb}^{ii}(y,r)$ \\
    $y\in$ & [2002,$y_0-1$] & [2002,2100] & [1996,2024]  \\
    $r,r_2\in$ & municipalities\_dist. & federalstates & federalstates  \\
    $s\in $ & $\{m,f\}$ & none & none  \\
    $a\in$ & none & none & none\\
    \hline
\end{tabular}
\end{center}
}

The main problem in processing of origin destination flows/probabilities from the available information is the availability of only two one-dimensional data-sets, internal emigrants $S_{mf}^{ie}$ (origin) and internal immigrants $S_{mf}^{ii}$ (destination), for the forecast. Not only do the data-sets lack the 2-nd dimension (origin+destination), but they are only on federalstates level and do not distinguish between male and female. To make things even worse, they do not incorporate emigrants which remained in the same federalstate.
\begin{enumerate}
\item In order to extend the data beyond $y_0-1$, we utilise the two forecasts for $[y_0,2100]$. We attempt to establish an origin destination mapping $X_1(y,r_1,r_2)$ between the different federalstates, so that the sum over all destinations (not equal to the origin) matches $S_{mf}^{ie}(y,r)$ and the sum over all origins (not equal to the destination) matches $S_{mf}^{ii}(y,r)$, that means, we aim to find $X_1$, so that
\begin{equation}\label{eq:odconstraints}\forall r_1: \sum_{r_2,r_2\neq r}X_1(y,r_1,r_2) = S_{mf}^{ie}(y,r_1),\quad \forall r_2: \sum_{r_1,r\neq r_2}X_1(y,r_1,r_2) = S_{mf}^{ii}(y,r_2).\end{equation}
Thus, we utilise IPF 2D using $\vec{a}_i:=S_{mf}^{ie}(y,r_i)$ and $\vec{b}_j:=S_{mf}^{ii}(y,r_j)$ as marginals.
\item  The algorithm requires a feasible initial estimate, since the problem with $2\cdot n$ equations ($n=9$ is equal to the number of federalstates) for finding  $n(n-1)$ variables is heavily underdetermined. Therefore, we use the most recent original OD-data from $S_{od}$. We aggregate sex, regions from municipalities\_districts to federalstates, and average the flows over the last three available years of the original data
\[X_2(r_1,r_2)=\frac{1}{3}\sum_{y=y_0-3}^{y_0-1}\sum_{d_1\in r_1}\sum_{s\in\{m,f\}}\sum_{d_2\in r_2}S_{od}(y,d_1,s,d_2).\]
Since the forecast data does not consider migrations within the same region, we define a new dataset $X_3$ where we set the corresponding diagonal entries to zero:
\[\forall r: X_3(r_1,r_2)=\left\lbrace\begin{array}{ll}X_2(r_1,r_2),&r_1\neq r_2,\\ 0,\text{else.}\end{array}\right\rbrace.\]
Note how the IDF 2D method only up and downscales values from the initial condition. Therefore any zeros in the initial matrix will remain as they are. Thus, $X_3$, is a feasible starting value for the IPF 2D, since the row- and column-sums do not include the diagonal elements.
\item We regard the result of the IPF 2D method (applied with convergence threshold of $10^{-10}$) as $X_4$:
\[X_4(y,r_1,r_2):y\in [y_0-1,2100],r_1\neq r_2\in \{\text{federalstates}\}.\]
\item The diagonal entries (currently $0$, by design) are still to be defined. We decided to assume that the migration within the federalstate grows/declines directly proportional to the migration into other federalstates. Therefore we simply upscale the values from the original distribution $X_2$: For every year $y\in [y_0,2100]$ and federalstate $r$ we define
\[X_4(y,r,r):=X_2(r,r)\frac{\sum_{r_2\neq r}X_4(y,r,r_2)}{\sum_{r_2\neq r}X_2(r,r_2)}.\]
\item In the next step, we disaggregate $X_4$ to match the resolution of $S_{od}$ using the Proportional Disaggregation method, Algorithm \ref{alg:prop_disagg}. As reference distribution we aggregate the last three available years:
\[X_5(r_1,s,r_2)=\sum_{y=y_0-3}^{y_0-1}S_{od}(y,d_1,s,d_2).\]
and disaggregate
\begin{multline*}
    X_4:([y_0,2100],\{\text{federalstates}\},\{\text{federalstates}\})\\
    \rightarrow ([y_0,2100],\{\text{municipalities\_districts}\},\{m,f\},\{\text{municipalities\_districts}\}) \\
    \text{ via }X_5\text{ key }(\{\text{federalstates}\},\{\text{federalstates}\}).
\end{multline*}
We merge the result with $S_{od}$ and get
\[X_6(y,r_1,s,r_2):y\in [2002,2100],r_1,r_2\in \{\text{municipalities\_districts}\},s\in \{m,f\}.\]
\item In a final step, we proceed analogously for the historic internal migration data $S_{mb}^{ii}$ and $S_{mf}^{ie}$ since it has the same structure as the forecast data. However, we use years $2002-2004$ to compute the distribution / initial condition for the IPF 2D. The results of all procedures are joined along the time-axis to get one final dataset
\[\hat{OD}(y,r_1,s,r_2):y\in [1996,2100],r_1,r_2\in \{\text{municipalities\_registrationdistricts}\},s\in \{m,f\}.\]
\item While this file will be regarded as the primary output of the processing, aggregation to coarser regional-levels will be done to become more flexible.
\end{enumerate}

\begin{parameterfile}{Interregional Flows $\hat{OD}$}{regional-level municipalities\_registrationdistricts}
\hline
\multicolumn{2}{c}{\textbf{Contents}\label{param:od}}\\
\hline
\multicolumn{2}{p{15cm}}{Total number of persons migrating from region $r_1$ into region $r_2$ with sex $s$ in year $y$.}\\
\hline
\multicolumn{2}{c}{\textbf{Resolution}}\\
\hline
\textbf{Time-Frame} & $[1996,2100]$\\
\textbf{Regional-Level } & municipalities\_districts\\
\textbf{Age-Resolution} & none\\
\hline
\textbf{Other Regional-Levels} & federalstates / districts\_districts\\
\hline
\end{parameterfile}
In comparison with the ground-truth provided by Statistics-Austria, we would score the validity of the parameter values (0 invalid, 10 fully valid) as follows:
\begin{center}
    \begin{tabular}{c|c|c|c|c|p{5cm}}
    years & regions & sex & age & score & description\\
    \hline
    $[2002,y_0-1]$ & all & all & all & 10 & raw data.\\
    $[y_0,2100]$ & all & all & all & 3 & Forecast based on immigration and emigration numbers on federalstate level and the IPF 2D method.\\
    $[1996,2001]$ & all & all & all & 3 & Historic flows are based on immigration and emigration numbers on federalstate level and the IPF 2D method.\\
    \end{tabular}
\end{center}

\subsubsection{Internal Emigrants \texorpdfstring{$\hat{IE}$}{IE}}
\label{sec:calculation_internal_emigration}
In the next step we compute the harmonised census for the internal emigrants $\hat{IE}$ which will be required to compute the internal emigration probability for all 3 IM models. That means, we focus on internal emigrants and their age and sex distribution without analysing their destination region. The Proportional Disaggregation method, Algorithm \ref{alg:prop_disagg}, is used for disaggregation of age-classes.

We will use the following sources:
{\setlength{\extrarowheight}{1mm}
\begin{center}
    \begin{tabular}{cccc}
    \hline
    \makecell{\textbf{name}\\\textbf{of source}}& 
    \makecell{Binnenwanderungen \\innerhalb Österreichs\\ ab 2002} & \makecell{Interregional Flows $\hat{OD}$}\\
    \textbf{source} & S. \ref{src:internalMigrantsStatCube} (i. emigrants) & Parameter Value \ref{param:od} \\
    \hline
    \textbf{variable} &  $S_{m}^{ie}(y,r,s,a)$ & $\hat{OD}(y,r,s,r_2)$ \\
    $y\in$ & [2002,$y_0-1$] & [1996,2100] \\
    $r,r_2\in$ & districts\_districts & districts\_districts\\
    $s\in $ & $\{m,f\}$ & $\{m,f\}$  \\
    $a\in$ & $[0,95^+]$ & none\\
    \hline
\end{tabular}
\end{center}
}
The core of the age-dependent internal-emigrant is source $S_m^{ie}$ which was collected from Statistics Austria's STATCube. To establish the forecast, we utilise the $\hat{OD}$ parameter values which were derived in the previous section. This way, we remain consistent w.r. to total numbers.
\begin{enumerate}
\item To compute the forecast, we use the age distribution of $S_{m}^{ie}$ from $y_0-3$ to $y_0-1$:
\[X_1(r,s,a)=\sum_{y=y_0-3}^{y_0-1} S_{m}^{ii}(y,r,s,a)\]
and the aggregated emigrants from $\hat{OD}$ (restricted to $[y_0,2100]$)
\[X_2(y,r,s)=\sum_{r_1}\hat{OD}(y,r,s,r_1),\]
and de-aggregate with the Proportional Method:
\begin{multline*}
    X_2:([y_0,2100],\{\text{districts\_districts}\},\{m,f\})\\
    \rightarrow ([y_0,2100],\{\text{districts\_districts}\},\{m,f\},[0,95^+])\\
    \text{ via }X_1\text{ key } (\{\text{districts\_districts}\},\{m,f\}).
\end{multline*}
\item We proceed identically with the historic $OD$ data between 1996 and 2001 using the distribution of 2002 to 2004. The results are merged with $S_{m}^{ie}$ into
\[\hat{IE}(y,r,s,a):y\in [1996,2100],r\in \{\text{districts\_districts}\},a\in [0,95^+].\]
\item Aggregation to federalstate level is considered as secondary output of this procedure.
\end{enumerate}
\newpage
\begin{parameterfile}{Internal Emigrants $\hat{IE}$}{districts\_districts}
\hline
\multicolumn{2}{c}{\textbf{Contents}\label{param:ie}}\\
\hline
\multicolumn{2}{p{15cm}}{Total number of persons to internally migrate from region $r$ with sex $s$ and age $a$ (at the date of migration) in year $y$.}\\
\hline
\multicolumn{2}{c}{\textbf{Resolution}}\\
\hline
\textbf{Time-Frame} & $[1996,2100]$\\
\textbf{Regional-Level} &  districts\_districts\\
\textbf{Age-Resolution} & $[0,95^+]$\\
\hline
\textbf{Other Regional-Levels} & federalstates\\
\hline
\end{parameterfile}
In comparison with the ground-truth provided by Statistics-Austria, we would score the validity of the parameter values (0 invalid, 10 fully valid) as follows:
\begin{center}
    \begin{tabular}{c|c|c|c|c|p{5cm}}
    years & regions & sex & age & score & description\\
    \hline
    $[2002,y_0]$ & all & all & all & 10 & raw data.\\
    $[y_0,2100]$ & all & all & all & 4 & Age distribution from last three available years, districts\_districts distribution from federalstates.\\
    $[1996,2001]$ & all & all & all & 4 & Age distribution from first three available years, districts\_districts distribution from federalstates.\\
    \end{tabular}
\end{center}

\subsubsection{Internal Immigrants \texorpdfstring{$\hat{II}$}{II}}
\label{sec:calculation_biregional}
To compute parameter $\hat{II}$ for the biregional model, we focus on internal immigrants and their age and sex distribution without analysing their origin region. The Proportional Disaggregation method, Algorithm \ref{alg:prop_disagg}, is used for disaggregation of age-classes.

We will use the following sources:
{\setlength{\extrarowheight}{1mm}
\begin{center}
    \begin{tabular}{cccc}
    \hline
    \makecell{\textbf{name}\\\textbf{of source}}& 
    \makecell{Binnenwanderungen \\innerhalb Österreichs\\ ab 2002} & \makecell{Interregional Flows $\hat{OD}$}& \makecell{Internal Emigrants $\hat{IE}$}\\
    \textbf{source} & S. \ref{src:internalMigrantsStatCube} (i. immigrants) & Parameter Value \ref{param:od} & Parameter Value \ref{param:ie} \\
    \hline
    \textbf{variable} &  $S_{m}^{ii}(y,r,s,a)$ & $\hat{OD}(y,r,s,r_2)$ & $\hat{IE}(y,r,s,a)$ \\
    $y\in$ & [2002,$y_0-1$] & [1996,2100] & [1996,2100] \\
    $r,r_2\in$ & districts\_districts & districts\_districts & districts\_districts\\
    $s\in $ & $\{m,f\}$ & $\{m,f\}$ & $\{m,f\}$  \\
    $a\in$ & $[0,95^+]$ & none & $[0,95^+]$\\
    \hline
\end{tabular}
\end{center}
}
With one exception, the computation for internal immigrants is identical to that for internal emigrants. However, a final adjustment needs to be made to guarantee that the age distributions of the internal emigrants and immigrants align.
\begin{enumerate}
\item We proceed analogous to the internal emigrant harmonisation, while the $OD$ data ($\hat{OD}$ parameter file) is aggregated w.r. to the destination dimension. We denote the result of this procedure as 
\[X_1(y,r,s,a):y\in [1996,2100],r\in \{\text{districts\_districts}\},a\in [0,95^+].\]
\item Unfortunately, this way, we receive a mismatch with age distribution of the computed internal emigrants $IE(y,r,s,a)$ since
\[\forall y\in [1996,2001]\cup [y_0,2100],s\in \{m,f\}: \sum_{r}IE(y,r,s,a)=\sum_{r}II(y,r,s,a)\]
is not guaranteed. To solve this problem, we once again apply the IPF 2D method: for any year $y\in [1996,2001]\cup [y_0,2100]$ and sex $s$, we use
\[\sum_{r}IE(y,r,s,a),\text{ and }\sum_{a}II(y,r,s,a)\]
as marginals, and $II(y,r,s,a)$ as initial matrix of the algorithm. Note that the algorithm is now applied for the region and age dimension, instead of the two region dimensions, as for the $OD$ computation. Therefore, the result is guaranteed to match the age distribution of the aggregated internal emigration data and to maintain the overall distribution of destination regions.
\item The results of the IPF 2D (tolerance $10^{-10}$) are merged with $S_{m}^{ii}$ into
\[\hat{II}(y,r,s,a):y\in [1996,2100],r\in \{\text{districts\_districts}\},a\in [0,95^+].\]
\item Aggregation to federalstates level is considered as secondary output of this procedure.
\end{enumerate}
\begin{parameterfile}{Internal Immigrants $\hat{II}$}{districts\_districts}
\hline
\multicolumn{2}{c}{\textbf{Contents}\label{param:ii}}\\
\hline
\multicolumn{2}{p{15cm}}{Total number of persons to migrate into region $r_2$ with sex $s$ and age $a$ (at the date of migration) in year $y$.}\\
\hline
\multicolumn{2}{c}{\textbf{Resolution}}\\
\hline
\textbf{Time-Frame} & $[1996,2100]$\\
\textbf{Regional-Level} &  districts\_districts\\
\textbf{Age-Resolution} & $[0,95^+]$\\
\hline
\textbf{Other Regional-Levels} & federalstates\\
\hline
\end{parameterfile}
In comparison with the ground-truth provided by Statistics-Austria, we would score the validity of the parameter values (0 invalid, 10 fully valid) as follows:
\begin{center}
    \begin{tabular}{c|c|c|c|c|p{5cm}}
    years & regions & sex & age & score & description\\
    \hline
    $[2002,y_0]$ & all & all & all & 10 & raw data.\\
    $[y_0,2100]$ & all & all & all & 4 & Age distribution from last three available years, districts\_districts distribution from federalstates.\\
    $[1996,2001]$ & all & all & all & 4 & Age distribution from first three available years, districts\_districts distribution from federalstates.\\
    \end{tabular}
\end{center}

\subsubsection{Internal Migrants by Age, Origin and Destination \texorpdfstring{$\hat{M}$}{M}}
\label{sec:calculation_fullregional}
In this section we will derive migration flows $\hat{M}$ which describe the number of persons with sex $s$ and age $a$ emigrating from a certain origin $r_1$ into a certain destination $r_2$. Note that all parameter values computed for internal migration so far have in common that they do not contain full information for parametrisation of the migration process. While the origin-destination parameters $\hat{OD}$ lack the age resolution, internal emi- and immigrants $\hat{IE}$ and $\hat{II}$ lack the corresponding destination/origin mapping. Core of this computation is the 3-dimensional Iterative Proportional Fitting, Algorithm \ref{alg:ipf_3d}.

We will use no additional sources but only derived parameter values:

{\setlength{\extrarowheight}{1mm}
\begin{center}
    \begin{tabular}{ccccc}
    \hline

    \makecell{\textbf{name}\\\textbf{of source}}&
    \makecell{Interregional Flows $\hat{OD}$} &
    \makecell{Internal Emigrants $\hat{IE}$} &
    \makecell{Internal Immigrants $\hat{II}$}

    \\

    \textbf{source} & Parameter Value \ref{param:od} & Parameter Value \ref{param:ie} & Parameter Value \ref{param:ii}\\
    \hline
    \textbf{variable} &  $\hat{OD}(y,r_1,s,r_2)$ & $\hat{IE}(y,r_1,s,a)$ & $\hat{II}(y,r_2,s,a)$\\
    $y\in$ & [1996,2100] & [1996,2100] & [1996,2100] \\
    \multirow{2}{*}{$r_1,r_2\in$}& districts\_districts/ & districts\_districts/ & districts\_districts/\\ & federalstates&
    federalstates&
    federalstates\\
    $s\in $ & $\{m,f\}$ & $\{m,f\}$ & $\{m,f\}$ \\
    $a\in$ & none & $[0,95^+]$ &  $[0,95^+]$\\
    \hline
\end{tabular}
\end{center}
}

Since no additional data from Statistics Austria is used, the only knowledge we have for computing $\hat{M}$ is the causal relation between the data: For all years $y$ and sex $s$, we know that
\begin{align}\forall r_1,a:& \sum_{r_2}\hat{M}(y,r_1,s,a,r_2)=\hat{IE}(y,r_1,s,a),\\
\forall r_2,a:&\sum_{r_1}\hat{M}(y,r_1,s,a,r_2)=\hat{II}(y,r_2,s,a),\\
\forall r_1,r_2:& \sum_{a}\hat{M}(y,r_1,s,a,r_2)=\hat{OD}(y,r_1,s,r_2).\end{align}
Let $n$ denote the total number of regions (in our case around 100 districts) and $m$ the total number of age-classes (in our case 95), then the problem has $2mn+n^2$ equations and $n^2m$ unknowns.

This provides the perfect setup for applying IPF 3D. We use the three datasets at districts\_districts level as marginals, the simple tensor $M_0(y,r_1,s,a,r_2)\equiv 1$ as initial condition, and tolerance $10^{-4}$ is used as stopping criterion. This choice for the initial condition turned out to be sufficiently good to result in convergence in less than 200 iterations.

The result 
\[\hat{M}(y,r_1,s,a,r_2):y\in [1996,2100],r_1,r_2\in \{\text{districts\_districts}\},s\in \{m,f\},a\in [0,95^+]\]
is regarded as the primary output of this procedure. Aggregation to the federalstates level is provided as a secondary outcome.

\begin{parameterfile}{Internal Migrants $\hat{M}$}{districts\_districts}
\hline
\multicolumn{2}{c}{\textbf{Contents}\label{param:im}}\\
\hline
\multicolumn{2}{p{15cm}}{Total number of persons migrating from region $r_1$ into region $r_2$ with sex $s$ and age $a$ (at the date of migration) within year $y$.}\\
\hline
\multicolumn{2}{c}{\textbf{Resolution}}\\
\hline
\textbf{Time-Frame} & $[1996,2100]$\\
\textbf{Regional-Level} & districts\_districts\\
\textbf{Age-Resolution} & $[0,95^+]$\\
\hline
\textbf{Other Regional-Levels} & federalstates
\end{parameterfile}
In comparison with the ground-truth provided by Statistics-Austria, we would score the validity of the parameter values (0 invalid, 10 fully valid) as follows:
\begin{center}
    \begin{tabular}{c|c|c|c|c|p{5cm}}
    years & regions & sex & age & score & description\\
    \hline
    $[2002,y_0-1]$ & all & all & all & 8 & Matched with IPF 3D.\\
    $[y_0,2100]$ & all & all & all & 2.5 & Forecast based on immigration and emigration numbers on federalstate level and IPF matching.\\
    $[1996,2001]$ & all & all & all & 2.5 & Historic information based on immigration and emigration numbers on federalstate level and IPF matching.\\
    \end{tabular}
\end{center}

\subsection{Internal Emigration Probability \texorpdfstring{$\hat{IE}^p$}{IEp}}
\label{sec:calculation_ie_probability}
Finally, after harmonising the internal migration census, the internal emigration probability which is required for all three internal migration models of GEPOC ABM IM is computed analogously to all other probabilities using the enhanced Farr formula, Corollary \ref{cor:farr_param}. 

{\setlength{\extrarowheight}{1mm}
\begin{center}
    \begin{tabular}{ccccc}
    \hline
    \makecell{\textbf{name}\\\textbf{of source}}&
    \makecell{Population $\hat{P}$} &
    \makecell{Deaths $\hat{D}$} &
    \makecell{Emigrants $\hat{E}$} &
    \makecell{Internal Emigrants $\hat{IE}$} \\
    \textbf{source} & Parameter Value \ref{param:population} & Parameter Value \ref{param:deaths}  & Parameter Value \ref{param:emigrants} & Parameter Value \ref{param:ie}\\
    \hline
    \textbf{variable} & $\hat{P}(y,r,s,a)$ & $\hat{D}(y,r,s,a)$ & $\hat{E}(y,r,s,a)$ & $\hat{IE}(y,r,s,a)$ \\
    $y\in$ & [1962,2101] & [1962,2100] & [1996,2100] & [1996,2100]\\
    $r\in$ & \multicolumn{4}{c}{federalstates / districts\_districts}\\
    $s\in $ & $\{m,f\}$ & $\{m,f\}$ & $\{m,f\}$ & $\{m,f\}$\\
    $a\in$ & $[0,100]$ & $[0,100]$ & $[0,100]$ & $[0,95+]$\\
    \hline
\end{tabular}
\end{center}
}

After aggregating all data to the same age resolution $[0,95+]$, again,
\begin{equation*}
    Q(y,r,s,a) = \hat{D}(y,r,s,a)+\hat{E}(y,r,s,a)
\end{equation*}
is considered the total number of individuals leaving the cohort, then we apply the enhanced Farr Formula, Corollary \ref{cor:farr_param}, for computation of an internal emigration probability which is compatible with Definition \ref{def:probability}. We want to emphasise that an internally migrated agent remains a component of the model and is still capable of having other events in the course of the remaining life-year. Therefore, internal migrants must not be considered as cohort-leavers.

This results in the following data-set:
\begin{parameterfile}{Internal Emigration Probabilities $\hat{IE}^p$}{districts\_districts}
\hline
\multicolumn{2}{c}{\textbf{Contents}\label{param:iemi_prob}}\\
\hline
\multicolumn{2}{p{15cm}}{Probability that an agent living in region $r$ who has its $a$-th birthday in year $y$ emigrates internally until its next birthday.}\\
\hline
\multicolumn{2}{c}{\textbf{Resolution}}\\
\hline
\textbf{Time-Frame} & $[1996,2100]$\\
\textbf{Regional-Level} & districts\_districts\\
\textbf{Age-Resolution} & $[0,95^+]$\\
\hline
\textbf{Other Regional-Levels} & federalstates\\
\hline
\end{parameterfile}

\newpage
\section{Validation}
\label{sec:validation}
\begin{figure}
    \centering
    \begin{tikzpicture}[scale=0.9, database/.style={
        inner xsep = 0.0,
        outer xsep = 0.0,
        path picture={
            \fill[fill=#1] (-0.5,-0.25) rectangle (0.5,0.25);
            \draw[fill=#1] (0.0, 0.25) circle [x radius=0.5,y radius=0.25];
            \draw (-0.5, 0.0) arc [start angle=180, end angle=360,x radius=0.5, y radius=0.25];
            \draw[fill=#1] (-0.5,-0.25) arc [start angle=180, end angle=360,x radius=0.5, y radius=0.25];
            \draw (-0.5,-0.25) -- (-0.5,0.25);
            \draw (0.5,-0.25) -- (0.5,0.25);
        },
        minimum width=1.5cm,
        minimum height=1.5cm,
        fill=white,
        scale=0.3
    },
    simstate/.style={circle,draw, fill=blue!20,minimum width=0.35cm,
        minimum height=0.35cm},
    simparam/.style={draw,fill=yellow!30,minimum width=0.35cm,
        minimum height=0.35cm},
    paramarrow/.style = {-Stealth,draw=yellow!90!blue,line width = 1},
    simdyn/.style={-Stealth,draw=blue, line width = 1.2},
    labelright/.style={align=right, anchor=east},
    parametercomp/.style={draw=red},
    validation/.style={Stealth-Stealth,draw=green,line width=1.1}
    ]
%\draw[dotted] (0,0) -- (0,-10);
%\draw[dotted] (2.5,0) -- (2.5,-10);
%\draw[dotted] (3.5,0) -- (3.5,-10);
%\draw[dotted] (6,0) -- (6,-10);
%\draw[dotted] (7,0) -- (7,-10);
%\draw[dotted] (9.5,0) -- (9.5,-10);
%\draw[dotted] (12,0) -- (12,-10);

%\draw[fill=green!10] (-0.6,0.8) rectangle (6.25,-1.2);
%\draw[fill=black!10] (6.25,0.8) rectangle (12.6,-1.2);

\node[labelright] at (-1,1) {date (Jan 1\textsuperscript{st})};
\node[rotate=45] at (0,1) {$y_{p,0}$};
\node[] at (1.25,1) {$\dots$};
\node[rotate=45] at (2.5,1) {$y_{0}$};
\node[rotate=45] at (3.5,1) {$y_{0}+1$};
\node[] at (4.75,1) {$\dots$};
\node[rotate=45] at (6,1) {$y_{now}$};
\node[rotate=45] at (7,1) {$y_{now}+1$};
\node[] at (8.25,1) {$\dots$};
\node[rotate=45] at (9.5,1) {$y_{end}$};
\node[] at (10.75,1) {$\dots$};
\node[rotate=45] at (12,1) {$y_{p,end}$};

\node[labelright] at (-1,0) {population data (Jan 1\textsuperscript{st})};
\node[database={white}] (p1) at (0,0) {};
\node[] at (1.25,0) {$\dots$};
\node[database={white}] (p2) at (2.5,0) {};
\node[database={white}] (p3) at (3.5,0) {};
\node[] at (4.75,0) {$\dots$};
\node[database={white}] (p4) at (6,0) {};
\node[database={black!20}] (p5) at (7,0) {};
\node[] at (8.25,0) {$\dots$};
\node[database={black!20}] (p6) at (9.5,0) {};
\node[] at (10.75,0) {$\dots$};
\node[database={black!20}] (p7) at (12,0) {};

\node[labelright] at (-1,-0.75) {births, deaths\\ migration data (year)};
\node[database={white}] (d1) at (0.5,-0.75) {};
\node[] at (1.25,-0.75) {$\dots$};
\node[database={white}] (d2) at (2,-0.75) {};
\node[database={white}] (d3) at (3,-0.75) {};
\node[database={white}] (d4) at (4,-0.75) {};
\node[] at (4.75,-0.75) {$\dots$};
\node[database={white}] (d5) at (5.5,-0.75) {};
\node[database={black!20}] (d6) at (6.5,-0.75) {};
\node[database={black!20}] (d7) at (7.5,-0.75) {};
\node[] at (8.25,-0.75) {$\dots$};
\node[database={black!20}]  (d8) at (9,-0.75) {};
\node[database={black!20}]  (d9) at (10,-0.75) {};
\node[] at (10.75,-0.75) {$\dots$};
\node[database={black!20}] (d10) at (11.5,-0.75) {};

\node[labelright] at (2,-2.5) {simulation};
\node[simstate] (s1) at (2.5,-2.5) {};
\node[simstate] (s2) at (3.5,-2.5) {};
\coordinate (s21) at (4.5,-2.5) {};
\coordinate (s22) at (5,-2.5) {};
\node[simstate] (s3) at (6,-2.5) {};
\node[simstate] (s4) at (7,-2.5) {};
\coordinate (s41) at (8,-2.5) {};
\coordinate (s42) at (8.5,-2.5) {};
\node[simstate] (s5) at (9.5,-2.5) {};
\draw[simdyn] (s1) to (s2);
\draw[simdyn] (s2) to (s21);
\draw[simdyn,dotted] (s21) to (s22);
\draw[simdyn] (s22) to (s3);
\draw[simdyn] (s3) to (s4);
\draw[simdyn] (s4) to (s41);
\draw[simdyn,dotted] (s41) to (s42);
\draw[simdyn] (s42) to (s5);

\node[labelright] at (2,-1.5) {initial population};
\node[simparam] (c1) at (2.5,-1.5) {};
\draw[paramarrow] (c1) to (s1);

\node[simparam] (c2) at (3,-3.5) {};
\node[simparam] (c3) at (4,-3.5) {};
\node[simparam] (c4) at (5.5,-3.5) {};
\node[simparam] (c5) at (6.5,-3.5) {};
\node[simparam] (c6) at (7.5,-3.5) {};
\node[simparam] (c7) at (9,-3.5) {};

\node[labelright] at (2,-3.5) {birth-, death-, \\migration- probabilities};
\draw[paramarrow] (c2) to ++ (0,0.8);
\draw[paramarrow] (c3) to ++ (0,0.8);
\draw[paramarrow] (c4) to ++ (0,0.8);
\draw[paramarrow] (c5) to ++ (0,0.8);
\draw[paramarrow] (c6) to ++ (0,0.8);
\draw[paramarrow] (c7) to ++ (0,0.8);

\node[parametercomp,minimum width=0.5cm,minimum height=0.5cm] (pp2) at (p2) {};
\draw[parametercomp,-Stealth] (pp2) to (c1);
\draw[parametercomp] ($(p1.north west)+(-0.2,0.1)$) rectangle ($(d10.south east)+(0.7,-0.1)$);
\draw[parametercomp,-Stealth] ($(d10.south east)+(0.7,1.0)$) -- ($(d10.south)+(1.5,1.0)$) -- ($(c7)+(4,-0.7)$) -- ($(c7)+(0,-0.7)$) -- (c7);
\draw[parametercomp,-Stealth] ($(c7)+(0,-0.7)$) -- ($(c6)+(0,-0.7)$) -- (c6);
\draw[parametercomp,-Stealth] ($(c6)+(0,-0.7)$) -- ($(c5)+(0,-0.7)$) -- (c5);
\draw[parametercomp,-Stealth] ($(c5)+(0,-0.7)$) -- ($(c4)+(0,-0.7)$) -- (c4);
\draw[parametercomp,-Stealth] ($(c4)+(0,-0.7)$) -- ($(c3)+(0,-0.7)$) -- (c3);
\draw[parametercomp,-Stealth] ($(c3)+(0,-0.7)$) -- ($(c2)+(0,-0.7)$) -- (c2);

\draw[validation] ($(s1.north)+(0.5,0.1)$) -- ($(d3.south)+(0,-0.1)$);
\draw[validation] ($(s2.north)+(0,0.1)$) -- ($(p3.south)+(0,-0.1)$);
\draw[validation] ($(s2.north)+(0.5,0.1)$) -- ($(d4.south)+(0,-0.1)$);
\draw[validation] ($(s3.north)+(-0.5,0.1)$) -- ($(d5.south)+(0,-0.1)$);
\draw[validation] ($(s3.north)+(0,0.1)$) -- ($(p4.south)+(0,-0.1)$);
\draw[validation] ($(s4.north)+(-0.5,0.1)$) -- ($(d6.south)+(0,-0.1)$);
\draw[validation] ($(s4.north)+(0,0.1)$) -- ($(p5.south)+(0,-0.1)$);
\draw[validation] ($(s4.north)+(0.5,0.1)$) -- ($(d7.south)+(0,-0.1)$);
\draw[validation] ($(s5.north)+(-0.5,0.1)$) -- ($(d8.south)+(0,-0.1)$);
\draw[validation] ($(s5.north)+(0.0,0.1)$) -- ($(p6.south)+(0,-0.1)$);
\node[text=green] at (4.75,-1.75) {$\dots$};
\node[text=green] at (8.25,-1.75) {$\dots$};
\node[text=green] at (10.5,-1.75) {validation};
\node[text=red] at (11.2,-3.9) {parameter calculation};
\end{tikzpicture}
    \caption{General parametrisation and validation scheme of GEPOC ABM. Population, death, birth, and migration counts in form of census counts (white) and forecasts (grey) pose the source for the parameter values and the validation reference at the same time.}
    \label{fig:validation}
\end{figure}
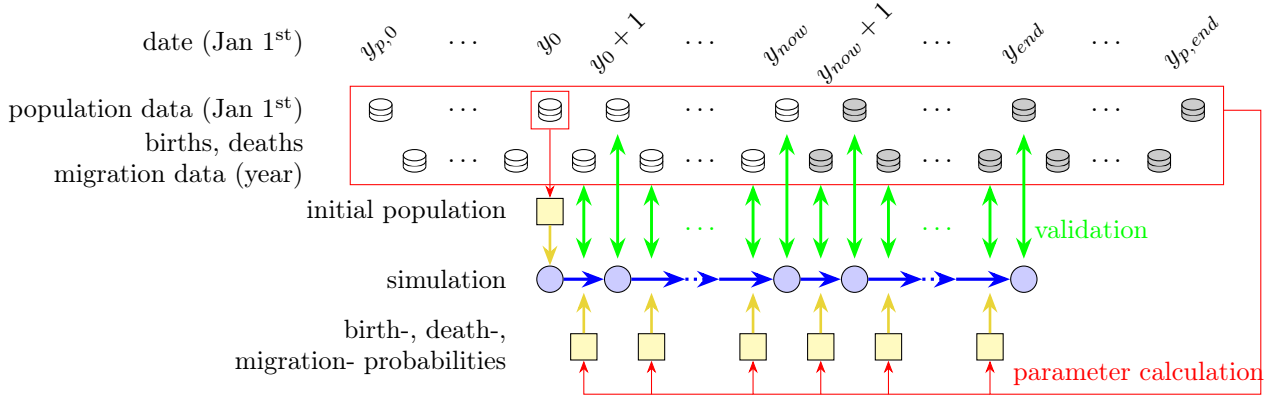

In the following we will quantitatively compare the fully parametrised model GEPOC ABM (see, \cite{bicher2025gepoc}) with the given information about the population of Austria as reference (see Figure \ref{fig:validation}).

\paragraph{Reference.} We will treat the sources described in Section \ref{sec:source_data} as calibration reference. That means that we consider the available data about the Austrian population as ground truth and do not question the accuracy of the data compared to the actual population. This is particularly relevant for the validation of the future population dynamics, where the main forecasting scenario of Statistik Austria (``Prognose Hauptszenario'') will be considered as ground truth, even though there is not yet any actual population to compare to. The model will be regarded as \textit{quantitatively valid} if the differences between the synthetic census, generated by the simulation, and the reference data is sufficiently small.

\paragraph{Error Metric.} For each simulation scenario (see below), we run Monte Carlo runs and average the results. The number of runs was carefully chosen by contrasting the sample standard deviation with the sample mean (using the Gaussian approach described in \cite{bicher_review_2019}). From the simulations, we will carry out all information about demographic quantities introduced in Section \ref{sec:quantities} on a yearly basis and using the highest possible regional, age, and sex resolution. Data for any lower resolution required for comparison with data is aggregated in post-processing.

The quantitative validation will be performed using the following error function for a time series $Y$ compared to a reference $X$:
\begin{equation}
    e_{max}(Y,X):=\max_{t\in[t_1,t_2]}\left(\frac{Y(t)-X(t)}{\max(1,X(t))}\right),\quad e_{min}(Y,X):=\min_{t\in[t_1,t_2]}\left(\frac{Y(t)-X(t)}{\max(1,X(t))}\right) .
\end{equation}
We hereby quantify the highest positive and negative relative distance from the outcome $Y$ to the reference $X$ on the time-frame $[t_1,t_2]$. The $\max$ expression is used to avoid divisions by zero. We argue that this quantity is comparably strict in contrast to usual scores, since it harshly punishes short term fluctuations, which would be averaged when using e.g. classical $R^2$ scores. Moreover, since signs are not cancelled out, it is also well suitable for analysing over- and underestimation.

\paragraph{Simulation Scenarios.} In total, seven simulation scenarios (SC1-SC7) are defined to analyse the validity of the model.

In the first part, we will investigate and compare the outcomes of one scenario for GEPOC ABM Geography (\textbf{SC1}) and one for GEPOC ABM IM (\textbf{SC2}) using the full-regional internal migration model. SC2 will be considered as the main-scenario and a high-accuracy fit is to be expected. For SC1, we want to analyse how well the model can fit the data without regarding internal migration. For both scenarios we will use January 2000 as start date and investigate the population change until 2050 in yearly steps. This way, the time-span covers three different intervals w.r. to the parametrization: for 1996 to 1999 (2001, for migration respectively), data for demographic change is only available on low-resolution. Then, from 2000 (2002) until 2024, the model parameters are estimated based on high-resolution data. Then again, from 2025, the parameters are based on low-resolution forecast data. For details of which parameter values and which regional-levels were used for which parameter in which scenario, we refer to Table \ref{tbl:parametrisation_scenarios}.

In the second part, two additional scenarios for GEPOC ABM IM are evaluated, one with the biregional (\textbf{SC3}) and one with the interregional (\textbf{SC4}) internal migration model. They are compared with the full regional model (\textbf{SC2}) to analyse the impact of the internal migration modelling approach. Finally, in the third part, we will investigate three additional scenarios where other fundamental model-parameters from SC2 are varied. Scenario \textbf{SC5} will extend the simulations beyond 2050 until 2100, \textbf{SC6} uses monthly instead of yearly macro-steps and \textbf{SC7} will use a downscaled population in which every agent represents $10$ real inhabitants of Austria. We refer to the corresponding sections for details about parametrisation and start with the evaluation of SC1 and SC2.

\begin{table}
\begin{center}
    \begin{tabular}{p{4cm}c|cc}
     & & \textbf{SC1} & \textbf{SC2} \\
     \hline
     GEPOC Model & & Geography & IM\\
     sim start-time&$t_{0}$&2000-01-01 & 2000-01-01\\
     time-step lengths & $\Delta t$ & years & years\\
     sim end-time&$t_{e}$&2050-01-01 & 2050-01-01\\
     scale & $\sigma$ & 1.0 & 1.0\\
     Monte-Carlo runs & & 9 & 9\\
     \hline
     probability for male person-agent at birth & $\hat{\alpha}_m$ & P.V. \ref{param:male_prob} & P.V.  \ref{param:male_prob}\\
maximum age & $\hat{a}_{max}$ & $100$ & $100$\\
regional-level for initial population & $r_0$ & municipalities\_registrationdistricts & municipalities\_registrationdistricts\\
data for initial population & $\hat{P}$ &  P.V. \ref{param:population} & P.V. \ref{param:population}\\
regional-level for initial population refinement & $r_{min}$ & 100x100[m] raster & 100x100[m] raster\\
regional-level for births & $r_b$ & federalstates & federalstates\\
birth probabilities & $\hat{B^p}$ & P.V. \ref{param:birth_prob} & P.V. \ref{param:birth_prob}\\
regional-level for deaths & $r_d$ & federalstates & federalstates\\
death probabilities & $\hat{D^p}$ & P.V. \ref{param:death_prob} & P.V. \ref{param:birth_prob}\\
regional-level for emigrants & $r_e$ & districts\_districts & districts\_districts\\
emigration probabilities & $\hat{E^p}$ & P.V. \ref{param:emigration_prob} & P.V. \ref{param:emigration_prob}\\
regional-level for immigrants & $r_i$ & municipalities\_registrationdistricts & municipalities\_registrationdistricts\\
immigrants & $\hat{I}$ & P.V. \ref{param:immigrants} & P.V. \ref{param:immigrants}\\
\hline
internal migration model & & & full regional\\
regional-level for internal migration & $r_{ie}$ & - &districts\_districts\\
internal emigration probabilities & $\hat{IE}^p$ & - & P.V. \ref{param:ie}\\
internal migrants & $\hat{M}^p$ & - & P.V. \ref{param:im}\\
\hline
region families for all regional-levels except $r_{min}$& $A_j^{r_0}$ & P.V. \ref{param:regions} & P.V. \ref{param:regions}\\
region families for population refinement (level $r_{min}$) & $A_j^{r_{min}}$ & P.V. \ref{param:regions_min} & P.V. \ref{param:regions_min}\\
\end{tabular}
\end{center}
\caption{Parametrisation setup for simulation scenarios SC1 and SC2.}
\label{tbl:parametrisation_scenarios}
\end{table}

\subsection{Population}
\label{sec:validation_population}
Since it is the most important variable for GEPOC, we start by comparing the simulation results from SC1 and SC2 with the data for the total population.

\subsubsection{Comparison with Source \ref{src:popBase} (Bevölkerungsstand)}
\begin{figure}
    \centering
    \includegraphics[width=0.7\linewidth]{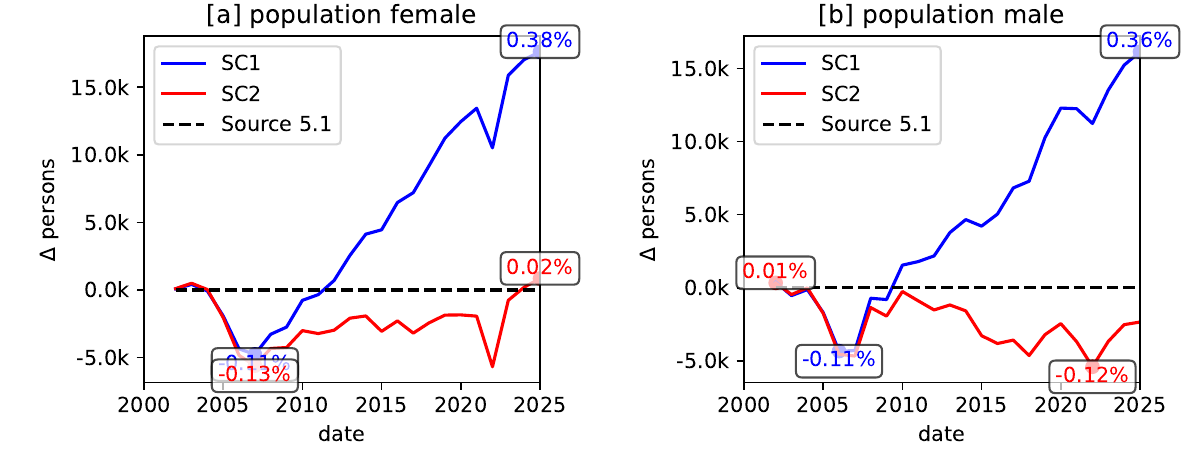}
    \caption{Differences between the male and female population data (Source \ref{src:popBase}) and the two simulation scenarios SC1 and SC2.}
    \label{fig:population_sex_diff}
\end{figure}

\begin{figure}
    \centering
    \includegraphics[width=\linewidth]{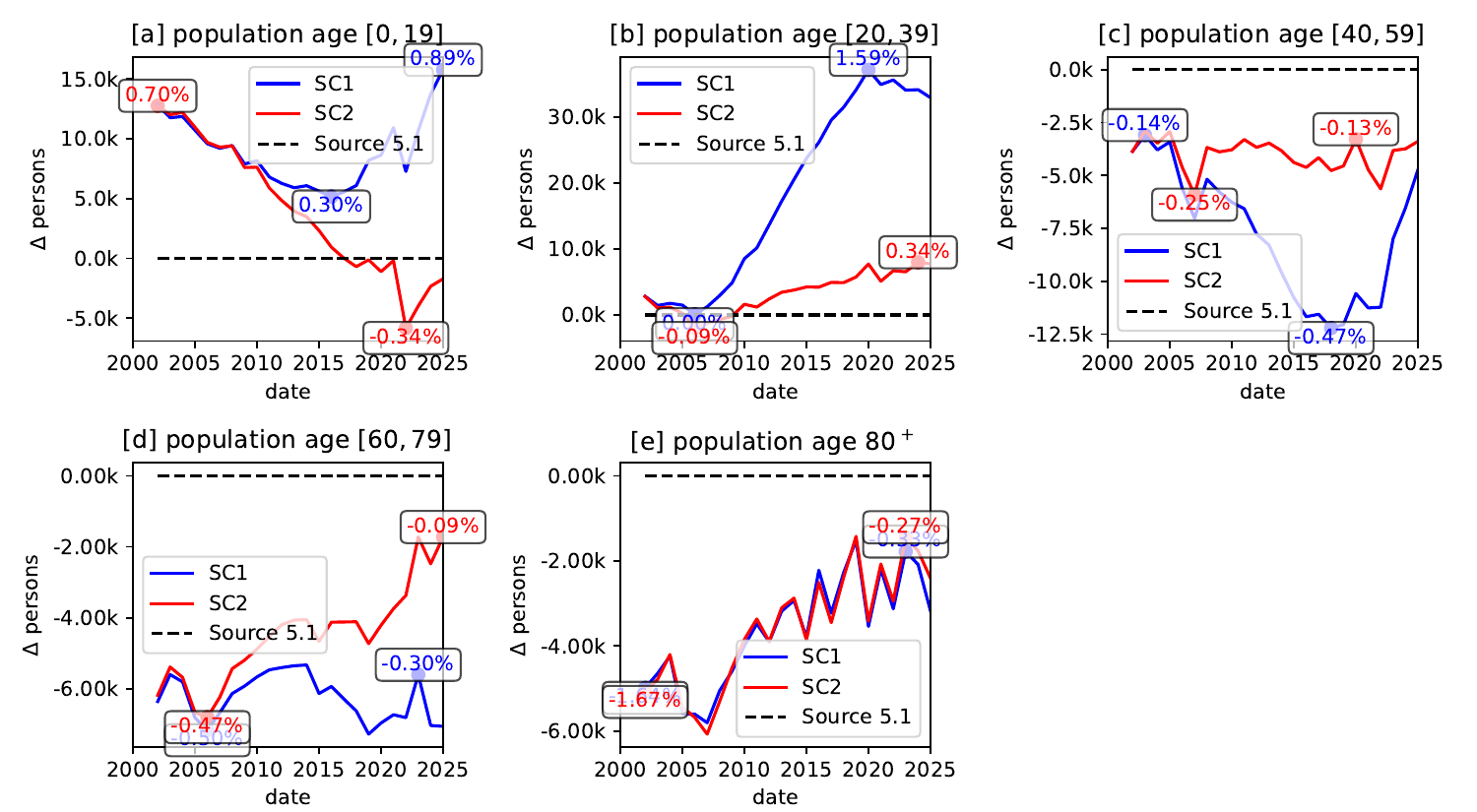}
    \caption{Differences between the population data (Source \ref{src:popBase}) and the two simulation scenarios SC1 and SC2 after aggregation to Austria and 20-year age classes.}
    \label{fig:population_age_diff}
\end{figure}

\begin{figure}
    \centering
    \includegraphics[width=\linewidth]{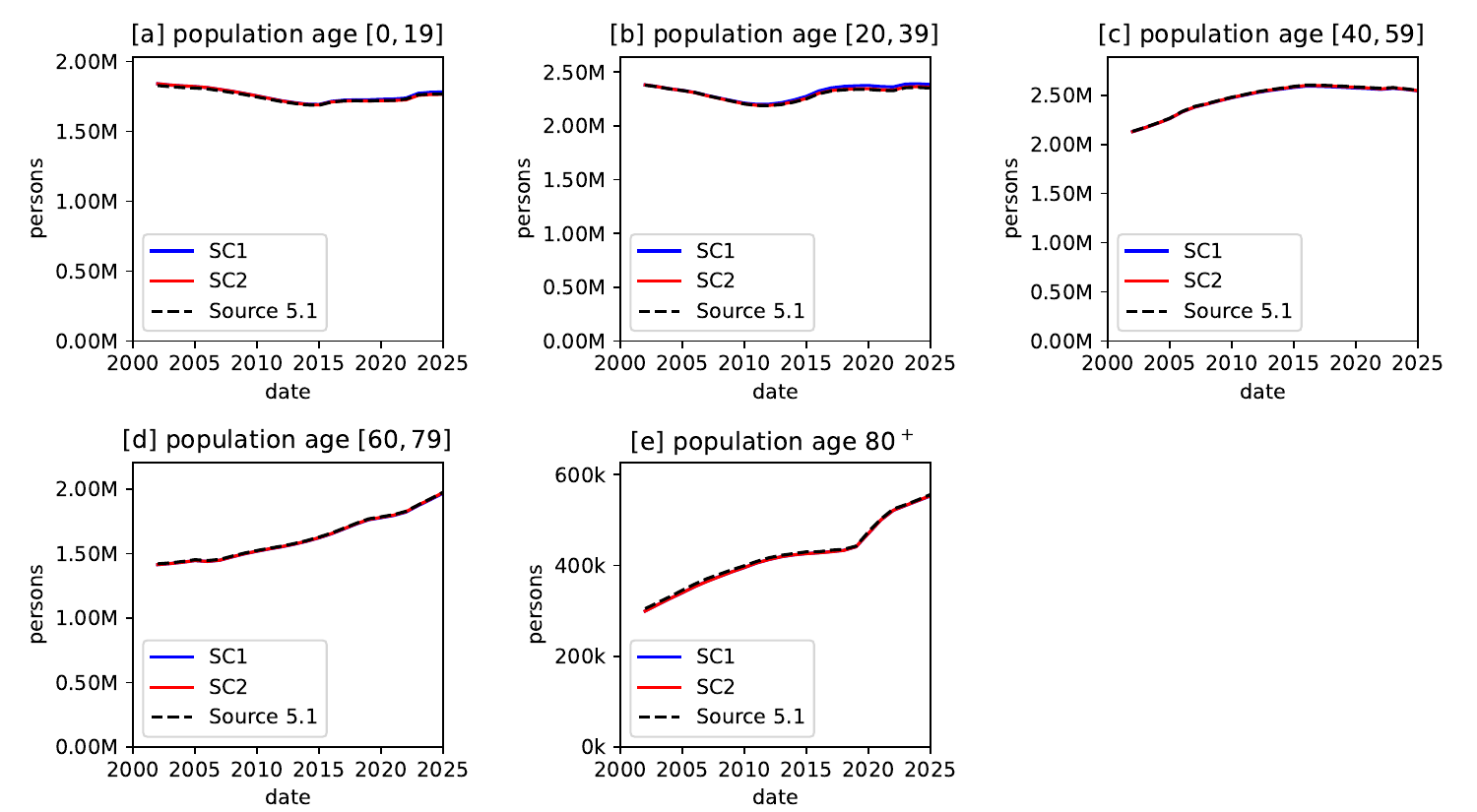}
    \caption{Comparison between the population data (Source \ref{src:popBase}) and the two simulation scenarios SC1 and SC2 after aggregation to Austria and 20-year age classes.}
    \label{fig:population_age}
\end{figure}

\begin{figure}
    \centering
    \includegraphics[width=\linewidth]{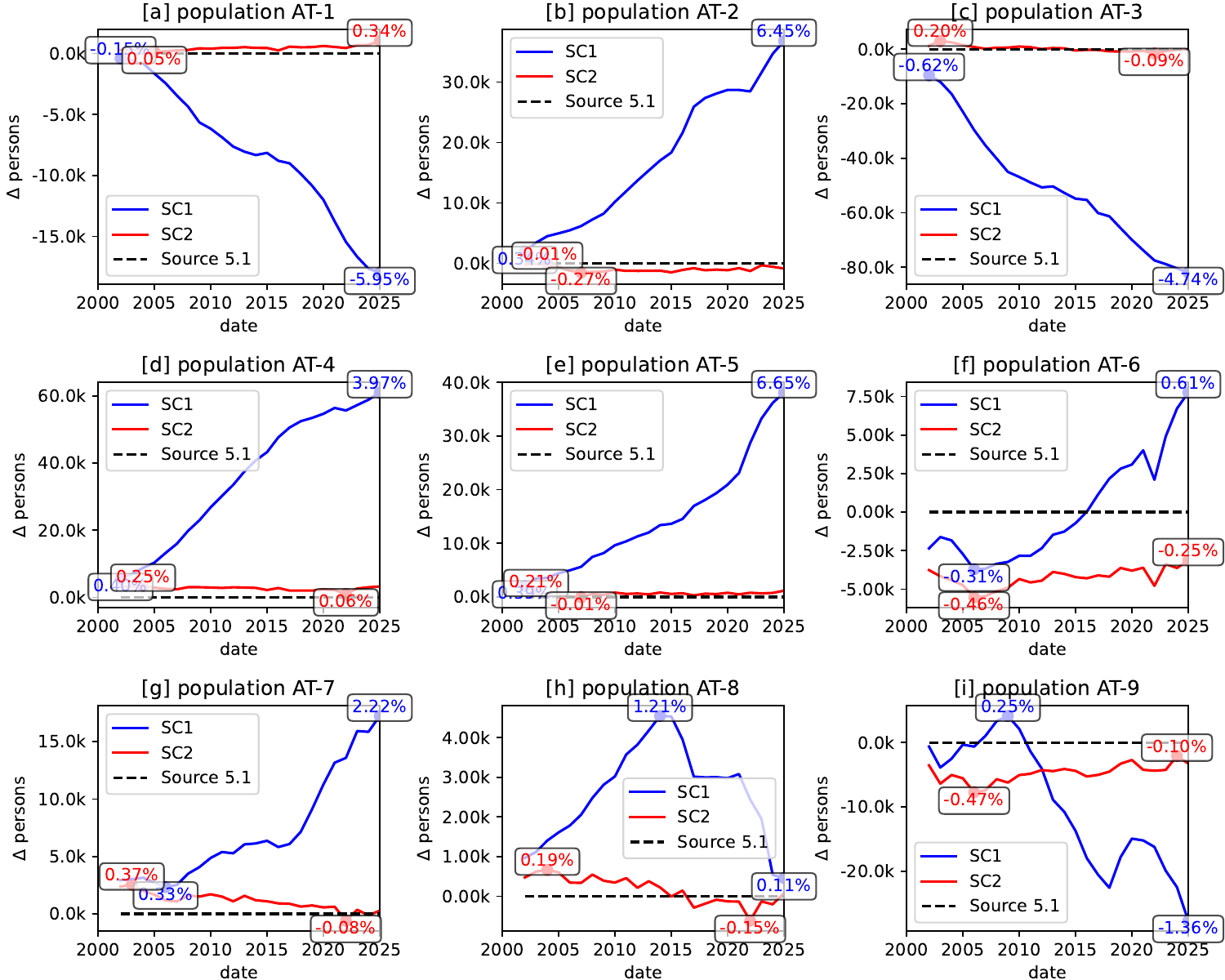}
    \caption{Differences between the population data (Source \ref{src:popBase}) and the two simulation scenarios SC1 and SC2 for all federalstates of Austria.}
    \label{fig:population_fed_diff}
\end{figure}

First of all, we compare the values with the highly resolved data given by Source \ref{src:popBase} for the period between 2002 and 2025.
\paragraph{Total and sex.} Figure \ref{fig:population_sex_diff} shows the a comparison between SC1, SC2 and the reference from Source \ref{src:popBase} for the total male and female population. After 25 years of simulation, the overall population per sex remained within $0.4\%$ deviation of the actual population and the two cohorts are less than 20000 persons off. The IM model performs even better, where the error remained within $0.15\%$. This is not surprising, since the model should, in theory, conserve the regional age structure better. Yet the figure also indicates a trend to slightly overestimate the total population for both models.
\paragraph{Age-classes.} Difference curves for twenty-year age-classes are displayed in Figure \ref{fig:population_age_diff}. The differences vary between -13000 and 40000 individuals, which may sound large at the first glance, but effectively correspond to relative differences within $\pm 1.6\%$. In contrast to Figure \ref{fig:population_age_diff}, Figure \ref{fig:population_age} displays the absolute quantities where, the curves from SC1 and SC2 are almost indistinguishable from the reference data. This type of plot is only being used this single time to visualise that an offset of 40000 persons might only seem large on the first glance. Anyhow, in particular for the age class 20-39, the internal migration model gives more accurate results.
\paragraph{Federal-states.} Figure \ref{fig:population_fed_diff} shows the analogous plot to Figure \ref{fig:population_age_diff} for the nine federal states of Austria. As expected, scenario SC1 without internal migration results in larger deviations, up to $7\%$. The two federal states with the highest deviations, AT-1 and AT-2, are well known for being subject to high internal migration-dynamics. Scenario SC2 with internal migration showed its highest deviation in AT-6 and AT-9 with roughly $-0.5\%$.

\paragraph{Finer aggregation levels.} The maximum deviations for the mentioned and also more detailed aggregation levels are summarised in Table \ref{tbl:deviations1}. Regionally, comparably large offsets can be found for elderly people in both models, however, SC2 performs, on average, much better.

\begin{table}
    \begin{center}
    \begin{scriptsize}
        \begin{tabular}{ccc|cc|cc}
        region & sex & age & \multicolumn{2}{c|}{\textbf{SC1} ($e_{min},e_{max}$)} & \multicolumn{2}{c}{\textbf{SC2} ($e_{min},e_{max}$)}\\
        \hline
-&-&-&\cellcolor{blue!1}$-0.11\%$&\cellcolor{red!3}$0.37\%$&\cellcolor{blue!1}$-0.12\%$&\cellcolor{red!1}$0.01\%$\\
\hline
-&female&-&\cellcolor{blue!1}$-0.11\%$&\cellcolor{red!3}$0.38\%$&\cellcolor{blue!1}$-0.13\%$&\cellcolor{red!1}$0.02\%$\\
-&male&-&\cellcolor{blue!1}$-0.11\%$&\cellcolor{red!3}$0.36\%$&\cellcolor{blue!1}$-0.12\%$&\cellcolor{red!1}$0.01\%$\\
\hline
-&-&$[0,19]$&\cellcolor{red!2}$0.30\%$&\cellcolor{red!6}$0.89\%$&\cellcolor{blue!3}$-0.34\%$&\cellcolor{red!5}$0.70\%$\\
-&-&$[20,39]$&\cellcolor{red!1}$0.00\%$&\cellcolor{red!10}$1.59\%$&\cellcolor{blue!1}$-0.09\%$&\cellcolor{red!3}$0.34\%$\\
-&-&$[40,59]$&\cellcolor{blue!4}$-0.47\%$&\cellcolor{blue!1}$-0.14\%$&\cellcolor{blue!2}$-0.25\%$&\cellcolor{blue!1}$-0.13\%$\\
-&-&$[60,79]$&\cellcolor{blue!4}$-0.50\%$&\cellcolor{blue!2}$-0.30\%$&\cellcolor{blue!4}$-0.47\%$&\cellcolor{blue!1}$-0.09\%$\\
-&-&$80^+$&\cellcolor{blue!11}$-1.64\%$&\cellcolor{blue!3}$-0.33\%$&\cellcolor{blue!11}$-1.67\%$&\cellcolor{blue!2}$-0.27\%$\\
\hline
AT-1&-&-&\cellcolor{blue!40}$-5.95\%$&\cellcolor{blue!1}$-0.15\%$&\cellcolor{red!1}$0.05\%$&\cellcolor{red!3}$0.34\%$\\
AT-2&-&-&\cellcolor{red!3}$0.34\%$&\cellcolor{red!41}$6.45\%$&\cellcolor{blue!2}$-0.27\%$&\cellcolor{blue!1}$-0.01\%$\\
AT-3&-&-&\cellcolor{blue!32}$-4.74\%$&\cellcolor{blue!5}$-0.62\%$&\cellcolor{blue!1}$-0.09\%$&\cellcolor{red!2}$0.20\%$\\
AT-4&-&-&\cellcolor{red!3}$0.40\%$&\cellcolor{red!25}$3.97\%$&\cellcolor{red!1}$0.06\%$&\cellcolor{red!2}$0.25\%$\\
AT-5&-&-&\cellcolor{red!3}$0.39\%$&\cellcolor{red!42}$6.65\%$&\cellcolor{blue!1}$-0.01\%$&\cellcolor{red!2}$0.21\%$\\
AT-6&-&-&\cellcolor{blue!3}$-0.31\%$&\cellcolor{red!4}$0.61\%$&\cellcolor{blue!4}$-0.46\%$&\cellcolor{blue!2}$-0.25\%$\\
AT-7&-&-&\cellcolor{red!3}$0.33\%$&\cellcolor{red!14}$2.22\%$&\cellcolor{blue!1}$-0.08\%$&\cellcolor{red!3}$0.37\%$\\
AT-8&-&-&\cellcolor{red!1}$0.11\%$&\cellcolor{red!8}$1.21\%$&\cellcolor{blue!2}$-0.15\%$&\cellcolor{red!2}$0.19\%$\\
AT-9&-&-&\cellcolor{blue!9}$-1.36\%$&\cellcolor{red!2}$0.25\%$&\cellcolor{blue!4}$-0.47\%$&\cellcolor{blue!1}$-0.10\%$\\
\hline
AT-1&-&$[0,19]$&\cellcolor{blue!44}$-6.57\%$&\cellcolor{red!2}$0.19\%$&\cellcolor{blue!3}$-0.34\%$&\cellcolor{red!8}$1.19\%$\\
AT-2&-&$[0,19]$&\cellcolor{red!5}$0.77\%$&\cellcolor{red!83}$13.23\%$&\cellcolor{blue!2}$-0.21\%$&\cellcolor{red!5}$0.78\%$\\
AT-3&-&$[0,19]$&\cellcolor{blue!45}$-6.86\%$&\cellcolor{blue!2}$-0.18\%$&\cellcolor{blue!4}$-0.49\%$&\cellcolor{red!6}$0.83\%$\\
AT-4&-&$[0,19]$&\cellcolor{red!5}$0.76\%$&\cellcolor{red!40}$6.33\%$&\cellcolor{blue!1}$-0.00\%$&\cellcolor{red!6}$0.83\%$\\
AT-5&-&$[0,19]$&\cellcolor{red!9}$1.33\%$&\cellcolor{red!55}$8.73\%$&\cellcolor{red!1}$0.14\%$&\cellcolor{red!6}$0.95\%$\\
AT-6&-&$[0,19]$&\cellcolor{red!3}$0.35\%$&\cellcolor{red!13}$2.00\%$&\cellcolor{blue!3}$-0.39\%$&\cellcolor{red!5}$0.71\%$\\
AT-7&-&$[0,19]$&\cellcolor{red!5}$0.67\%$&\cellcolor{red!14}$2.18\%$&\cellcolor{blue!5}$-0.64\%$&\cellcolor{red!8}$1.14\%$\\
AT-8&-&$[0,19]$&\cellcolor{blue!8}$-1.15\%$&\cellcolor{red!4}$0.63\%$&\cellcolor{blue!5}$-0.71\%$&\cellcolor{red!4}$0.60\%$\\
AT-9&-&$[0,19]$&\cellcolor{blue!18}$-2.73\%$&\cellcolor{red!25}$3.92\%$&\cellcolor{blue!5}$-0.73\%$&\cellcolor{red!3}$0.36\%$\\
AT-1&-&$[20,39]$&\cellcolor{red!7}$1.01\%$&\cellcolor{red!58}$9.21\%$&\cellcolor{blue!2}$-0.16\%$&\cellcolor{red!7}$0.97\%$\\
AT-2&-&$[20,39]$&\cellcolor{red!8}$1.19\%$&\cellcolor{red!100}$15.99\%$&\cellcolor{blue!3}$-0.40\%$&\cellcolor{red!2}$0.25\%$\\
AT-3&-&$[20,39]$&\cellcolor{blue!5}$-0.74\%$&\cellcolor{red!32}$5.00\%$&\cellcolor{red!1}$0.12\%$&\cellcolor{red!4}$0.49\%$\\
AT-4&-&$[20,39]$&\cellcolor{red!8}$1.20\%$&\cellcolor{red!56}$8.82\%$&\cellcolor{red!1}$0.14\%$&\cellcolor{red!4}$0.63\%$\\
AT-5&-&$[20,39]$&\cellcolor{red!5}$0.77\%$&\cellcolor{red!75}$11.85\%$&\cellcolor{blue!1}$-0.09\%$&\cellcolor{red!3}$0.43\%$\\
AT-6&-&$[20,39]$&\cellcolor{blue!2}$-0.25\%$&\cellcolor{red!15}$2.38\%$&\cellcolor{blue!6}$-0.80\%$&\cellcolor{red!3}$0.35\%$\\
AT-7&-&$[20,39]$&\cellcolor{blue!1}$-0.08\%$&\cellcolor{red!19}$2.90\%$&\cellcolor{blue!1}$-0.03\%$&\cellcolor{red!2}$0.32\%$\\
AT-8&-&$[20,39]$&\cellcolor{red!1}$0.06\%$&\cellcolor{red!26}$4.12\%$&\cellcolor{red!1}$0.11\%$&\cellcolor{red!3}$0.38\%$\\
AT-9&-&$[20,39]$&\cellcolor{blue!74}$-11.26\%$&\cellcolor{blue!7}$-0.92\%$&\cellcolor{blue!3}$-0.36\%$&\cellcolor{red!5}$0.71\%$\\
AT-1&-&$[40,59]$&\cellcolor{blue!54}$-8.15\%$&\cellcolor{blue!5}$-0.74\%$&\cellcolor{blue!1}$-0.01\%$&\cellcolor{red!2}$0.30\%$\\
AT-2&-&$[40,59]$&\cellcolor{blue!3}$-0.36\%$&\cellcolor{red!29}$4.50\%$&\cellcolor{blue!2}$-0.24\%$&\cellcolor{blue!1}$-0.01\%$\\
AT-3&-&$[40,59]$&\cellcolor{blue!66}$-10.05\%$&\cellcolor{blue!5}$-0.74\%$&\cellcolor{blue!1}$-0.07\%$&\cellcolor{red!1}$0.09\%$\\
AT-4&-&$[40,59]$&\cellcolor{red!1}$0.11\%$&\cellcolor{red!20}$3.06\%$&\cellcolor{red!1}$0.13\%$&\cellcolor{red!2}$0.25\%$\\
AT-5&-&$[40,59]$&\cellcolor{red!1}$0.02\%$&\cellcolor{red!30}$4.79\%$&\cellcolor{blue!2}$-0.20\%$&\cellcolor{red!2}$0.23\%$\\
AT-6&-&$[40,59]$&\cellcolor{blue!5}$-0.65\%$&\cellcolor{red!5}$0.66\%$&\cellcolor{blue!6}$-0.81\%$&\cellcolor{blue!3}$-0.43\%$\\
AT-7&-&$[40,59]$&\cellcolor{red!3}$0.35\%$&\cellcolor{red!14}$2.18\%$&\cellcolor{blue!1}$-0.03\%$&\cellcolor{red!2}$0.30\%$\\
AT-8&-&$[40,59]$&\cellcolor{blue!3}$-0.43\%$&\cellcolor{red!2}$0.18\%$&\cellcolor{blue!1}$-0.07\%$&\cellcolor{red!1}$0.16\%$\\
AT-9&-&$[40,59]$&\cellcolor{red!1}$0.02\%$&\cellcolor{red!31}$4.84\%$&\cellcolor{blue!6}$-0.84\%$&\cellcolor{blue!3}$-0.45\%$\\
AT-1&-&$[60,79]$&\cellcolor{blue!80}$-12.19\%$&\cellcolor{blue!6}$-0.86\%$&\cellcolor{blue!3}$-0.45\%$&\cellcolor{red!2}$0.22\%$\\
AT-2&-&$[60,79]$&\cellcolor{blue!15}$-2.27\%$&\cellcolor{blue!3}$-0.40\%$&\cellcolor{blue!4}$-0.55\%$&\cellcolor{blue!2}$-0.23\%$\\
AT-3&-&$[60,79]$&\cellcolor{blue!41}$-6.11\%$&\cellcolor{blue!6}$-0.81\%$&\cellcolor{blue!3}$-0.42\%$&\cellcolor{red!1}$0.02\%$\\
AT-4&-&$[60,79]$&\cellcolor{blue!4}$-0.61\%$&\cellcolor{red!3}$0.35\%$&\cellcolor{blue!4}$-0.60\%$&\cellcolor{red!2}$0.18\%$\\
AT-5&-&$[60,79]$&\cellcolor{blue!4}$-0.57\%$&\cellcolor{red!18}$2.73\%$&\cellcolor{blue!5}$-0.63\%$&\cellcolor{red!1}$0.13\%$\\
AT-6&-&$[60,79]$&\cellcolor{blue!12}$-1.78\%$&\cellcolor{blue!3}$-0.41\%$&\cellcolor{blue!4}$-0.60\%$&\cellcolor{blue!3}$-0.36\%$\\
AT-7&-&$[60,79]$&\cellcolor{blue!1}$-0.05\%$&\cellcolor{red!12}$1.91\%$&\cellcolor{blue!1}$-0.14\%$&\cellcolor{red!2}$0.27\%$\\
AT-8&-&$[60,79]$&\cellcolor{blue!2}$-0.23\%$&\cellcolor{red!10}$1.50\%$&\cellcolor{blue!2}$-0.27\%$&\cellcolor{red!2}$0.29\%$\\
AT-9&-&$[60,79]$&\cellcolor{blue!1}$-0.07\%$&\cellcolor{red!47}$7.50\%$&\cellcolor{blue!5}$-0.72\%$&\cellcolor{blue!3}$-0.38\%$\\
AT-1&-&$80^+$&\cellcolor{blue!55}$-8.35\%$&\cellcolor{blue!16}$-2.38\%$&\cellcolor{blue!12}$-1.80\%$&\cellcolor{red!5}$0.73\%$\\
AT-2&-&$80^+$&\cellcolor{blue!24}$-3.60\%$&\cellcolor{blue!7}$-1.01\%$&\cellcolor{blue!21}$-3.06\%$&\cellcolor{blue!2}$-0.21\%$\\
AT-3&-&$80^+$&\cellcolor{blue!38}$-5.73\%$&\cellcolor{blue!18}$-2.67\%$&\cellcolor{blue!13}$-1.84\%$&\cellcolor{blue!2}$-0.21\%$\\
AT-4&-&$80^+$&\cellcolor{blue!16}$-2.43\%$&\cellcolor{blue!4}$-0.48\%$&\cellcolor{blue!17}$-2.45\%$&\cellcolor{blue!4}$-0.46\%$\\
AT-5&-&$80^+$&\cellcolor{blue!16}$-2.33\%$&\cellcolor{red!5}$0.74\%$&\cellcolor{blue!15}$-2.23\%$&\cellcolor{blue!1}$-0.05\%$\\
AT-6&-&$80^+$&\cellcolor{blue!16}$-2.29\%$&\cellcolor{blue!5}$-0.75\%$&\cellcolor{blue!18}$-2.70\%$&\cellcolor{blue!3}$-0.42\%$\\
AT-7&-&$80^+$&\cellcolor{blue!11}$-1.65\%$&\cellcolor{red!6}$0.83\%$&\cellcolor{blue!10}$-1.40\%$&\cellcolor{red!4}$0.54\%$\\
AT-8&-&$80^+$&\cellcolor{blue!18}$-2.66\%$&\cellcolor{red!4}$0.52\%$&\cellcolor{blue!17}$-2.58\%$&\cellcolor{blue!1}$-0.07\%$\\
AT-9&-&$80^+$&\cellcolor{blue!1}$-0.01\%$&\cellcolor{red!44}$6.99\%$&\cellcolor{blue!9}$-1.28\%$&\cellcolor{blue!2}$-0.22\%$\\
        \end{tabular}
    \end{scriptsize}
    \end{center}
    \caption{Relative maximum differences between the total population data (Source \ref{src:popBase} and the simulations SC1 and SC2 between 2002 and 2025.}
    \label{tbl:deviations1}
\end{table}

\subsubsection{Comparison with Source \ref{src:popForecast} (Bevölkerung zum Jahresanfang 1952 bis 2101)}
\begin{figure}
    \centering
    \includegraphics[width=0.8\linewidth]{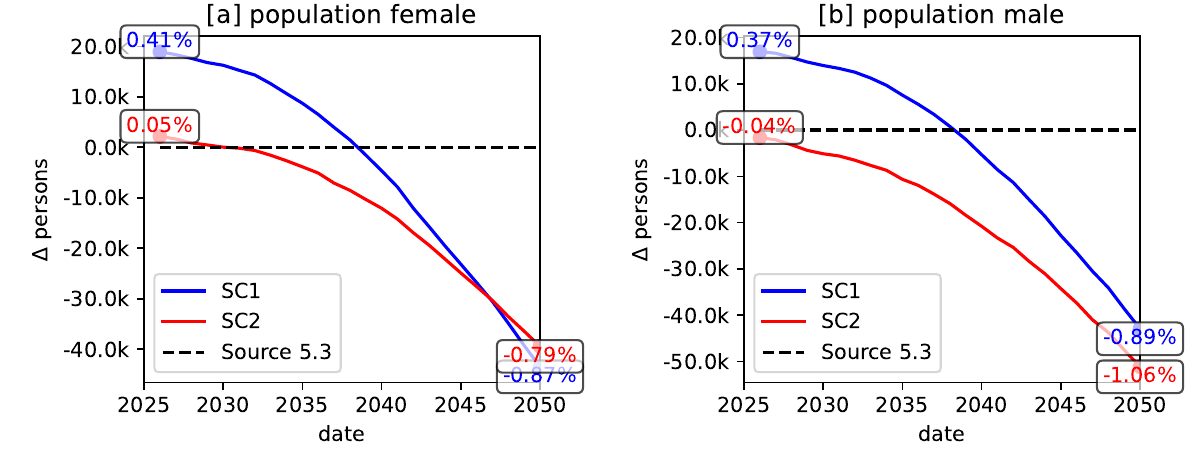}
    \caption{Differences between the male and female population forecast data (Source \ref{src:popForecast}) and the two simulation scenarios SC1 and SC2.}
    \label{fig:population_fc_sex_diff}
\end{figure}

\begin{figure}
    \centering
    \includegraphics[width=\linewidth]{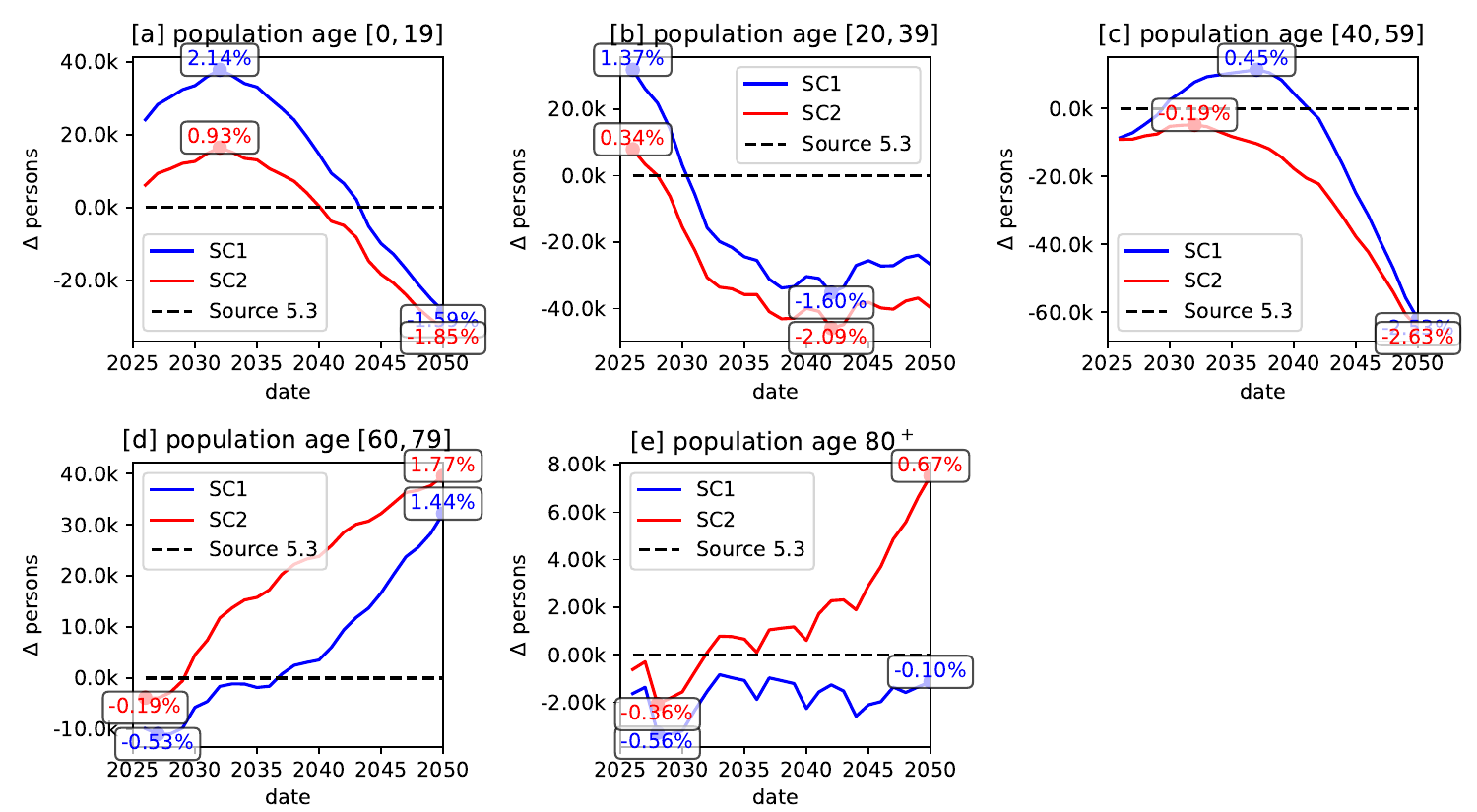}
    \caption{Differences between the population forecast data (Source \ref{src:popForecast}) and the two simulation scenarios SC1 and SC2 after aggregation to Austria and 20-year age classes.}
    \label{fig:population_fc_age_diff}
\end{figure}

\begin{figure}
    \centering
    \includegraphics[width=\linewidth]{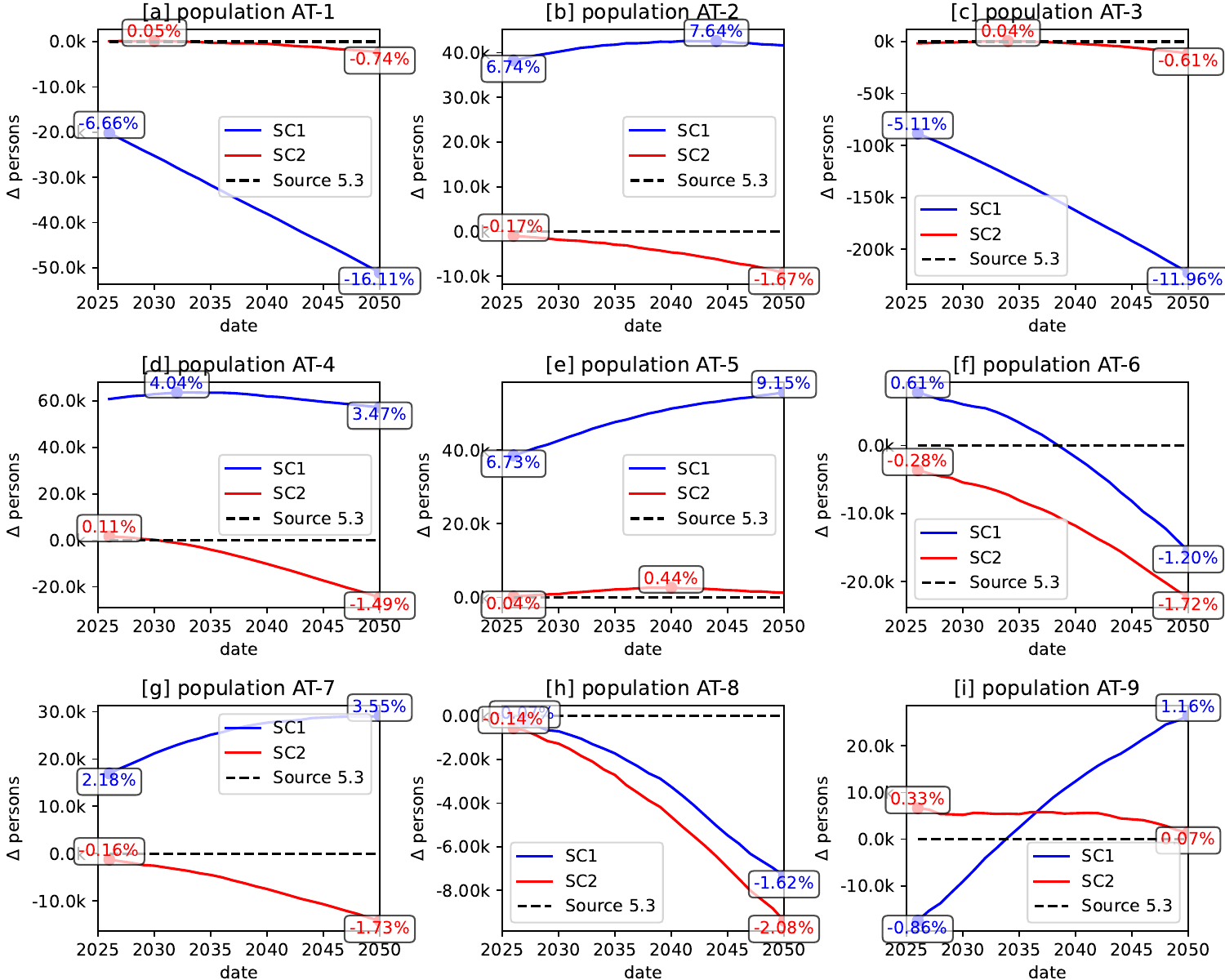}
    \caption{Differences between the population forecast data (Source \ref{src:popForecast}) and the two simulation scenarios SC1 and SC2 for all federalstates of Austria.}
    \label{fig:population_fc_fed_diff}
\end{figure}

For analysing the quality beyond the period, for which actual data is available, we use Source \ref{src:popForecast} between 2026 to 2050. Note that the age and regional resolution given by the source is sufficient to perform the same analysis as in the prior section. Figures \ref{fig:population_fc_sex_diff}, \ref{fig:population_fc_age_diff}, and \ref{fig:population_fc_fed_diff} provide the analogous plots to the ones shown in the previous section, but for the time-frame between 2026-01-01 and 2050-01-01. 
\paragraph{Total and sex.}
Figure \ref{fig:population_fc_sex_diff} shows that the trend towards overestimation indicated by Figure \ref{fig:population_sex_diff} is not prolonged, but turns into a clear underestimation with $-1\%$ at the end of the period. Here, both models perform roughly equally well. In general, the trend towards underestimation can be explained by inaccurate disaggregation of the low-resolution forecast data. At least on this high aggregation level, the problem seems to be more severe than the missing internal migration and causes both model scenarios to become inaccurate eventually.

\paragraph{Federal-states, age-classes and finer.}
On the regional-level, the results from SC1 start to differ more heavily from the ground truth, whereas SC2 captures the regional trends much better and keeps them within the $\pm 2\%$ region. The differences for the age-classes remain in a  similar magnitude as for 2002-2025 for both models. On a less aggregated scope (see Table \ref{tbl:deviations2}), we find the largest differences for the SC1 model for the [20,39] cohort in AT-2, the elderly in AT-9 (Vienna) and the [0,19] cohort in AT-1. The latter two can be explained by the high internal migration dynamics between AT-1, AT-3 and the city of Vienna AT-9. 

\begin{table}
    \begin{center}
    \begin{scriptsize}
        \begin{tabular}{ccc|cc|cc}
        region & sex & age & \multicolumn{2}{c|}{\textbf{SC1} ($e_{min},e_{max}$)} & \multicolumn{2}{c}{\textbf{SC2} ($e_{min},e_{max}$)}\\
        \hline
-&-&-&\cellcolor{blue!4}$-0.88\%$&\cellcolor{red!2}$0.39\%$&\cellcolor{blue!4}$-0.92\%$&\cellcolor{red!1}$0.01\%$\\
\hline
-&female&-&\cellcolor{blue!4}$-0.87\%$&\cellcolor{red!2}$0.41\%$&\cellcolor{blue!4}$-0.79\%$&\cellcolor{red!1}$0.05\%$\\
-&male&-&\cellcolor{blue!4}$-0.89\%$&\cellcolor{red!2}$0.37\%$&\cellcolor{blue!5}$-1.06\%$&\cellcolor{blue!1}$-0.04\%$\\
\hline
-&-&$[0,19]$&\cellcolor{blue!7}$-1.59\%$&\cellcolor{red!11}$2.14\%$&\cellcolor{blue!8}$-1.85\%$&\cellcolor{red!5}$0.93\%$\\
-&-&$[20,39]$&\cellcolor{blue!7}$-1.60\%$&\cellcolor{red!7}$1.37\%$&\cellcolor{blue!9}$-2.09\%$&\cellcolor{red!2}$0.34\%$\\
-&-&$[40,59]$&\cellcolor{blue!11}$-2.53\%$&\cellcolor{red!3}$0.45\%$&\cellcolor{blue!11}$-2.63\%$&\cellcolor{blue!1}$-0.19\%$\\
-&-&$[60,79]$&\cellcolor{blue!3}$-0.53\%$&\cellcolor{red!7}$1.44\%$&\cellcolor{blue!1}$-0.19\%$&\cellcolor{red!9}$1.77\%$\\
-&-&$80^+$&\cellcolor{blue!3}$-0.56\%$&\cellcolor{blue!1}$-0.10\%$&\cellcolor{blue!2}$-0.36\%$&\cellcolor{red!4}$0.67\%$\\
\hline
AT-1&-&-&\cellcolor{blue!65}$-16.11\%$&\cellcolor{blue!27}$-6.66\%$&\cellcolor{blue!3}$-0.74\%$&\cellcolor{red!1}$0.05\%$\\
AT-2&-&-&\cellcolor{red!33}$6.74\%$&\cellcolor{red!37}$7.64\%$&\cellcolor{blue!7}$-1.67\%$&\cellcolor{blue!1}$-0.17\%$\\
AT-3&-&-&\cellcolor{blue!48}$-11.96\%$&\cellcolor{blue!21}$-5.11\%$&\cellcolor{blue!3}$-0.61\%$&\cellcolor{red!1}$0.04\%$\\
AT-4&-&-&\cellcolor{red!17}$3.47\%$&\cellcolor{red!20}$4.04\%$&\cellcolor{blue!6}$-1.49\%$&\cellcolor{red!1}$0.11\%$\\
AT-5&-&-&\cellcolor{red!33}$6.73\%$&\cellcolor{red!45}$9.15\%$&\cellcolor{red!1}$0.04\%$&\cellcolor{red!3}$0.44\%$\\
AT-6&-&-&\cellcolor{blue!5}$-1.20\%$&\cellcolor{red!3}$0.61\%$&\cellcolor{blue!7}$-1.72\%$&\cellcolor{blue!2}$-0.28\%$\\
AT-7&-&-&\cellcolor{red!11}$2.18\%$&\cellcolor{red!18}$3.55\%$&\cellcolor{blue!7}$-1.73\%$&\cellcolor{blue!1}$-0.16\%$\\
AT-8&-&-&\cellcolor{blue!7}$-1.62\%$&\cellcolor{blue!1}$-0.07\%$&\cellcolor{blue!9}$-2.08\%$&\cellcolor{blue!1}$-0.14\%$\\
AT-9&-&-&\cellcolor{blue!4}$-0.86\%$&\cellcolor{red!6}$1.16\%$&\cellcolor{red!1}$0.07\%$&\cellcolor{red!2}$0.33\%$\\
\hline
AT-1&-&$[0,19]$&\cellcolor{blue!80}$-19.98\%$&\cellcolor{blue!27}$-6.59\%$&\cellcolor{blue!11}$-2.59\%$&\cellcolor{red!3}$0.56\%$\\
AT-2&-&$[0,19]$&\cellcolor{red!70}$14.40\%$&\cellcolor{red!83}$17.09\%$&\cellcolor{blue!10}$-2.38\%$&\cellcolor{red!1}$0.09\%$\\
AT-3&-&$[0,19]$&\cellcolor{blue!65}$-16.06\%$&\cellcolor{blue!24}$-5.94\%$&\cellcolor{blue!13}$-3.24\%$&\cellcolor{red!3}$0.46\%$\\
AT-4&-&$[0,19]$&\cellcolor{red!7}$1.33\%$&\cellcolor{red!33}$6.68\%$&\cellcolor{blue!15}$-3.63\%$&\cellcolor{red!4}$0.68\%$\\
AT-5&-&$[0,19]$&\cellcolor{red!39}$8.06\%$&\cellcolor{red!50}$10.18\%$&\cellcolor{red!2}$0.35\%$&\cellcolor{red!13}$2.66\%$\\
AT-6&-&$[0,19]$&\cellcolor{blue!3}$-0.52\%$&\cellcolor{red!17}$3.38\%$&\cellcolor{blue!12}$-2.89\%$&\cellcolor{red!2}$0.35\%$\\
AT-7&-&$[0,19]$&\cellcolor{red!9}$1.81\%$&\cellcolor{red!21}$4.25\%$&\cellcolor{blue!20}$-4.86\%$&\cellcolor{blue!1}$-0.14\%$\\
AT-8&-&$[0,19]$&\cellcolor{blue!17}$-4.01\%$&\cellcolor{red!2}$0.33\%$&\cellcolor{blue!16}$-3.88\%$&\cellcolor{red!2}$0.34\%$\\
AT-9&-&$[0,19]$&\cellcolor{blue!5}$-1.03\%$&\cellcolor{red!16}$3.14\%$&\cellcolor{red!6}$1.21\%$&\cellcolor{red!16}$3.21\%$\\
AT-1&-&$[20,39]$&\cellcolor{blue!29}$-7.14\%$&\cellcolor{red!21}$4.34\%$&\cellcolor{blue!16}$-3.87\%$&\cellcolor{blue!2}$-0.44\%$\\
AT-2&-&$[20,39]$&\cellcolor{red!76}$15.65\%$&\cellcolor{red!100}$20.71\%$&\cellcolor{blue!4}$-0.77\%$&\cellcolor{red!3}$0.42\%$\\
AT-3&-&$[20,39]$&\cellcolor{blue!21}$-5.22\%$&\cellcolor{red!20}$4.01\%$&\cellcolor{blue!14}$-3.44\%$&\cellcolor{blue!2}$-0.29\%$\\
AT-4&-&$[20,39]$&\cellcolor{red!21}$4.22\%$&\cellcolor{red!35}$7.05\%$&\cellcolor{blue!13}$-3.20\%$&\cellcolor{blue!1}$-0.14\%$\\
AT-5&-&$[20,39]$&\cellcolor{red!46}$9.43\%$&\cellcolor{red!61}$12.58\%$&\cellcolor{blue!13}$-3.16\%$&\cellcolor{red!3}$0.42\%$\\
AT-6&-&$[20,39]$&\cellcolor{blue!4}$-0.80\%$&\cellcolor{red!12}$2.36\%$&\cellcolor{blue!15}$-3.61\%$&\cellcolor{red!1}$0.10\%$\\
AT-7&-&$[20,39]$&\cellcolor{red!5}$0.99\%$&\cellcolor{red!17}$3.40\%$&\cellcolor{blue!12}$-2.96\%$&\cellcolor{red!1}$0.18\%$\\
AT-8&-&$[20,39]$&\cellcolor{blue!14}$-3.48\%$&\cellcolor{blue!3}$-0.51\%$&\cellcolor{blue!8}$-1.78\%$&\cellcolor{blue!1}$-0.16\%$\\
AT-9&-&$[20,39]$&\cellcolor{blue!43}$-10.73\%$&\cellcolor{blue!40}$-9.84\%$&\cellcolor{blue!1}$-0.19\%$&\cellcolor{red!7}$1.37\%$\\
AT-1&-&$[40,59]$&\cellcolor{blue!74}$-18.26\%$&\cellcolor{blue!29}$-7.06\%$&\cellcolor{blue!18}$-4.42\%$&\cellcolor{red!5}$0.89\%$\\
AT-2&-&$[40,59]$&\cellcolor{red!23}$4.71\%$&\cellcolor{red!47}$9.57\%$&\cellcolor{blue!11}$-2.69\%$&\cellcolor{blue!2}$-0.40\%$\\
AT-3&-&$[40,59]$&\cellcolor{blue!61}$-15.20\%$&\cellcolor{blue!39}$-9.59\%$&\cellcolor{blue!11}$-2.57\%$&\cellcolor{red!2}$0.27\%$\\
AT-4&-&$[40,59]$&\cellcolor{red!13}$2.52\%$&\cellcolor{red!31}$6.29\%$&\cellcolor{blue!12}$-2.85\%$&\cellcolor{red!4}$0.63\%$\\
AT-5&-&$[40,59]$&\cellcolor{red!24}$4.90\%$&\cellcolor{red!56}$11.45\%$&\cellcolor{blue!5}$-1.11\%$&\cellcolor{red!4}$0.66\%$\\
AT-6&-&$[40,59]$&\cellcolor{blue!12}$-2.75\%$&\cellcolor{red!8}$1.51\%$&\cellcolor{blue!15}$-3.61\%$&\cellcolor{blue!2}$-0.39\%$\\
AT-7&-&$[40,59]$&\cellcolor{red!10}$2.05\%$&\cellcolor{red!23}$4.74\%$&\cellcolor{blue!11}$-2.62\%$&\cellcolor{blue!1}$-0.06\%$\\
AT-8&-&$[40,59]$&\cellcolor{blue!18}$-4.27\%$&\cellcolor{red!1}$0.07\%$&\cellcolor{blue!17}$-4.16\%$&\cellcolor{blue!1}$-0.05\%$\\
AT-9&-&$[40,59]$&\cellcolor{blue!4}$-0.77\%$&\cellcolor{red!13}$2.59\%$&\cellcolor{blue!8}$-1.91\%$&\cellcolor{blue!3}$-0.65\%$\\
AT-1&-&$[60,79]$&\cellcolor{blue!69}$-17.06\%$&\cellcolor{blue!52}$-12.91\%$&\cellcolor{blue!1}$-0.05\%$&\cellcolor{red!23}$4.58\%$\\
AT-2&-&$[60,79]$&\cellcolor{blue!12}$-2.84\%$&\cellcolor{red!8}$1.53\%$&\cellcolor{blue!6}$-1.31\%$&\cellcolor{blue!1}$-0.20\%$\\
AT-3&-&$[60,79]$&\cellcolor{blue!47}$-11.56\%$&\cellcolor{blue!27}$-6.50\%$&\cellcolor{red!1}$0.09\%$&\cellcolor{red!22}$4.36\%$\\
AT-4&-&$[60,79]$&\cellcolor{red!1}$0.12\%$&\cellcolor{red!31}$6.26\%$&\cellcolor{red!1}$0.00\%$&\cellcolor{red!12}$2.40\%$\\
AT-5&-&$[60,79]$&\cellcolor{red!14}$2.70\%$&\cellcolor{red!45}$9.27\%$&\cellcolor{blue!2}$-0.29\%$&\cellcolor{red!11}$2.24\%$\\
AT-6&-&$[60,79]$&\cellcolor{blue!9}$-2.15\%$&\cellcolor{red!1}$0.20\%$&\cellcolor{blue!2}$-0.41\%$&\cellcolor{red!8}$1.47\%$\\
AT-7&-&$[60,79]$&\cellcolor{red!8}$1.62\%$&\cellcolor{red!30}$6.09\%$&\cellcolor{blue!2}$-0.33\%$&\cellcolor{red!7}$1.29\%$\\
AT-8&-&$[60,79]$&\cellcolor{red!4}$0.76\%$&\cellcolor{red!13}$2.49\%$&\cellcolor{blue!2}$-0.37\%$&\cellcolor{red!1}$0.14\%$\\
AT-9&-&$[60,79]$&\cellcolor{red!38}$7.86\%$&\cellcolor{red!61}$12.50\%$&\cellcolor{blue!3}$-0.57\%$&\cellcolor{blue!1}$-0.07\%$\\
AT-1&-&$80^+$&\cellcolor{blue!79}$-19.53\%$&\cellcolor{blue!37}$-9.04\%$&\cellcolor{red!2}$0.38\%$&\cellcolor{red!10}$1.91\%$\\
AT-2&-&$80^+$&\cellcolor{blue!19}$-4.50\%$&\cellcolor{blue!6}$-1.41\%$&\cellcolor{blue!6}$-1.27\%$&\cellcolor{blue!1}$-0.01\%$\\
AT-3&-&$80^+$&\cellcolor{blue!46}$-11.33\%$&\cellcolor{blue!25}$-6.18\%$&\cellcolor{blue!2}$-0.45\%$&\cellcolor{red!9}$1.69\%$\\
AT-4&-&$80^+$&\cellcolor{blue!2}$-0.26\%$&\cellcolor{red!2}$0.30\%$&\cellcolor{blue!3}$-0.56\%$&\cellcolor{red!1}$0.06\%$\\
AT-5&-&$80^+$&\cellcolor{red!5}$0.86\%$&\cellcolor{red!25}$5.03\%$&\cellcolor{blue!1}$-0.07\%$&\cellcolor{red!8}$1.64\%$\\
AT-6&-&$80^+$&\cellcolor{blue!14}$-3.26\%$&\cellcolor{blue!5}$-1.03\%$&\cellcolor{blue!1}$-0.05\%$&\cellcolor{red!2}$0.40\%$\\
AT-7&-&$80^+$&\cellcolor{red!6}$1.17\%$&\cellcolor{red!18}$3.55\%$&\cellcolor{blue!1}$-0.22\%$&\cellcolor{red!7}$1.28\%$\\
AT-8&-&$80^+$&\cellcolor{red!3}$0.56\%$&\cellcolor{red!14}$2.82\%$&\cellcolor{blue!3}$-0.62\%$&\cellcolor{red!4}$0.81\%$\\
AT-9&-&$80^+$&\cellcolor{red!36}$7.34\%$&\cellcolor{red!94}$19.45\%$&\cellcolor{blue!9}$-2.03\%$&\cellcolor{red!7}$1.45\%$\\
\end{tabular}
\end{scriptsize}
\end{center}
    \caption{Relative maximum differences between the total population forecast (Source \ref{src:popForecast} and the simulations SC1 and SC2 between 2025 and 2050.}
    \label{tbl:deviations2}
\end{table}
\FloatBarrier

\subsection{Births and Deaths}
\label{sec:validation_birth_deaths}
In the next step, we will compare total births and deaths. For both, we utilise Source \ref{src:indicators}.

\subsubsection{Comparison with Source \ref{src:indicators} (Demographische Zeitreihenindikatoren) - Births}
\begin{figure}
    \centering
    \includegraphics[width=0.8\linewidth]{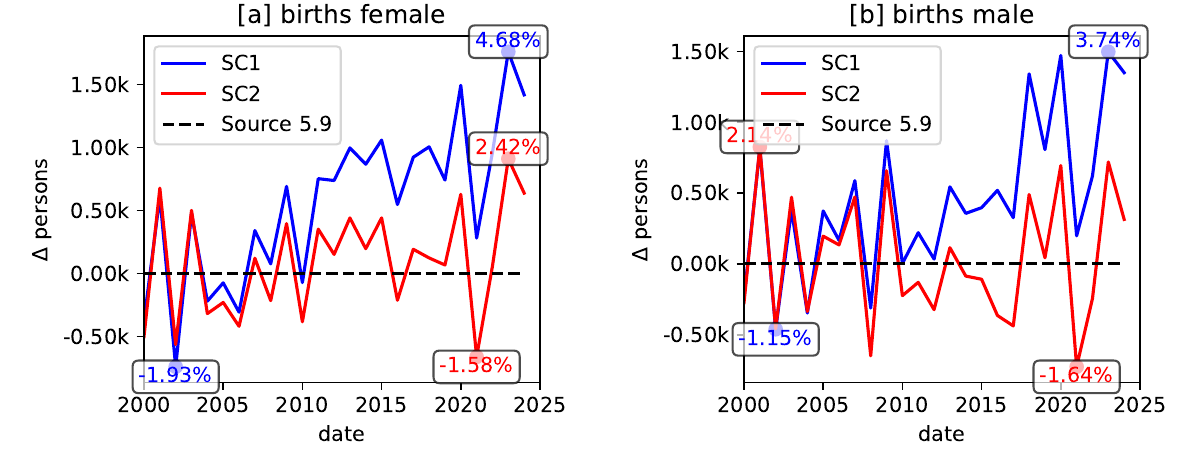}
    \caption{Differences between the male and female births (Source \ref{src:indicators}) and the two simulation scenarios SC1 and SC2.}
    \label{fig:births_sex_diff}
\end{figure}

\begin{figure}
    \centering
    \includegraphics[width=\linewidth]{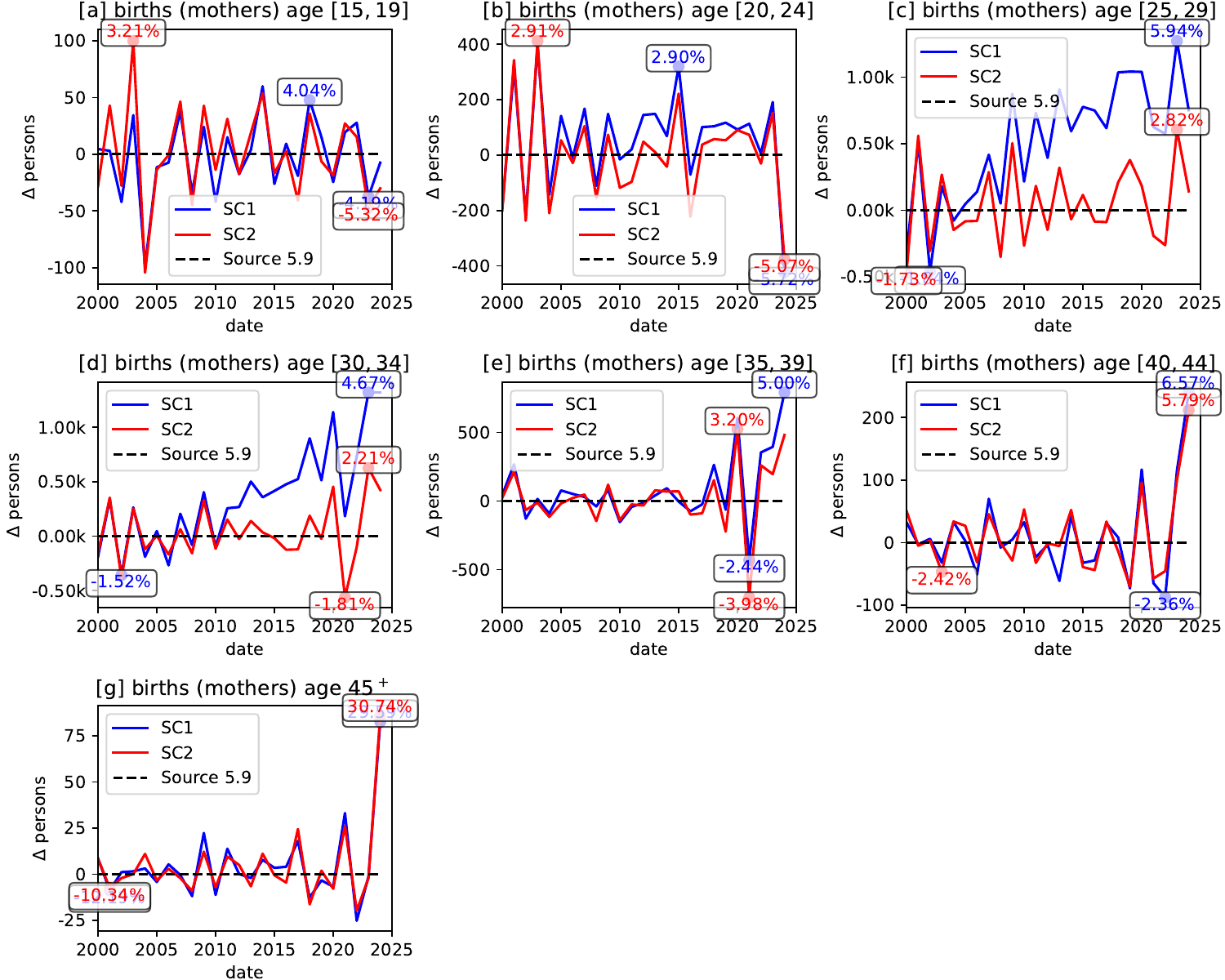}
    \caption{Differences between the total births (Source \ref{src:indicators}) and the two simulation scenarios SC1 and SC2 after aggregation to Austria and 5-year age classes of the mother.}
    \label{fig:births_age_diff}
\end{figure}

Source \ref{src:indicators} provides a reference for births in the time-frame between 2000 and 2024. Table 1 in Source \ref{src:indicators} contains the total number of births per sex of the child and per age of the mother on the federal-state level.
\paragraph{Total and sex.} As seen in Figure \ref{fig:births_sex_diff}, the births of male and female children in scenarios SC1 and SC2 remain within a $\pm 4.7\%$ range around the given data, with the highest deviation for year 2022. In absolute numbers, this corresponds to around 1700 newborn children per year. SC2 performs slightly better than SC1.
\paragraph{Age-classes.} 
Figure \ref{fig:births_age_diff} shows, how the births correspond to the age of the mother. An unexpectedly large deviation can be observed for the last simulation year 2024, in particular for the $45^+$ age class with roughly $30\%$. This is due to a problem with computing the age-dependent forecast (see Section \ref{sec:calculation_birth}): The Gauss distribution used for the forecast is fully symmetric, whereas the actual age-distribution between 2015 and 2024 is skewed to the right (see Figure \ref{fig:gauss_fit}). As a result, the naive birthrates forecast for 2025 is too high for the older age classes and too low for the younger ones. Due to technical reasons (compare with Corollary \ref{cor:farr_param}), this also influences the final probabilities for 2024.
\paragraph{Federal-states.} 
Table \ref{tbl:deviations_births} shows summary of the maximum offsets from SC1 and SC2 compared to the total births from Source \ref{src:indicators}. Comparing the federalstates, in particular model SC1 shows high differences due to mentioned role of internal migration for those regions which particularly impacts young families. A figure analogous to the ones for comparing age-classes and sex can be found in the Appendix section (Figure \ref{fig:births_fed_diff}).

\begin{table}
    \begin{center}
    \begin{scriptsize}
        \begin{tabular}{ccc|cc|cc}
        region & sex & age (mother) & \multicolumn{2}{c|}{\textbf{SC1} ($e_{min},e_{max}$)} & \multicolumn{2}{c}{\textbf{SC2} ($e_{min},e_{max}$)}\\
        \hline
-&-&-&\cellcolor{blue!9}$-1.53\%$&\cellcolor{red!21}$4.20\%$&\cellcolor{blue!9}$-1.61\%$&\cellcolor{red!11}$2.10\%$\\
\hline
-&female&-&\cellcolor{blue!11}$-1.93\%$&\cellcolor{red!24}$4.68\%$&\cellcolor{blue!9}$-1.58\%$&\cellcolor{red!13}$2.42\%$\\
-&male&-&\cellcolor{blue!7}$-1.15\%$&\cellcolor{red!19}$3.74\%$&\cellcolor{blue!9}$-1.64\%$&\cellcolor{red!11}$2.14\%$\\
\hline
\hline
-&-&$[15,19]$&\cellcolor{blue!23}$-4.19\%$&\cellcolor{red!21}$4.04\%$&\cellcolor{blue!29}$-5.32\%$&\cellcolor{red!17}$3.21\%$\\
-&-&$[20,24]$&\cellcolor{blue!31}$-5.72\%$&\cellcolor{red!15}$2.90\%$&\cellcolor{blue!28}$-5.07\%$&\cellcolor{red!15}$2.91\%$\\
-&-&$[25,29]$&\cellcolor{blue!10}$-1.84\%$&\cellcolor{red!30}$5.94\%$&\cellcolor{blue!10}$-1.73\%$&\cellcolor{red!15}$2.82\%$\\
-&-&$[30,34]$&\cellcolor{blue!9}$-1.52\%$&\cellcolor{red!24}$4.67\%$&\cellcolor{blue!10}$-1.81\%$&\cellcolor{red!12}$2.21\%$\\
-&-&$[35,39]$&\cellcolor{blue!14}$-2.44\%$&\cellcolor{red!26}$5.00\%$&\cellcolor{blue!22}$-3.98\%$&\cellcolor{red!17}$3.20\%$\\
-&-&$[40,44]$&\cellcolor{blue!13}$-2.36\%$&\cellcolor{red!33}$6.57\%$&\cellcolor{blue!13}$-2.42\%$&\cellcolor{red!29}$5.79\%$\\
-&-&$45^+$&\cellcolor{blue!66}$-12.19\%$&\cellcolor{red!100}$29.59\%$&\cellcolor{blue!56}$-10.34\%$&\cellcolor{red!100}$30.74\%$\\
\hline
AT-1&-&-&\cellcolor{blue!1}$-0.15\%$&\cellcolor{red!63}$12.50\%$&\cellcolor{blue!14}$-2.58\%$&\cellcolor{red!38}$7.44\%$\\
AT-2&-&-&\cellcolor{red!3}$0.53\%$&\cellcolor{red!100}$20.65\%$&\cellcolor{blue!16}$-2.98\%$&\cellcolor{red!15}$2.83\%$\\
AT-3&-&-&\cellcolor{blue!16}$-2.83\%$&\cellcolor{red!35}$6.83\%$&\cellcolor{blue!12}$-2.15\%$&\cellcolor{red!16}$3.02\%$\\
AT-4&-&-&\cellcolor{blue!7}$-1.15\%$&\cellcolor{red!57}$11.38\%$&\cellcolor{blue!11}$-1.99\%$&\cellcolor{red!16}$3.17\%$\\
AT-5&-&-&\cellcolor{blue!8}$-1.37\%$&\cellcolor{red!86}$17.10\%$&\cellcolor{blue!8}$-1.49\%$&\cellcolor{red!39}$7.67\%$\\
AT-6&-&-&\cellcolor{blue!7}$-1.17\%$&\cellcolor{red!37}$7.35\%$&\cellcolor{blue!14}$-2.51\%$&\cellcolor{red!22}$4.23\%$\\
AT-7&-&-&\cellcolor{blue!8}$-1.50\%$&\cellcolor{red!26}$5.13\%$&\cellcolor{blue!16}$-2.89\%$&\cellcolor{red!15}$2.83\%$\\
AT-8&-&-&\cellcolor{blue!7}$-1.18\%$&\cellcolor{red!29}$5.66\%$&\cellcolor{blue!12}$-2.20\%$&\cellcolor{red!18}$3.59\%$\\
AT-9&-&-&\cellcolor{blue!73}$-13.68\%$&\cellcolor{red!10}$1.94\%$&\cellcolor{blue!23}$-4.23\%$&\cellcolor{red!18}$3.56\%$\\
\end{tabular}
\end{scriptsize}
\end{center}
    \caption{Relative maximum differences between the birth data (Source \ref{src:indicators}, Table 1) and the simulations SC1 and SC2 between 2000 and 2024.}
    \label{tbl:deviations_births}
\end{table}

\subsubsection{Comparison with Source \ref{src:indicators} (Demographische Zeitreihenindikatoren) - Deaths}
\begin{figure}
    \centering
    \includegraphics[width=0.8\linewidth]{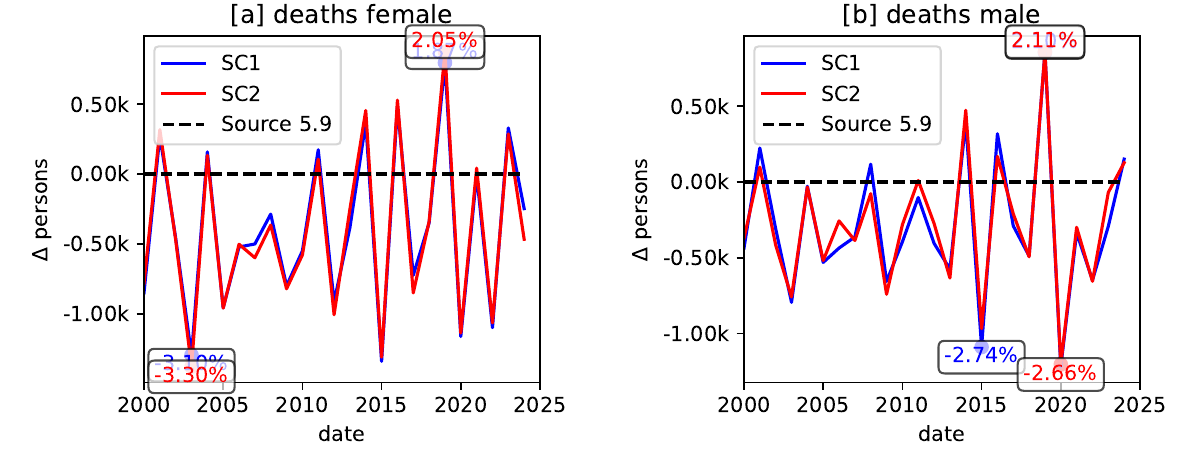}
    \caption{Differences between the male and female deaths (Source \ref{src:indicators}) and the two simulation scenarios SC1 and SC2.}
    \label{fig:deaths_sex_diff}
\end{figure}

Source \ref{src:indicators} provides a reference for deaths in the time-frame between 2000 and 2024. Table 3 in Source \ref{src:indicators} contains the total number of deaths per sex and age on the federalstate level.
\paragraph{Total and sex.} As seen in Figure \ref{fig:deaths_sex_diff}, the male and female deaths in scenarios SC1 and SC2 remain within a $\pm 3.5\%$ range around the given data. In absolute numbers, this corresponds to around 1000 persons per year. No obvious difference between SC1 and SC2 can be found. In general, deaths seem to be slightly underestimated.

\paragraph{Age-classes and federal-states.}
With respect to age (see Table \ref{tbl:deviations_death}), the models lie within $\pm6\%$ around the reference data for 20-year age-classes. With respect to federal-states, both models perform similarly well, with the exception of AT-1 and AT-9, where, again, the internal migration effect seems to be important for the validity of the model.
Result plots are found in the Appendix (Figures \ref{fig:deaths_age_diff} and \ref{fig:deaths_fed_diff}).

\begin{table}
    \begin{center}
    \begin{scriptsize}
        \begin{tabular}{ccc|cc|cc}
        region & sex & age & \multicolumn{2}{c|}{\textbf{SC1} ($e_{min},e_{max}$)} & \multicolumn{2}{c}{\textbf{SC2} ($e_{min},e_{max}$)}\\
        \hline
-&-&-&\cellcolor{blue!23}$-2.93\%$&\cellcolor{red!21}$1.98\%$&\cellcolor{blue!22}$-2.77\%$&\cellcolor{red!22}$2.08\%$\\
\hline
-&female&-&\cellcolor{blue!25}$-3.10\%$&\cellcolor{red!20}$1.87\%$&\cellcolor{blue!26}$-3.30\%$&\cellcolor{red!22}$2.05\%$\\
-&male&-&\cellcolor{blue!22}$-2.74\%$&\cellcolor{red!22}$2.10\%$&\cellcolor{blue!21}$-2.66\%$&\cellcolor{red!22}$2.11\%$\\
\hline
-&-&$[0,19]$&\cellcolor{blue!22}$-2.74\%$&\cellcolor{red!60}$5.70\%$&\cellcolor{blue!14}$-1.70\%$&\cellcolor{red!54}$5.19\%$\\
-&-&$[20,39]$&\cellcolor{blue!28}$-3.54\%$&\cellcolor{red!45}$4.26\%$&\cellcolor{blue!32}$-3.99\%$&\cellcolor{red!45}$4.30\%$\\
-&-&$[40,59]$&\cellcolor{blue!18}$-2.28\%$&\cellcolor{red!22}$2.02\%$&\cellcolor{blue!19}$-2.39\%$&\cellcolor{red!22}$2.07\%$\\
-&-&$[60,79]$&\cellcolor{blue!18}$-2.24\%$&\cellcolor{red!23}$2.16\%$&\cellcolor{blue!16}$-1.95\%$&\cellcolor{red!24}$2.24\%$\\
-&-&$80^+$&\cellcolor{blue!33}$-4.13\%$&\cellcolor{red!20}$1.92\%$&\cellcolor{blue!31}$-3.92\%$&\cellcolor{red!23}$2.14\%$\\
\hline
AT-1&-&-&\cellcolor{blue!80}$-10.27\%$&\cellcolor{red!10}$0.92\%$&\cellcolor{blue!26}$-3.26\%$&\cellcolor{red!35}$3.34\%$\\
AT-2&-&-&\cellcolor{blue!40}$-5.13\%$&\cellcolor{red!22}$2.03\%$&\cellcolor{blue!35}$-4.37\%$&\cellcolor{red!29}$2.76\%$\\
AT-3&-&-&\cellcolor{blue!58}$-7.38\%$&\cellcolor{red!9}$0.83\%$&\cellcolor{blue!19}$-2.41\%$&\cellcolor{red!18}$1.65\%$\\
AT-4&-&-&\cellcolor{blue!30}$-3.78\%$&\cellcolor{red!28}$2.65\%$&\cellcolor{blue!31}$-3.94\%$&\cellcolor{red!29}$2.70\%$\\
AT-5&-&-&\cellcolor{blue!36}$-4.62\%$&\cellcolor{red!24}$2.23\%$&\cellcolor{blue!25}$-3.19\%$&\cellcolor{red!19}$1.75\%$\\
AT-6&-&-&\cellcolor{blue!45}$-5.72\%$&\cellcolor{red!18}$1.72\%$&\cellcolor{blue!41}$-5.14\%$&\cellcolor{red!30}$2.85\%$\\
AT-7&-&-&\cellcolor{blue!25}$-3.11\%$&\cellcolor{red!21}$1.98\%$&\cellcolor{blue!21}$-2.64\%$&\cellcolor{red!19}$1.76\%$\\
AT-8&-&-&\cellcolor{blue!27}$-3.45\%$&\cellcolor{red!44}$4.22\%$&\cellcolor{blue!33}$-4.14\%$&\cellcolor{red!37}$3.53\%$\\
AT-9&-&-&\cellcolor{blue!11}$-1.35\%$&\cellcolor{red!100}$9.61\%$&\cellcolor{blue!25}$-3.08\%$&\cellcolor{red!32}$2.99\%$\\
\end{tabular}
\end{scriptsize}
\end{center}
    \caption{Relative maximum differences between the death data (Source \ref{src:indicators}, Table 3) and the simulations SC1 and SC2 between 2000 and 2024.}
    \label{tbl:deviations_death}
\end{table}

\FloatBarrier
\subsubsection{Comparison with Source \ref{src:migrationForeacst} (Bevölkerungsbewegung 1961 bis 2100)}
\begin{figure}
    \centering
    \includegraphics[width=\linewidth]{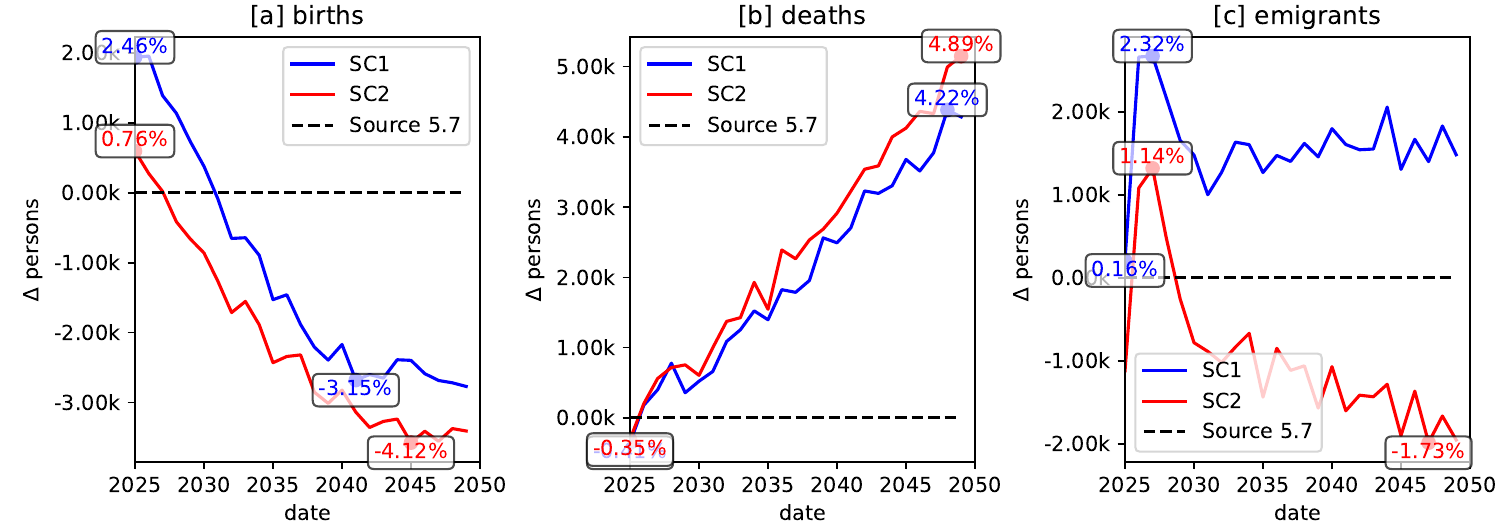}
    \caption{Differences between the total births deaths and emigrants from Source \ref{src:migrationForeacst} and the two simulation scenarios SC1 and SC2.}
    \label{fig:allthree_fc_diff}
\end{figure}

To validate the model beyond 2024, we investigate the migration forecast Source \ref{src:migrationForeacst}. The forecast includes a total number of births and deaths per year and federalstate until 2100, without any further age and sex resolution.
\paragraph{Total.} Figure \ref{fig:allthree_fc_diff}, panels [a] and [b], show a comparison between the total births and deaths from Source \ref{src:migrationForeacst} and SC1 and SC2. For births and deaths, both models perform roughly equally well and the error is within $\pm5\%$ around the reference. Births tend to become underestimated, deaths trend towards overestimation.

\paragraph{Federalstates.} 
Focusing on federalstates, the model with internal migration results is more valid with respect to births and deaths than the one without. The highest deviations are found on the end of the simulation period, indicating that certain errors are dynamically accumulating when internal migration is not considered. A summary of the differences is given in Table \ref{tbl:deviations_birthdeath}. Figures showing the differences per federal-state are found in the Appendix (Figures \ref{fig:births_fc_fed_diff} and \ref{fig:deaths_fc_fed_diff}).

\begin{table}
    \begin{center}
    \begin{scriptsize}
        \begin{tabular}{ccc|cc|cc}
        region & sex & age & \multicolumn{2}{c|}{\textbf{SC1} ($e_{min},e_{max}$)} & \multicolumn{2}{c}{\textbf{SC2} ($e_{min},e_{max}$)}\\
        \hline
        \multicolumn{7}{c}{births}\\
        \hline
-&-&-&\cellcolor{blue!21}$-3.15\%$&\cellcolor{red!15}$2.46\%$&\cellcolor{blue!27}$-4.12\%$&\cellcolor{red!5}$0.76\%$\\
\hline
AT-1&-&-&\cellcolor{blue!80}$-12.57\%$&\cellcolor{red!36}$5.98\%$&\cellcolor{blue!29}$-4.41\%$&\cellcolor{red!6}$0.97\%$\\
AT-2&-&-&\cellcolor{red!67}$11.38\%$&\cellcolor{red!100}$17.07\%$&\cellcolor{blue!21}$-3.22\%$&\cellcolor{red!1}$0.17\%$\\
AT-3&-&-&\cellcolor{blue!65}$-10.14\%$&\cellcolor{red!28}$4.70\%$&\cellcolor{blue!22}$-3.31\%$&\cellcolor{red!11}$1.75\%$\\
AT-4&-&-&\cellcolor{red!16}$2.57\%$&\cellcolor{red!55}$9.26\%$&\cellcolor{blue!28}$-4.38\%$&\cellcolor{red!12}$1.95\%$\\
AT-5&-&-&\cellcolor{red!34}$5.79\%$&\cellcolor{red!71}$11.98\%$&\cellcolor{red!11}$1.87\%$&\cellcolor{red!30}$4.98\%$\\
AT-6&-&-&\cellcolor{blue!12}$-1.79\%$&\cellcolor{red!22}$3.60\%$&\cellcolor{blue!27}$-4.21\%$&\cellcolor{red!4}$0.60\%$\\
AT-7&-&-&\cellcolor{blue!3}$-0.43\%$&\cellcolor{red!24}$4.02\%$&\cellcolor{blue!38}$-5.87\%$&\cellcolor{blue!2}$-0.25\%$\\
AT-8&-&-&\cellcolor{blue!43}$-6.66\%$&\cellcolor{red!8}$1.29\%$&\cellcolor{blue!53}$-8.28\%$&\cellcolor{blue!6}$-0.79\%$\\
AT-9&-&-&\cellcolor{blue!73}$-11.35\%$&\cellcolor{blue!59}$-9.14\%$&\cellcolor{blue!40}$-6.19\%$&\cellcolor{blue!3}$-0.39\%$\\
\hline
\multicolumn{7}{c}{deaths}\\
 \hline
-&-&-&\cellcolor{blue!3}$-0.41\%$&\cellcolor{red!24}$4.22\%$&\cellcolor{blue!3}$-0.35\%$&\cellcolor{red!27}$4.89\%$\\
\hline
AT-1&-&-&\cellcolor{blue!80}$-14.14\%$&\cellcolor{blue!45}$-7.86\%$&\cellcolor{blue!2}$-0.22\%$&\cellcolor{red!34}$6.23\%$\\
AT-2&-&-&\cellcolor{blue!12}$-2.09\%$&\cellcolor{red!10}$1.79\%$&\cellcolor{blue!6}$-1.03\%$&\cellcolor{red!20}$3.66\%$\\
AT-3&-&-&\cellcolor{blue!43}$-7.53\%$&\cellcolor{blue!33}$-5.72\%$&\cellcolor{blue!3}$-0.44\%$&\cellcolor{red!35}$6.29\%$\\
AT-4&-&-&\cellcolor{blue!2}$-0.32\%$&\cellcolor{red!32}$5.69\%$&\cellcolor{blue!1}$-0.08\%$&\cellcolor{red!27}$4.86\%$\\
AT-5&-&-&\cellcolor{red!5}$0.88\%$&\cellcolor{red!61}$11.10\%$&\cellcolor{blue!8}$-1.34\%$&\cellcolor{red!40}$7.18\%$\\
AT-6&-&-&\cellcolor{blue!11}$-1.94\%$&\cellcolor{red!15}$2.62\%$&\cellcolor{blue!7}$-1.19\%$&\cellcolor{red!28}$5.11\%$\\
AT-7&-&-&\cellcolor{red!5}$0.86\%$&\cellcolor{red!49}$8.87\%$&\cellcolor{blue!1}$-0.06\%$&\cellcolor{red!33}$5.88\%$\\
AT-8&-&-&\cellcolor{blue!1}$-0.11\%$&\cellcolor{red!48}$8.62\%$&\cellcolor{blue!1}$-0.17\%$&\cellcolor{red!36}$6.53\%$\\
AT-9&-&-&\cellcolor{red!41}$7.49\%$&\cellcolor{red!100}$18.32\%$&\cellcolor{blue!11}$-1.94\%$&\cellcolor{red!16}$2.88\%$\\
\end{tabular}
\end{scriptsize}
\end{center}
    \caption{Relative maximum differences between the birth and death forecast (Source \ref{src:migrationForeacst}) and the simulations SC1 and SC2 between 2000 and 2049.}
    \label{tbl:deviations_birthdeath}
\end{table}

\FloatBarrier

\subsection{Emigrants and Immigrants}
\label{sec:validation_emigrants_immigrants}
With respect to external migration, we only need to validate emigrants. Immigrants are sampled into the model based on processed parameter data, which matches the given raw data on the respective level of aggregation (see calculation of immigration data, Section \ref{sec:calculation_immigration}). Therefore, alike the initial population, immigrant counts in the model cannot differ from the parameter table (as long as the model's implementation is correct).

As before, we validate the time period before and after 2024 separately from each other.

\subsubsection{Comparison with Source \ref{src:migrationCountry} (Wanderungen mit dem Ausland ab 2002 nach Alter, Geschlecht und Staatsangehörigkeit)}
\begin{figure}
    \centering
    \includegraphics[width=0.9\linewidth]{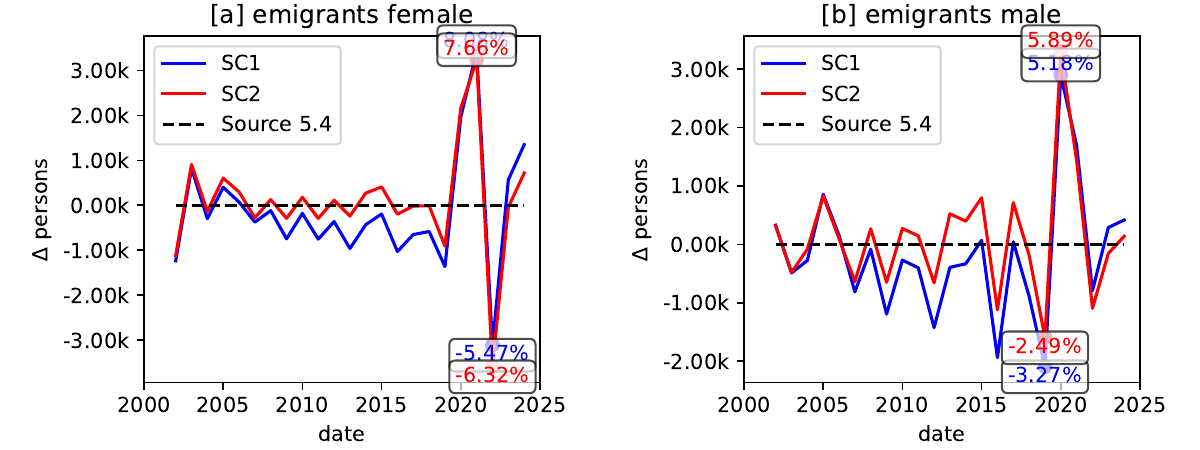}
    \caption{Differences between the male and female emigrants (Source \ref{src:migrationCountry}) and the two simulation scenarios SC1 and SC2.}
    \label{fig:emigrants_sex_diff}
\end{figure}

\begin{figure}
\centering
\includegraphics[width=0.9\linewidth]{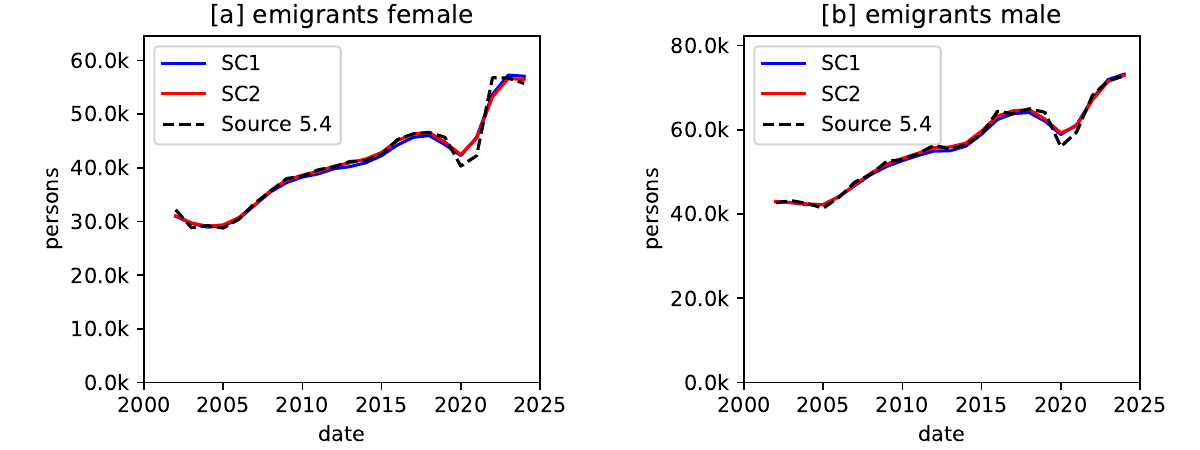}
\caption{Comparison between emigrant data from Source \ref{src:migrationCountry} and the simulation scenarios SC1 and SC2 for male and female persons.}
\label{fig:emigrants_sex}
\end{figure}

 Source \ref{src:migrationCountry} contains migration data with sex and single age resolution for the whole of Austria from 2002 to 2024. Hence, we use it to evaluate the simulation results on the country level w.r. sex and age-classes. 
 \paragraph{Total and sex.} As indicated by Figure \ref{fig:emigrants_sex_diff}, emigrants is the highest fluctuating quantity in the model with deviations of up to $\pm 8\%$ per sex. Corresponding large errors, however, are only found between 2019 and 2023. Before and after this period, the offsets are in the $\pm 2\%$ region. Having a look at the total numbers, Figure \ref{fig:emigrants_sex} reveals that the comparably high errors originate from a temporary smoothing effect of the model. Hence, the sudden decline of the numbers during the COVID pandemic could not be depicted properly. On average, SC1 seems to slightly underestimate the quantity.
 
 \paragraph{Age-classes.}
 Table \ref{tbl:deviations_emigrants_1} indicates that the emigrants for high age cohorts $80^+$ stand out with huge errors. The origin of this problem is, again, that emigration itself highly fluctuates between years which are smoothed by the model. Considering the low number of elderly emigrants, we do not consider this problem as very severe.

\begin{table}
    \begin{center}
    \begin{scriptsize}
        \begin{tabular}{ccc|cc|cc}
        region & sex & age & \multicolumn{2}{c|}{\textbf{SC1} ($e_{min},e_{max}$)} & \multicolumn{2}{c}{\textbf{SC2} ($e_{min},e_{max}$)}\\
-&-&-&\cellcolor{blue!10}$-3.15\%$&\cellcolor{red!11}$5.05\%$&\cellcolor{blue!11}$-3.75\%$&\cellcolor{red!13}$5.67\%$\\
\hline
-&female&-&\cellcolor{blue!16}$-5.47\%$&\cellcolor{red!18}$8.08\%$&\cellcolor{blue!19}$-6.32\%$&\cellcolor{red!17}$7.66\%$\\
-&male&-&\cellcolor{blue!10}$-3.27\%$&\cellcolor{red!12}$5.18\%$&\cellcolor{blue!8}$-2.49\%$&\cellcolor{red!13}$5.89\%$\\
\hline
-&-&$[0,19]$&\cellcolor{blue!10}$-3.21\%$&\cellcolor{red!36}$16.53\%$&\cellcolor{blue!21}$-6.93\%$&\cellcolor{red!27}$12.36\%$\\
-&-&$[20,39]$&\cellcolor{blue!19}$-6.51\%$&\cellcolor{red!8}$3.50\%$&\cellcolor{blue!8}$-2.75\%$&\cellcolor{red!12}$5.33\%$\\
-&-&$[40,59]$&\cellcolor{blue!12}$-3.81\%$&\cellcolor{red!20}$9.07\%$&\cellcolor{blue!20}$-6.74\%$&\cellcolor{red!13}$6.08\%$\\
-&-&$[60,79]$&\cellcolor{blue!39}$-13.39\%$&\cellcolor{red!25}$11.40\%$&\cellcolor{blue!38}$-12.81\%$&\cellcolor{red!23}$10.47\%$\\
-&-&$80^+$&\cellcolor{blue!79}$-27.05\%$&\cellcolor{red!100}$46.83\%$&\cellcolor{blue!80}$-27.68\%$&\cellcolor{red!100}$46.80\%$
\end{tabular}
\end{scriptsize}
\end{center}
    \caption{Relative maximum differences between the emigration data (Source \ref{src:migrationCountry}) and the simulations SC1 and SC2 between 2002 and 2024.}
    \label{tbl:deviations_emigrants_1}
\end{table}

\subsubsection{Comparison with Sources \ref{src:migrationMuni_1} and \ref{src:migrationMuni_2} (Wanderungen mit dem Ausland von 2002 bis 2014 (ab 2015) nach Altersgruppen, Gemeinde und
Staatsangehörigkeit)}
\begin{figure}
    \centering
    \includegraphics[width=\linewidth]{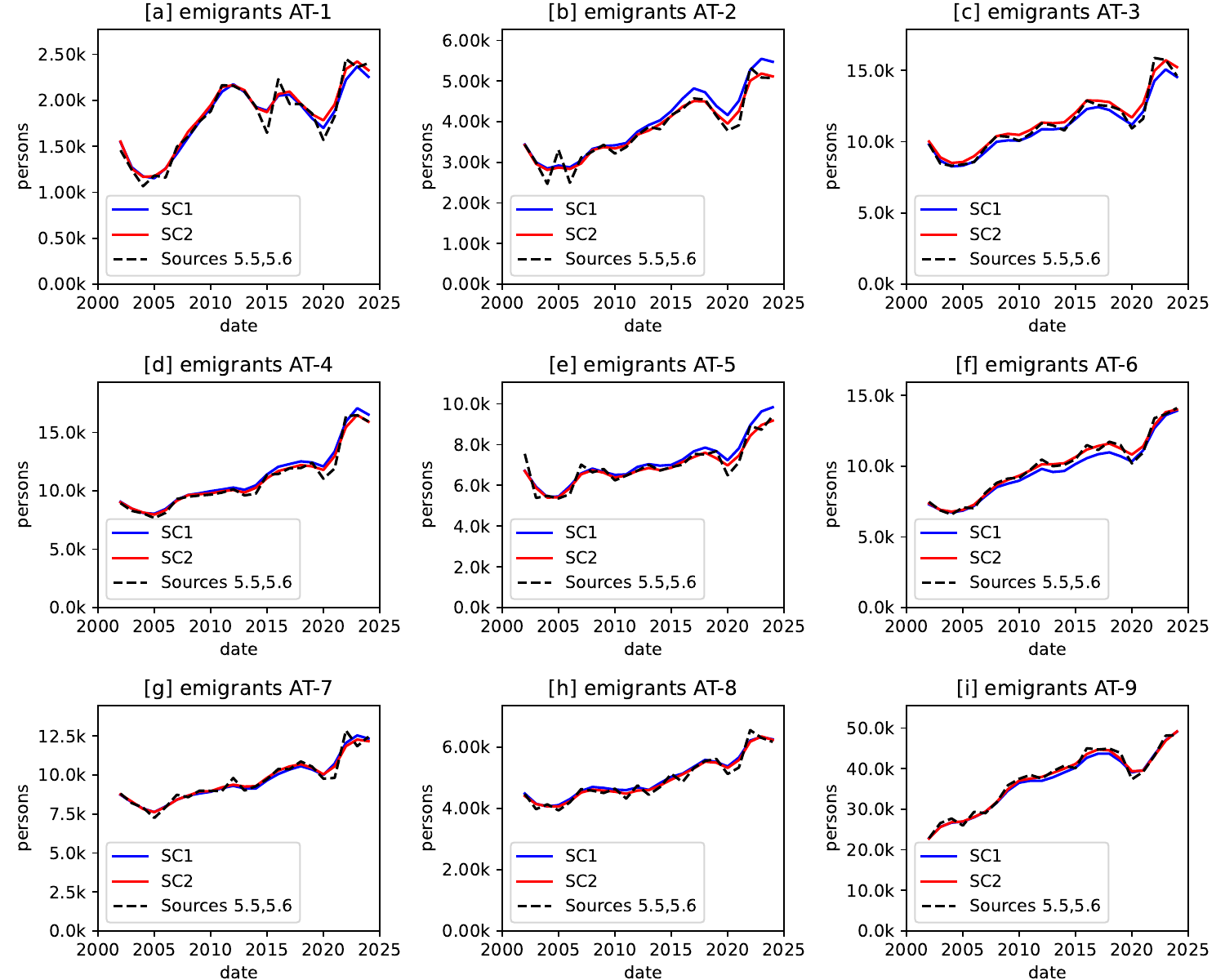}
    \caption{Comparison between the emigrants from Sources \ref{src:migrationMuni_1} and \ref{src:migrationMuni_2} and the two simulation scenarios SC1 and SC2 for all nine federalstates.}
    \label{fig:emigrants_fed}
\end{figure}

Sources \ref{src:migrationMuni_1} and \ref{src:migrationMuni_2} contain total number of migrants per municipality and five year-age classes between 2002 and 2024. We will use this source for validation w.r. to federalstates. As seen in Table \ref{tbl:deviations_emigrants_2}, the maximum relative differences are comparably large (up to $15\%$ in AT-1). The large differences are put into perspective when analysing Figure \ref{fig:emigrants_fed}, which shows the time-series of the emigrations in the model and in the data for the nine federal-states. The charts show how the target value itself fluctuates and how the model depicts the fluctuations. Even if the behaviour corresponds qualitatively to the data, the model reacts less sensitively and fluctuates with lower amplitude. A plot showing the differences, analogous to the ones displayed before, is found in the Appendix (Figure \ref{fig:emigrants_fed_diff}).

\begin{table}
    \begin{center}
    \begin{scriptsize}
        \begin{tabular}{ccc|cc|cc}
        region & sex & age & \multicolumn{2}{c|}{\textbf{SC1} ($e_{min},e_{max}$)} & \multicolumn{2}{c}{\textbf{SC2} ($e_{min},e_{max}$)}\\
AT-1&-&-&\cellcolor{blue!57}$-9.24\%$&\cellcolor{red!96}$14.65\%$&\cellcolor{blue!45}$-7.28\%$&\cellcolor{red!89}$13.65\%$\\
AT-2&-&-&\cellcolor{blue!71}$-11.56\%$&\cellcolor{red!100}$15.38\%$&\cellcolor{blue!80}$-13.09\%$&\cellcolor{red!91}$13.85\%$\\
AT-3&-&-&\cellcolor{blue!62}$-10.11\%$&\cellcolor{red!29}$4.33\%$&\cellcolor{blue!35}$-5.67\%$&\cellcolor{red!61}$9.35\%$\\
AT-4&-&-&\cellcolor{blue!18}$-2.81\%$&\cellcolor{red!80}$12.20\%$&\cellcolor{blue!36}$-5.76\%$&\cellcolor{red!58}$8.84\%$\\
AT-5&-&-&\cellcolor{blue!68}$-11.03\%$&\cellcolor{red!74}$11.34\%$&\cellcolor{blue!69}$-11.21\%$&\cellcolor{red!60}$9.10\%$\\
AT-6&-&-&\cellcolor{blue!50}$-8.09\%$&\cellcolor{red!16}$2.44\%$&\cellcolor{blue!23}$-3.65\%$&\cellcolor{red!40}$6.10\%$\\
AT-7&-&-&\cellcolor{blue!38}$-6.19\%$&\cellcolor{red!62}$9.42\%$&\cellcolor{blue!48}$-7.83\%$&\cellcolor{red!49}$7.51\%$\\
AT-8&-&-&\cellcolor{blue!32}$-5.17\%$&\cellcolor{red!42}$6.37\%$&\cellcolor{blue!36}$-5.77\%$&\cellcolor{red!34}$5.15\%$\\
AT-9&-&-&\cellcolor{blue!32}$-5.18\%$&\cellcolor{red!30}$4.55\%$&\cellcolor{blue!27}$-4.32\%$&\cellcolor{red!35}$5.32\%$
\end{tabular}
\end{scriptsize}
\end{center}
    \caption{Relative maximum differences between the emigration data (Sources \ref{src:migrationMuni_1} and \ref{src:migrationMuni_2}) and the simulations SC1 and SC2 between 2002 and 2024.}
    \label{tbl:deviations_emigrants_2}
\end{table}
\FloatBarrier

\subsubsection{Comparison with Source \ref{src:migrationForeacst} (Bevölkerungsbewegung 1961 bis 2100 nach Bundesland, Bewegungsarten und Szenarien)}
To validate the model beyond 2024, we investigate the migration forecast Source \ref{src:migrationForeacst}. The forecast includes a total number of emigrants per year and federalstate until 2100 without any further age and sex resolution.

\paragraph{Total.} Figure \ref{fig:allthree_fc_diff} panel [c] shows a comparison between the total emigrants from Source \ref{src:migrationForeacst} and SC1 and SC2. Since the open data forecast does not fluctuate so much anymore, also the differences become more stable for the forecast. Interestingly, the model without internal migration (SC1) shows a slightly different trend than the one with internal migration, which underestimates the data slightly. Up to some extent, SC1 can be considered more accurate here. However, looking at the differences on federal-state level (see below) gives a different picture, so that the comparably good fit of SC1 is estimated to be rather coincidental.

\paragraph{Federal-states.} 
For SC1, numbers fluctuate up to $15\%$ around the data (AT-2,AT-1) in the first few forecasting years (2027), thereafter, the maximum offsets decrease to around $4\%$. SC2 performs better with differences in the $3\%$ region. A summary of the differences is given in Table \ref{tbl:deviations_emigrants_3}. Figures showing the differences per federal-state are found in the Appendix (Figure \ref{fig:emigrants_fc_fed_diff}).

\begin{table}
    \begin{center}
    \begin{scriptsize}
        \begin{tabular}{ccc|cc|cc}
        region & sex & age & \multicolumn{2}{c|}{\textbf{SC1} ($e_{min},e_{max}$)} & \multicolumn{2}{c}{\textbf{SC2} ($e_{min},e_{max}$)}\\
-&-&-&\cellcolor{red!2}$0.16\%$&\cellcolor{red!16}$2.32\%$&\cellcolor{blue!11}$-1.73\%$&\cellcolor{red!8}$1.14\%$\\
\hline
AT-1&-&-&\cellcolor{blue!80}$-12.57\%$&\cellcolor{red!7}$0.97\%$&\cellcolor{blue!9}$-1.38\%$&\cellcolor{red!33}$4.78\%$\\
AT-2&-&-&\cellcolor{red!59}$8.69\%$&\cellcolor{red!100}$14.83\%$&\cellcolor{blue!7}$-1.05\%$&\cellcolor{red!18}$2.54\%$\\
AT-3&-&-&\cellcolor{blue!62}$-9.69\%$&\cellcolor{blue!12}$-1.87\%$&\cellcolor{blue!1}$-0.01\%$&\cellcolor{red!22}$3.23\%$\\
AT-4&-&-&\cellcolor{red!28}$4.05\%$&\cellcolor{red!43}$6.34\%$&\cellcolor{blue!9}$-1.40\%$&\cellcolor{red!16}$2.34\%$\\
AT-5&-&-&\cellcolor{red!49}$7.14\%$&\cellcolor{red!67}$9.84\%$&\cellcolor{blue!12}$-1.76\%$&\cellcolor{red!7}$0.99\%$\\
AT-6&-&-&\cellcolor{blue!4}$-0.62\%$&\cellcolor{red!34}$4.91\%$&\cellcolor{blue!8}$-1.15\%$&\cellcolor{red!16}$2.30\%$\\
AT-7&-&-&\cellcolor{red!19}$2.77\%$&\cellcolor{red!42}$6.13\%$&\cellcolor{blue!14}$-2.06\%$&\cellcolor{red!9}$1.29\%$\\
AT-8&-&-&\cellcolor{blue!24}$-3.70\%$&\cellcolor{red!1}$0.13\%$&\cellcolor{blue!13}$-1.89\%$&\cellcolor{red!10}$1.36\%$\\
AT-9&-&-&\cellcolor{blue!17}$-2.65\%$&\cellcolor{red!5}$0.74\%$&\cellcolor{blue!21}$-3.29\%$&\cellcolor{red!1}$0.01\%$
\end{tabular}
\end{scriptsize}
\end{center}
    \caption{Relative maximum differences between the emigration forecast (Source \ref{src:migrationForeacst}) and the simulations SC1 and SC2 between 2025 and 2049.}
    \label{tbl:deviations_emigrants_3}
\end{table}
\FloatBarrier

\subsection{Internal Migrants}
\label{sec:validation_internal_migrants}
In the next step, we will evaluate the validity of internal migration processes. Hereby, we will add two additional model scenarios for GEPOC IM using the two other available IM models, i.e. the biregional and the interregional model. The parametrisation is summarised in Table \ref{tbl:parametrisation_scenarios_im}.

\begin{table}[h]
\begin{center}
    \begin{tabular}{p{4cm}c|cc}
     & & \textbf{SC3} & \textbf{SC4} \\
     \hline
internal migration model & & biregional & interregional \\
regional-level for internal migration & $r_{ie}$ & districts\_districts &districts\_districts\\
internal emigration probabilities & $\hat{IE}^p$ & P.V. \ref{param:ie} & P.V. \ref{param:ie}\\
internal migrants & $\hat{II}$/$\hat{OD}$ & P.V. \ref{param:ii} & P.V. \ref{param:od}\\
\end{tabular}
\end{center}
\caption{Parametrisation setup for simulation scenarios SC3 and SC4. All other parameters are equal to the SC2 scenario.}
\label{tbl:parametrisation_scenarios_im}
\end{table}
\subsubsection{Comparison with Source \ref{src:popBase} (Bevölkerungsstand) and \ref{src:popForecast} (Bevölkerung zum Jahresanfang 1952 bis 2101)}
Before investigating the impact of the three internal migration models, we first of all investigate if SC3 and SC4 properly depict the population.

Table \ref{tbl:deviations_im_pop} shows the differences between the model results of SC2-SC4 compared to the population data, Source \ref{src:popBase} (2000-2025), and the population forecast, Source \ref{src:popForecast} (2025-2050). First of all, all three simulation scenarios can be considered as valid population models, for their maximum deviation from the total population lies around $1\%$. In general, SC4 shows a slightly different trend than the other two. This is due to the problem that the model does not match the age-distribution of the internal immigrants. As a result, the age-distributions of the total population will start to evolve differently, which, in direct consequence, causes differences in births, deaths and emigrations. The corresponding plots have been left out of the report since they do not provide any new insights.

\begin{table}
    \begin{center}
    \begin{scriptsize}
        \begin{tabular}{ccc|cc|cc|cc}
region & sex & age & \multicolumn{2}{c|}{\textbf{SC2} ($e_{min},e_{max}$)} & \multicolumn{2}{c}{\textbf{SC3} ($e_{min},e_{max}$)} & \multicolumn{2}{c}{\textbf{SC4} ($e_{min},e_{max}$)}\\
\hline
\multicolumn{9}{c}{2000-2024}\\
\hline
-&-&-&\cellcolor{blue!4}$-0.12\%$&\cellcolor{red!1}$0.01\%$&\cellcolor{blue!4}$-0.12\%$&\cellcolor{red!1}$0.01\%$&\cellcolor{blue!3}$-0.08\%$&\cellcolor{red!21}$0.45\%$\\
\hline
-&female&-&\cellcolor{blue!4}$-0.13\%$&\cellcolor{red!1}$0.02\%$&\cellcolor{blue!4}$-0.13\%$&\cellcolor{red!2}$0.02\%$&\cellcolor{blue!3}$-0.09\%$&\cellcolor{red!22}$0.47\%$\\
-&male&-&\cellcolor{blue!4}$-0.12\%$&\cellcolor{red!1}$0.01\%$&\cellcolor{blue!4}$-0.13\%$&\cellcolor{red!1}$0.02\%$&\cellcolor{blue!3}$-0.07\%$&\cellcolor{red!20}$0.43\%$\\
\hline
-&-&$[0,19]$&\cellcolor{blue!10}$-0.34\%$&\cellcolor{red!32}$0.70\%$&\cellcolor{blue!11}$-0.35\%$&\cellcolor{red!33}$0.71\%$&\cellcolor{red!14}$0.29\%$&\cellcolor{red!36}$0.77\%$\\
-&-&$[20,39]$&\cellcolor{blue!3}$-0.09\%$&\cellcolor{red!16}$0.34\%$&\cellcolor{blue!3}$-0.08\%$&\cellcolor{red!16}$0.35\%$&\cellcolor{red!2}$0.04\%$&\cellcolor{red!61}$1.33\%$\\
-&-&$[40,59]$&\cellcolor{blue!8}$-0.25\%$&\cellcolor{blue!4}$-0.13\%$&\cellcolor{blue!8}$-0.24\%$&\cellcolor{blue!3}$-0.09\%$&\cellcolor{blue!9}$-0.27\%$&\cellcolor{red!9}$0.18\%$\\
-&-&$[60,79]$&\cellcolor{blue!14}$-0.47\%$&\cellcolor{blue!3}$-0.09\%$&\cellcolor{blue!15}$-0.48\%$&\cellcolor{blue!3}$-0.09\%$&\cellcolor{blue!15}$-0.47\%$&\cellcolor{blue!7}$-0.23\%$\\
-&-&$80^+$&\cellcolor{blue!50}$-1.67\%$&\cellcolor{blue!9}$-0.27\%$&\cellcolor{blue!49}$-1.64\%$&\cellcolor{blue!10}$-0.32\%$&\cellcolor{blue!49}$-1.63\%$&\cellcolor{blue!11}$-0.37\%$\\
\hline
AT-1&-&-&\cellcolor{red!3}$0.05\%$&\cellcolor{red!16}$0.34\%$&\cellcolor{blue!4}$-0.10\%$&\cellcolor{red!18}$0.38\%$&\cellcolor{red!4}$0.08\%$&\cellcolor{red!36}$0.79\%$\\
AT-2&-&-&\cellcolor{blue!9}$-0.27\%$&\cellcolor{blue!1}$-0.01\%$&\cellcolor{blue!12}$-0.38\%$&\cellcolor{red!8}$0.17\%$&\cellcolor{blue!6}$-0.19\%$&\cellcolor{red!21}$0.44\%$\\
AT-3&-&-&\cellcolor{blue!3}$-0.09\%$&\cellcolor{red!9}$0.20\%$&\cellcolor{red!1}$0.00\%$&\cellcolor{red!11}$0.23\%$&\cellcolor{blue!2}$-0.04\%$&\cellcolor{red!10}$0.20\%$\\
AT-4&-&-&\cellcolor{red!3}$0.06\%$&\cellcolor{red!12}$0.25\%$&\cellcolor{blue!6}$-0.18\%$&\cellcolor{red!19}$0.42\%$&\cellcolor{red!12}$0.25\%$&\cellcolor{red!28}$0.61\%$\\
AT-5&-&-&\cellcolor{blue!1}$-0.01\%$&\cellcolor{red!10}$0.21\%$&\cellcolor{blue!12}$-0.39\%$&\cellcolor{red!11}$0.23\%$&\cellcolor{blue!1}$-0.01\%$&\cellcolor{red!14}$0.31\%$\\
AT-6&-&-&\cellcolor{blue!14}$-0.46\%$&\cellcolor{blue!8}$-0.25\%$&\cellcolor{blue!13}$-0.42\%$&\cellcolor{red!1}$0.00\%$&\cellcolor{blue!13}$-0.43\%$&\cellcolor{red!31}$0.68\%$\\
AT-7&-&-&\cellcolor{blue!3}$-0.08\%$&\cellcolor{red!18}$0.37\%$&\cellcolor{blue!12}$-0.38\%$&\cellcolor{red!34}$0.73\%$&\cellcolor{red!8}$0.17\%$&\cellcolor{red!19}$0.42\%$\\
AT-8&-&-&\cellcolor{blue!5}$-0.15\%$&\cellcolor{red!9}$0.19\%$&\cellcolor{blue!13}$-0.44\%$&\cellcolor{red!12}$0.25\%$&\cellcolor{blue!11}$-0.35\%$&\cellcolor{red!8}$0.18\%$\\
AT-9&-&-&\cellcolor{blue!14}$-0.47\%$&\cellcolor{blue!4}$-0.10\%$&\cellcolor{blue!22}$-0.72\%$&\cellcolor{red!7}$0.15\%$&\cellcolor{blue!13}$-0.41\%$&\cellcolor{red!29}$0.62\%$\\
\hline
\multicolumn{9}{c}{2025-2049}\\
\hline
-&-&-&\cellcolor{blue!28}$-0.92\%$&\cellcolor{red!1}$0.01\%$&\cellcolor{blue!28}$-0.92\%$&\cellcolor{red!1}$0.01\%$&\cellcolor{red!6}$0.11\%$&\cellcolor{red!25}$0.54\%$\\
\hline
-&female&-&\cellcolor{blue!24}$-0.79\%$&\cellcolor{red!3}$0.05\%$&\cellcolor{blue!25}$-0.83\%$&\cellcolor{red!3}$0.05\%$&\cellcolor{red!9}$0.20\%$&\cellcolor{red!27}$0.58\%$\\
-&male&-&\cellcolor{blue!32}$-1.06\%$&\cellcolor{blue!2}$-0.04\%$&\cellcolor{blue!30}$-1.00\%$&\cellcolor{blue!2}$-0.04\%$&\cellcolor{red!2}$0.03\%$&\cellcolor{red!23}$0.50\%$\\
\hline
-&-&$[0,19]$&\cellcolor{blue!55}$-1.85\%$&\cellcolor{red!43}$0.93\%$&\cellcolor{blue!58}$-1.94\%$&\cellcolor{red!38}$0.83\%$&\cellcolor{blue!13}$-0.41\%$&\cellcolor{red!98}$2.14\%$\\
-&-&$[20,39]$&\cellcolor{blue!62}$-2.09\%$&\cellcolor{red!16}$0.34\%$&\cellcolor{blue!62}$-2.07\%$&\cellcolor{red!17}$0.37\%$&\cellcolor{blue!25}$-0.81\%$&\cellcolor{red!62}$1.35\%$\\
-&-&$[40,59]$&\cellcolor{blue!78}$-2.63\%$&\cellcolor{blue!6}$-0.19\%$&\cellcolor{blue!76}$-2.55\%$&\cellcolor{blue!6}$-0.19\%$&\cellcolor{blue!36}$-1.20\%$&\cellcolor{red!36}$0.79\%$\\
-&-&$[60,79]$&\cellcolor{blue!6}$-0.19\%$&\cellcolor{red!81}$1.77\%$&\cellcolor{blue!6}$-0.18\%$&\cellcolor{red!81}$1.78\%$&\cellcolor{blue!12}$-0.39\%$&\cellcolor{red!100}$2.19\%$\\
-&-&$80^+$&\cellcolor{blue!11}$-0.36\%$&\cellcolor{red!31}$0.67\%$&\cellcolor{blue!12}$-0.39\%$&\cellcolor{red!31}$0.68\%$&\cellcolor{blue!16}$-0.54\%$&\cellcolor{red!10}$0.20\%$\\
\hline
AT-1&-&-&\cellcolor{blue!22}$-0.74\%$&\cellcolor{red!3}$0.05\%$&\cellcolor{blue!17}$-0.56\%$&\cellcolor{red!17}$0.36\%$&\cellcolor{red!24}$0.53\%$&\cellcolor{red!77}$1.69\%$\\
AT-2&-&-&\cellcolor{blue!50}$-1.67\%$&\cellcolor{blue!6}$-0.17\%$&\cellcolor{blue!58}$-1.94\%$&\cellcolor{blue!22}$-0.72\%$&\cellcolor{blue!2}$-0.06\%$&\cellcolor{red!19}$0.41\%$\\
AT-3&-&-&\cellcolor{blue!19}$-0.61\%$&\cellcolor{red!2}$0.04\%$&\cellcolor{blue!15}$-0.49\%$&\cellcolor{red!15}$0.31\%$&\cellcolor{blue!6}$-0.19\%$&\cellcolor{red!9}$0.19\%$\\
AT-4&-&-&\cellcolor{blue!45}$-1.49\%$&\cellcolor{red!5}$0.11\%$&\cellcolor{blue!45}$-1.52\%$&\cellcolor{blue!14}$-0.45\%$&\cellcolor{blue!21}$-0.68\%$&\cellcolor{red!24}$0.52\%$\\
AT-5&-&-&\cellcolor{red!2}$0.04\%$&\cellcolor{red!21}$0.44\%$&\cellcolor{blue!29}$-0.95\%$&\cellcolor{blue!10}$-0.34\%$&\cellcolor{red!6}$0.12\%$&\cellcolor{red!23}$0.49\%$\\
AT-6&-&-&\cellcolor{blue!51}$-1.72\%$&\cellcolor{blue!9}$-0.28\%$&\cellcolor{blue!51}$-1.70\%$&\cellcolor{blue!14}$-0.46\%$&\cellcolor{red!17}$0.37\%$&\cellcolor{red!35}$0.76\%$\\
AT-7&-&-&\cellcolor{blue!52}$-1.73\%$&\cellcolor{blue!5}$-0.16\%$&\cellcolor{blue!44}$-1.46\%$&\cellcolor{blue!7}$-0.23\%$&\cellcolor{blue!33}$-1.11\%$&\cellcolor{red!11}$0.22\%$\\
AT-8&-&-&\cellcolor{blue!62}$-2.08\%$&\cellcolor{blue!5}$-0.14\%$&\cellcolor{blue!44}$-1.48\%$&\cellcolor{red!19}$0.40\%$&\cellcolor{blue!73}$-2.44\%$&\cellcolor{blue!11}$-0.35\%$\\
AT-9&-&-&\cellcolor{red!3}$0.07\%$&\cellcolor{red!15}$0.33\%$&\cellcolor{red!5}$0.09\%$&\cellcolor{red!44}$0.95\%$&\cellcolor{red!52}$1.14\%$&\cellcolor{red!71}$1.56\%$\\
\hline
\end{tabular}
\end{scriptsize}
\end{center}
    \caption{Relative maximum differences between the population data and forecast (Sources \ref{src:popBase} and \ref{src:popForecast}) and the simulations SC2-SC4.}
    \label{tbl:deviations_im_pop}
\end{table}

\subsubsection{Comparison with Source \ref{src:internalMigrantsStatCube} (BWanderungen innerhalb Österreichs ab 2002)}
In the next step, we will evaluate the internal emigrants and immigrants compared to Source \ref{src:internalMigrantsStatCube}. We will use 20 year age classes and compare the internal emigrants and immigrants between model result and data for all nine federal-states.

\begin{figure}
\centering
\includegraphics[width=0.7\linewidth]{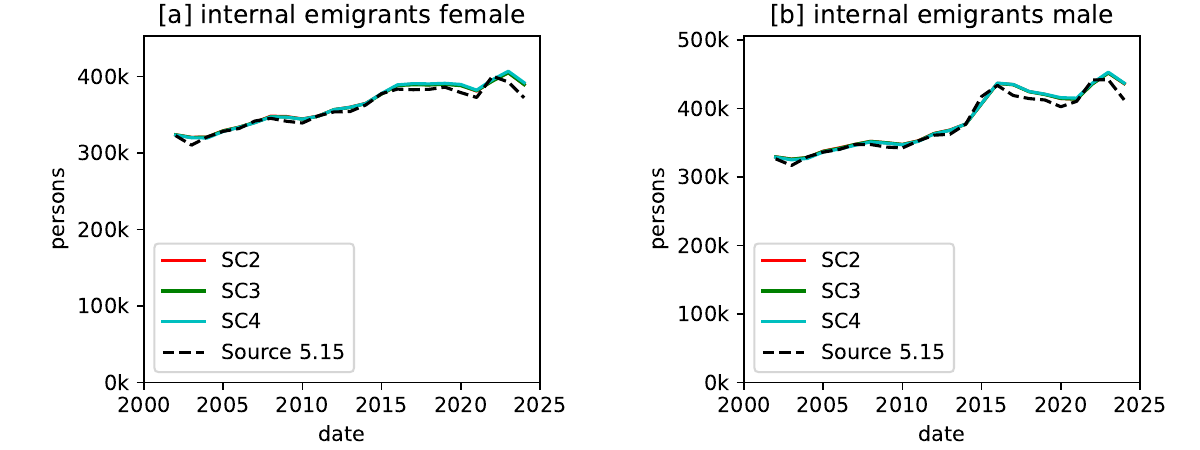}
\caption{Comparison between internal emigrants data from Source \ref{src:internalMigrantsStatCube} and the simulation scenarios SC2-SC4 for male and female persons.}
\label{fig:internal_emigrants_sex}
\end{figure}

Table \ref{tbl:deviations_internalemigrants} shows that the offsets from the actual data are very similar to the ones for the external emigration. The offsets vary in the range between $-6\%$ to $10\%$ w.r. to sex, age and federal-states. As the number of people is around ten times higher than in the case of external emigration, the percentage deviations are more serious. On the high aggregation level, the largest offset is found for the male internal emigrants with around $6\%$. As seen in Figure \ref{fig:internal_emigrants_sex}, the differences can be considered to be the result of a time-lag effect rather than systematic underestimation. Further result plots are found in the Appendix (Figures \ref{fig:internal_emigrants_sex_diff} to \ref{fig:internal_emigrants_fed_diff}).

Comparing the three scenarios SC2 to SC4 with each other, we do not find any larger differences. This was expected, since all three use the same parameters and modelling concept for deciding about internal emigration. We expect this to change when looking into internal immigration.

Table \ref{tbl:deviations_internalimmigrants} shows the offset-table for the internal immigrants. First of all, the overall deviations for sex and age was left out, since it would be equivalent to the ones from the internal emigration. Even investigating the outcomes on the federal-state level for the destination region, no systematic differences between the model results can be observed: By aggregation over the origin regions and the age distribution, the potential weaknesses of the interregional and the biregional models do not show. The maximum offsets all vary between $\pm 9\%$ and are very similar to the ones for the internal emigration. 

The similarity between the models stops, having a look at the mixed outputs for destination federal-states and age-classes, which are shown in the lower part of Table \ref{tbl:deviations_internalimmigrants}. While it is surprising that the errors for SC3 are smaller than the ones for SC2, in general, SC4 performs by far worse than the other two models.

\begin{figure}
\centering
\includegraphics[width=0.6\linewidth]{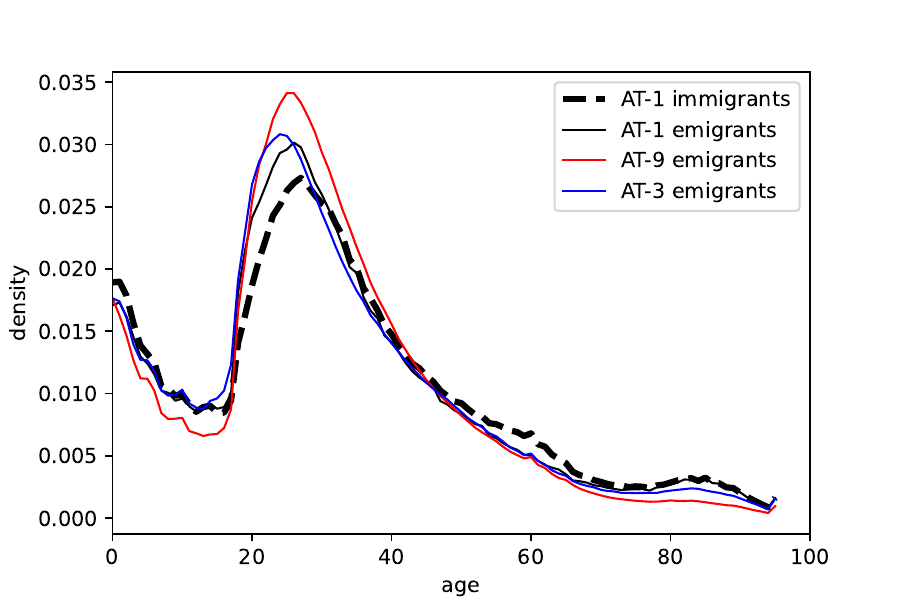}
\caption{Comparison between the age profiles of internal immigrants into AT-1 and the internal emigrants of the most important origin regions AT-1,AT-3 and AT-9. All data is from Source \ref{src:internalMigrantsStatCube} for the years between 2002 and 2024 and normed to an area of one below the curve.}
\label{fig:comparison_AT1}
\end{figure}

As an illustrative example, we can point out federal-state AT-1, in which SC4 heavily overestimates for age-class $[20,39]$ and underestimates for older age-classes (up to $-30\%$ off for $[60,79]$). This is a direct result of the concept of the interregional model: the age profile of the immigrants into a certain destination region originates from the average of the age-profiles of the internal emigrants of the contributing origin regions, weighted by their contribution to the migration. That means the age profile of the immigrants into AT-1 is primarily caused by the age profile of the emigrants from AT-1 itself (i.e. persons who move within the federal-state) and the ones from AT-3 and AT-9, which, in sum, are responsible for over $75\%$ of all internal immigrations beyond the federal-state border. Figure \ref{fig:comparison_AT1} shows a layover of the internal immigration age-profile of AT-1 and the age-profiles of internal emigrants of AT-1, AT-3 and AT-9. It is clearly visible that the emigration age-profiles are all much ``younger'' than the immigration age-profile, in particular the one from Vienna (AT-9). Hence, the interregional model can never properly depict the real situation.

The real situation is more complex and requires to make destination regions age-dependent: Primarily older internal emigrants from AT-3 and AT-9 tend to move to AT-1, whereas younger ones usually move within AT-3, AT-9 or someplace else. From the perspective of federal-state AT-1, young internal emigrants tend to leave the region, whereas older ones are more likely to stay.

In summary, the interregional model shows expected weaknesses when it comes to depicting the correct age-resolution of internal immigrants into certain regions, since the age resolution is ignored when choosing a new destination region. In this light, it is up to some extent even surprising that the results for the other quantities (population, births, etc.) are still within a reasonable error margin. However, since a correct regional age distribution is relevant for all other model processes, SC4 leads to the largest errors to the reference data, not only for the population but in general for all other dimensions (deaths, births, etc.).

\begin{table}
    \begin{center}
    \begin{scriptsize}
        \begin{tabular}{ccc|cc|cc|cc}
        region (origin) & sex & age & \multicolumn{2}{c|}{\textbf{SC2} ($e_{min},e_{max}$)} & \multicolumn{2}{c}{\textbf{SC3} ($e_{min},e_{max}$)} & \multicolumn{2}{c}{\textbf{SC4} ($e_{min},e_{max}$)}\\
-&-&-&\cellcolor{blue!21}$-1.38\%$&\cellcolor{red!37}$5.24\%$&\cellcolor{blue!22}$-1.49\%$&\cellcolor{red!37}$5.26\%$&\cellcolor{blue!19}$-1.25\%$&\cellcolor{red!40}$5.65\%$\\
\hline
-&female&-&\cellcolor{blue!22}$-1.48\%$&\cellcolor{red!33}$4.68\%$&\cellcolor{blue!25}$-1.64\%$&\cellcolor{red!33}$4.64\%$&\cellcolor{blue!17}$-1.15\%$&\cellcolor{red!38}$5.32\%$\\
-&male&-&\cellcolor{blue!37}$-2.51\%$&\cellcolor{red!40}$5.74\%$&\cellcolor{blue!37}$-2.51\%$&\cellcolor{red!41}$5.81\%$&\cellcolor{blue!36}$-2.43\%$&\cellcolor{red!42}$5.94\%$\\
\hline
-&-&$[0,19]$&\cellcolor{blue!40}$-2.68\%$&\cellcolor{red!43}$6.16\%$&\cellcolor{blue!41}$-2.78\%$&\cellcolor{red!42}$5.91\%$&\cellcolor{blue!25}$-1.64\%$&\cellcolor{red!58}$8.32\%$\\
-&-&$[20,39]$&\cellcolor{blue!13}$-0.88\%$&\cellcolor{red!42}$5.93\%$&\cellcolor{blue!14}$-0.90\%$&\cellcolor{red!42}$5.97\%$&\cellcolor{blue!21}$-1.36\%$&\cellcolor{red!37}$5.28\%$\\
-&-&$[40,59]$&\cellcolor{blue!47}$-3.15\%$&\cellcolor{red!22}$3.08\%$&\cellcolor{blue!50}$-3.34\%$&\cellcolor{red!23}$3.18\%$&\cellcolor{blue!37}$-2.46\%$&\cellcolor{red!29}$4.03\%$\\
-&-&$[60,79]$&\cellcolor{blue!57}$-3.83\%$&\cellcolor{red!67}$9.54\%$&\cellcolor{blue!58}$-3.92\%$&\cellcolor{red!64}$9.11\%$&\cellcolor{blue!44}$-2.99\%$&\cellcolor{red!64}$9.14\%$\\
-&-&$80^+$&\cellcolor{red!5}$0.71\%$&\cellcolor{red!62}$8.85\%$&\cellcolor{red!3}$0.42\%$&\cellcolor{red!59}$8.33\%$&\cellcolor{red!3}$0.41\%$&\cellcolor{red!62}$8.84\%$\\
\hline
AT-1&-&-&\cellcolor{blue!42}$-2.81\%$&\cellcolor{red!43}$6.07\%$&\cellcolor{blue!41}$-2.73\%$&\cellcolor{red!43}$6.17\%$&\cellcolor{red!9}$1.17\%$&\cellcolor{red!100}$14.36\%$\\
AT-2&-&-&\cellcolor{blue!55}$-3.73\%$&\cellcolor{red!55}$7.89\%$&\cellcolor{blue!60}$-4.04\%$&\cellcolor{red!52}$7.36\%$&\cellcolor{blue!18}$-1.20\%$&\cellcolor{red!74}$10.50\%$\\
AT-3&-&-&\cellcolor{blue!40}$-2.68\%$&\cellcolor{red!32}$4.48\%$&\cellcolor{blue!40}$-2.71\%$&\cellcolor{red!34}$4.77\%$&\cellcolor{blue!27}$-1.79\%$&\cellcolor{red!61}$8.66\%$\\
AT-4&-&-&\cellcolor{blue!36}$-2.41\%$&\cellcolor{red!41}$5.79\%$&\cellcolor{blue!35}$-2.36\%$&\cellcolor{red!40}$5.70\%$&\cellcolor{blue!36}$-2.38\%$&\cellcolor{red!49}$7.02\%$\\
AT-5&-&-&\cellcolor{blue!48}$-3.21\%$&\cellcolor{red!43}$6.15\%$&\cellcolor{blue!43}$-2.89\%$&\cellcolor{red!38}$5.41\%$&\cellcolor{blue!35}$-2.36\%$&\cellcolor{red!49}$6.98\%$\\
AT-6&-&-&\cellcolor{blue!40}$-2.68\%$&\cellcolor{red!35}$4.97\%$&\cellcolor{blue!39}$-2.62\%$&\cellcolor{red!34}$4.79\%$&\cellcolor{blue!39}$-2.63\%$&\cellcolor{red!41}$5.78\%$\\
AT-7&-&-&\cellcolor{blue!70}$-4.74\%$&\cellcolor{red!66}$9.48\%$&\cellcolor{blue!68}$-4.61\%$&\cellcolor{red!68}$9.70\%$&\cellcolor{blue!73}$-4.95\%$&\cellcolor{red!64}$9.17\%$\\
AT-8&-&-&\cellcolor{blue!75}$-5.04\%$&\cellcolor{red!33}$4.73\%$&\cellcolor{blue!80}$-5.44\%$&\cellcolor{red!35}$5.02\%$&\cellcolor{blue!74}$-5.00\%$&\cellcolor{red!37}$5.23\%$\\
AT-9&-&-&\cellcolor{blue!22}$-1.47\%$&\cellcolor{red!57}$8.16\%$&\cellcolor{blue!21}$-1.42\%$&\cellcolor{red!58}$8.26\%$&\cellcolor{blue!69}$-4.64\%$&\cellcolor{red!49}$7.02\%$
\end{tabular}
\end{scriptsize}
\end{center}
    \caption{Relative maximum differences between the internal emigrants (Source \ref{src:internalMigrantsStatCube}) and the simulations SC2 to SC4 between 2002 and 2024.}
    \label{tbl:deviations_internalemigrants}
\end{table}

\begin{table}
    \begin{center}
    \begin{scriptsize}
        \begin{tabular}{ccc|cc|cc|cc}
        region (destination) & sex & age & \multicolumn{2}{c|}{\textbf{SC2} ($e_{min},e_{max}$)} & \multicolumn{2}{c}{\textbf{SC3} ($e_{min},e_{max}$)}&\multicolumn{2}{c}{\textbf{SC4} ($e_{min},e_{max}$)}\\
\hline
AT-1&-&-&\cellcolor{blue!7}$-2.43\%$&\cellcolor{red!11}$5.86\%$&\cellcolor{blue!6}$-1.98\%$&\cellcolor{red!10}$5.13\%$&\cellcolor{red!2}$1.06\%$&\cellcolor{red!21}$11.32\%$\\
AT-2&-&-&\cellcolor{blue!8}$-3.05\%$&\cellcolor{red!13}$7.15\%$&\cellcolor{blue!4}$-1.44\%$&\cellcolor{red!10}$5.04\%$&\cellcolor{blue!3}$-0.88\%$&\cellcolor{red!18}$9.52\%$\\
AT-3&-&-&\cellcolor{blue!8}$-2.77\%$&\cellcolor{red!9}$4.57\%$&\cellcolor{blue!5}$-1.67\%$&\cellcolor{red!10}$5.13\%$&\cellcolor{blue!2}$-0.64\%$&\cellcolor{red!14}$7.34\%$\\
AT-4&-&-&\cellcolor{blue!6}$-2.30\%$&\cellcolor{red!11}$5.87\%$&\cellcolor{blue!4}$-1.41\%$&\cellcolor{red!10}$5.24\%$&\cellcolor{blue!6}$-2.20\%$&\cellcolor{red!13}$7.01\%$\\
AT-5&-&-&\cellcolor{blue!9}$-3.15\%$&\cellcolor{red!11}$5.88\%$&\cellcolor{blue!5}$-1.75\%$&\cellcolor{red!10}$5.31\%$&\cellcolor{blue!6}$-2.26\%$&\cellcolor{red!13}$6.78\%$\\
AT-6&-&-&\cellcolor{blue!7}$-2.38\%$&\cellcolor{red!9}$4.96\%$&\cellcolor{blue!4}$-1.44\%$&\cellcolor{red!10}$5.26\%$&\cellcolor{blue!6}$-2.23\%$&\cellcolor{red!11}$5.88\%$\\
AT-7&-&-&\cellcolor{blue!12}$-4.46\%$&\cellcolor{red!17}$9.06\%$&\cellcolor{blue!4}$-1.46\%$&\cellcolor{red!10}$5.28\%$&\cellcolor{blue!13}$-4.68\%$&\cellcolor{red!16}$8.77\%$\\
AT-8&-&-&\cellcolor{blue!13}$-4.68\%$&\cellcolor{red!8}$4.39\%$&\cellcolor{blue!4}$-1.46\%$&\cellcolor{red!10}$5.07\%$&\cellcolor{blue!13}$-4.76\%$&\cellcolor{red!9}$4.88\%$\\
AT-9&-&-&\cellcolor{blue!4}$-1.41\%$&\cellcolor{red!13}$6.71\%$&\cellcolor{blue!5}$-1.57\%$&\cellcolor{red!10}$5.39\%$&\cellcolor{blue!10}$-3.68\%$&\cellcolor{red!11}$5.66\%$\\
\hline
AT-1&-&$[0,19]$&\cellcolor{blue!14}$-5.37\%$&\cellcolor{red!12}$6.49\%$&\cellcolor{blue!9}$-3.33\%$&\cellcolor{red!10}$5.51\%$&\cellcolor{blue!16}$-6.10\%$&\cellcolor{red!13}$6.73\%$\\
AT-2&-&$[0,19]$&\cellcolor{blue!12}$-4.33\%$&\cellcolor{red!24}$12.96\%$&\cellcolor{blue!8}$-2.89\%$&\cellcolor{red!10}$5.22\%$&\cellcolor{blue!8}$-2.73\%$&\cellcolor{red!29}$15.79\%$\\
AT-3&-&$[0,19]$&\cellcolor{blue!11}$-3.89\%$&\cellcolor{red!11}$6.05\%$&\cellcolor{blue!7}$-2.65\%$&\cellcolor{red!11}$6.04\%$&\cellcolor{blue!18}$-6.80\%$&\cellcolor{red!10}$5.36\%$\\
AT-4&-&$[0,19]$&\cellcolor{blue!9}$-3.36\%$&\cellcolor{red!11}$5.83\%$&\cellcolor{blue!7}$-2.68\%$&\cellcolor{red!11}$5.84\%$&\cellcolor{blue!13}$-4.93\%$&\cellcolor{red!13}$7.12\%$\\
AT-5&-&$[0,19]$&\cellcolor{blue!12}$-4.26\%$&\cellcolor{red!14}$7.72\%$&\cellcolor{blue!10}$-3.73\%$&\cellcolor{red!12}$6.33\%$&\cellcolor{blue!3}$-1.14\%$&\cellcolor{red!26}$14.22\%$\\
AT-6&-&$[0,19]$&\cellcolor{blue!8}$-2.88\%$&\cellcolor{red!13}$6.93\%$&\cellcolor{blue!8}$-2.78\%$&\cellcolor{red!11}$5.80\%$&\cellcolor{blue!8}$-2.83\%$&\cellcolor{red!21}$11.38\%$\\
AT-7&-&$[0,19]$&\cellcolor{blue!15}$-5.41\%$&\cellcolor{red!17}$9.24\%$&\cellcolor{blue!9}$-3.13\%$&\cellcolor{red!12}$6.32\%$&\cellcolor{blue!13}$-4.98\%$&\cellcolor{red!20}$10.71\%$\\
AT-8&-&$[0,19]$&\cellcolor{blue!14}$-5.25\%$&\cellcolor{red!9}$4.94\%$&\cellcolor{blue!7}$-2.38\%$&\cellcolor{red!12}$6.25\%$&\cellcolor{blue!18}$-6.87\%$&\cellcolor{red!10}$5.23\%$\\
AT-9&-&$[0,19]$&\cellcolor{blue!9}$-3.24\%$&\cellcolor{red!11}$6.06\%$&\cellcolor{blue!8}$-3.08\%$&\cellcolor{red!11}$5.91\%$&\cellcolor{red!4}$1.74\%$&\cellcolor{red!20}$10.77\%$\\
AT-1&-&$[20,39]$&\cellcolor{blue!4}$-1.38\%$&\cellcolor{red!14}$7.43\%$&\cellcolor{blue!4}$-1.17\%$&\cellcolor{red!11}$6.08\%$&\cellcolor{red!14}$7.31\%$&\cellcolor{red!67}$36.76\%$\\
AT-2&-&$[20,39]$&\cellcolor{blue!8}$-2.75\%$&\cellcolor{red!13}$7.20\%$&\cellcolor{blue!3}$-0.79\%$&\cellcolor{red!12}$6.11\%$&\cellcolor{red!5}$2.73\%$&\cellcolor{red!26}$14.36\%$\\
AT-3&-&$[20,39]$&\cellcolor{blue!4}$-1.37\%$&\cellcolor{red!10}$5.28\%$&\cellcolor{blue!3}$-1.05\%$&\cellcolor{red!11}$5.94\%$&\cellcolor{red!7}$3.53\%$&\cellcolor{red!33}$18.18\%$\\
AT-4&-&$[20,39]$&\cellcolor{blue!4}$-1.52\%$&\cellcolor{red!13}$6.99\%$&\cellcolor{blue!3}$-0.89\%$&\cellcolor{red!11}$5.99\%$&\cellcolor{red!1}$0.16\%$&\cellcolor{red!18}$9.50\%$\\
AT-5&-&$[20,39]$&\cellcolor{blue!10}$-3.60\%$&\cellcolor{red!12}$6.35\%$&\cellcolor{blue!3}$-0.95\%$&\cellcolor{red!11}$5.83\%$&\cellcolor{blue!8}$-2.93\%$&\cellcolor{red!12}$6.29\%$\\
AT-6&-&$[20,39]$&\cellcolor{blue!6}$-2.19\%$&\cellcolor{red!11}$6.10\%$&\cellcolor{blue!2}$-0.57\%$&\cellcolor{red!12}$6.11\%$&\cellcolor{blue!6}$-2.30\%$&\cellcolor{red!12}$6.41\%$\\
AT-7&-&$[20,39]$&\cellcolor{blue!9}$-3.14\%$&\cellcolor{red!12}$6.13\%$&\cellcolor{blue!3}$-0.84\%$&\cellcolor{red!11}$5.94\%$&\cellcolor{blue!11}$-3.92\%$&\cellcolor{red!9}$4.56\%$\\
AT-8&-&$[20,39]$&\cellcolor{blue!12}$-4.52\%$&\cellcolor{red!9}$4.86\%$&\cellcolor{blue!3}$-1.00\%$&\cellcolor{red!11}$5.64\%$&\cellcolor{blue!11}$-4.08\%$&\cellcolor{red!10}$5.46\%$\\
AT-9&-&$[20,39]$&\cellcolor{blue!3}$-1.01\%$&\cellcolor{red!13}$7.16\%$&\cellcolor{blue!3}$-1.00\%$&\cellcolor{red!11}$5.98\%$&\cellcolor{blue!31}$-11.66\%$&\cellcolor{blue!1}$-0.08\%$\\
AT-1&-&$[40,59]$&\cellcolor{blue!13}$-4.65\%$&\cellcolor{red!9}$4.95\%$&\cellcolor{blue!10}$-3.51\%$&\cellcolor{red!7}$3.63\%$&\cellcolor{blue!40}$-15.46\%$&\cellcolor{red!5}$2.55\%$\\
AT-2&-&$[40,59]$&\cellcolor{blue!12}$-4.52\%$&\cellcolor{red!10}$5.53\%$&\cellcolor{blue!10}$-3.57\%$&\cellcolor{red!7}$3.38\%$&\cellcolor{blue!22}$-8.39\%$&\cellcolor{red!4}$1.82\%$\\
AT-3&-&$[40,59]$&\cellcolor{blue!12}$-4.51\%$&\cellcolor{red!5}$2.38\%$&\cellcolor{blue!9}$-3.39\%$&\cellcolor{red!6}$3.02\%$&\cellcolor{blue!28}$-10.54\%$&\cellcolor{red!3}$1.42\%$\\
AT-4&-&$[40,59]$&\cellcolor{blue!7}$-2.54\%$&\cellcolor{red!9}$4.88\%$&\cellcolor{blue!9}$-3.21\%$&\cellcolor{red!5}$2.76\%$&\cellcolor{blue!10}$-3.73\%$&\cellcolor{red!8}$4.05\%$\\
AT-5&-&$[40,59]$&\cellcolor{blue!8}$-3.00\%$&\cellcolor{red!6}$3.20\%$&\cellcolor{blue!8}$-2.90\%$&\cellcolor{red!6}$3.25\%$&\cellcolor{blue!13}$-4.92\%$&\cellcolor{red!5}$2.34\%$\\
AT-6&-&$[40,59]$&\cellcolor{blue!10}$-3.61\%$&\cellcolor{red!6}$3.23\%$&\cellcolor{blue!11}$-4.05\%$&\cellcolor{red!6}$3.18\%$&\cellcolor{blue!8}$-2.98\%$&\cellcolor{red!8}$3.89\%$\\
AT-7&-&$[40,59]$&\cellcolor{blue!14}$-5.29\%$&\cellcolor{red!23}$12.30\%$&\cellcolor{blue!8}$-2.95\%$&\cellcolor{red!7}$3.69\%$&\cellcolor{blue!13}$-4.73\%$&\cellcolor{red!25}$13.39\%$\\
AT-8&-&$[40,59]$&\cellcolor{blue!13}$-4.91\%$&\cellcolor{red!6}$3.27\%$&\cellcolor{blue!7}$-2.39\%$&\cellcolor{red!8}$3.95\%$&\cellcolor{blue!13}$-4.81\%$&\cellcolor{red!9}$4.71\%$\\
AT-9&-&$[40,59]$&\cellcolor{blue!8}$-3.04\%$&\cellcolor{red!11}$6.09\%$&\cellcolor{blue!9}$-3.33\%$&\cellcolor{red!6}$3.29\%$&\cellcolor{red!4}$1.85\%$&\cellcolor{red!24}$13.00\%$\\
AT-1&-&$[60,79]$&\cellcolor{blue!11}$-4.20\%$&\cellcolor{red!14}$7.52\%$&\cellcolor{blue!13}$-4.78\%$&\cellcolor{red!25}$13.34\%$&\cellcolor{blue!80}$-30.92\%$&\cellcolor{blue!20}$-7.46\%$\\
AT-2&-&$[60,79]$&\cellcolor{blue!13}$-4.64\%$&\cellcolor{red!32}$17.22\%$&\cellcolor{blue!9}$-3.19\%$&\cellcolor{red!17}$9.00\%$&\cellcolor{blue!45}$-17.06\%$&\cellcolor{red!21}$11.63\%$\\
AT-3&-&$[60,79]$&\cellcolor{blue!17}$-6.23\%$&\cellcolor{red!12}$6.38\%$&\cellcolor{blue!12}$-4.30\%$&\cellcolor{red!18}$9.79\%$&\cellcolor{blue!48}$-18.34\%$&\cellcolor{blue!2}$-0.50\%$\\
AT-4&-&$[60,79]$&\cellcolor{blue!20}$-7.50\%$&\cellcolor{red!26}$14.00\%$&\cellcolor{blue!9}$-3.43\%$&\cellcolor{red!17}$8.96\%$&\cellcolor{blue!28}$-10.48\%$&\cellcolor{red!21}$11.51\%$\\
AT-5&-&$[60,79]$&\cellcolor{blue!10}$-3.80\%$&\cellcolor{red!17}$8.89\%$&\cellcolor{blue!11}$-3.89\%$&\cellcolor{red!17}$9.12\%$&\cellcolor{blue!17}$-6.49\%$&\cellcolor{red!13}$6.93\%$\\
AT-6&-&$[60,79]$&\cellcolor{blue!13}$-5.00\%$&\cellcolor{red!12}$6.22\%$&\cellcolor{blue!10}$-3.71\%$&\cellcolor{red!17}$8.94\%$&\cellcolor{blue!21}$-7.76\%$&\cellcolor{red!9}$4.80\%$\\
AT-7&-&$[60,79]$&\cellcolor{blue!32}$-12.21\%$&\cellcolor{red!59}$32.33\%$&\cellcolor{blue!11}$-4.09\%$&\cellcolor{red!17}$9.04\%$&\cellcolor{blue!33}$-12.57\%$&\cellcolor{red!62}$33.91\%$\\
AT-8&-&$[60,79]$&\cellcolor{blue!23}$-8.74\%$&\cellcolor{red!24}$13.15\%$&\cellcolor{blue!11}$-4.23\%$&\cellcolor{red!12}$6.19\%$&\cellcolor{blue!21}$-8.01\%$&\cellcolor{red!31}$16.67\%$\\
AT-9&-&$[60,79]$&\cellcolor{blue!14}$-5.13\%$&\cellcolor{red!11}$5.72\%$&\cellcolor{blue!11}$-4.00\%$&\cellcolor{red!16}$8.76\%$&\cellcolor{red!22}$11.89\%$&\cellcolor{red!62}$34.23\%$\\
AT-1&-&$80^+$&\cellcolor{red!1}$0.28\%$&\cellcolor{red!26}$13.92\%$&\cellcolor{blue!2}$-0.50\%$&\cellcolor{red!20}$10.64\%$&\cellcolor{blue!55}$-20.93\%$&\cellcolor{red!12}$6.60\%$\\
AT-2&-&$80^+$&\cellcolor{blue!12}$-4.41\%$&\cellcolor{red!32}$17.71\%$&\cellcolor{blue!2}$-0.62\%$&\cellcolor{red!18}$9.48\%$&\cellcolor{blue!28}$-10.68\%$&\cellcolor{red!21}$11.44\%$\\
AT-3&-&$80^+$&\cellcolor{red!3}$1.63\%$&\cellcolor{red!17}$9.13\%$&\cellcolor{blue!1}$-0.27\%$&\cellcolor{red!15}$8.11\%$&\cellcolor{blue!36}$-13.84\%$&\cellcolor{blue!8}$-3.08\%$\\
AT-4&-&$80^+$&\cellcolor{blue!6}$-2.24\%$&\cellcolor{red!20}$11.01\%$&\cellcolor{blue!2}$-0.42\%$&\cellcolor{red!15}$8.06\%$&\cellcolor{blue!17}$-6.36\%$&\cellcolor{red!16}$8.44\%$\\
AT-5&-&$80^+$&\cellcolor{blue!6}$-2.32\%$&\cellcolor{red!26}$13.88\%$&\cellcolor{red!1}$0.51\%$&\cellcolor{red!17}$8.97\%$&\cellcolor{blue!22}$-8.42\%$&\cellcolor{red!21}$11.35\%$\\
AT-6&-&$80^+$&\cellcolor{blue!2}$-0.48\%$&\cellcolor{red!22}$11.88\%$&\cellcolor{blue!2}$-0.39\%$&\cellcolor{red!15}$8.29\%$&\cellcolor{blue!9}$-3.24\%$&\cellcolor{red!18}$9.80\%$\\
AT-7&-&$80^+$&\cellcolor{blue!5}$-1.59\%$&\cellcolor{red!27}$14.58\%$&\cellcolor{red!1}$0.23\%$&\cellcolor{red!20}$10.77\%$&\cellcolor{blue!7}$-2.55\%$&\cellcolor{red!22}$12.08\%$\\
AT-8&-&$80^+$&\cellcolor{blue!9}$-3.29\%$&\cellcolor{red!29}$16.04\%$&\cellcolor{blue!1}$-0.15\%$&\cellcolor{red!19}$10.26\%$&\cellcolor{blue!6}$-2.24\%$&\cellcolor{red!31}$16.78\%$\\
AT-9&-&$80^+$&\cellcolor{red!2}$0.72\%$&\cellcolor{red!26}$14.00\%$&\cellcolor{red!5}$2.33\%$&\cellcolor{red!20}$10.75\%$&\cellcolor{red!51}$28.05\%$&\cellcolor{red!100}$55.46\%$\end{tabular}
\end{scriptsize}
\end{center}
    \caption{Relative maximum differences between the internal immigrants (Source \ref{src:internalMigrantsStatCube}) and the simulations SC2 to SC4 between 2002 and 2024.}
    \label{tbl:deviations_internalimmigrants}
\end{table}

\subsubsection{Comparison with Source \ref{src:internalMigrants} (Wanderungen innerhalb Österreichs ab 2002)}
\begin{figure}
\centering
\includegraphics[width=0.8\linewidth]{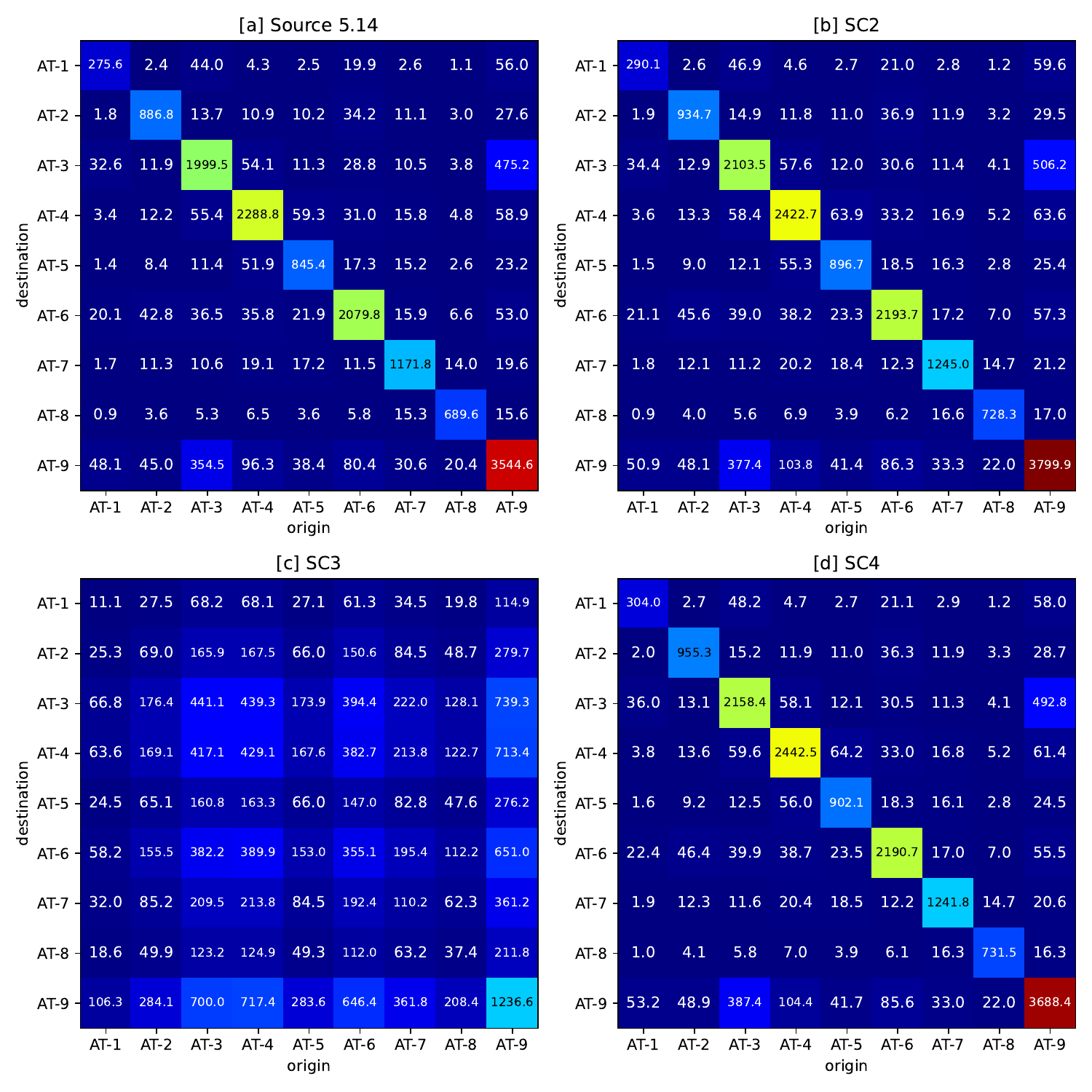}
\caption{Comparison between origin-destination flow data between 2002 and 2024 (Source \ref{src:internalMigrants}) and the simulation scenarios SC2-SC4. The numbers are in units of thousand migrants over the whole period.}
\label{fig:internal_migrants_od}
\end{figure}

Finally, we investigate the origin-destination behaviour of the three internal-migration models. Table \ref{tbl:deviations_od} shows the maximum offsets of the model results to the data. It is well visible that the biregional model (SC3) does not at all represent a valid behaviour in this regard. The maximum differences are beyond any reasonable margin. As a result, the model should only be used if the correct representation of regional flows are irrelevant. The other two models nicely depict the flows between the federal-states with rare outliers up to $\pm 20\%$ deviation (e.g. between AT-1 and AT-8), which can be explained by small numbers of migrants. Figure \ref{fig:internal_migrants_od} shows the total number of internal migrants between the regions between 2002 and 2024. It is easily seen that the internal migrants within the same region are the most relevant ones w.r. to absolute numbers. Flows between the eastern federalstates AT-3, AT-9, AT-4 and AT-6 are also by several margins larger than flows between western federalstates. It is nicely seen in panel [c] with the biregional model that all rows and all columns of the origin-destination matrix are a multiple of each other. This is a result of the concept that the bidirectional model samples all destinations independent of the origin regions.

\begin{figure}
\centering
\includegraphics[width=0.8\linewidth]{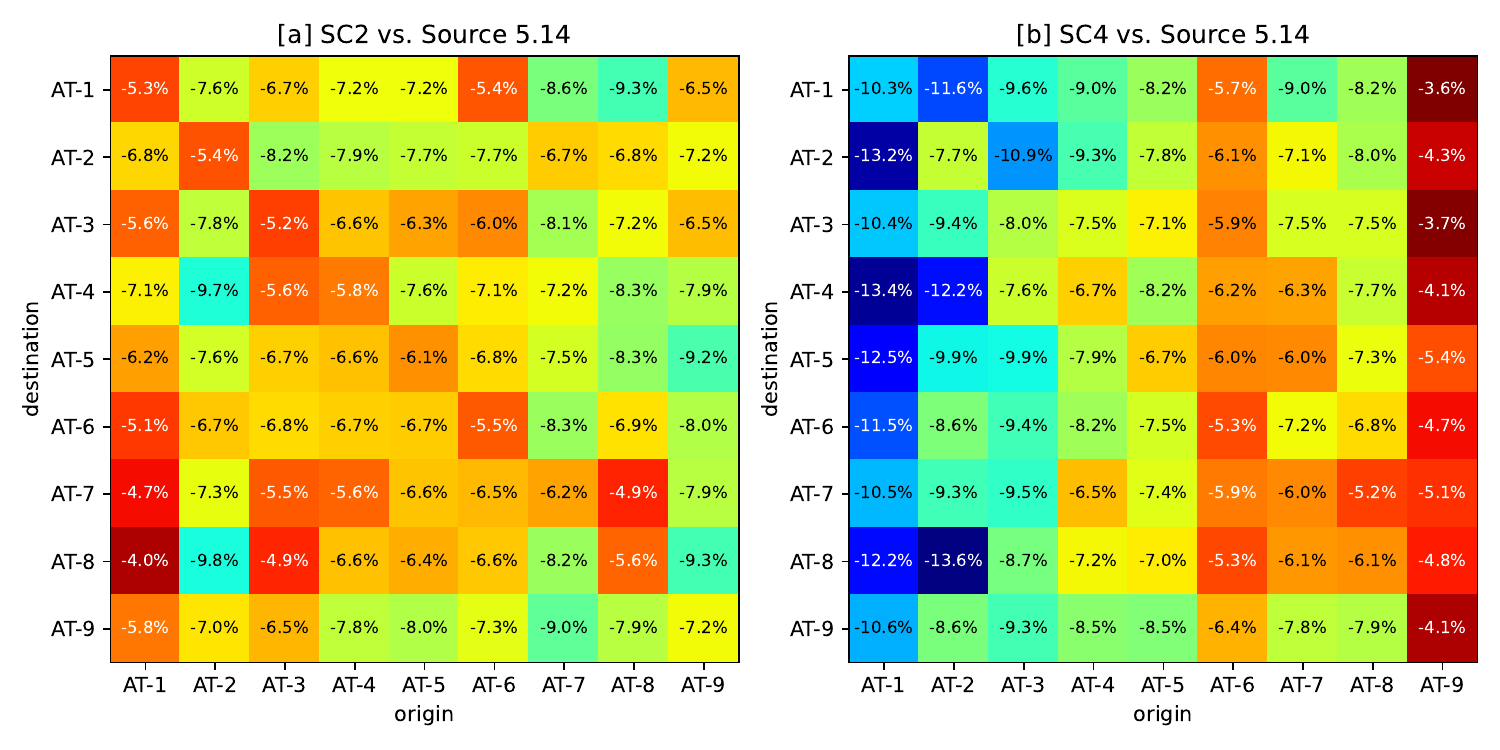}
\caption{Differences between the total internal migrants between 2002 and 2024 (Source \ref{src:internalMigrants}) and the simulation scenarios SC2 and SC4.}
\label{fig:internal_migrants_od_diff}
\end{figure}

Figure \ref{fig:internal_migrants_od_diff} puts the numbers of SC2 and SC4 in relation with Source \ref{src:internalMigrants} (SC3 was left out, since it does not validly represent the situation). Both panels show the mentioned systematic underestimation of the internal emigrants (discussed before). Part [b] of the figure reveals bias with respect to the internal migrants from Vienna (AT-9) and Burgenland (AT-1) by the interregional model SC4. We estimate that this is a result of the problem of the interregional model to validly depict the dynamics of the age structure of the population, in particular in large cities.

\begin{table}
    \begin{center}
    \begin{tiny}
        \begin{tabular}{ccc|cc|cc|cc}
        region (origin) & sex & region (destination) & \multicolumn{2}{c|}{\textbf{SC2} ($e_{min},e_{max}$)} & \multicolumn{2}{c}{\textbf{SC3} ($e_{min},e_{max}$)}&\multicolumn{2}{c}{\textbf{SC4} ($e_{min},e_{max}$)}\\
AT-1&-&AT-1&\cellcolor{blue!12}$-2.53\%$&\cellcolor{red!17}$3.50\%$&\cellcolor{blue!80}$-96.60\%$&\cellcolor{blue!80}$-95.57\%$&\cellcolor{red!7}$1.26\%$&\cellcolor{red!49}$10.09\%$\\
AT-1&-&AT-2&\cellcolor{blue!33}$-7.41\%$&\cellcolor{red!33}$6.91\%$&\cellcolor{red!100}$573.09\%$&\cellcolor{red!100}$2605.71\%$&\cellcolor{red!2}$0.29\%$&\cellcolor{red!48}$9.91\%$\\
AT-1&-&AT-3&\cellcolor{blue!18}$-4.02\%$&\cellcolor{red!21}$4.23\%$&\cellcolor{red!100}$72.18\%$&\cellcolor{red!100}$127.14\%$&\cellcolor{red!2}$0.24\%$&\cellcolor{red!48}$10.01\%$\\
AT-1&-&AT-4&\cellcolor{blue!20}$-4.30\%$&\cellcolor{red!35}$7.25\%$&\cellcolor{red!100}$805.26\%$&\cellcolor{red!100}$2945.74\%$&\cellcolor{blue!18}$-3.90\%$&\cellcolor{red!84}$17.50\%$\\
AT-1&-&AT-5&\cellcolor{blue!80}$-17.92\%$&\cellcolor{red!34}$7.06\%$&\cellcolor{red!100}$965.59\%$&\cellcolor{red!100}$2922.58\%$&\cellcolor{blue!29}$-6.45\%$&\cellcolor{red!62}$12.84\%$\\
AT-1&-&AT-6&\cellcolor{blue!20}$-4.35\%$&\cellcolor{red!21}$4.26\%$&\cellcolor{red!100}$146.76\%$&\cellcolor{red!100}$271.70\%$&\cellcolor{red!1}$0.16\%$&\cellcolor{red!54}$11.32\%$\\
AT-1&-&AT-7&\cellcolor{blue!48}$-10.75\%$&\cellcolor{red!36}$7.47\%$&\cellcolor{red!100}$832.37\%$&\cellcolor{red!100}$3815.05\%$&\cellcolor{blue!16}$-3.43\%$&\cellcolor{red!54}$11.26\%$\\
AT-1&-&AT-8&\cellcolor{blue!66}$-14.81\%$&\cellcolor{red!43}$8.99\%$&\cellcolor{red!100}$484.18\%$&\cellcolor{red!100}$4252.47\%$&\cellcolor{blue!14}$-3.03\%$&\cellcolor{red!98}$20.54\%$\\
AT-1&-&AT-9&\cellcolor{blue!18}$-3.92\%$&\cellcolor{red!28}$5.69\%$&\cellcolor{red!100}$64.78\%$&\cellcolor{red!100}$148.22\%$&\cellcolor{red!4}$0.78\%$&\cellcolor{red!48}$9.91\%$\\
\hline
AT-2&-&AT-1&\cellcolor{blue!62}$-13.84\%$&\cellcolor{red!44}$9.04\%$&\cellcolor{red!100}$384.06\%$&\cellcolor{red!100}$1976.08\%$&\cellcolor{blue!22}$-4.94\%$&\cellcolor{red!69}$14.37\%$\\
AT-2&-&AT-2&\cellcolor{blue!15}$-3.33\%$&\cellcolor{red!37}$7.67\%$&\cellcolor{blue!80}$-93.07\%$&\cellcolor{blue!80}$-92.00\%$&\cellcolor{blue!4}$-0.76\%$&\cellcolor{red!50}$10.49\%$\\
AT-2&-&AT-3&\cellcolor{blue!21}$-4.61\%$&\cellcolor{red!45}$9.26\%$&\cellcolor{red!100}$431.26\%$&\cellcolor{red!100}$2257.50\%$&\cellcolor{blue!17}$-3.75\%$&\cellcolor{red!52}$10.84\%$\\
AT-2&-&AT-4&\cellcolor{blue!33}$-7.30\%$&\cellcolor{red!59}$12.24\%$&\cellcolor{red!100}$355.64\%$&\cellcolor{red!100}$2256.78\%$&\cellcolor{blue!23}$-5.01\%$&\cellcolor{red!69}$14.38\%$\\
AT-2&-&AT-5&\cellcolor{blue!25}$-5.62\%$&\cellcolor{red!48}$10.05\%$&\cellcolor{red!100}$314.73\%$&\cellcolor{red!100}$864.57\%$&\cellcolor{blue!14}$-3.01\%$&\cellcolor{red!37}$7.66\%$\\
AT-2&-&AT-6&\cellcolor{blue!23}$-5.16\%$&\cellcolor{red!41}$8.44\%$&\cellcolor{red!100}$142.32\%$&\cellcolor{red!100}$370.92\%$&\cellcolor{blue!13}$-2.89\%$&\cellcolor{red!49}$10.18\%$\\
AT-2&-&AT-7&\cellcolor{blue!27}$-6.00\%$&\cellcolor{red!35}$7.20\%$&\cellcolor{red!100}$319.00\%$&\cellcolor{red!100}$821.17\%$&\cellcolor{blue!7}$-1.56\%$&\cellcolor{red!32}$6.60\%$\\
AT-2&-&AT-8&\cellcolor{blue!31}$-6.80\%$&\cellcolor{red!42}$8.82\%$&\cellcolor{red!100}$358.03\%$&\cellcolor{red!100}$2315.89\%$&\cellcolor{blue!25}$-5.49\%$&\cellcolor{red!69}$14.29\%$\\
AT-2&-&AT-9&\cellcolor{blue!20}$-4.35\%$&\cellcolor{red!47}$9.71\%$&\cellcolor{red!100}$382.76\%$&\cellcolor{red!100}$639.40\%$&\cellcolor{blue!13}$-2.87\%$&\cellcolor{red!52}$10.88\%$\\
\hline
AT-3&-&AT-1&\cellcolor{blue!10}$-2.14\%$&\cellcolor{red!18}$3.64\%$&\cellcolor{red!100}$24.81\%$&\cellcolor{red!100}$91.42\%$&\cellcolor{blue!4}$-0.72\%$&\cellcolor{red!36}$7.45\%$\\
AT-3&-&AT-2&\cellcolor{blue!19}$-4.23\%$&\cellcolor{red!20}$4.16\%$&\cellcolor{red!100}$530.81\%$&\cellcolor{red!100}$1926.22\%$&\cellcolor{blue!12}$-2.63\%$&\cellcolor{red!36}$7.42\%$\\
AT-3&-&AT-3&\cellcolor{blue!14}$-3.05\%$&\cellcolor{red!15}$3.07\%$&\cellcolor{blue!80}$-79.73\%$&\cellcolor{blue!80}$-78.18\%$&\cellcolor{blue!8}$-1.78\%$&\cellcolor{red!34}$6.96\%$\\
AT-3&-&AT-4&\cellcolor{blue!12}$-2.61\%$&\cellcolor{red!22}$4.53\%$&\cellcolor{red!100}$394.47\%$&\cellcolor{red!100}$806.89\%$&\cellcolor{blue!13}$-2.81\%$&\cellcolor{red!24}$4.93\%$\\
AT-3&-&AT-5&\cellcolor{blue!21}$-4.71\%$&\cellcolor{red!23}$4.80\%$&\cellcolor{red!100}$776.94\%$&\cellcolor{red!100}$1740.48\%$&\cellcolor{blue!14}$-2.94\%$&\cellcolor{red!43}$8.86\%$\\
AT-3&-&AT-6&\cellcolor{blue!11}$-2.27\%$&\cellcolor{red!17}$3.50\%$&\cellcolor{red!100}$634.89\%$&\cellcolor{red!100}$1319.82\%$&\cellcolor{blue!5}$-0.91\%$&\cellcolor{red!28}$5.72\%$\\
AT-3&-&AT-7&\cellcolor{blue!18}$-4.04\%$&\cellcolor{red!30}$6.09\%$&\cellcolor{red!100}$1153.56\%$&\cellcolor{red!100}$2683.44\%$&\cellcolor{blue!11}$-2.36\%$&\cellcolor{red!36}$7.41\%$\\
AT-3&-&AT-8&\cellcolor{blue!35}$-7.67\%$&\cellcolor{red!27}$5.52\%$&\cellcolor{red!100}$823.94\%$&\cellcolor{red!100}$3532.88\%$&\cellcolor{blue!16}$-3.43\%$&\cellcolor{red!33}$6.74\%$\\
AT-3&-&AT-9&\cellcolor{blue!9}$-1.92\%$&\cellcolor{red!18}$3.66\%$&\cellcolor{red!100}$70.61\%$&\cellcolor{red!100}$105.14\%$&\cellcolor{blue!9}$-2.01\%$&\cellcolor{red!35}$7.33\%$\\
\hline
AT-4&-&AT-1&\cellcolor{blue!28}$-6.28\%$&\cellcolor{red!30}$6.29\%$&\cellcolor{red!100}$703.62\%$&\cellcolor{red!100}$2308.07\%$&\cellcolor{blue!27}$-5.93\%$&\cellcolor{red!55}$11.47\%$\\
AT-4&-&AT-2&\cellcolor{blue!26}$-5.66\%$&\cellcolor{red!35}$7.32\%$&\cellcolor{red!100}$704.75\%$&\cellcolor{red!100}$2388.73\%$&\cellcolor{blue!13}$-2.82\%$&\cellcolor{red!44}$9.24\%$\\
AT-4&-&AT-3&\cellcolor{blue!15}$-3.37\%$&\cellcolor{red!47}$9.74\%$&\cellcolor{red!100}$477.78\%$&\cellcolor{red!100}$892.33\%$&\cellcolor{blue!15}$-3.16\%$&\cellcolor{red!40}$8.37\%$\\
AT-4&-&AT-4&\cellcolor{blue!12}$-2.50\%$&\cellcolor{red!22}$4.44\%$&\cellcolor{blue!80}$-82.63\%$&\cellcolor{blue!80}$-81.50\%$&\cellcolor{blue!11}$-2.38\%$&\cellcolor{red!27}$5.49\%$\\
AT-4&-&AT-5&\cellcolor{blue!14}$-3.01\%$&\cellcolor{red!36}$7.49\%$&\cellcolor{red!100}$139.69\%$&\cellcolor{red!100}$235.58\%$&\cellcolor{blue!9}$-2.01\%$&\cellcolor{red!40}$8.21\%$\\
AT-4&-&AT-6&\cellcolor{blue!15}$-3.26\%$&\cellcolor{red!21}$4.26\%$&\cellcolor{red!100}$636.90\%$&\cellcolor{red!100}$1393.97\%$&\cellcolor{blue!9}$-1.80\%$&\cellcolor{red!27}$5.58\%$\\
AT-4&-&AT-7&\cellcolor{blue!22}$-4.93\%$&\cellcolor{red!26}$5.28\%$&\cellcolor{red!100}$523.74\%$&\cellcolor{red!100}$1499.34\%$&\cellcolor{blue!14}$-3.11\%$&\cellcolor{red!36}$7.54\%$\\
AT-4&-&AT-8&\cellcolor{blue!25}$-5.53\%$&\cellcolor{red!28}$5.82\%$&\cellcolor{red!100}$995.88\%$&\cellcolor{red!100}$2592.43\%$&\cellcolor{blue!21}$-4.72\%$&\cellcolor{red!34}$7.14\%$\\
AT-4&-&AT-9&\cellcolor{blue!8}$-1.65\%$&\cellcolor{red!28}$5.85\%$&\cellcolor{red!100}$448.04\%$&\cellcolor{red!100}$824.86\%$&\cellcolor{blue!11}$-2.26\%$&\cellcolor{red!28}$5.76\%$\\
\hline
AT-5&-&AT-1&\cellcolor{blue!22}$-4.93\%$&\cellcolor{red!73}$15.28\%$&\cellcolor{red!100}$395.21\%$&\cellcolor{red!100}$2404.72\%$&\cellcolor{blue!25}$-5.56\%$&\cellcolor{red!56}$11.71\%$\\
AT-5&-&AT-2&\cellcolor{blue!23}$-5.15\%$&\cellcolor{red!31}$6.50\%$&\cellcolor{red!100}$171.12\%$&\cellcolor{red!100}$869.42\%$&\cellcolor{blue!20}$-4.37\%$&\cellcolor{red!25}$5.21\%$\\
AT-5&-&AT-3&\cellcolor{blue!22}$-4.83\%$&\cellcolor{red!24}$4.91\%$&\cellcolor{red!100}$745.28\%$&\cellcolor{red!100}$1948.04\%$&\cellcolor{blue!23}$-5.17\%$&\cellcolor{red!30}$6.10\%$\\
AT-5&-&AT-4&\cellcolor{blue!17}$-3.82\%$&\cellcolor{red!42}$8.76\%$&\cellcolor{red!100}$70.81\%$&\cellcolor{red!100}$229.58\%$&\cellcolor{blue!15}$-3.30\%$&\cellcolor{red!50}$10.32\%$\\
AT-5&-&AT-5&\cellcolor{blue!15}$-3.27\%$&\cellcolor{red!21}$4.30\%$&\cellcolor{blue!80}$-92.89\%$&\cellcolor{blue!80}$-92.14\%$&\cellcolor{blue!11}$-2.30\%$&\cellcolor{red!25}$5.08\%$\\
AT-5&-&AT-6&\cellcolor{blue!15}$-3.33\%$&\cellcolor{red!24}$5.03\%$&\cellcolor{red!100}$352.83\%$&\cellcolor{red!100}$671.61\%$&\cellcolor{blue!11}$-2.40\%$&\cellcolor{red!29}$5.95\%$\\
AT-5&-&AT-7&\cellcolor{blue!19}$-4.18\%$&\cellcolor{red!28}$5.70\%$&\cellcolor{red!100}$205.77\%$&\cellcolor{red!100}$582.35\%$&\cellcolor{blue!12}$-2.56\%$&\cellcolor{red!40}$8.28\%$\\
AT-5&-&AT-8&\cellcolor{blue!21}$-4.63\%$&\cellcolor{red!33}$6.82\%$&\cellcolor{red!100}$624.34\%$&\cellcolor{red!100}$1917.04\%$&\cellcolor{blue!16}$-3.46\%$&\cellcolor{red!43}$8.98\%$\\
AT-5&-&AT-9&\cellcolor{blue!14}$-3.09\%$&\cellcolor{red!28}$5.80\%$&\cellcolor{red!100}$418.89\%$&\cellcolor{red!100}$816.09\%$&\cellcolor{blue!12}$-2.55\%$&\cellcolor{red!30}$6.28\%$\\
\hline
AT-6&-&AT-1&\cellcolor{blue!22}$-4.83\%$&\cellcolor{red!20}$4.01\%$&\cellcolor{red!100}$143.06\%$&\cellcolor{red!100}$224.95\%$&\cellcolor{blue!17}$-3.67\%$&\cellcolor{red!23}$4.76\%$\\
AT-6&-&AT-2&\cellcolor{blue!18}$-3.96\%$&\cellcolor{red!24}$4.96\%$&\cellcolor{red!100}$140.58\%$&\cellcolor{red!100}$498.55\%$&\cellcolor{blue!19}$-4.20\%$&\cellcolor{red!12}$2.44\%$\\
AT-6&-&AT-3&\cellcolor{blue!27}$-5.98\%$&\cellcolor{red!23}$4.73\%$&\cellcolor{red!100}$776.42\%$&\cellcolor{red!100}$1462.22\%$&\cellcolor{blue!24}$-5.38\%$&\cellcolor{red!21}$4.29\%$\\
AT-6&-&AT-4&\cellcolor{blue!30}$-6.69\%$&\cellcolor{red!26}$5.44\%$&\cellcolor{red!100}$590.99\%$&\cellcolor{red!100}$1522.12\%$&\cellcolor{blue!34}$-7.57\%$&\cellcolor{red!22}$4.59\%$\\
AT-6&-&AT-5&\cellcolor{blue!15}$-3.36\%$&\cellcolor{red!24}$4.88\%$&\cellcolor{red!100}$481.71\%$&\cellcolor{red!100}$844.50\%$&\cellcolor{blue!26}$-5.82\%$&\cellcolor{red!27}$5.49\%$\\
AT-6&-&AT-6&\cellcolor{blue!11}$-2.44\%$&\cellcolor{red!15}$3.06\%$&\cellcolor{blue!80}$-84.51\%$&\cellcolor{blue!80}$-83.02\%$&\cellcolor{blue!11}$-2.34\%$&\cellcolor{red!16}$3.35\%$\\
AT-6&-&AT-7&\cellcolor{blue!19}$-4.22\%$&\cellcolor{red!21}$4.36\%$&\cellcolor{red!100}$1047.38\%$&\cellcolor{red!100}$1881.01\%$&\cellcolor{blue!25}$-5.47\%$&\cellcolor{red!22}$4.56\%$\\
AT-6&-&AT-8&\cellcolor{blue!23}$-5.09\%$&\cellcolor{red!38}$7.94\%$&\cellcolor{red!100}$998.91\%$&\cellcolor{red!100}$2495.83\%$&\cellcolor{blue!30}$-6.73\%$&\cellcolor{red!22}$4.58\%$\\
AT-6&-&AT-9&\cellcolor{blue!15}$-3.31\%$&\cellcolor{red!22}$4.41\%$&\cellcolor{red!100}$518.10\%$&\cellcolor{red!100}$840.34\%$&\cellcolor{blue!17}$-3.76\%$&\cellcolor{red!15}$2.99\%$\\
\hline
AT-7&-&AT-1&\cellcolor{blue!33}$-7.26\%$&\cellcolor{red!55}$11.37\%$&\cellcolor{red!100}$580.98\%$&\cellcolor{red!100}$2211.11\%$&\cellcolor{blue!27}$-6.07\%$&\cellcolor{red!46}$9.50\%$\\
AT-7&-&AT-2&\cellcolor{blue!18}$-3.96\%$&\cellcolor{red!54}$11.31\%$&\cellcolor{red!100}$450.64\%$&\cellcolor{red!100}$877.00\%$&\cellcolor{blue!16}$-3.57\%$&\cellcolor{red!67}$13.87\%$\\
AT-7&-&AT-3&\cellcolor{blue!23}$-5.15\%$&\cellcolor{red!42}$8.79\%$&\cellcolor{red!100}$1319.39\%$&\cellcolor{red!100}$2639.77\%$&\cellcolor{blue!24}$-5.21\%$&\cellcolor{red!59}$12.34\%$\\
AT-7&-&AT-4&\cellcolor{blue!17}$-3.68\%$&\cellcolor{red!50}$10.36\%$&\cellcolor{red!100}$822.28\%$&\cellcolor{red!100}$1694.53\%$&\cellcolor{blue!19}$-4.16\%$&\cellcolor{red!45}$9.26\%$\\
AT-7&-&AT-5&\cellcolor{blue!13}$-2.75\%$&\cellcolor{red!38}$7.88\%$&\cellcolor{red!100}$345.72\%$&\cellcolor{red!100}$603.06\%$&\cellcolor{blue!17}$-3.75\%$&\cellcolor{red!40}$8.38\%$\\
AT-7&-&AT-6&\cellcolor{blue!31}$-6.84\%$&\cellcolor{red!47}$9.85\%$&\cellcolor{red!100}$771.46\%$&\cellcolor{red!100}$1519.47\%$&\cellcolor{blue!24}$-5.24\%$&\cellcolor{red!32}$6.54\%$\\
AT-7&-&AT-7&\cellcolor{blue!22}$-4.79\%$&\cellcolor{red!46}$9.47\%$&\cellcolor{blue!80}$-91.43\%$&\cellcolor{blue!80}$-90.13\%$&\cellcolor{blue!23}$-5.03\%$&\cellcolor{red!44}$9.23\%$\\
AT-7&-&AT-8&\cellcolor{blue!14}$-3.09\%$&\cellcolor{red!54}$11.13\%$&\cellcolor{red!100}$169.65\%$&\cellcolor{red!100}$543.18\%$&\cellcolor{blue!19}$-4.06\%$&\cellcolor{red!42}$8.76\%$\\
AT-7&-&AT-9&\cellcolor{blue!19}$-4.21\%$&\cellcolor{red!40}$8.29\%$&\cellcolor{red!100}$742.74\%$&\cellcolor{red!100}$1403.29\%$&\cellcolor{blue!13}$-2.85\%$&\cellcolor{red!27}$5.56\%$\\
\hline
AT-8&-&AT-1&\cellcolor{blue!39}$-8.59\%$&\cellcolor{red!56}$11.73\%$&\cellcolor{red!100}$911.23\%$&\cellcolor{red!100}$3573.23\%$&\cellcolor{blue!41}$-9.09\%$&\cellcolor{red!71}$14.81\%$\\
AT-8&-&AT-2&\cellcolor{blue!45}$-10.08\%$&\cellcolor{red!34}$7.09\%$&\cellcolor{red!100}$1155.17\%$&\cellcolor{red!100}$2225.48\%$&\cellcolor{blue!32}$-7.00\%$&\cellcolor{red!56}$11.76\%$\\
AT-8&-&AT-3&\cellcolor{blue!19}$-4.12\%$&\cellcolor{red!32}$6.52\%$&\cellcolor{red!100}$2242.18\%$&\cellcolor{red!100}$4243.00\%$&\cellcolor{blue!20}$-4.40\%$&\cellcolor{red!48}$9.98\%$\\
AT-8&-&AT-4&\cellcolor{blue!46}$-10.22\%$&\cellcolor{red!38}$7.97\%$&\cellcolor{red!100}$1679.42\%$&\cellcolor{red!100}$3298.58\%$&\cellcolor{blue!43}$-9.55\%$&\cellcolor{red!46}$9.54\%$\\
AT-8&-&AT-5&\cellcolor{blue!21}$-4.54\%$&\cellcolor{red!42}$8.73\%$&\cellcolor{red!100}$1178.51\%$&\cellcolor{red!100}$1976.35\%$&\cellcolor{blue!28}$-6.11\%$&\cellcolor{red!34}$7.12\%$\\
AT-8&-&AT-6&\cellcolor{blue!28}$-6.10\%$&\cellcolor{red!28}$5.81\%$&\cellcolor{red!100}$1247.75\%$&\cellcolor{red!100}$1918.14\%$&\cellcolor{blue!28}$-6.23\%$&\cellcolor{red!31}$6.34\%$\\
AT-8&-&AT-7&\cellcolor{blue!22}$-4.94\%$&\cellcolor{red!30}$6.12\%$&\cellcolor{red!100}$198.49\%$&\cellcolor{red!100}$529.85\%$&\cellcolor{blue!19}$-4.11\%$&\cellcolor{red!30}$6.26\%$\\
AT-8&-&AT-8&\cellcolor{blue!23}$-5.04\%$&\cellcolor{red!23}$4.71\%$&\cellcolor{blue!80}$-95.20\%$&\cellcolor{blue!80}$-94.23\%$&\cellcolor{blue!23}$-5.00\%$&\cellcolor{red!26}$5.29\%$\\
AT-8&-&AT-9&\cellcolor{blue!16}$-3.44\%$&\cellcolor{red!26}$5.35\%$&\cellcolor{red!100}$626.82\%$&\cellcolor{red!100}$1392.04\%$&\cellcolor{blue!14}$-3.11\%$&\cellcolor{red!28}$5.78\%$\\
\hline
AT-9&-&AT-1&\cellcolor{blue!6}$-1.30\%$&\cellcolor{red!32}$6.69\%$&\cellcolor{red!100}$75.29\%$&\cellcolor{red!100}$113.12\%$&\cellcolor{blue!20}$-4.44\%$&\cellcolor{red!34}$7.05\%$\\
AT-9&-&AT-2&\cellcolor{blue!13}$-2.89\%$&\cellcolor{red!46}$9.57\%$&\cellcolor{red!100}$746.32\%$&\cellcolor{red!100}$1362.15\%$&\cellcolor{blue!22}$-4.94\%$&\cellcolor{red!38}$7.84\%$\\
AT-9&-&AT-3&\cellcolor{blue!8}$-1.69\%$&\cellcolor{red!35}$7.31\%$&\cellcolor{red!100}$39.89\%$&\cellcolor{red!100}$58.26\%$&\cellcolor{blue!21}$-4.59\%$&\cellcolor{red!31}$6.47\%$\\
AT-9&-&AT-4&\cellcolor{blue!10}$-2.17\%$&\cellcolor{red!28}$5.81\%$&\cellcolor{red!100}$877.14\%$&\cellcolor{red!100}$1574.97\%$&\cellcolor{blue!23}$-5.05\%$&\cellcolor{red!30}$6.12\%$\\
AT-9&-&AT-5&\cellcolor{blue!4}$-0.74\%$&\cellcolor{red!41}$8.60\%$&\cellcolor{red!100}$807.26\%$&\cellcolor{red!100}$1361.14\%$&\cellcolor{blue!25}$-5.56\%$&\cellcolor{red!34}$7.08\%$\\
AT-9&-&AT-6&\cellcolor{blue!9}$-1.85\%$&\cellcolor{red!36}$7.55\%$&\cellcolor{red!100}$847.95\%$&\cellcolor{red!100}$1383.86\%$&\cellcolor{blue!24}$-5.36\%$&\cellcolor{red!31}$6.50\%$\\
AT-9&-&AT-7&\cellcolor{blue!9}$-1.82\%$&\cellcolor{red!40}$8.28\%$&\cellcolor{red!100}$1281.58\%$&\cellcolor{red!100}$2321.11\%$&\cellcolor{blue!24}$-5.23\%$&\cellcolor{red!43}$9.02\%$\\
AT-9&-&AT-8&\cellcolor{blue!12}$-2.53\%$&\cellcolor{red!33}$6.93\%$&\cellcolor{red!100}$740.15\%$&\cellcolor{red!100}$2247.42\%$&\cellcolor{blue!29}$-6.50\%$&\cellcolor{red!34}$7.07\%$\\
AT-9&-&AT-9&\cellcolor{blue!7}$-1.46\%$&\cellcolor{red!40}$8.33\%$&\cellcolor{blue!80}$-70.14\%$&\cellcolor{blue!80}$-63.50\%$&\cellcolor{blue!21}$-4.65\%$&\cellcolor{red!34}$7.10\%$
\end{tabular}
\end{tiny}
\end{center}
    \caption{Relative maximum differences between the origin destination data (Source \ref{src:internalMigrants}) and the simulations SC2 to SC4 between 2002 and 2024.}
    \label{tbl:deviations_od}
\end{table}

\subsubsection{Comparison with Source \ref{src:migrationForeacst} (Bevölkerungsbewegung 1961 bis 2100)}
For the forecast, we compare with Source \ref{src:migrationForeacst}, which contains the absolute number of internal immigrants and emigrants per federal-state. Since the numbers exclusively show migration beyond the federal-state border, we need to filter the simulation scenarios accordingly.

Table \ref{tbl:deviations_ieii_fc} shows the maximum positive and negative relative differences between the simulation scenarios and the forecasts. Clearly, SC3 cannot be validated with this strategy, since the biregional model does not preserve the origin-destination flows (which includes the flow from one region into itself).

\begin{table}
    \begin{center}
    \begin{scriptsize}
        \begin{tabular}{ccc|cc|cc|cc}
        region (origin) & sex & region (destination) & \multicolumn{2}{c|}{\textbf{SC2} ($e_{min},e_{max}$)} & \multicolumn{2}{c}{\textbf{SC3} ($e_{min},e_{max}$)}&\multicolumn{2}{c}{\textbf{SC4} ($e_{min},e_{max}$)}\\
-&-&-&\cellcolor{blue!10}$-0.50\%$&\cellcolor{red!22}$6.43\%$&\cellcolor{red!100}$383.39\%$&\cellcolor{red!100}$428.51\%$&\cellcolor{red!3}$0.66\%$&\cellcolor{red!26}$7.67\%$\\
\hline
AT-1&-&-&\cellcolor{blue!57}$-2.85\%$&\cellcolor{red!10}$2.96\%$&\cellcolor{red!100}$215.92\%$&\cellcolor{red!100}$230.58\%$&\cellcolor{red!25}$7.42\%$&\cellcolor{red!39}$11.63\%$\\
AT-2&-&-&\cellcolor{blue!6}$-0.27\%$&\cellcolor{red!71}$21.02\%$&\cellcolor{red!100}$424.57\%$&\cellcolor{red!100}$539.93\%$&\cellcolor{red!13}$3.67\%$&\cellcolor{red!88}$26.25\%$\\
AT-3&-&-&\cellcolor{blue!29}$-1.42\%$&\cellcolor{red!11}$3.07\%$&\cellcolor{red!100}$246.24\%$&\cellcolor{red!100}$265.24\%$&\cellcolor{red!13}$3.63\%$&\cellcolor{red!26}$7.64\%$\\
AT-4&-&-&\cellcolor{blue!26}$-1.29\%$&\cellcolor{red!37}$11.04\%$&\cellcolor{red!100}$565.43\%$&\cellcolor{red!100}$657.30\%$&\cellcolor{red!1}$0.00\%$&\cellcolor{red!46}$13.71\%$\\
AT-5&-&-&\cellcolor{blue!66}$-3.30\%$&\cellcolor{red!46}$13.53\%$&\cellcolor{red!100}$314.26\%$&\cellcolor{red!100}$395.55\%$&\cellcolor{blue!21}$-1.00\%$&\cellcolor{red!54}$16.14\%$\\
AT-6&-&-&\cellcolor{blue!38}$-1.85\%$&\cellcolor{red!42}$12.33\%$&\cellcolor{red!100}$602.45\%$&\cellcolor{red!100}$699.32\%$&\cellcolor{blue!1}$-0.02\%$&\cellcolor{red!44}$13.10\%$\\
AT-7&-&-&\cellcolor{blue!25}$-1.22\%$&\cellcolor{red!23}$6.79\%$&\cellcolor{red!100}$734.73\%$&\cellcolor{red!100}$787.40\%$&\cellcolor{blue!24}$-1.15\%$&\cellcolor{red!21}$6.27\%$\\
AT-8&-&-&\cellcolor{blue!29}$-1.41\%$&\cellcolor{red!9}$2.65\%$&\cellcolor{red!100}$1060.59\%$&\cellcolor{red!100}$1103.24\%$&\cellcolor{blue!49}$-2.43\%$&\cellcolor{red!7}$2.04\%$\\
AT-9&-&-&\cellcolor{red!2}$0.48\%$&\cellcolor{red!10}$2.81\%$&\cellcolor{red!100}$286.22\%$&\cellcolor{red!100}$303.82\%$&\cellcolor{blue!64}$-3.15\%$&\cellcolor{blue!1}$-0.02\%$\\
\hline
AT-1&-&-&\cellcolor{red!1}$0.08\%$&\cellcolor{red!17}$5.09\%$&\cellcolor{red!100}$160.44\%$&\cellcolor{red!100}$191.19\%$&\cellcolor{red!2}$0.51\%$&\cellcolor{red!18}$5.31\%$\\
AT-2&-&-&\cellcolor{blue!54}$-2.66\%$&\cellcolor{red!25}$7.48\%$&\cellcolor{red!100}$411.59\%$&\cellcolor{red!100}$497.17\%$&\cellcolor{blue!33}$-1.64\%$&\cellcolor{red!24}$6.98\%$\\
AT-3&-&-&\cellcolor{red!3}$0.79\%$&\cellcolor{red!16}$4.53\%$&\cellcolor{red!100}$193.21\%$&\cellcolor{red!100}$228.69\%$&\cellcolor{blue!21}$-1.02\%$&\cellcolor{red!11}$3.06\%$\\
AT-4&-&-&\cellcolor{blue!72}$-3.60\%$&\cellcolor{red!20}$5.97\%$&\cellcolor{red!100}$594.18\%$&\cellcolor{red!100}$633.66\%$&\cellcolor{blue!44}$-2.18\%$&\cellcolor{red!26}$7.73\%$\\
AT-5&-&-&\cellcolor{blue!46}$-2.27\%$&\cellcolor{red!29}$8.52\%$&\cellcolor{red!100}$374.30\%$&\cellcolor{red!100}$419.50\%$&\cellcolor{blue!24}$-1.19\%$&\cellcolor{red!31}$9.03\%$\\
AT-6&-&-&\cellcolor{blue!41}$-2.00\%$&\cellcolor{red!25}$7.43\%$&\cellcolor{red!100}$572.75\%$&\cellcolor{red!100}$626.84\%$&\cellcolor{blue!2}$-0.07\%$&\cellcolor{red!33}$9.83\%$\\
AT-7&-&-&\cellcolor{blue!63}$-3.13\%$&\cellcolor{red!26}$7.62\%$&\cellcolor{red!100}$929.89\%$&\cellcolor{red!100}$1039.90\%$&\cellcolor{blue!38}$-1.87\%$&\cellcolor{red!28}$8.12\%$\\
AT-8&-&-&\cellcolor{blue!40}$-1.97\%$&\cellcolor{red!25}$7.43\%$&\cellcolor{red!100}$1056.88\%$&\cellcolor{red!100}$1138.04\%$&\cellcolor{blue!44}$-2.15\%$&\cellcolor{red!25}$7.34\%$\\
AT-9&-&-&\cellcolor{blue!2}$-0.08\%$&\cellcolor{red!26}$7.50\%$&\cellcolor{red!100}$336.21\%$&\cellcolor{red!100}$373.57\%$&\cellcolor{red!13}$3.76\%$&\cellcolor{red!39}$11.56\%$
\end{tabular}
\end{scriptsize}
\end{center}
    \caption{Relative maximum differences between the internal migration forecasts (Source \ref{src:migrationForeacst}) and the simulations SC2 to SC4 between 2025 and 2049. Only internal migrants into different regions are counted.}
    \label{tbl:deviations_ieii_fc}
\end{table}

\FloatBarrier

\subsection{Additional Validation Scenarios}
\label{sec:validation_special}
In this final validation step, we will define several scenarios to evaluate very specific features about GEPOC ABM IM, using SC2 as reference scenario. First of all, SC5 will use [2000,2100] as simulation time-frame. This will increase the simulation duration from 50 to 100 years and we expect larger errors to the validation data. SC6 will be run with monthly steps (instead of yearly). We expect that, due to the concept with planning for birth-day to birth-day, the influence of the time-step parameter is minimal. Finally, SC7 will use a scale factor $\sigma=0.1$ instead of $1$. We expect that the (up-scaled) mean values still matches with the data. 
%SC8 will use the raw-parameters without correcting bias caused by simultaneous events. We expect that agent-events are sampled slightly not often enough. 
The scenarios are summarised in Table \ref{tbl:parametrisation_special}.

\begin{table}[h]
\begin{center}
    \begin{tabular}{p{4cm}c|c|ccc}
     & & \textbf{SC2} & \textbf{SC5} & \textbf{SC6} & \textbf{SC7}  \\
     \hline
sim start-time  &$t_0$& 2000-01-01 & 2000-01-01 & 2000-01-01 & 2000-01-01 \\
time-step lengths &  & years & years & months & years\\
sim end-time & $t_e$ & 2050-01-01 & 2100-01-01 & 2050-01-01 & 2050-01-01 \\
scale & $\sigma$ & $1.0$ & $1.0$ & $1.0$ & $0.1$ \\
Monthe-Carlo runs & & 9 & 9 & 9 & 27 
\end{tabular}
\end{center}
\caption{Parametrisation setup for simulation scenarios SC5 - SC7. All other parameters are equal to the SC2 scenario.}
\label{tbl:parametrisation_special}
\end{table}

\subsubsection{Simulation Scenario SC5 (Long Run).}

\begin{figure}
    \centering
    \includegraphics[width=\linewidth]{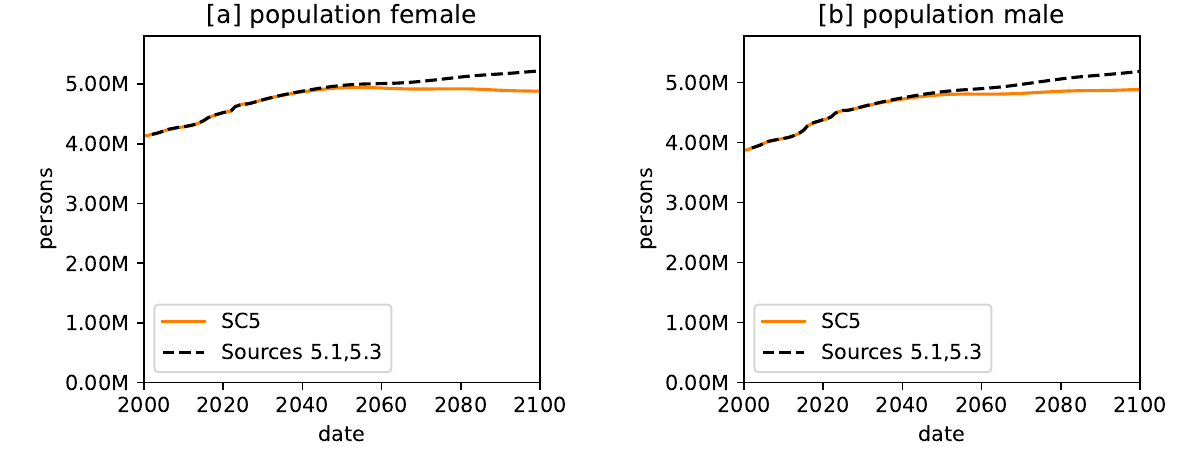}
    \caption{Comparison between the population data and forecast (Sources \ref{src:popBase} and \ref{src:popForecast}) and the simulation scenario SC5.}
    \label{fig:pop_long}
\end{figure}

\begin{figure}
    \centering
    \includegraphics[width=\linewidth]{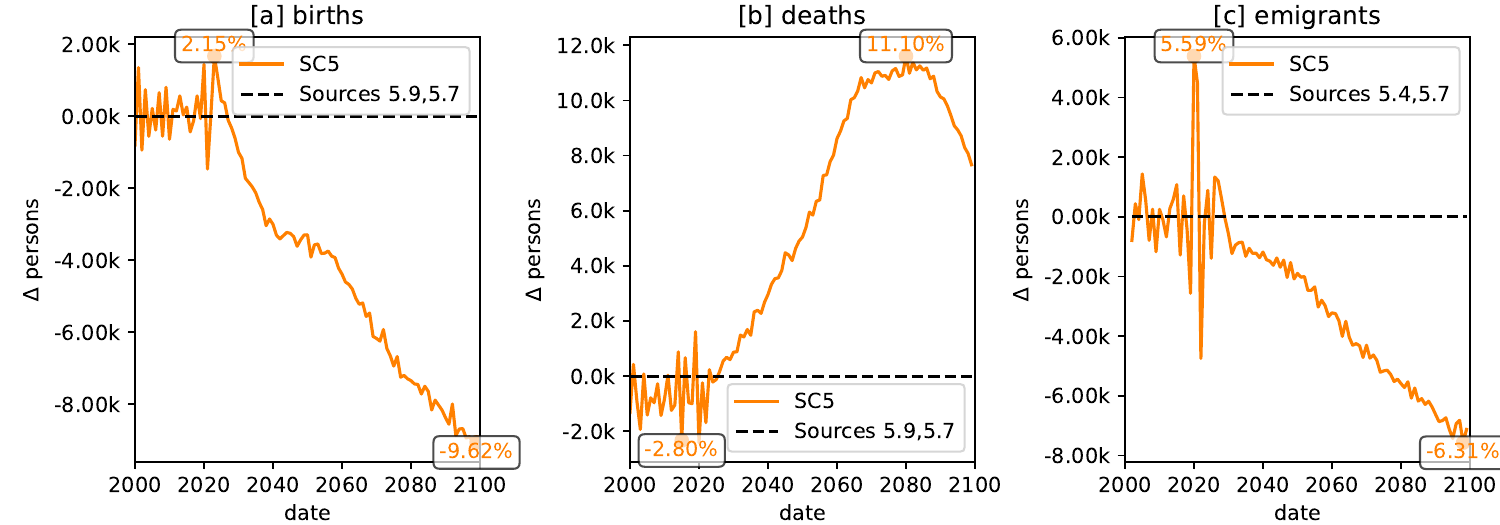}
    \caption{Differences between the different data and forecasts for births, deaths and emigrants and the corresponding values in simulation scenario SC5.}
    \label{fig:allThree_diff}
\end{figure}

The long simulation scenario between 2003 and 2100 shows a reasonably good fit with the population data until about year 2060. Thereafter, the simulated numbers underestimate the Statistics Austria forecast from Source \ref{src:popForecast} (see Figure \ref{fig:pop_long}). In 2100, the male and female population lie around $5\%$ below the reference. The underestimation originates from an underestimated net-balance between births, deaths and emigrants between 2030 and 2100. As seen in Figure \ref{fig:allThree_diff}, births are underestimated by up to $10\%$ (around 8500 births per year), deaths are overestimated by up to $11\%$ (about 10000 deaths in 2080), and emigrants are underestimated by up to $6\%$ (about 8000 emigrants in 2100). While deviations from deaths and emigrants roughly cancel out, the underestimated number of births causes the population to continuously drop below the reference. The origin of this problem lies in the assumptions made for the computation of the forecast from the low-resolution data.

\subsubsection{Simulation Scenarios SC6 (Monthly Steps) and SC7 (Scale 0.1).}
\begin{figure}
    \centering
    \includegraphics[width=0.9\linewidth]{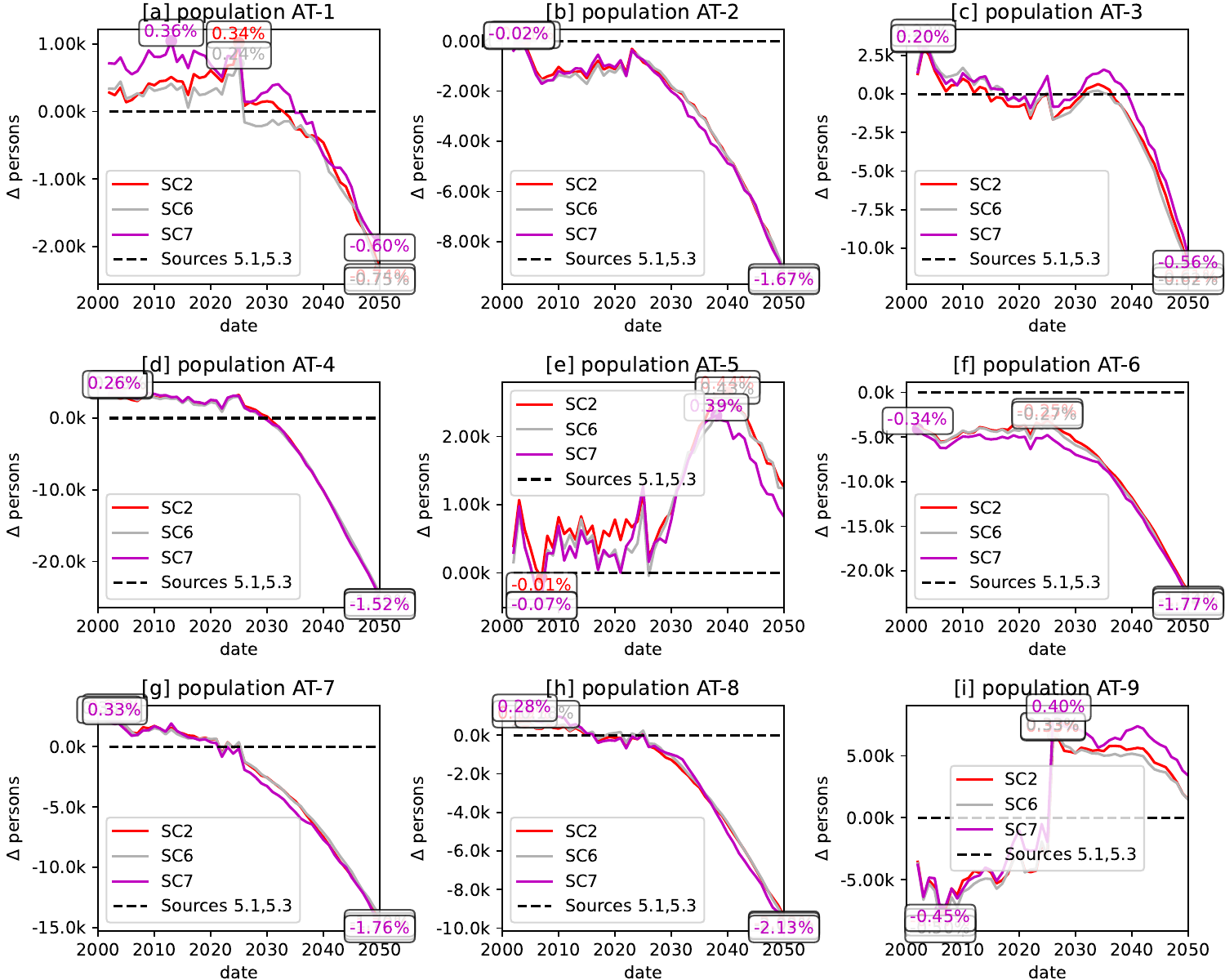}
    \caption{Differences between the population data/forecast (Source \ref{src:popBase}, \ref{src:popForecast}) and simulation scenarios SC2, SC6 and SC7 for the nine federal-states of Austria.}
    \label{fig:population_fed_diff_special}
\end{figure}

For both simulation scenarios, SC6 and SC7, we expect that the numbers only deviate minimally from SC2, if aggregated/scaled accordingly. That means, births, deaths, etc. must be aggregated from months to years in SC6, and numbers have to be scaled up by $1/\sigma=10$ in SC7. Figure \ref{fig:population_fed_diff_special} shows the differences between SC2, SC6 and SC7 to the reference data and forecast for the population for the nine federal-states of Austria. The upscaled results from scenario SC7 hardly differ from SC2. Since the base version of GEPOC ABM does not involve interaction between the agents, i.e. the agents can be expressed as independent stochastic processes, this behaviour was expected by direct application of the Law of Large Numbers. Scenario SC6, which uses different time-step lengths, shows small differences to SC2. This is a result of how newly created agents are added in GEPOC. Both immigrants and newborns are added to the dynamics of the model at the end of the discrete timesteps, independent of their scheduled immigration-/birth-date. Given the results shown in Figure \ref{fig:population_fed_diff_special}, there is no clear answer to the question of whether reduced time-step lengths improve or harm the validity of the results. However, this confirms the idea that the quality of the GEPOC ABM (IM) simulations does not (much) depend on the step-lengths.

\FloatBarrier

\newpage
\section*{Funding}
The methods presented are the result of many years of research in the field of population modelling and its application (e.g. epidemiology, logistics, etc.). We would therefore like to express our gratitude to several funding agencies, whose support for research projects has enabled important contributions to the methods presented in this work:
\begin{itemize}
    \item Austrian Research Promotion Agency (FFG), projects POPSICLE (FO999918405, 2024-2026), CIDS (881665, 2020-2021), AundO (867178, 2018-2020), DEXHELPP (843550, 2014-2018)
    \item Austrian Science Fund (FWF), project DynOptTestControl (I 5908-G,2022-2026)
    \item Vienna Science and Technology Fund (WWTF), projects BETTER (EICOV20028, 2023-2026), HPVienna (2025-2027), Syd19 (COV20-035,2020)
    \item Society for Medical Decision Making (SMDM), project TAV-COVID (gbmf9634, 2020-2021)
\end{itemize}
\bibliographystyle{plain}
\bibliography{References}

\newpage
\appendix 
\section{Appendix}
\label{sec:appendix}
\FloatBarrier
\subsection{Validation Plots - Population}
\begin{figure}
\centering
\includegraphics[width=0.7\linewidth]{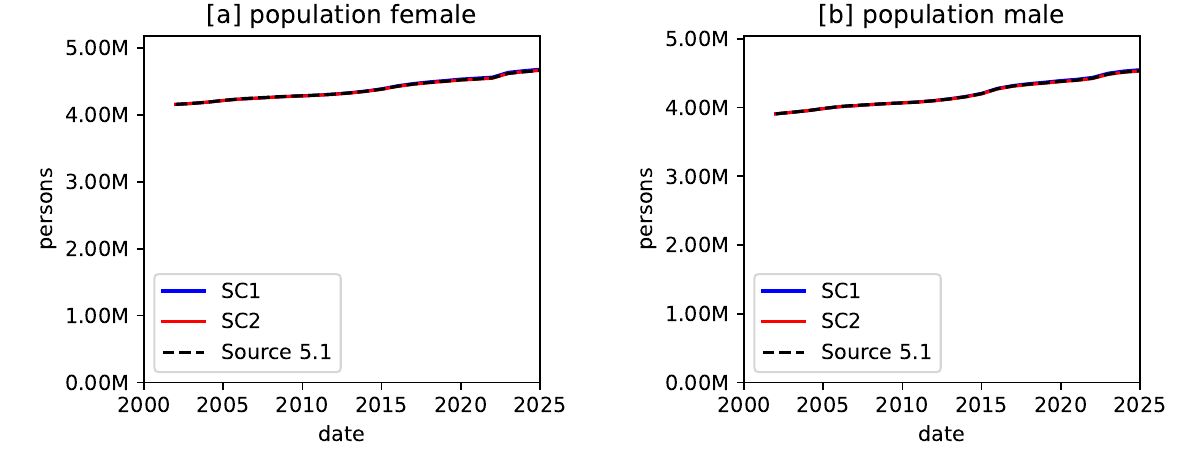}
\caption{Comparison between population data from Source \ref{src:popBase} and the simulation scenarios SC1 and SC2 for male and female persons.}
\label{fig:population_sex}
\end{figure}
\begin{figure}
\centering
\includegraphics[width=0.7\linewidth]{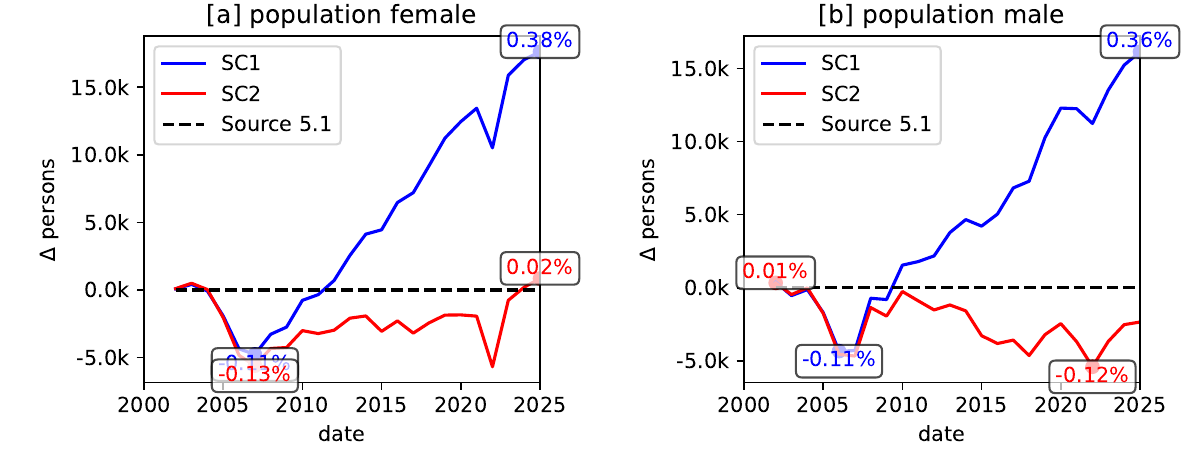}
\caption{Differences between population data from Source \ref{src:popBase} and the simulation scenarios SC1 and SC2 for male and female persons.}
\label{fig:population_sex_diff_b}
\end{figure}
\begin{figure}
\centering
\includegraphics[width=\linewidth]{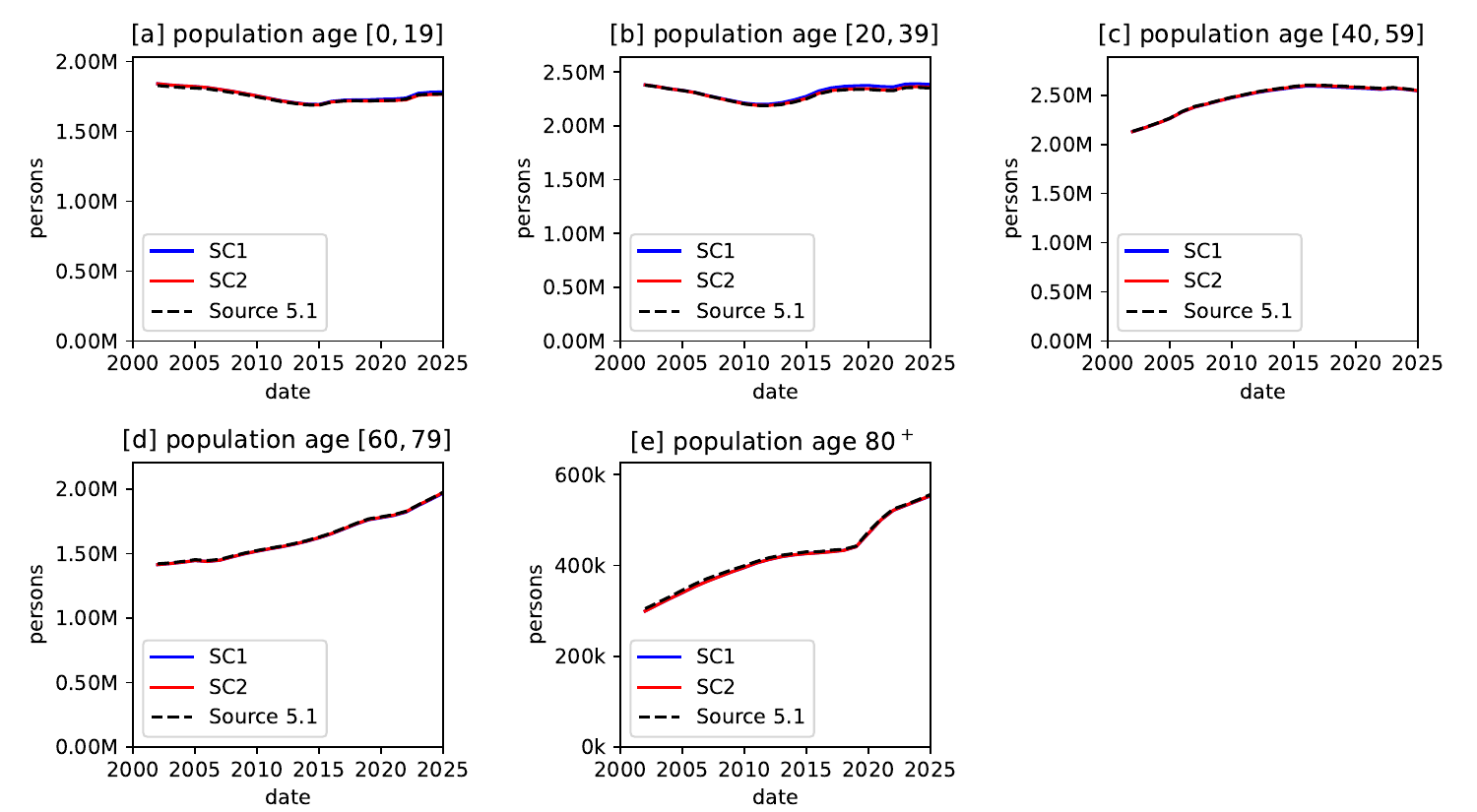}
\caption{Comparison between population data from Source \ref{src:popBase} and the simulation scenarios SC1 and SC2 for different age cohorts.}
\label{fig:population_age_b}
\end{figure}
\begin{figure}
\centering
\includegraphics[width=\linewidth]{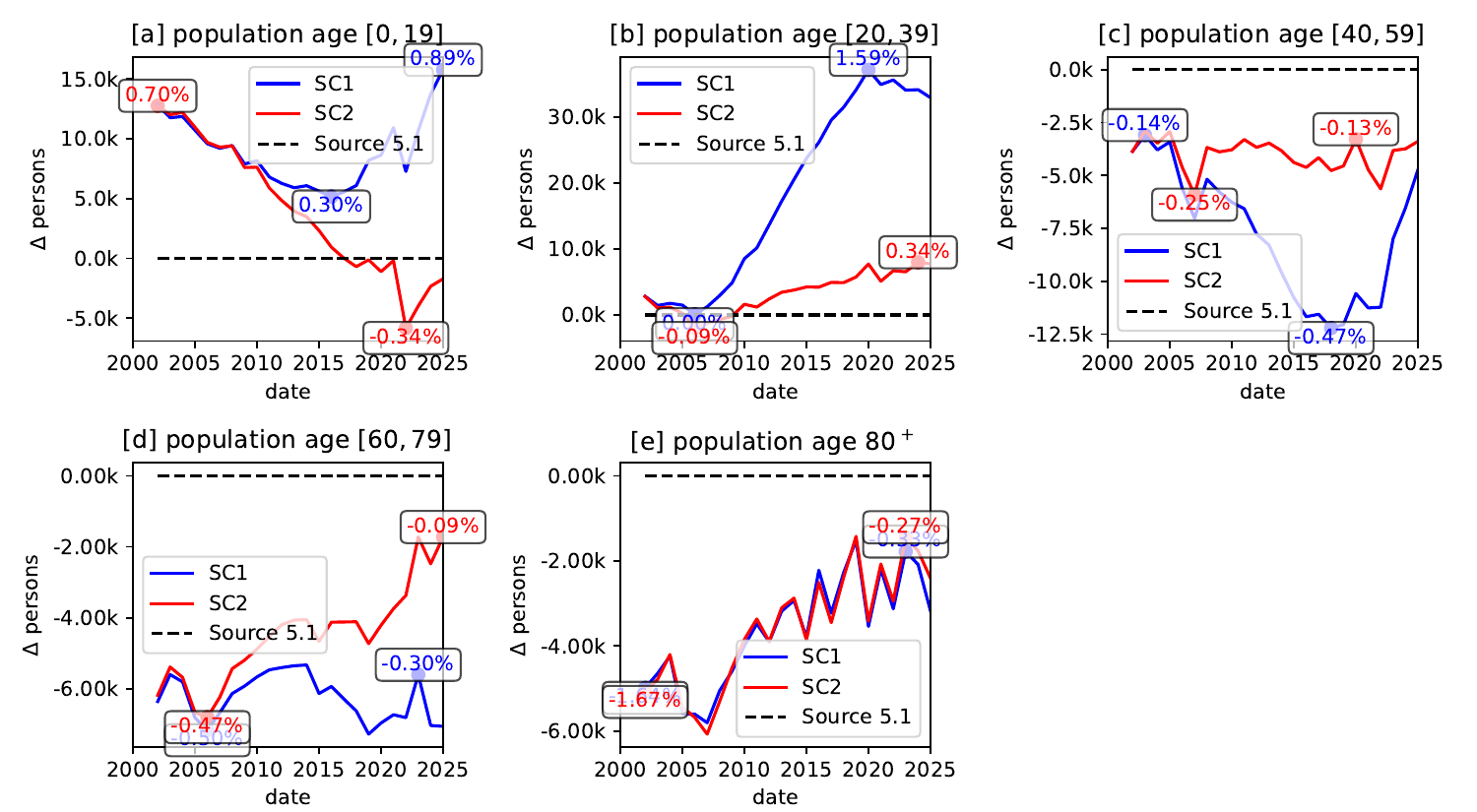}
\caption{Differences between population data from Source \ref{src:popBase} and the simulation scenarios SC1 and SC2 for different age cohorts.}
\label{fig:population_age_diff_b}
\end{figure}
\begin{figure}
\centering
\includegraphics[width=\linewidth]{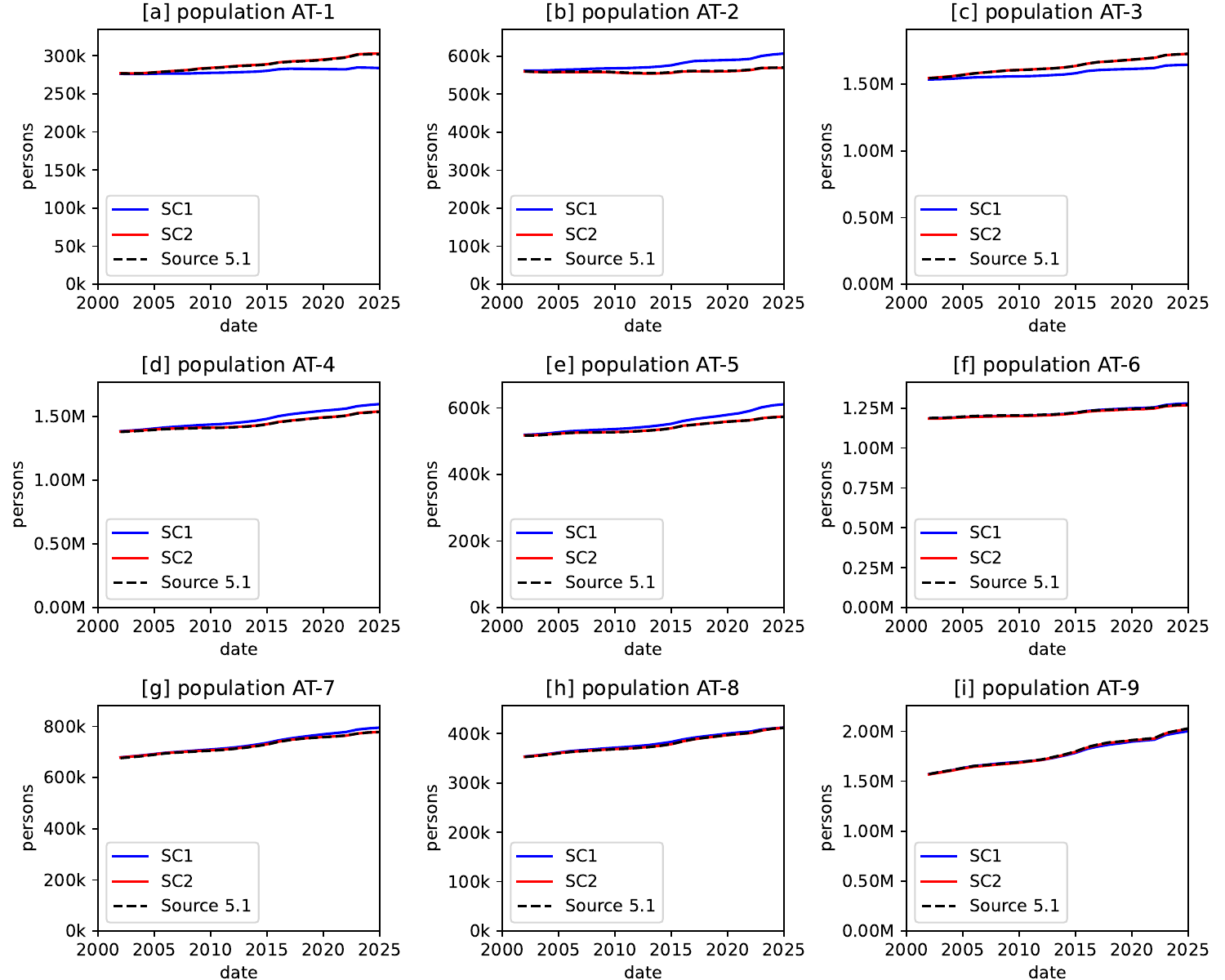}
\caption{Comparison between population data from Source \ref{src:popBase} and the simulation scenarios SC1 and SC2 for the nine federal-states of Austria.}
\label{fig:population_fed}
\end{figure}
\begin{figure}
\centering
\includegraphics[width=\linewidth]{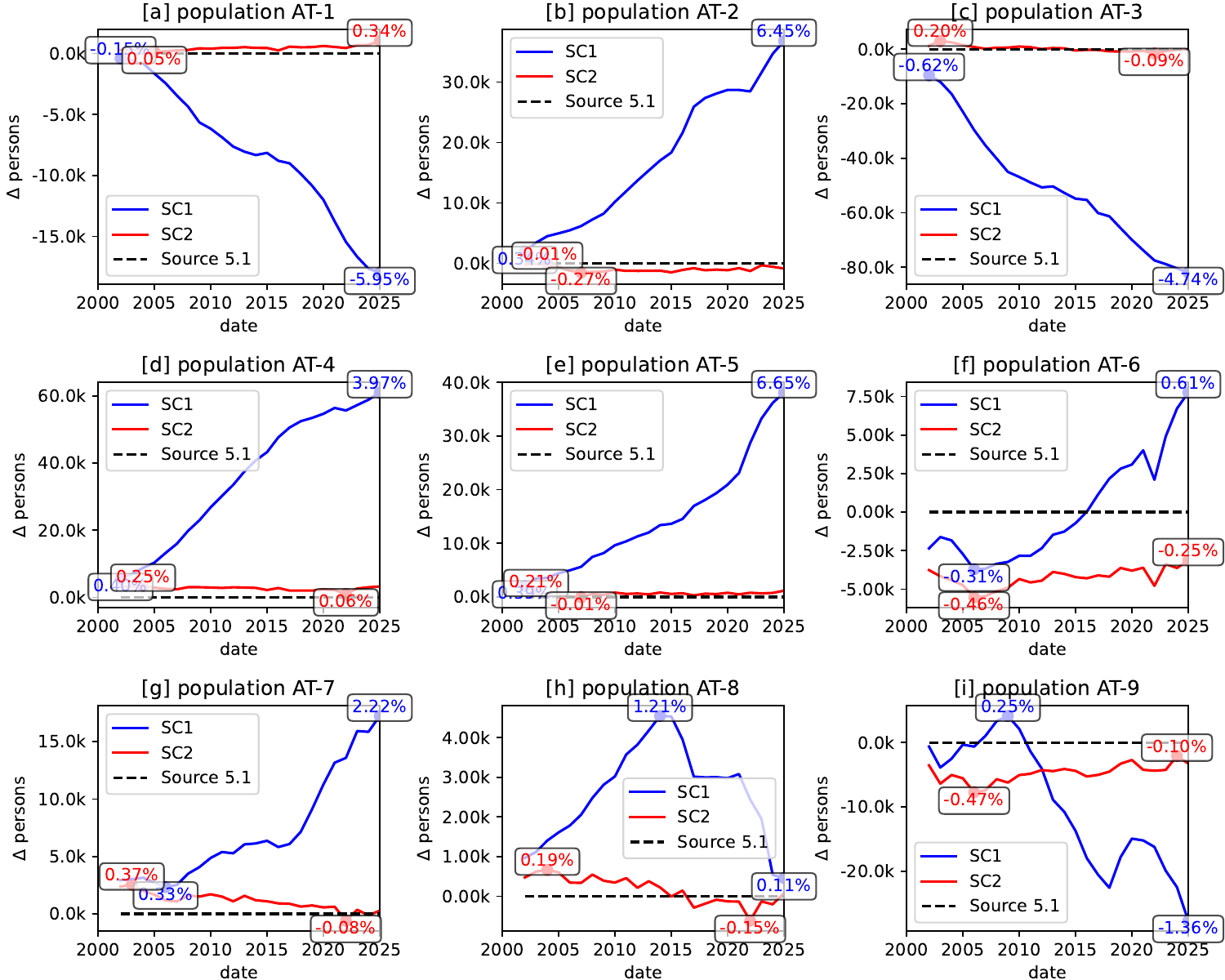}
\caption{Differences between population data from Source \ref{src:popBase} and the simulation scenarios SC1 and SC2 for the nine federal-states of Austria.}
\label{fig:population_fed_diff_b}
\end{figure}
\begin{figure}
\centering
\includegraphics[width=0.7\linewidth]{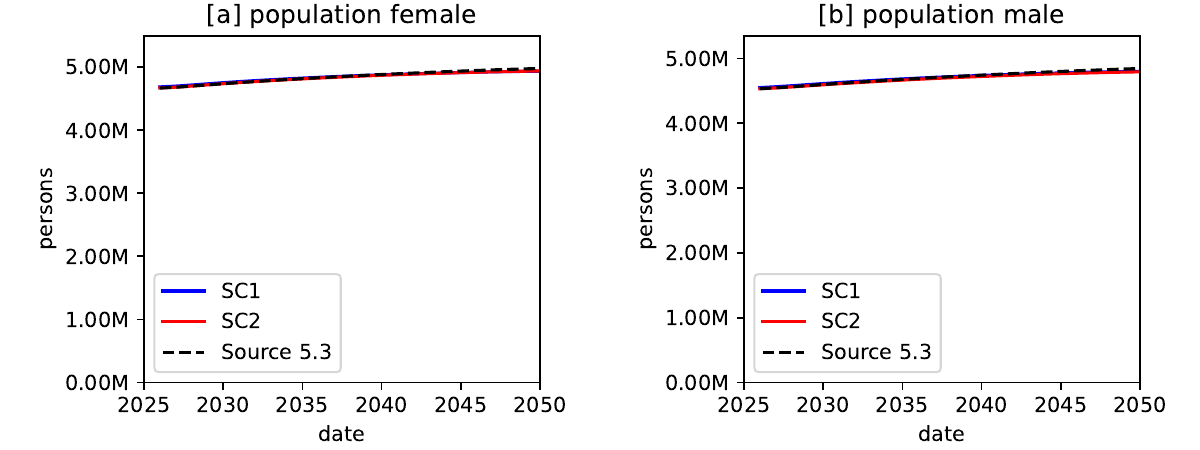}
\caption{Comparison between population forecast from Source \ref{src:popForecast} and the simulation scenarios SC1 and SC2 for male and female persons.}
\label{fig:population_fc_sex}
\end{figure}
\begin{figure}
\centering
\includegraphics[width=0.7\linewidth]{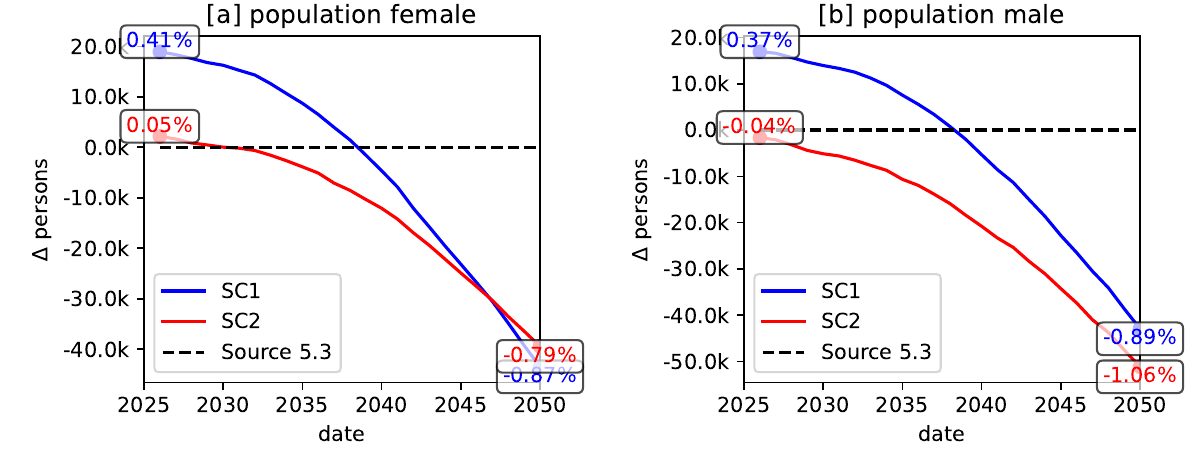}
\caption{Differences between population forecast from Source \ref{src:popForecast} and the simulation scenarios SC1 and SC2 for male and female persons.}
\label{fig:population_fc_sex_diff_b}
\end{figure}
\begin{figure}
\centering
\includegraphics[width=\linewidth]{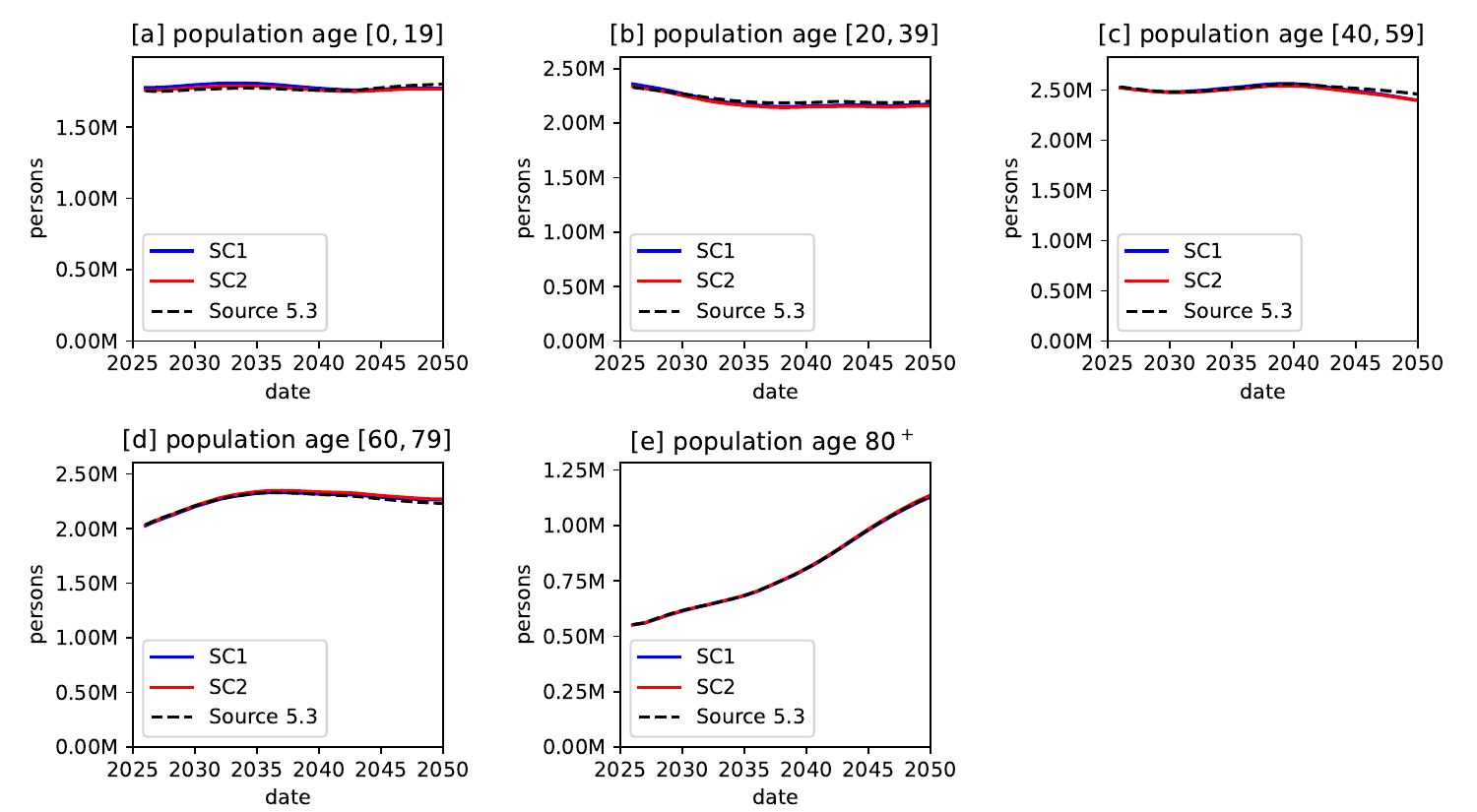}
\caption{Comparison between population forecast from Source \ref{src:popForecast} and the simulation scenarios SC1 and SC2 for different age cohorts.}
\label{fig:population_fc_age}
\end{figure}
\begin{figure}
\centering
\includegraphics[width=\linewidth]{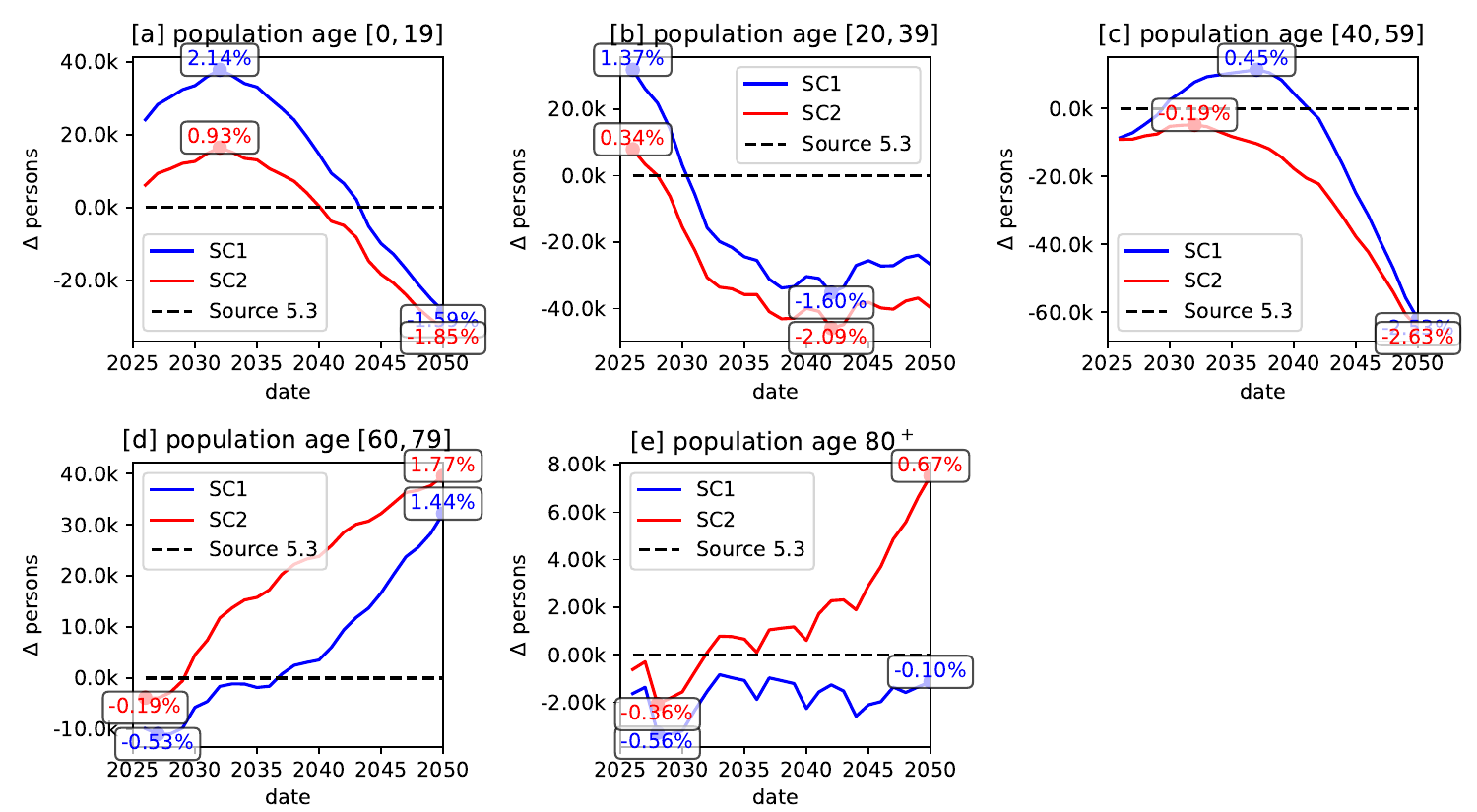}
\caption{Differences between population forecast from Source \ref{src:popForecast} and the simulation scenarios SC1 and SC2 for different age cohorts.}
\label{fig:population_fc_age_diff_b}
\end{figure}
\begin{figure}
\centering
\includegraphics[width=\linewidth]{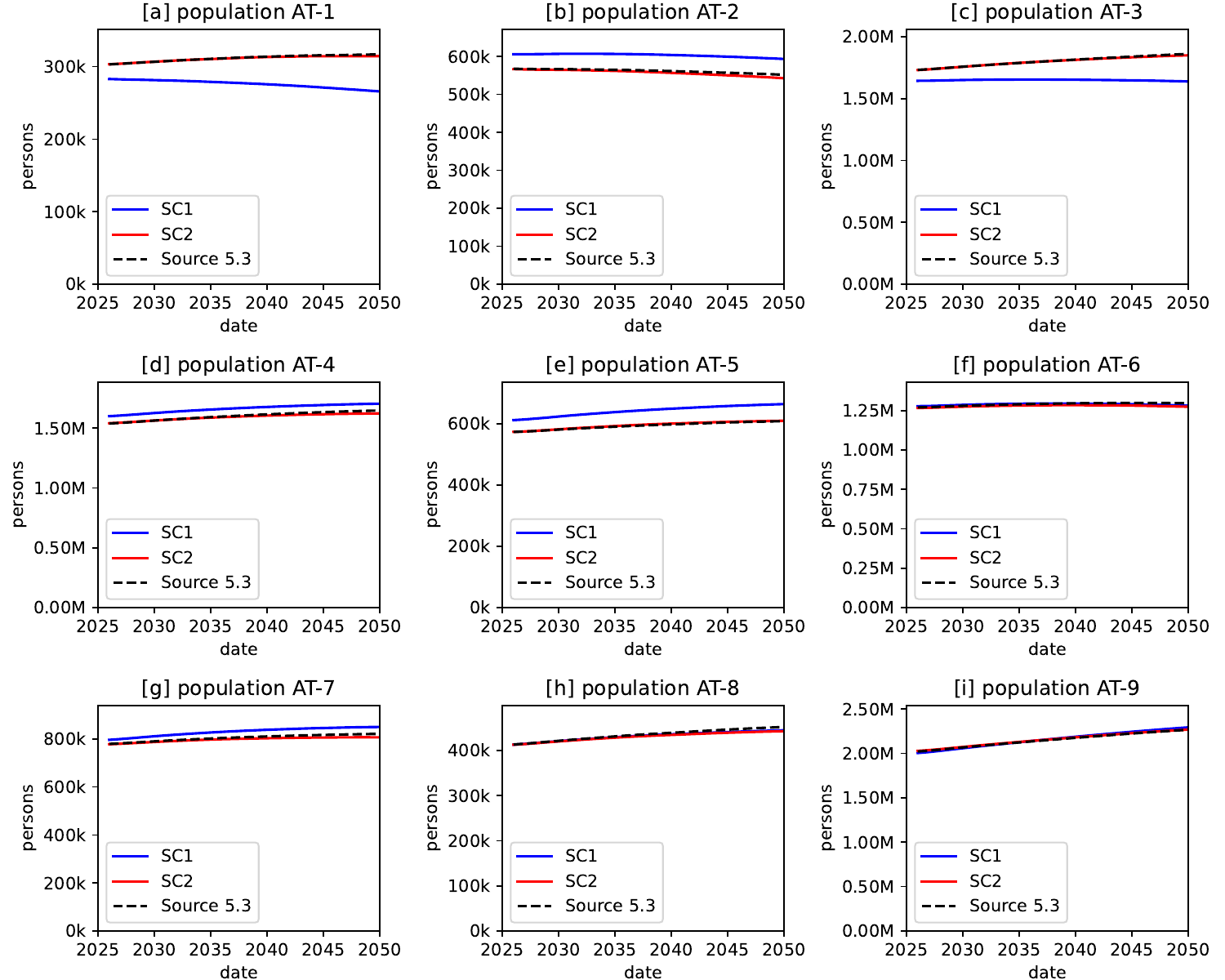}
\caption{Comparison between population forecast from Source \ref{src:popForecast} and the simulation scenarios SC1 and SC2 for the nine federal-states of Austria.}
\label{fig:population_fc_fed}
\end{figure}
\begin{figure}
\centering
\includegraphics[width=\linewidth]{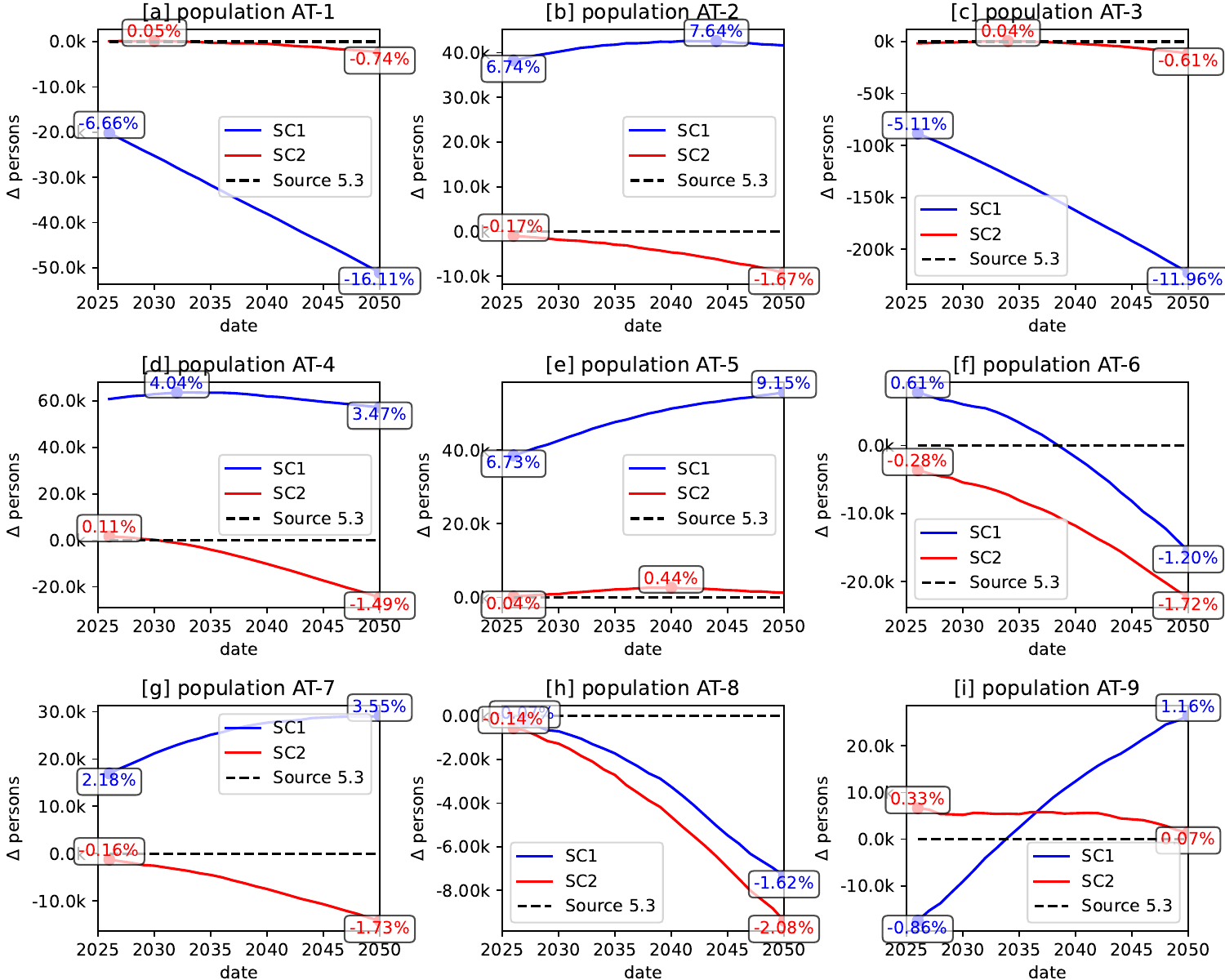}
\caption{Differences between population forecast from Source \ref{src:popForecast} and the simulation scenarios SC1 and SC2 for the nine federal-states of Austria.}
\label{fig:population_fc_fed_diff_b}
\end{figure}
\FloatBarrier
\subsection{Validation Plots - Births}
\begin{figure}
\centering
\includegraphics[width=0.7\linewidth]{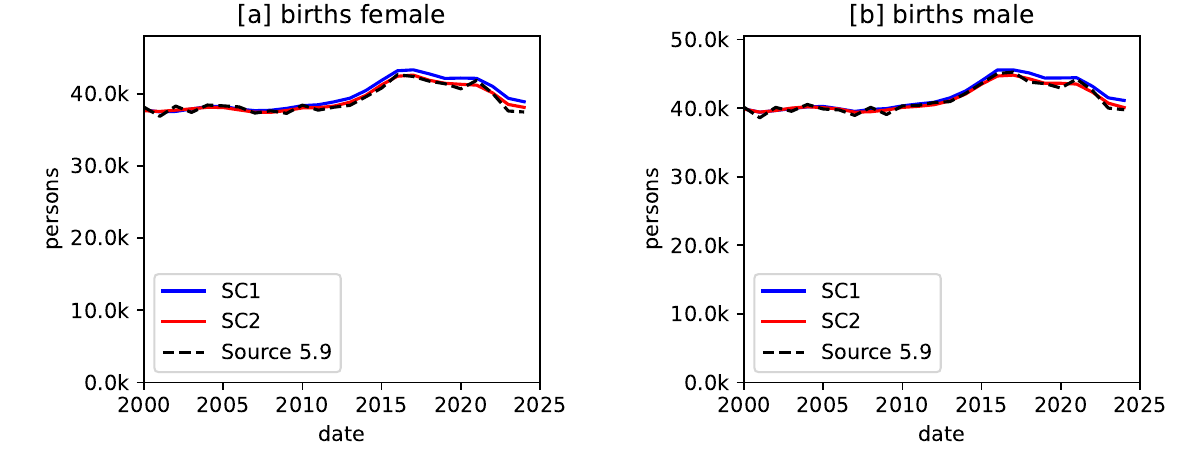}
\caption{Comparison between births data from Source \ref{src:indicators} and the simulation scenarios SC1 and SC2 for male and female persons.}
\label{fig:births_sex}
\end{figure}
\begin{figure}
\centering
\includegraphics[width=0.7\linewidth]{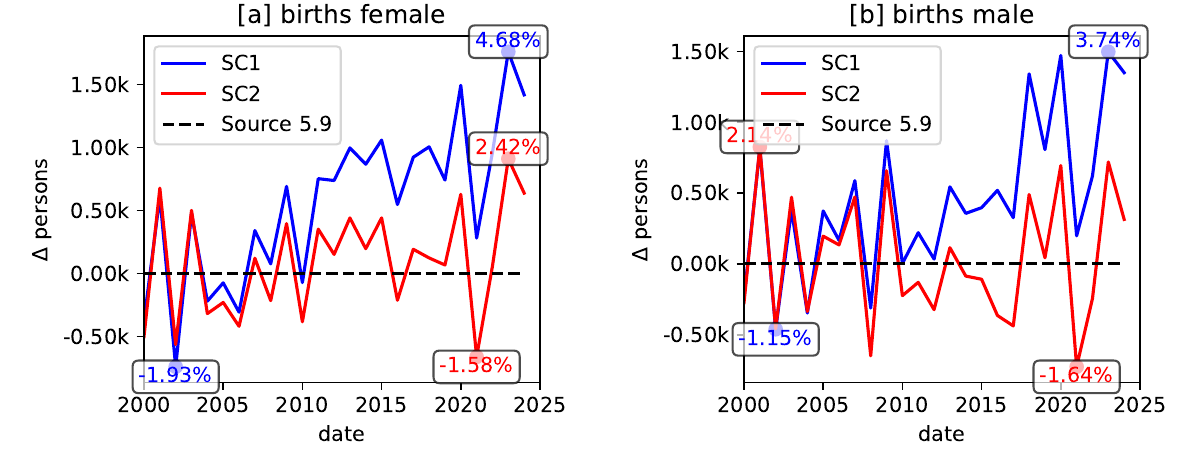}
\caption{Differences between births data from Source \ref{src:indicators} and the simulation scenarios SC1 and SC2 for male and female persons.}
\label{fig:births_sex_diff_b}
\end{figure}
\begin{figure}
\centering
\includegraphics[width=\linewidth]{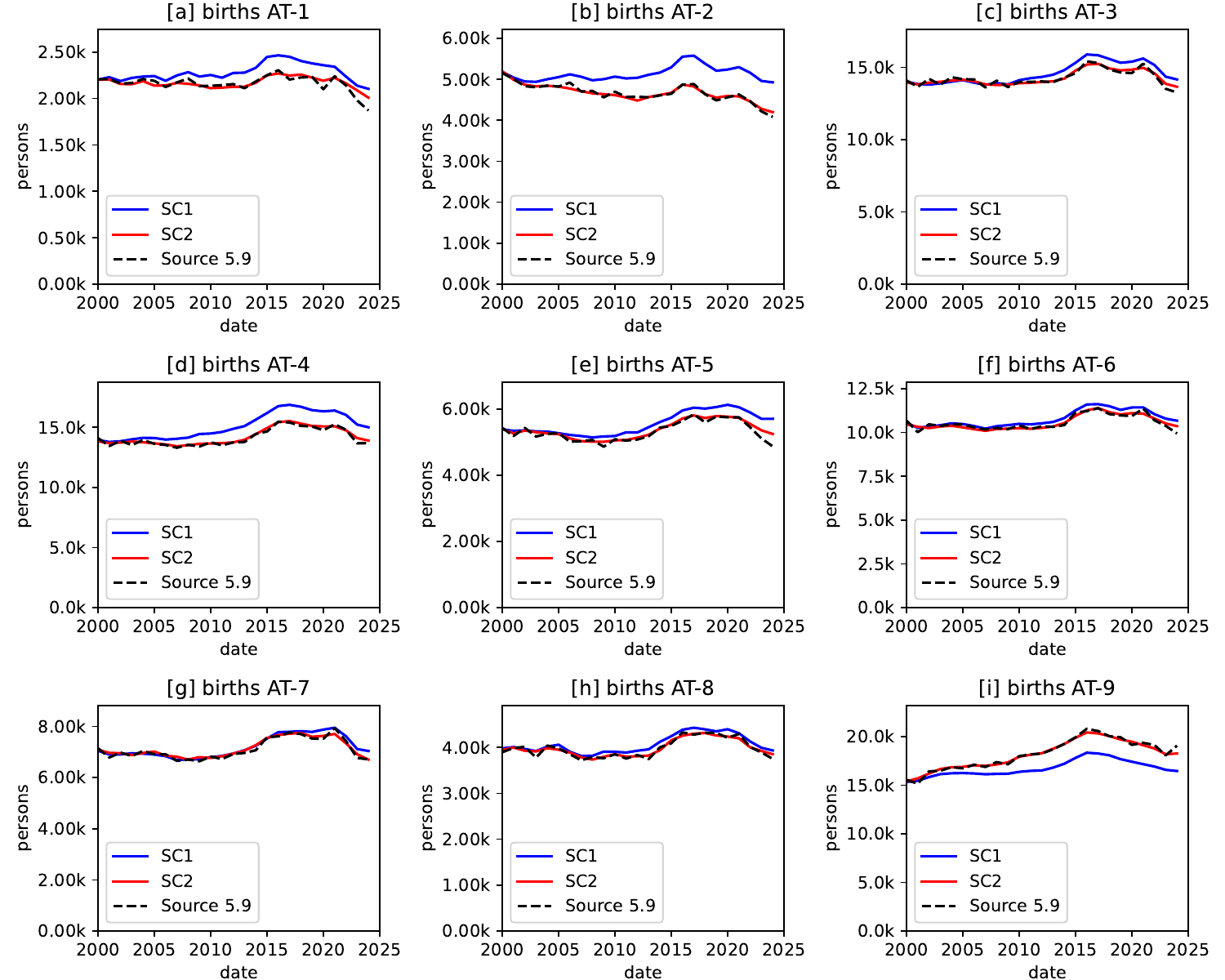}
\caption{Comparison between births data from Source \ref{src:indicators} and the simulation scenarios SC1 and SC2 for the nine federal-states of Austria.}
\label{fig:births_fed}
\end{figure}
\begin{figure}
\centering
\includegraphics[width=\linewidth]{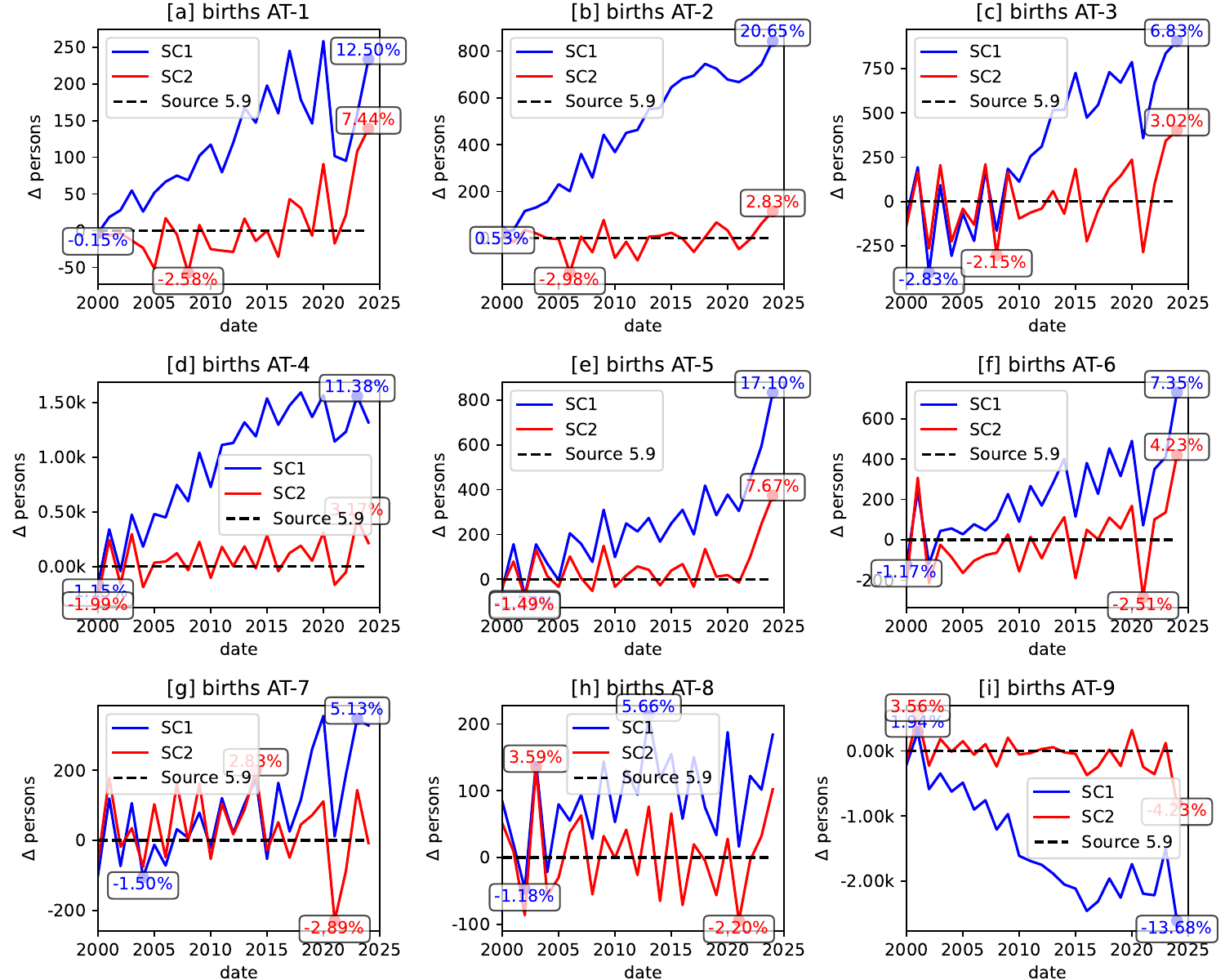}
\caption{Differences between births data from Source \ref{src:indicators} and the simulation scenarios SC1 and SC2 for the nine federal-states of Austria.}
\label{fig:births_fed_diff}
\end{figure}
\begin{figure}
\centering
\includegraphics[width=\linewidth]{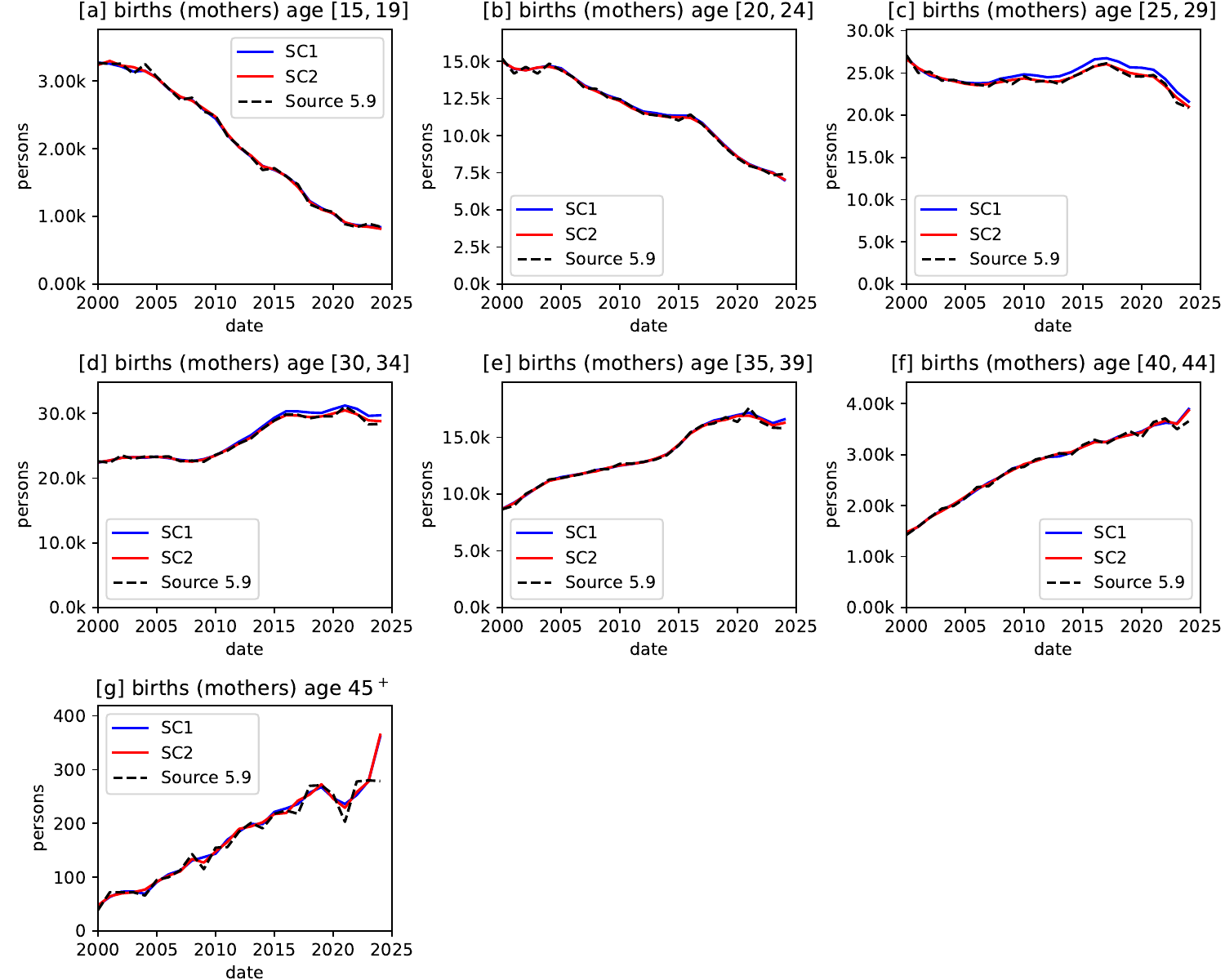}
\caption{Comparison between births data from Source \ref{src:indicators} and the simulation scenarios SC1 and SC2 for different age cohorts.}
\label{fig:births_age}
\end{figure}
\begin{figure}
\centering
\includegraphics[width=\linewidth]{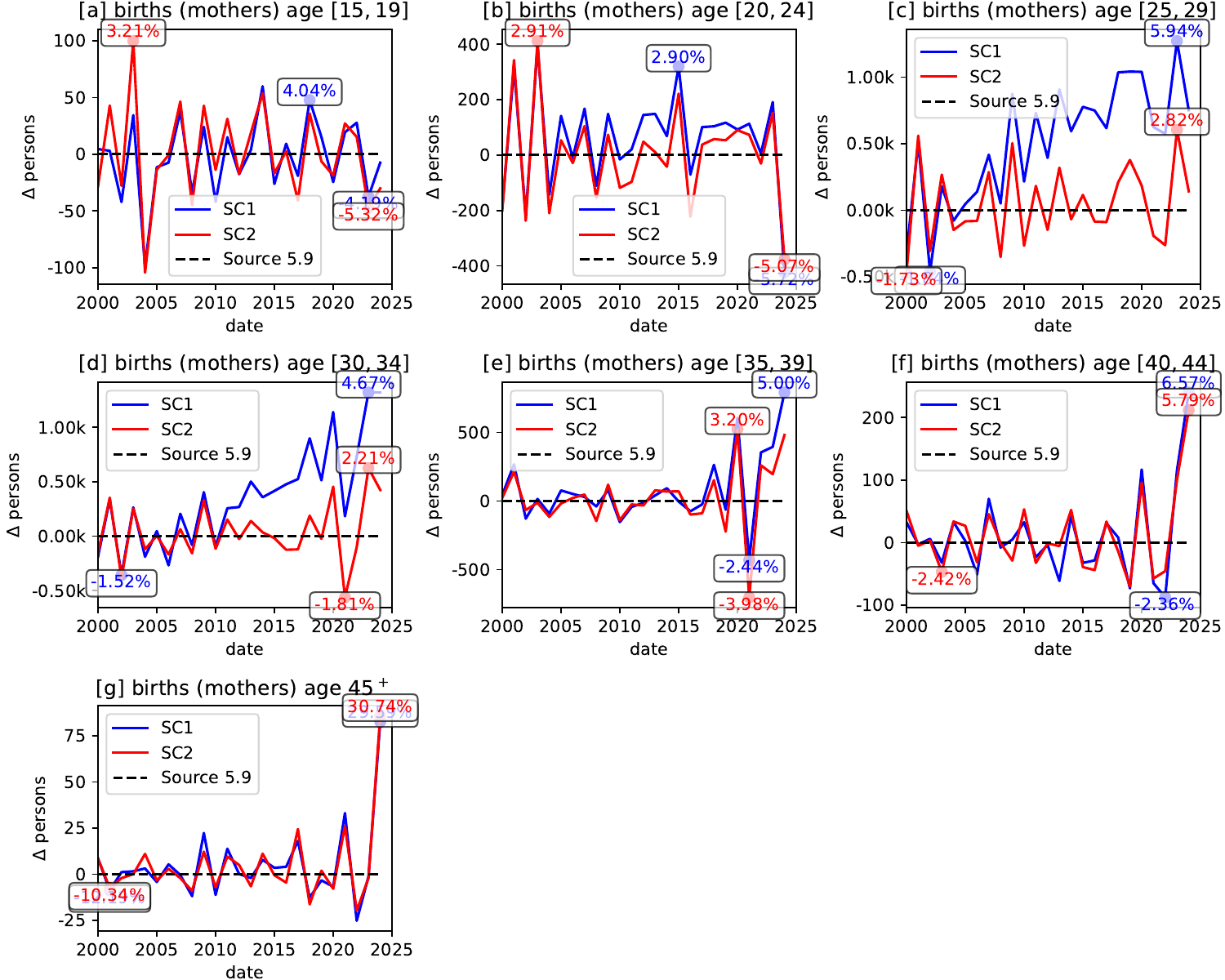}
\caption{Differences between births data from Source \ref{src:indicators} and the simulation scenarios SC1 and SC2 for different age cohorts.}
\label{fig:births_age_diff_b}
\end{figure}
\begin{figure}
\centering
\includegraphics[width=\linewidth]{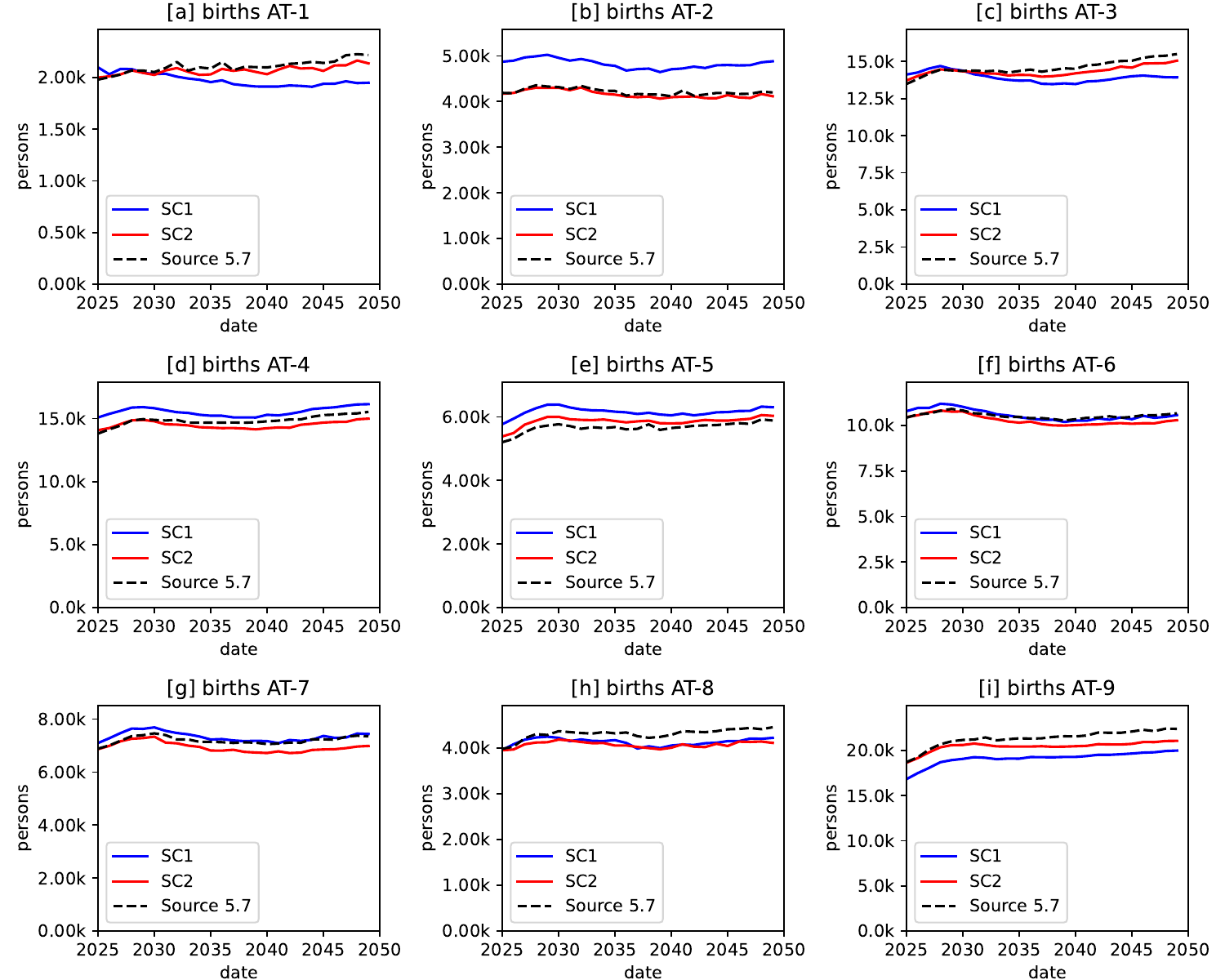}
\caption{Comparison between births forecast from Source \ref{src:migrationForeacst} and the simulation scenarios SC1 and SC2 for the nine federal-states of Austria.}
\label{fig:births_fc_fed}
\end{figure}
\begin{figure}
\centering
\includegraphics[width=\linewidth]{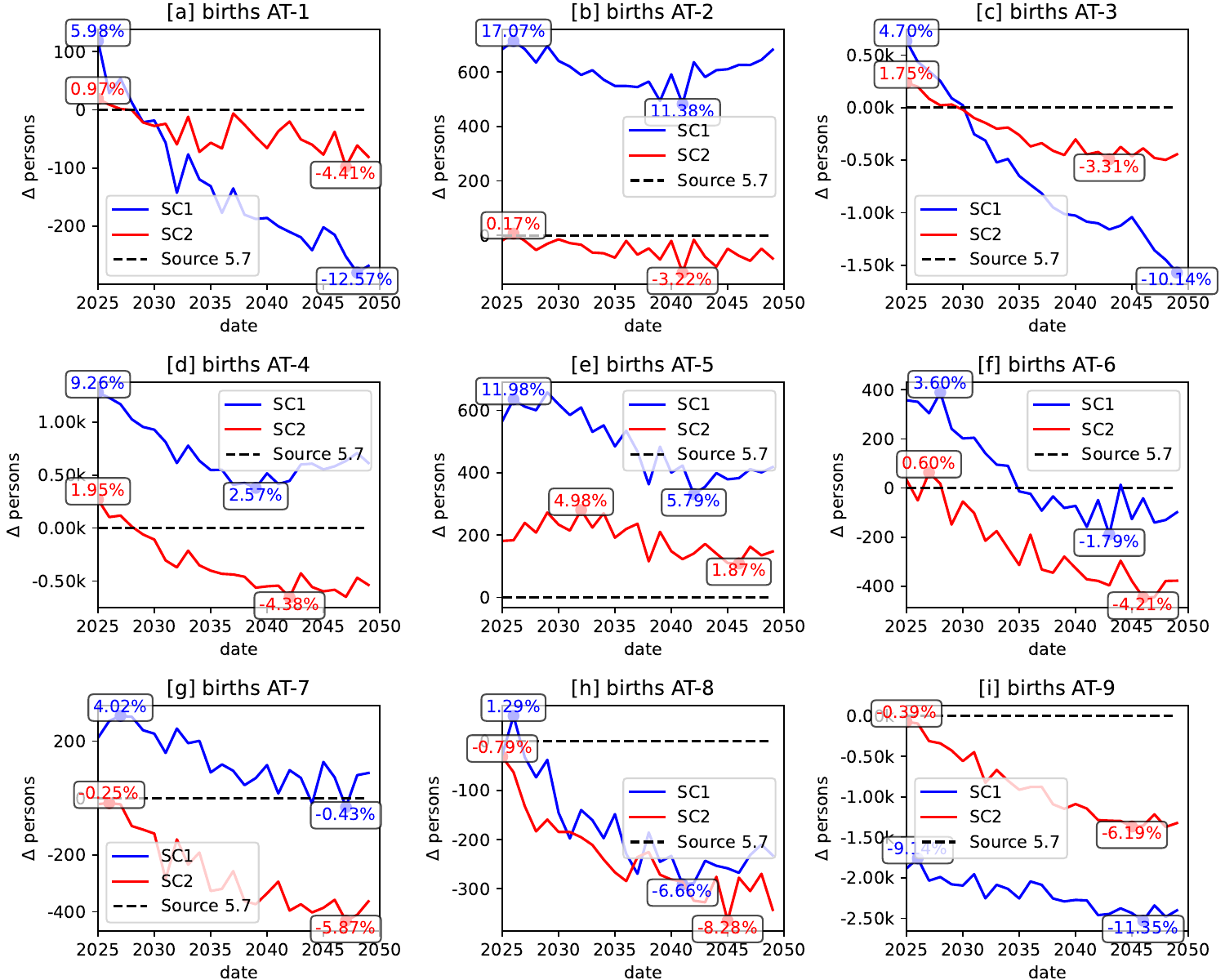}
\caption{Differences between births forecast from Source \ref{src:migrationForeacst} and the simulation scenarios SC1 and SC2 for the nine federal-states of Austria.}
\label{fig:births_fc_fed_diff}
\end{figure}
\FloatBarrier
\subsection{Validation Plots - Deaths}
\begin{figure}
\centering
\includegraphics[width=0.7\linewidth]{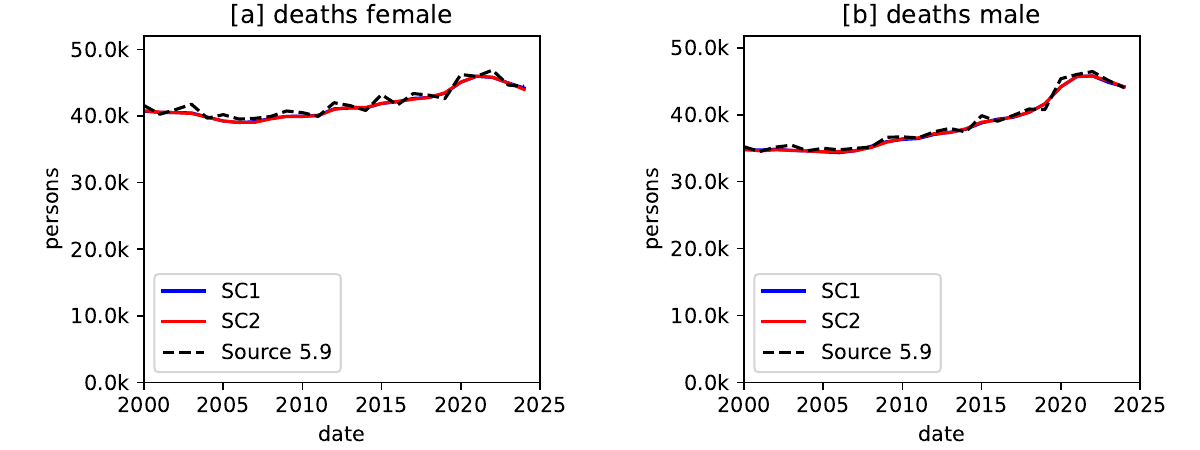}
\caption{Comparison between deaths data from Source \ref{src:indicators} and the simulation scenarios SC1 and SC2 for male and female persons.}
\label{fig:deaths_sex}
\end{figure}
\begin{figure}
\centering
\includegraphics[width=0.7\linewidth]{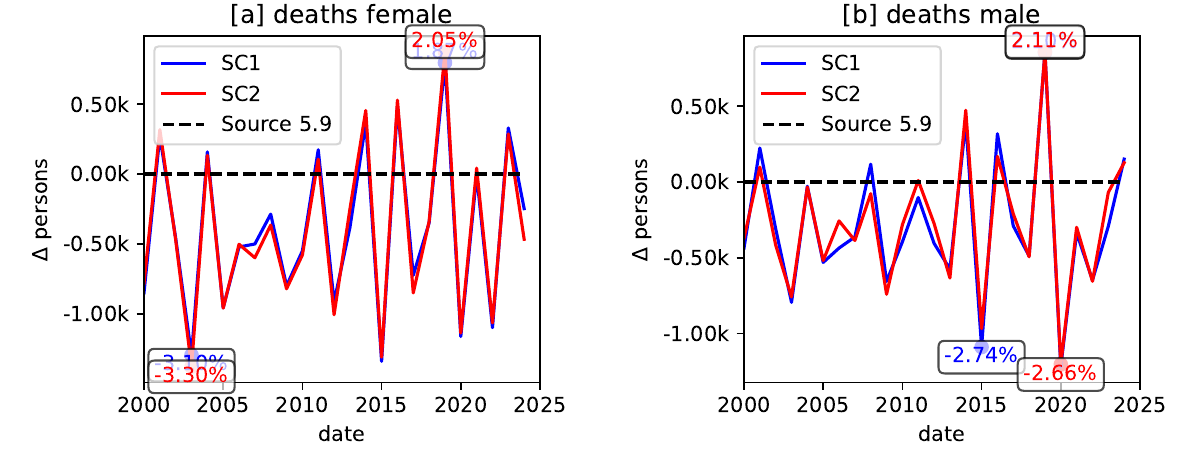}
\caption{Differences between deaths data from Source \ref{src:indicators} and the simulation scenarios SC1 and SC2 for male and female persons.}
\label{fig:deaths_sex_diff_b}
\end{figure}
\begin{figure}
\centering
\includegraphics[width=\linewidth]{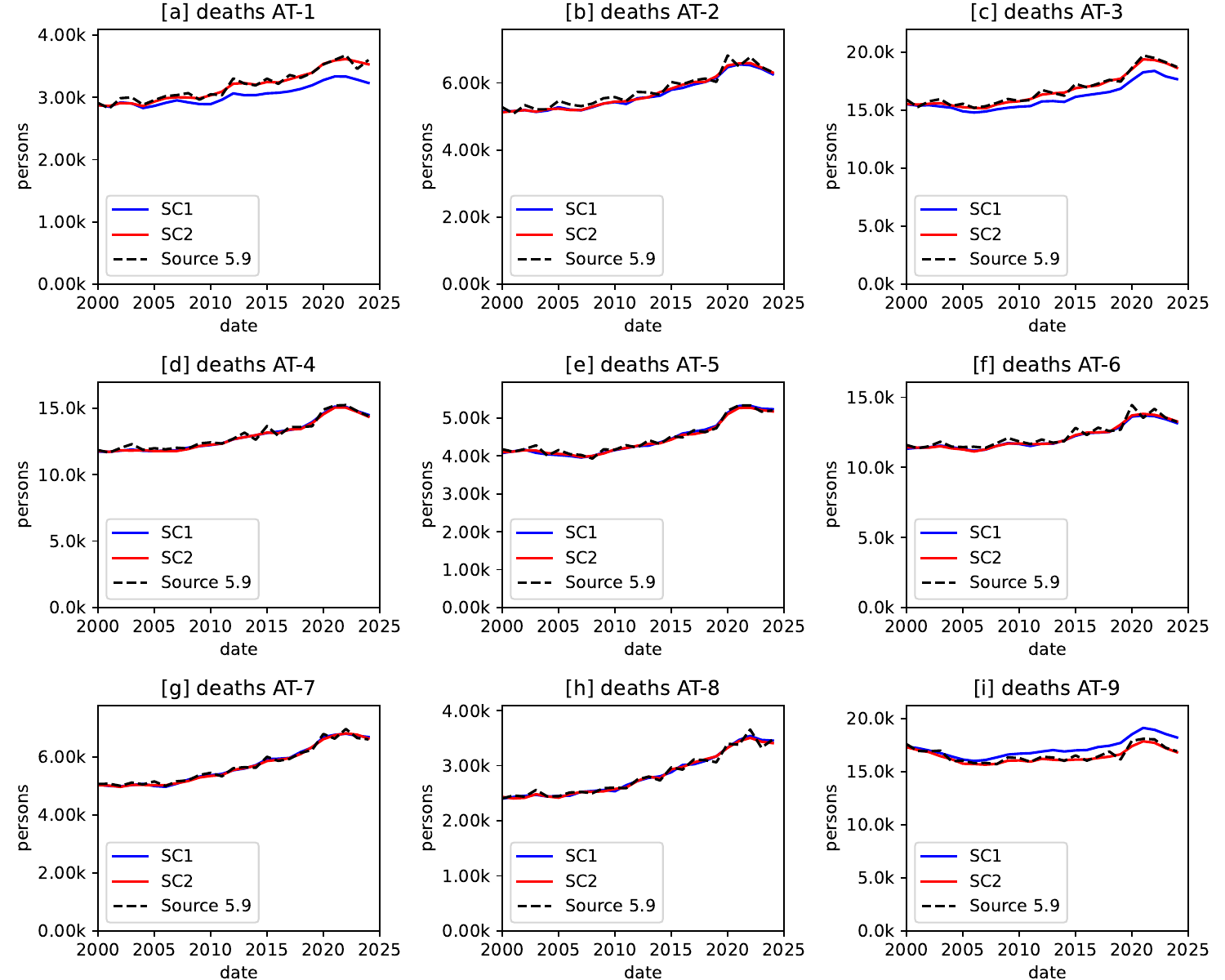}
\caption{Comparison between deaths data from Source \ref{src:indicators} and the simulation scenarios SC1 and SC2 for the nine federal-states of Austria.}
\label{fig:deaths_fed}
\end{figure}
\begin{figure}
\centering
\includegraphics[width=\linewidth]{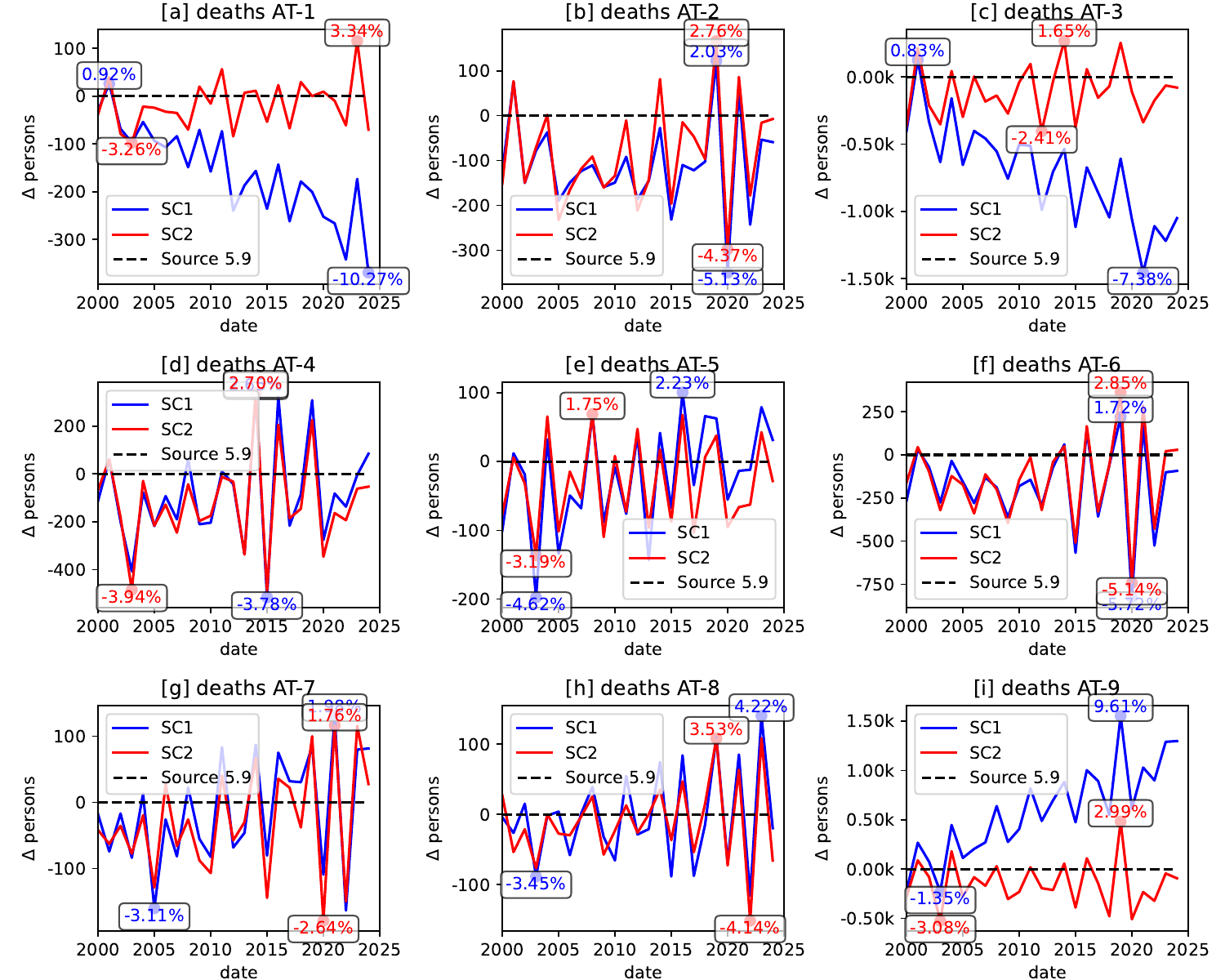}
\caption{Differences between deaths data from Source \ref{src:indicators} and the simulation scenarios SC1 and SC2 for the nine federal-states of Austria.}
\label{fig:deaths_fed_diff}
\end{figure}
\begin{figure}
\centering
\includegraphics[width=\linewidth]{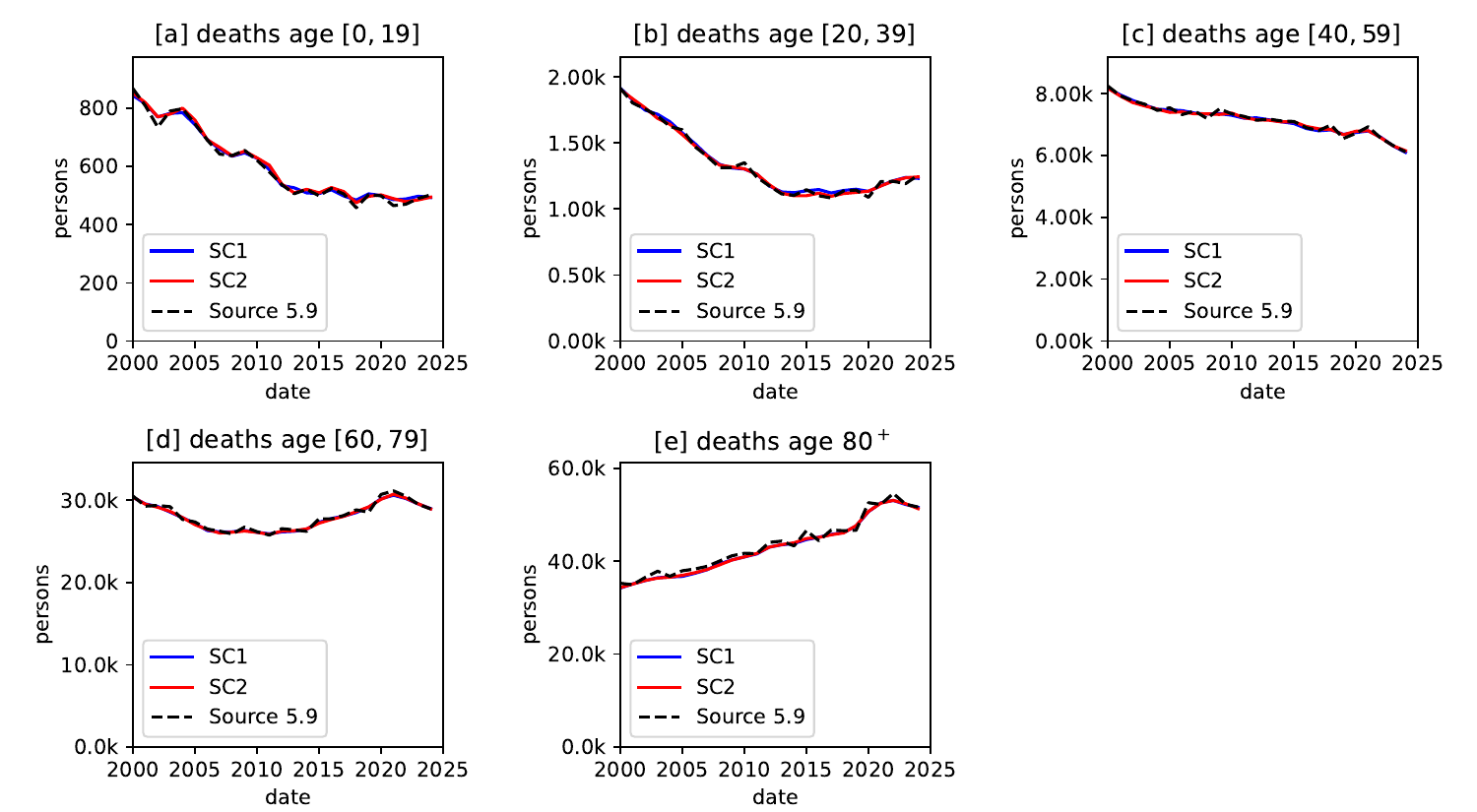}
\caption{Comparison between deaths data from Source \ref{src:indicators} and the simulation scenarios SC1 and SC2 for different age cohorts.}
\label{fig:deaths_age}
\end{figure}
\begin{figure}
\centering
\includegraphics[width=\linewidth]{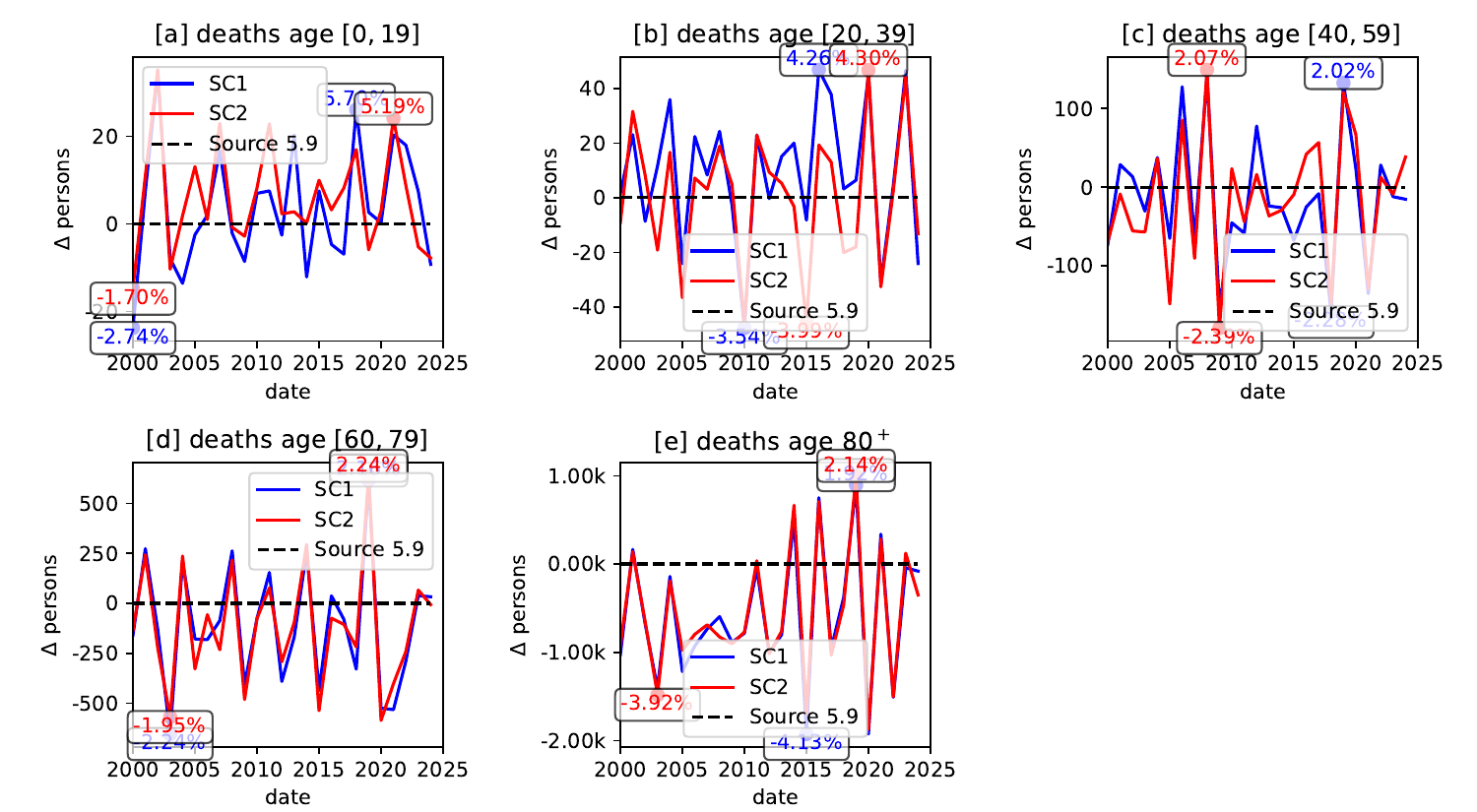}
\caption{Differences between deaths data from Source \ref{src:indicators} and the simulation scenarios SC1 and SC2 for different age cohorts.}
\label{fig:deaths_age_diff}
\end{figure}
\begin{figure}
\centering
\includegraphics[width=\linewidth]{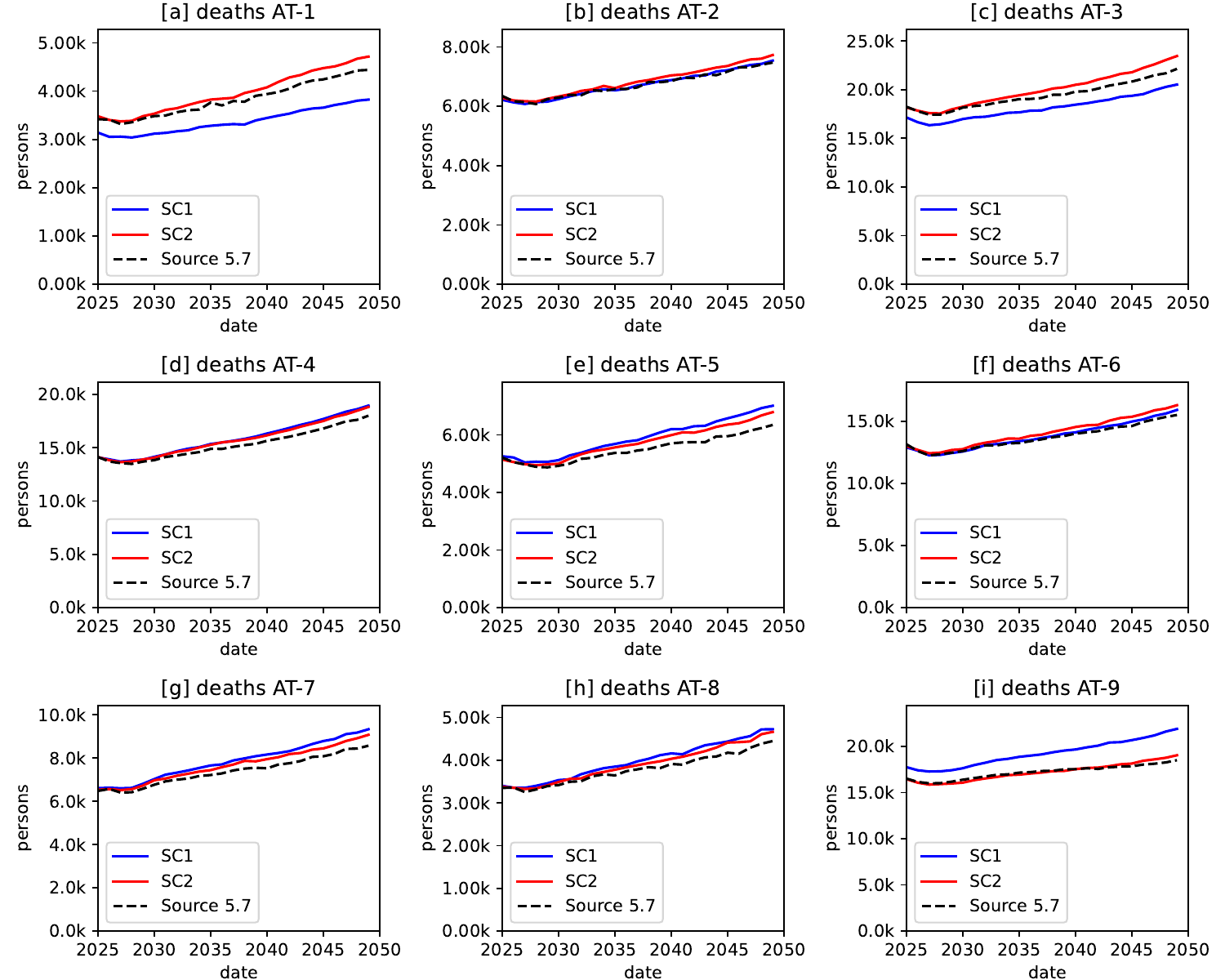}
\caption{Comparison between deaths forecast from Source \ref{src:migrationForeacst} and the simulation scenarios SC1 and SC2 for the nine federal-states of Austria.}
\label{fig:deaths_fc_fed}
\end{figure}
\begin{figure}
\centering
\includegraphics[width=\linewidth]{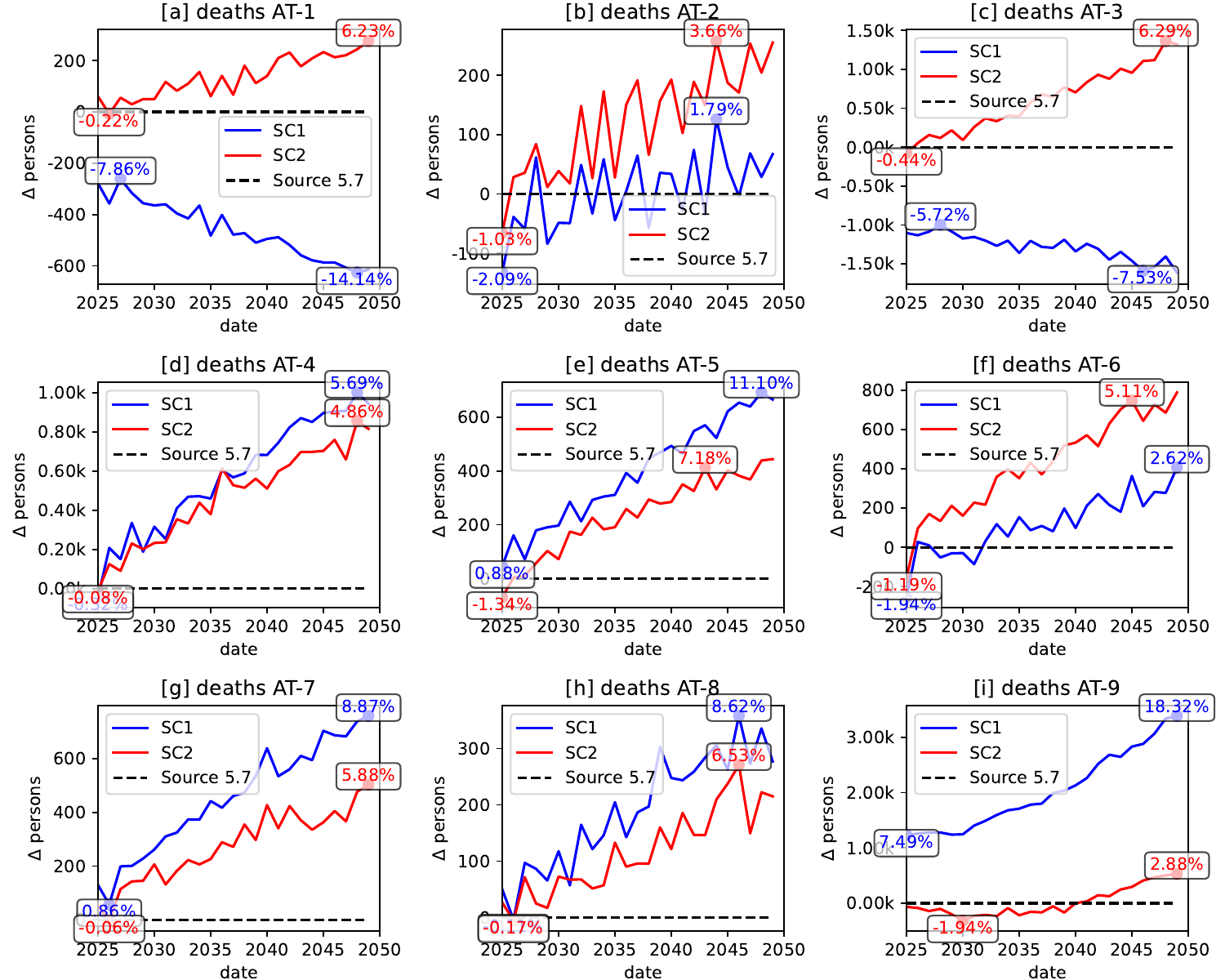}
\caption{Differences between deaths forecast from Source \ref{src:migrationForeacst} and the simulation scenarios SC1 and SC2 for the nine federal-states of Austria.}
\label{fig:deaths_fc_fed_diff}
\end{figure}
\FloatBarrier
\subsection{Validation Plots - Emigrants}
\begin{figure}
\centering
\includegraphics[width=0.7\linewidth]{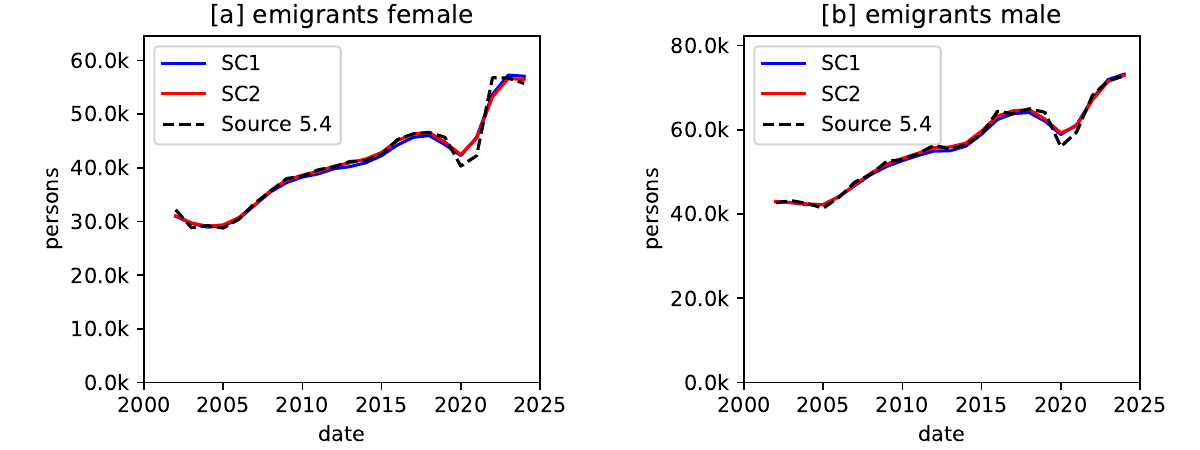}
\caption{Comparison between emigrant data from Source \ref{src:migrationCountry} and the simulation scenarios SC1 and SC2 for male and female persons.}
\label{fig:emigrants_sex_2}
\end{figure}
\begin{figure}
\centering
\includegraphics[width=0.7\linewidth]{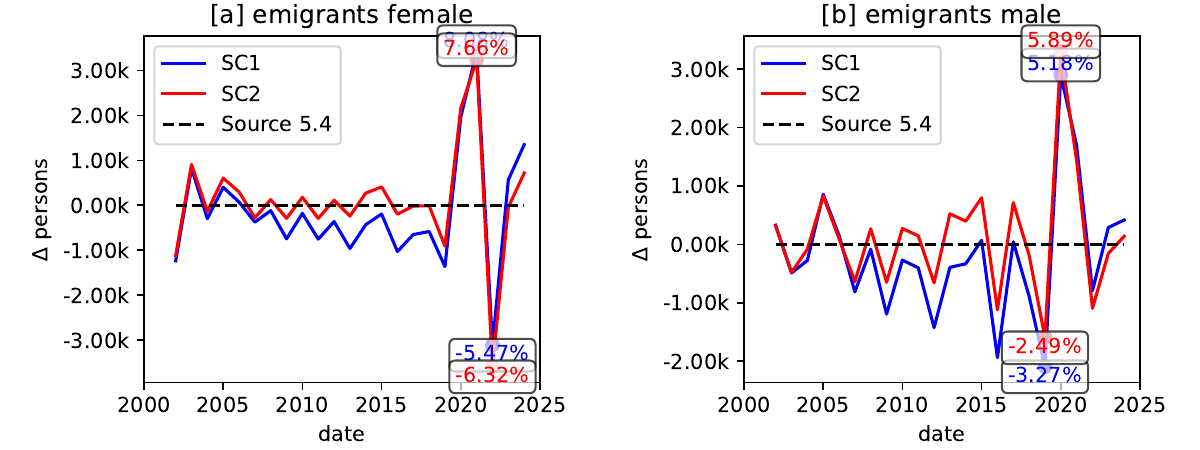}
\caption{Differences between emigrants data from Source \ref{src:migrationCountry} and the simulation scenarios SC1 and SC2 for male and female persons.}
\label{fig:emigrants_sex_diff_b}
\end{figure}
\begin{figure}
\centering
\includegraphics[width=\linewidth]{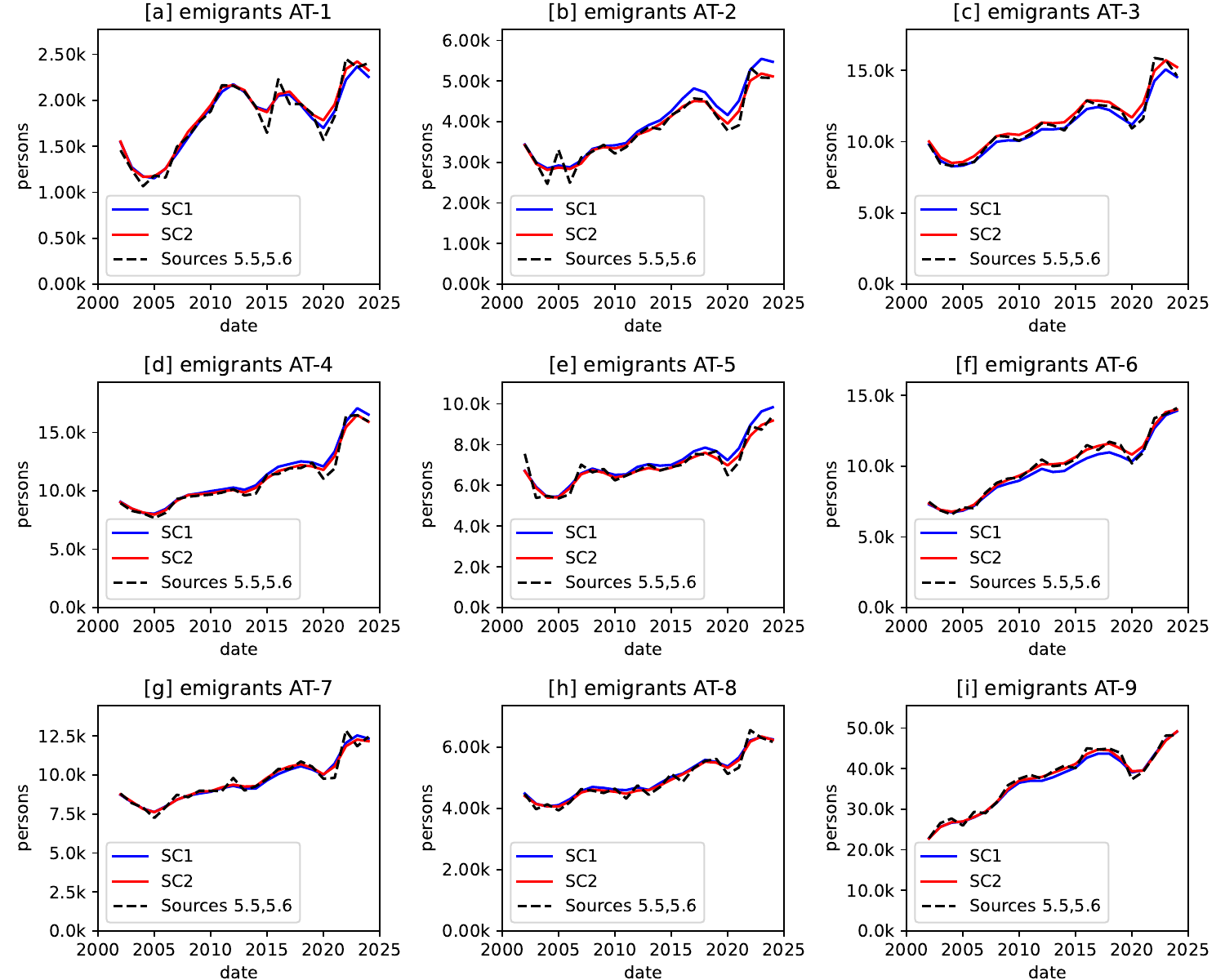}
\caption{Comparison between emigrants data from Sources \ref{src:migrationMuni_1}, \ref{src:migrationMuni_1} and the simulation scenarios SC1 and SC2 for the nine federal-states of Austria.}
\label{fig:emigrants_fed_b}
\end{figure}
\begin{figure}
\centering
\includegraphics[width=\linewidth]{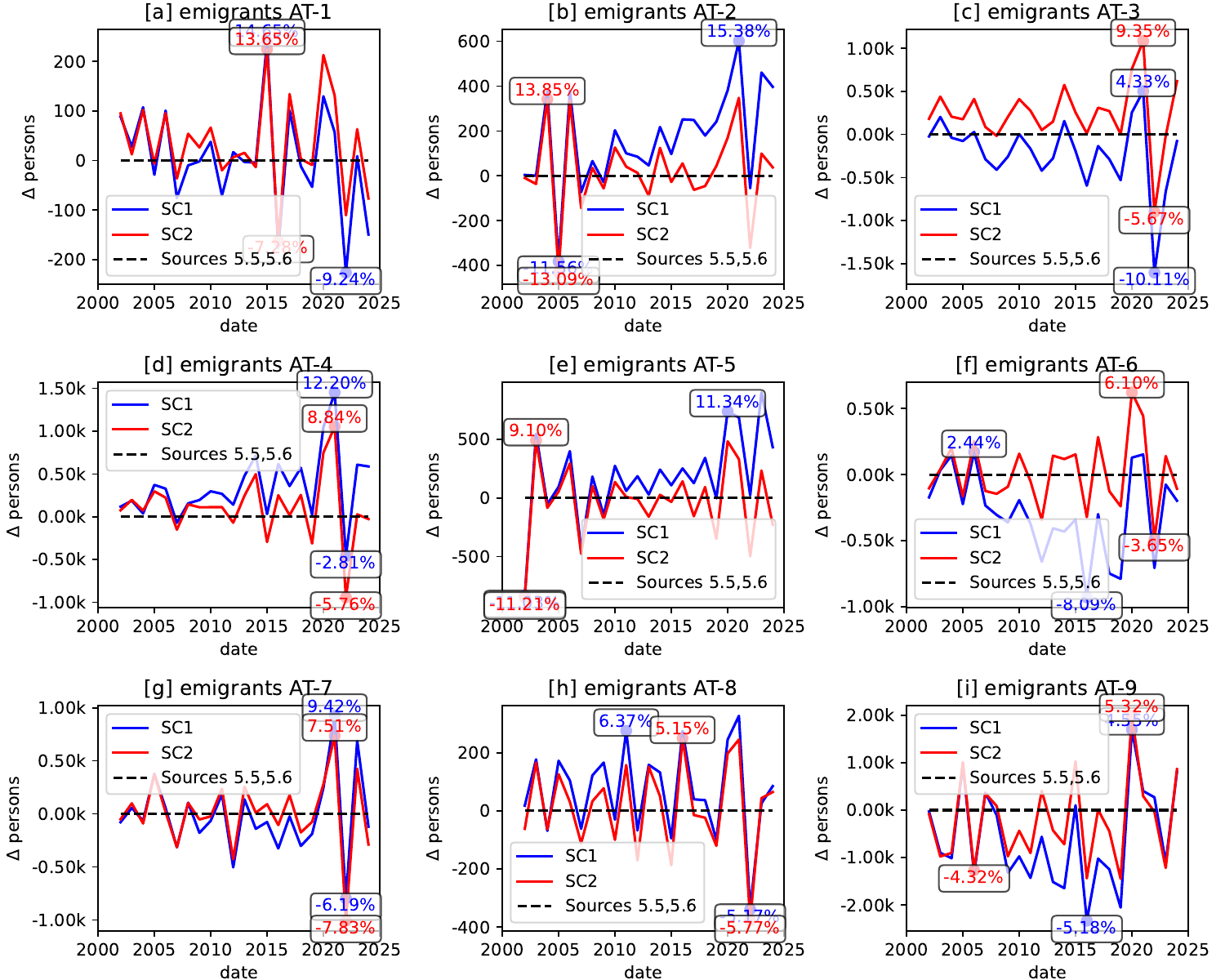}
\caption{Differences between emigrants data from Sources \ref{src:migrationMuni_1}, \ref{src:migrationMuni_1} and the simulation scenarios SC1 and SC2 for the nine federal-states of Austria.}
\label{fig:emigrants_fed_diff}
\end{figure}
\begin{figure}
\centering
\includegraphics[width=\linewidth]{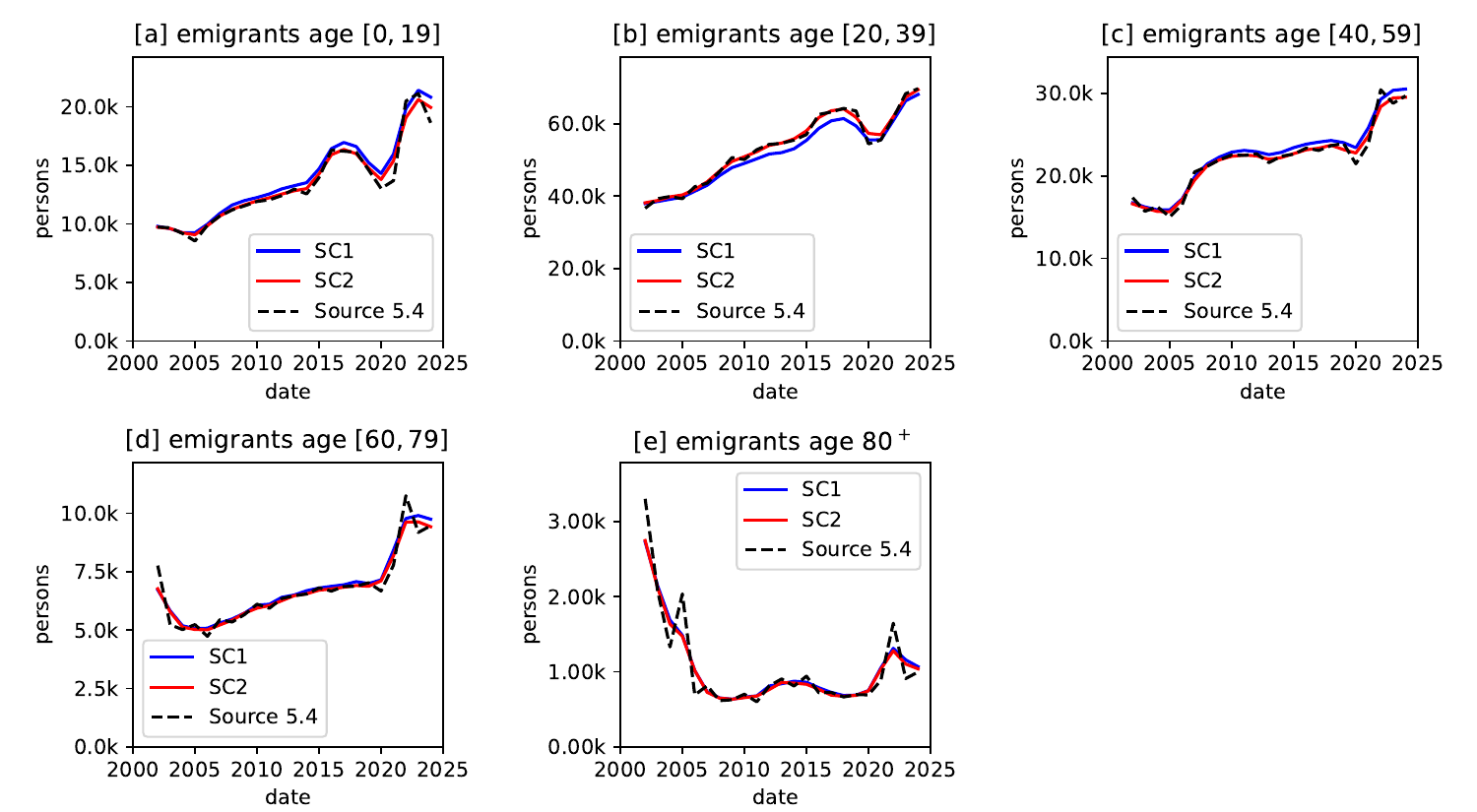}
\caption{Comparison between emigrants data from Source \ref{src:migrationCountry} and the simulation scenarios SC1 and SC2 for different age cohorts.}
\label{fig:emigrants_age}
\end{figure}
\begin{figure}
\centering
\includegraphics[width=\linewidth]{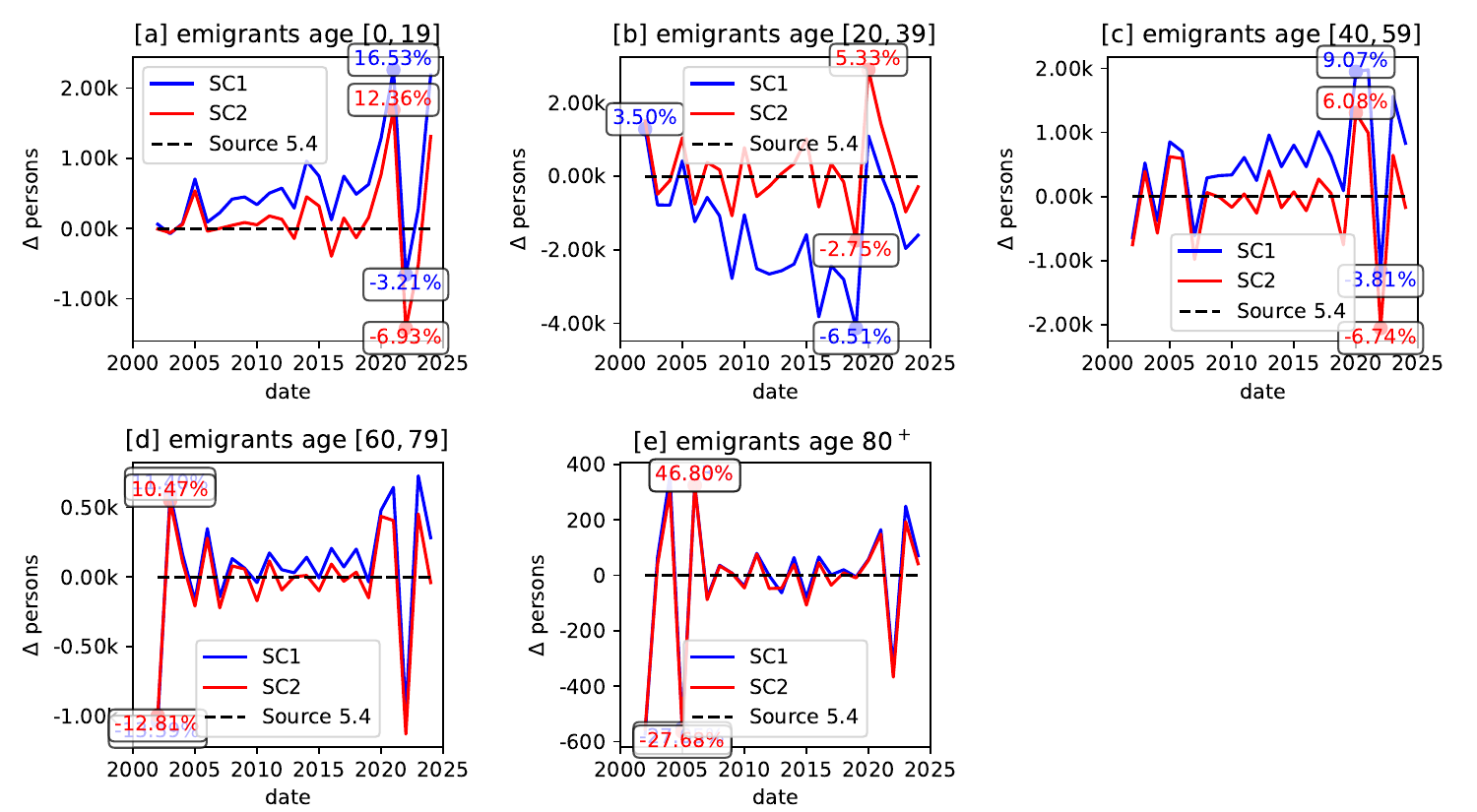}
\caption{Differences between emigrants data from Source \ref{src:migrationCountry} and the simulation scenarios SC1 and SC2 for different age cohorts.}
\label{fig:emigrants_age_diff}
\end{figure}
\begin{figure}
\centering
\includegraphics[width=\linewidth]{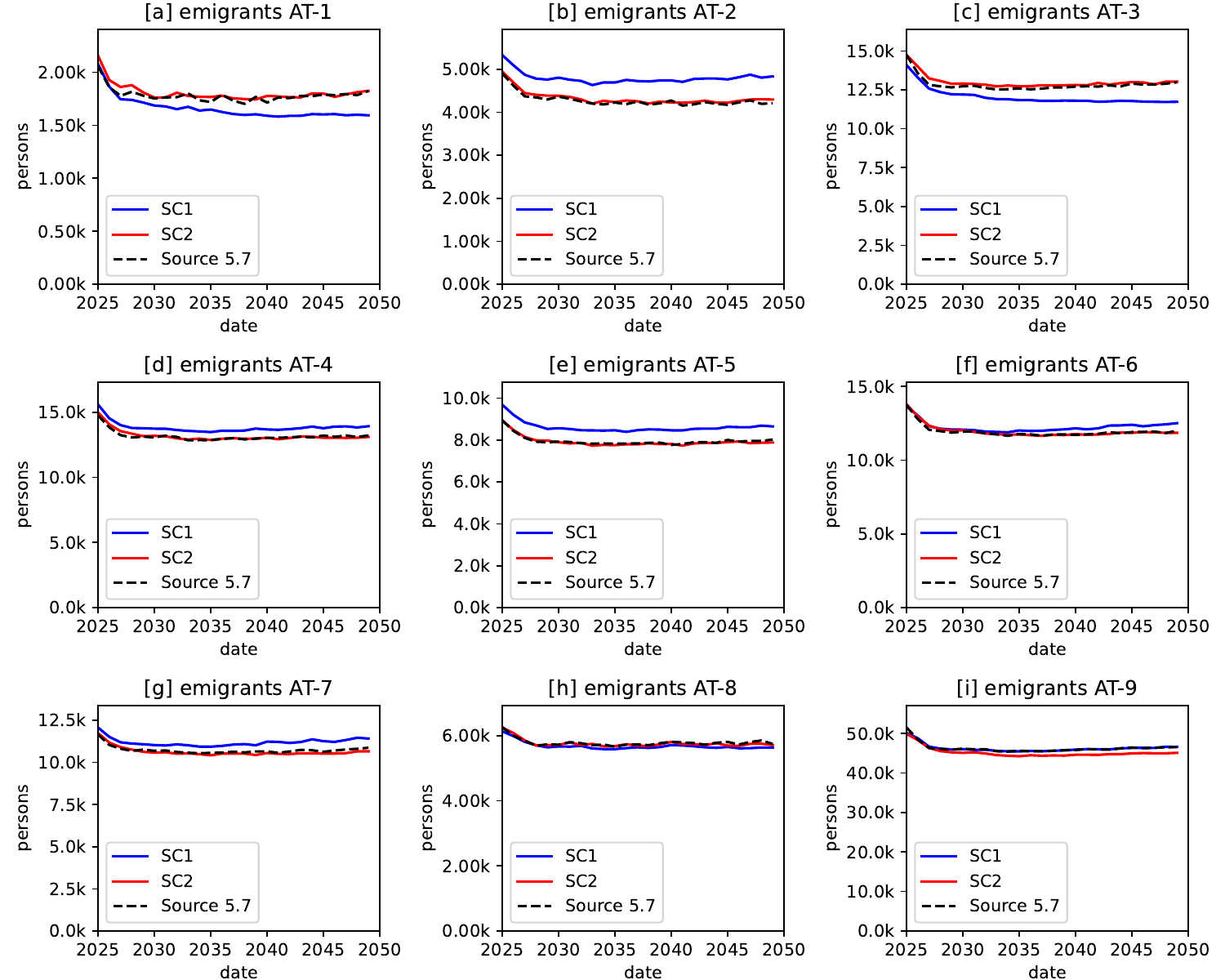}
\caption{Comparison between emigrants forecast from Source \ref{src:migrationForeacst} and the simulation scenarios SC1 and SC2 for the nine federal-states of Austria.}
\label{fig:emigrants_fc_fed}
\end{figure}
\begin{figure}
\centering
\includegraphics[width=\linewidth]{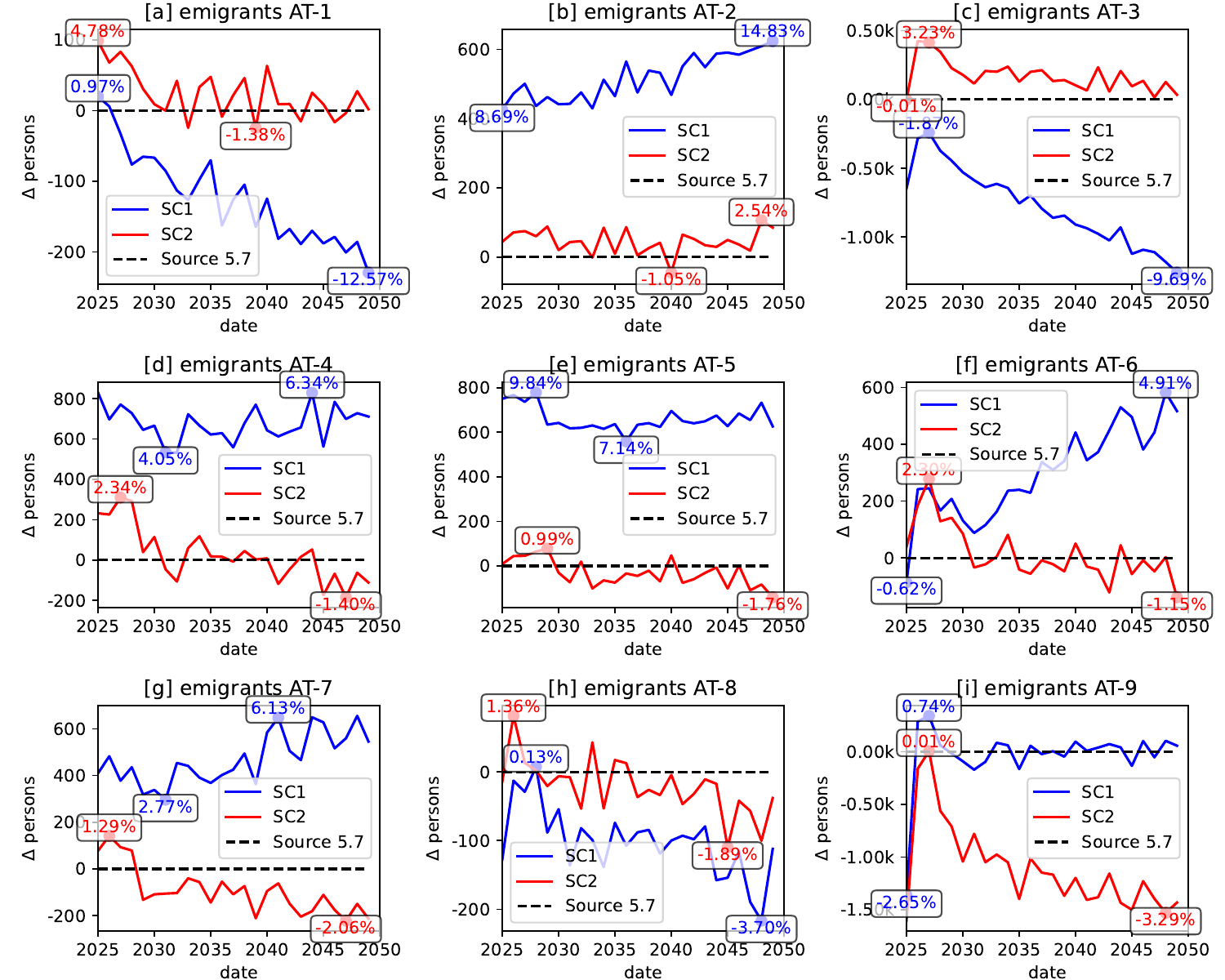}
\caption{Differences between emigrants forecast from Source \ref{src:migrationForeacst} and the simulation scenarios SC1 and SC2 for the nine federal-states of Austria.}
\label{fig:emigrants_fc_fed_diff}
\end{figure}
\FloatBarrier
\subsection{Validation Plots - Internal Migration}
\begin{figure}
\centering
\includegraphics[width=0.7\linewidth]{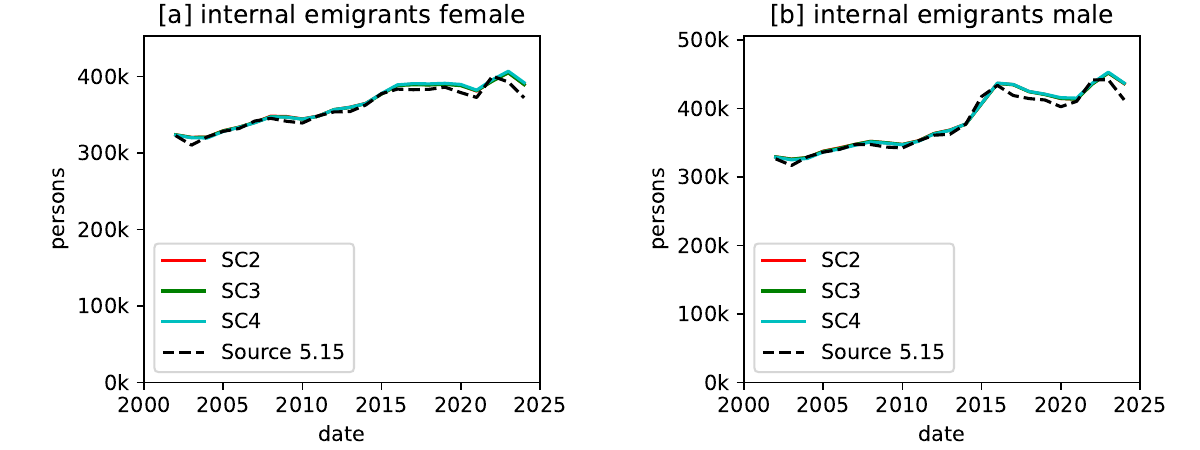}
\caption{Comparison between internal emigrants data from Source \ref{src:internalMigrantsStatCube} and the simulation scenarios SC2-SC4 for male and female persons.}
\label{fig:internal_emigrants_sex_b}
\end{figure}
\begin{figure}
\centering
\includegraphics[width=0.7\linewidth]{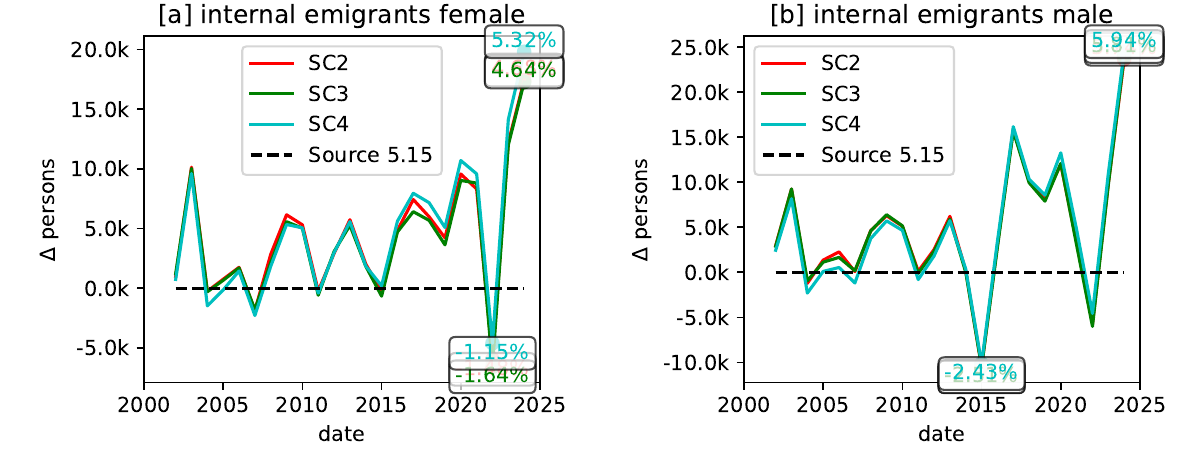}
\caption{Differences between internal emigrants data from Source \ref{src:internalMigrantsStatCube} and the simulation scenarios SC2-SC4 for male and female persons.}
\label{fig:internal_emigrants_sex_diff}
\end{figure}
\begin{figure}
\centering
\includegraphics[width=\linewidth]{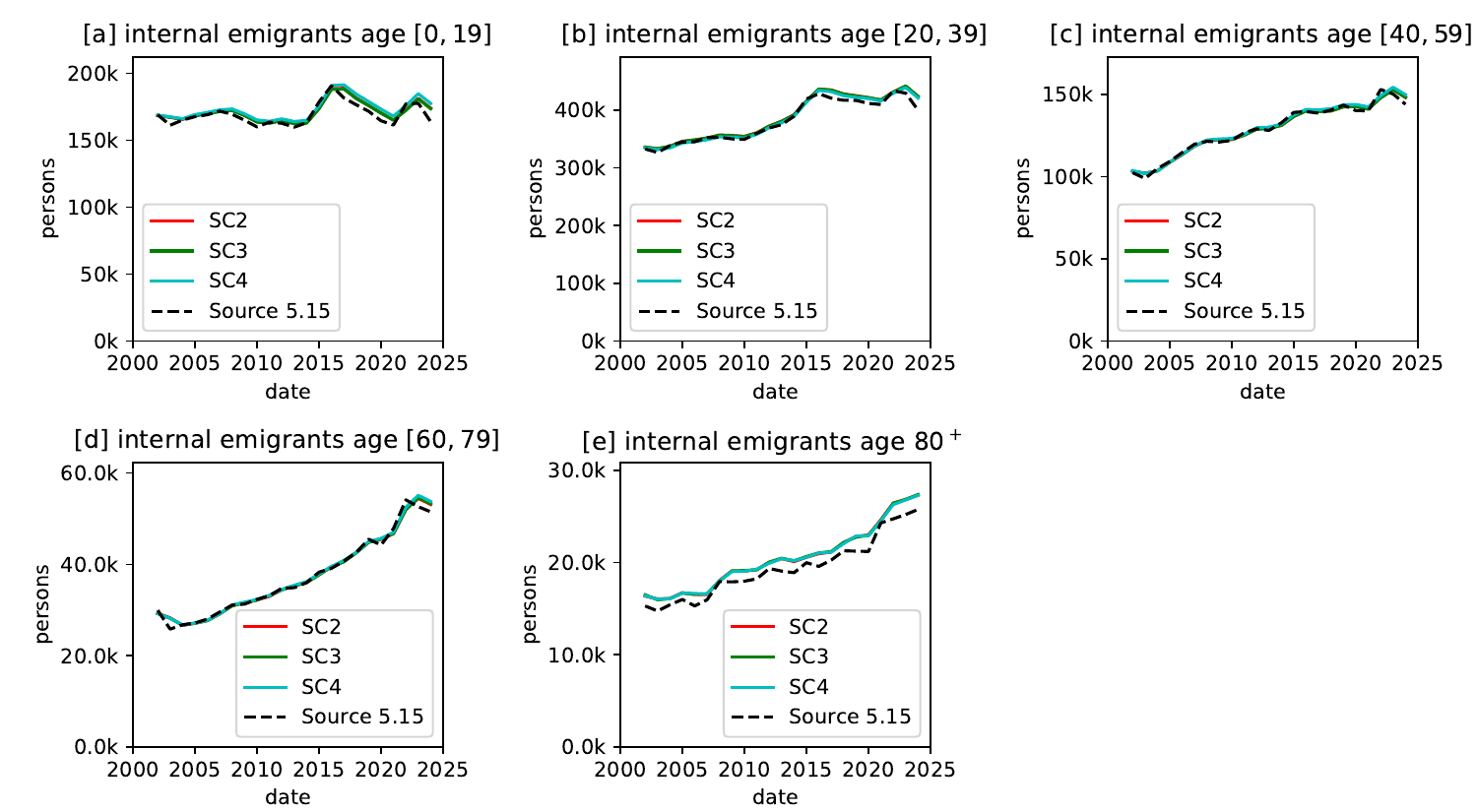}
\caption{Comparison between internal emigrants data from Source \ref{src:internalMigrantsStatCube} and the simulation scenarios SC2-SC4 for different age cohorts.}
\label{fig:internal_emigrants_age}
\end{figure}
\begin{figure}
\centering
\includegraphics[width=\linewidth]{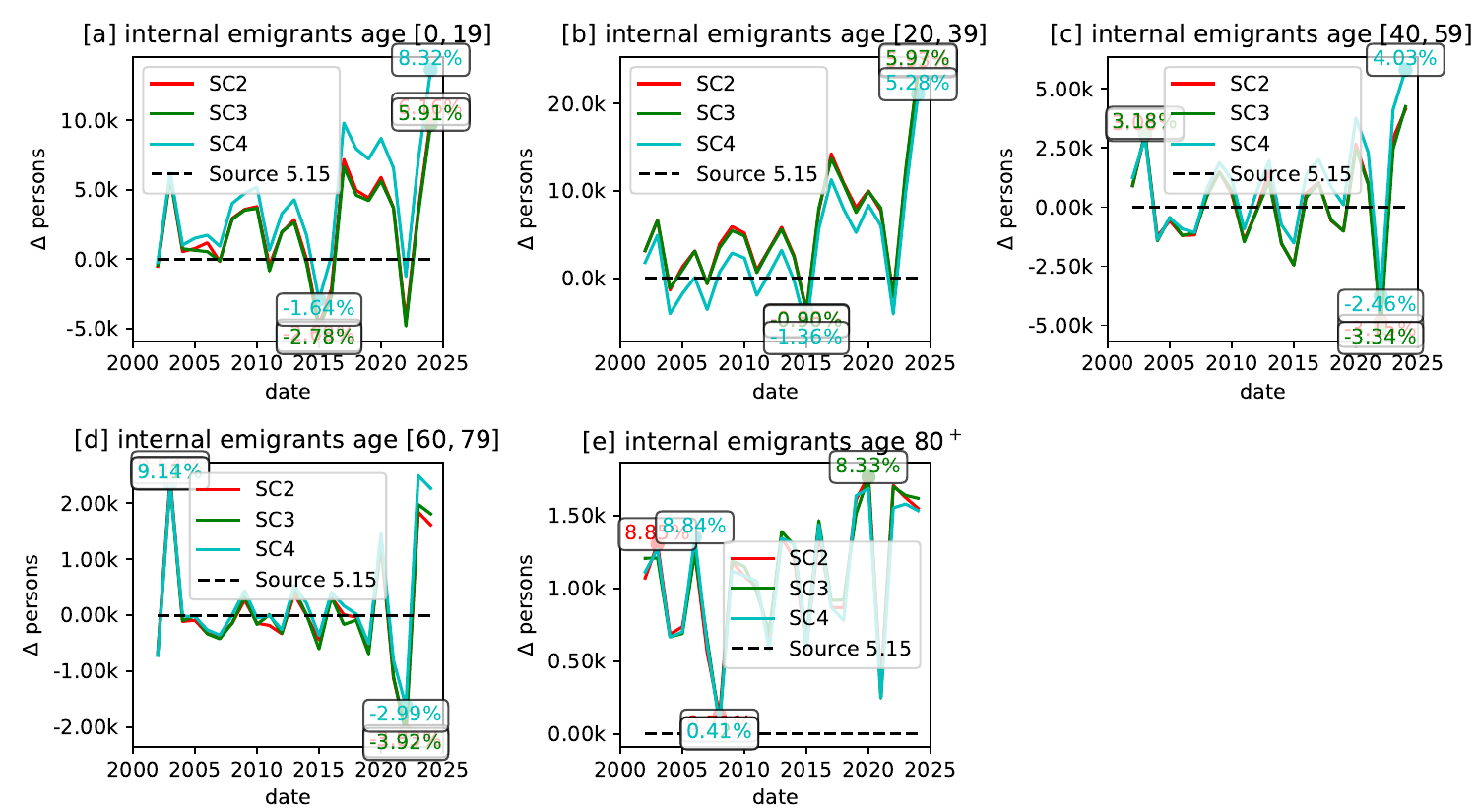}
\caption{Differences between internal emigrants data from Source \ref{src:internalMigrantsStatCube} and the simulation scenarios SC2-SC4 for different age cohorts.}
\label{fig:internal_emigrants_age_diff}
\end{figure}
\begin{figure}
\centering
\includegraphics[width=\linewidth]{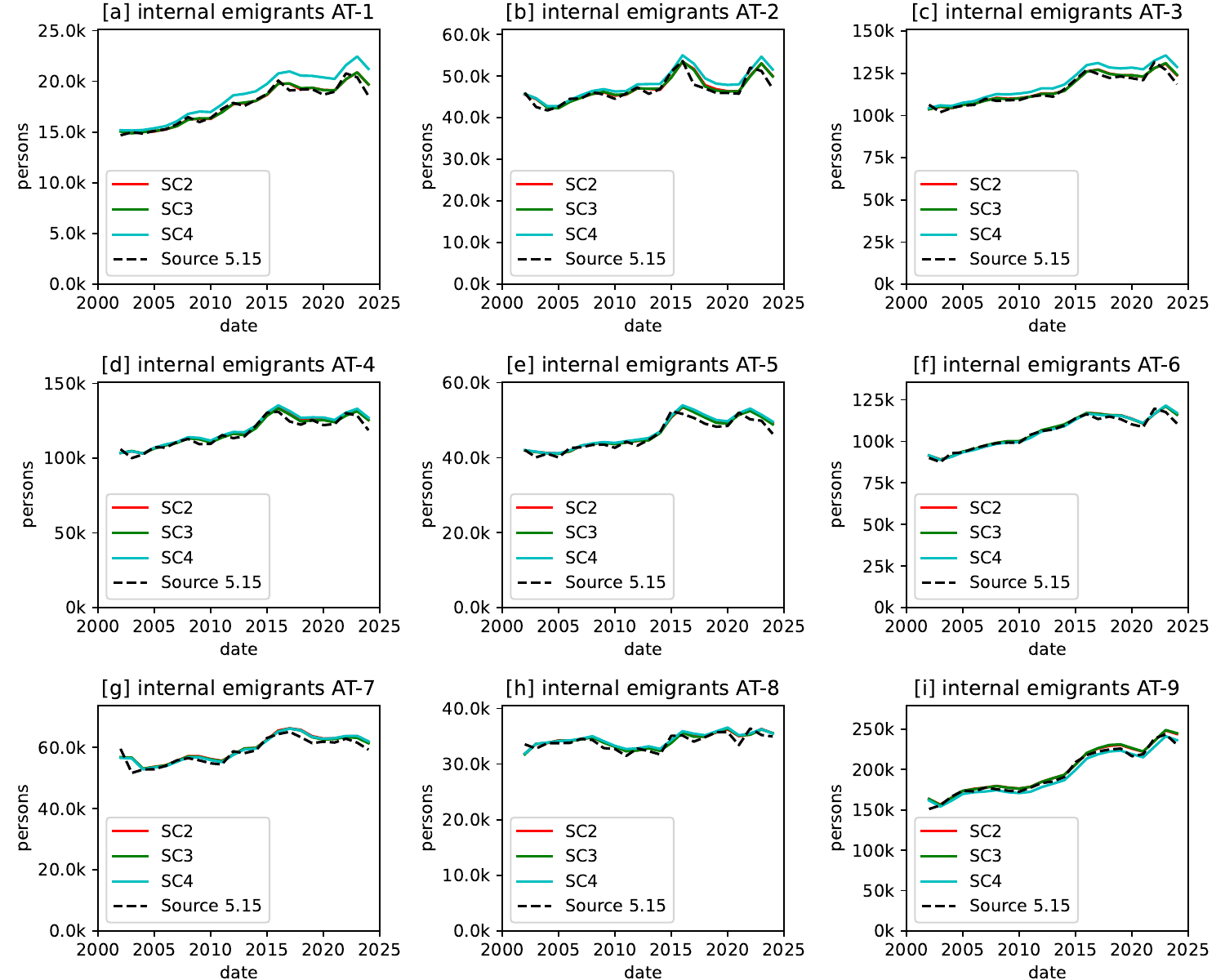}
\caption{Comparison between internal emigrants data from Source \ref{src:internalMigrantsStatCube} and the simulation scenarios SC2-SC4 for the nine federal-states of Austria.}
\label{fig:internal_emigrants_fed}
\end{figure}
\begin{figure}
\centering
\includegraphics[width=\linewidth]{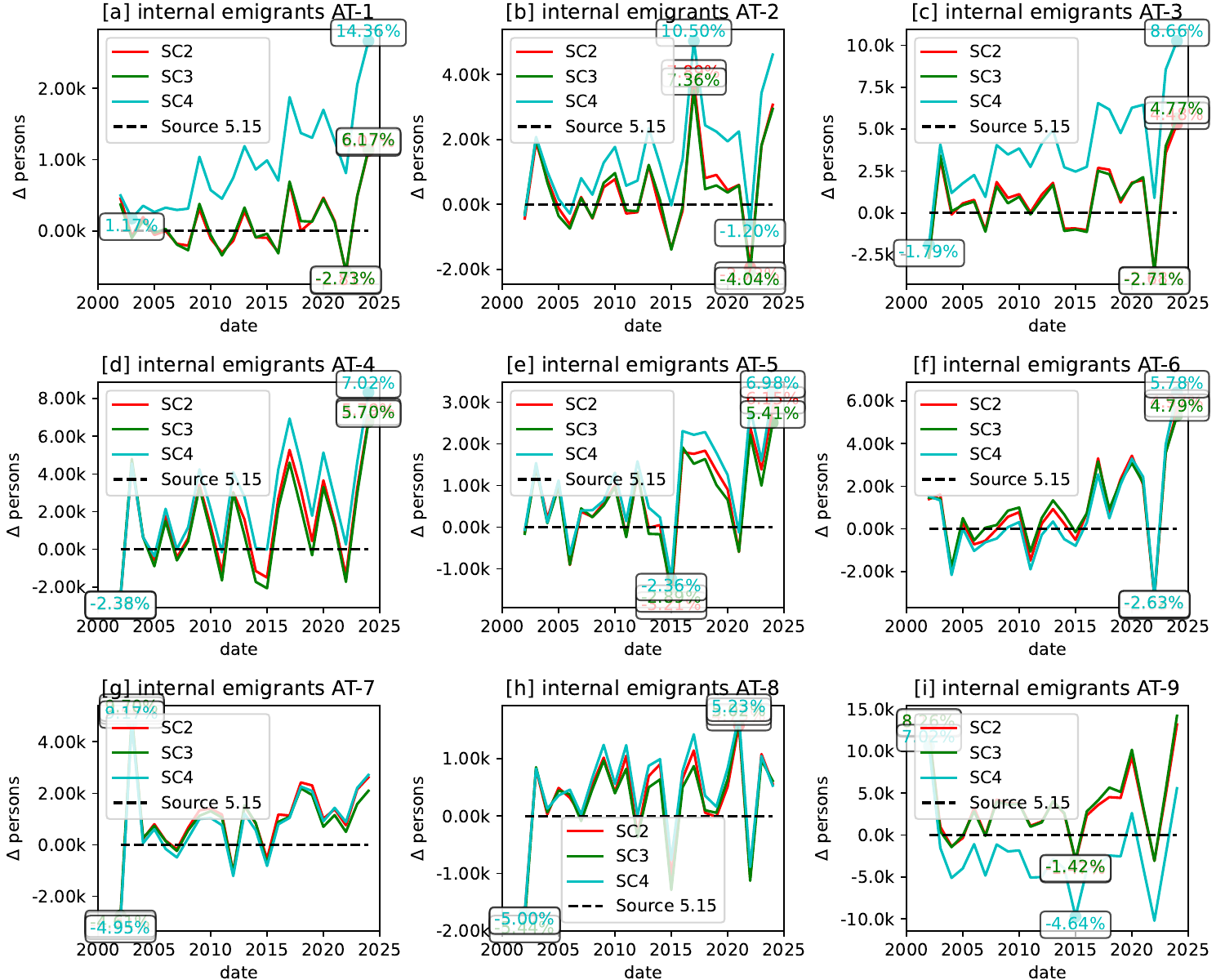}
\caption{Differences between internal emigrants data from Source \ref{src:internalMigrantsStatCube} and the simulation scenarios SC2-SC4 for the nine federal-states of Austria.}
\label{fig:internal_emigrants_fed_diff}
\end{figure}

\begin{figure}
\centering
\includegraphics[width=0.7\linewidth]{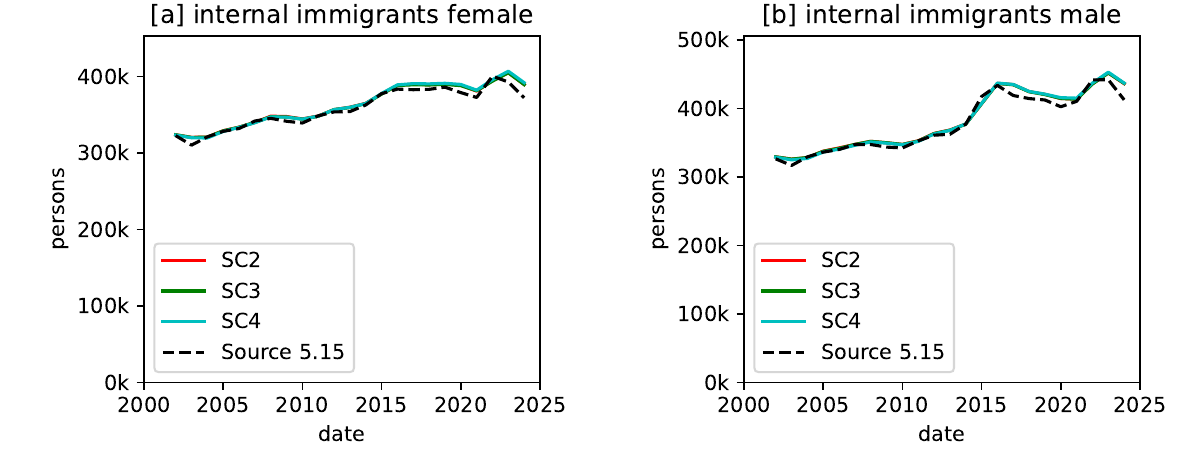}
\caption{Comparison between internal immigrants data from Source \ref{src:internalMigrantsStatCube} and the simulation scenarios SC2-SC4 for male and female persons.}
\label{fig:internal_immigrants_sex}
\end{figure}
\begin{figure}
\centering
\includegraphics[width=0.7\linewidth]{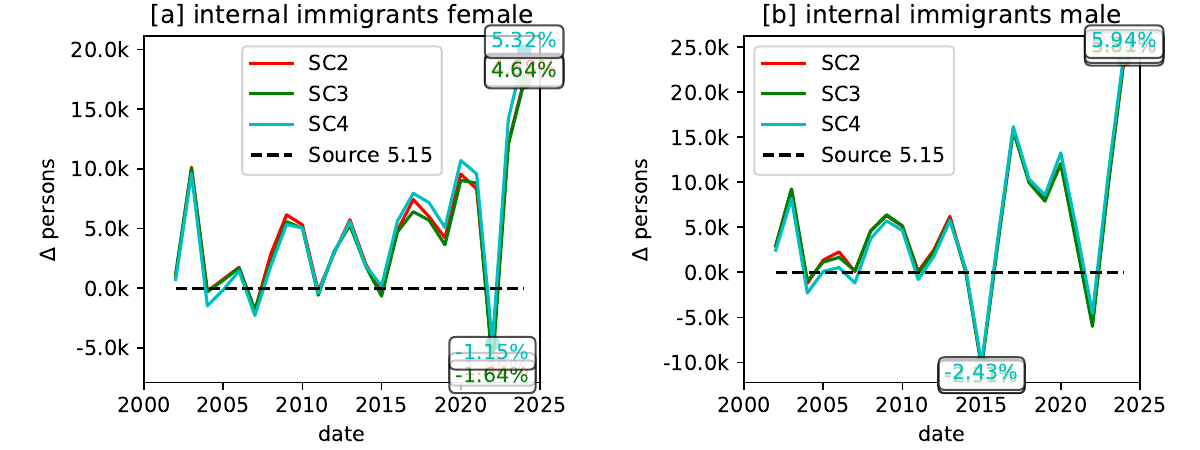}
\caption{Differences between internal immigrants data from Source \ref{src:internalMigrantsStatCube} and the simulation scenarios SC2-SC4 for male and female persons.}
\label{fig:internal_immigrants_sex_diff}
\end{figure}
\begin{figure}
\centering
\includegraphics[width=\linewidth]{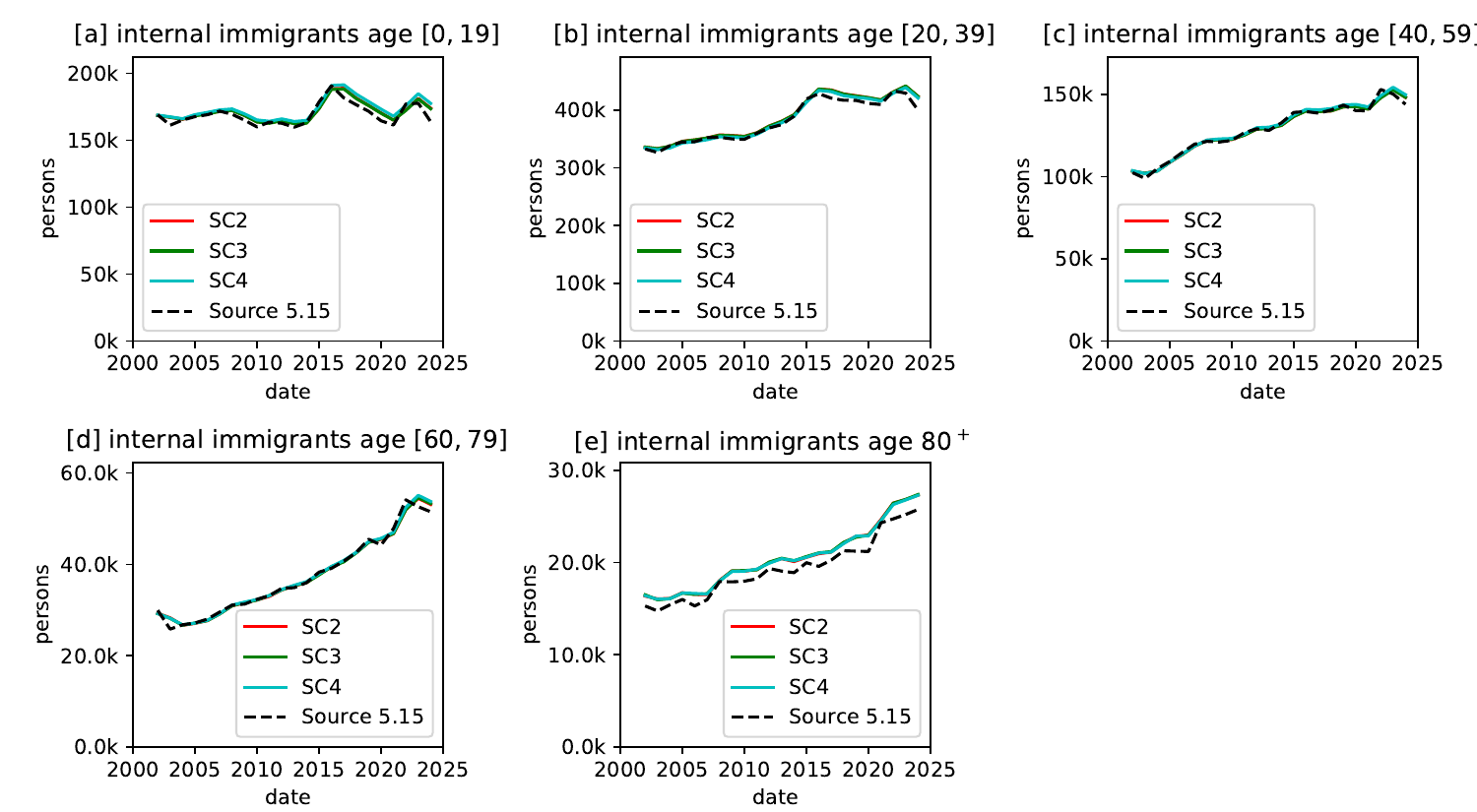}
\caption{Comparison between internal immigrants data from Source \ref{src:internalMigrantsStatCube} and the simulation scenarios SC2-SC4 for different age cohorts.}
\label{fig:internal_immigrants_age}
\end{figure}
\begin{figure}
\centering
\includegraphics[width=\linewidth]{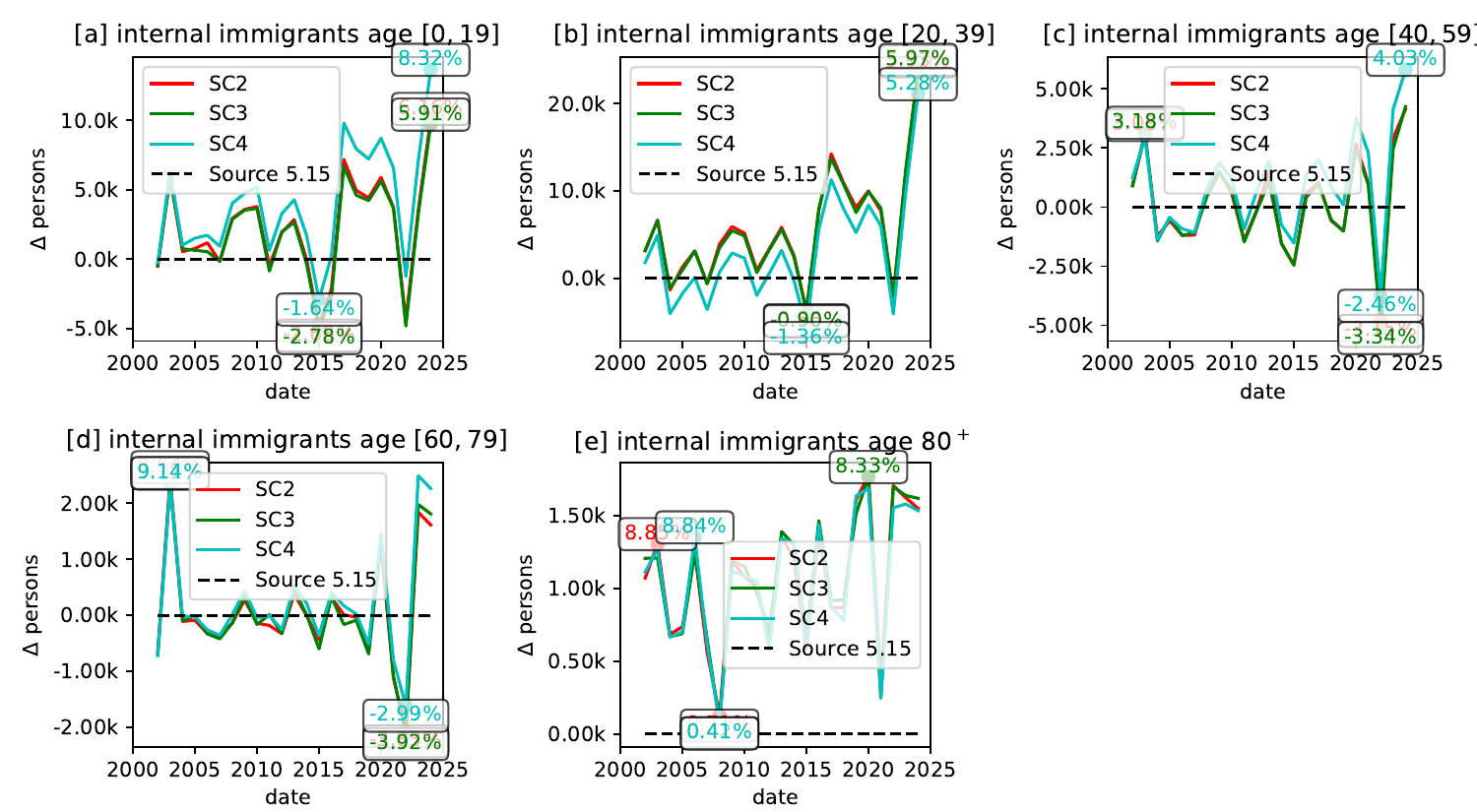}
\caption{Differences between internal immigrants data from Source \ref{src:internalMigrantsStatCube} and the simulation scenarios SC2-SC4 for different age cohorts.}
\label{fig:internal_immigrants_age_diff}
\end{figure}
\begin{figure}
\centering
\includegraphics[width=\linewidth]{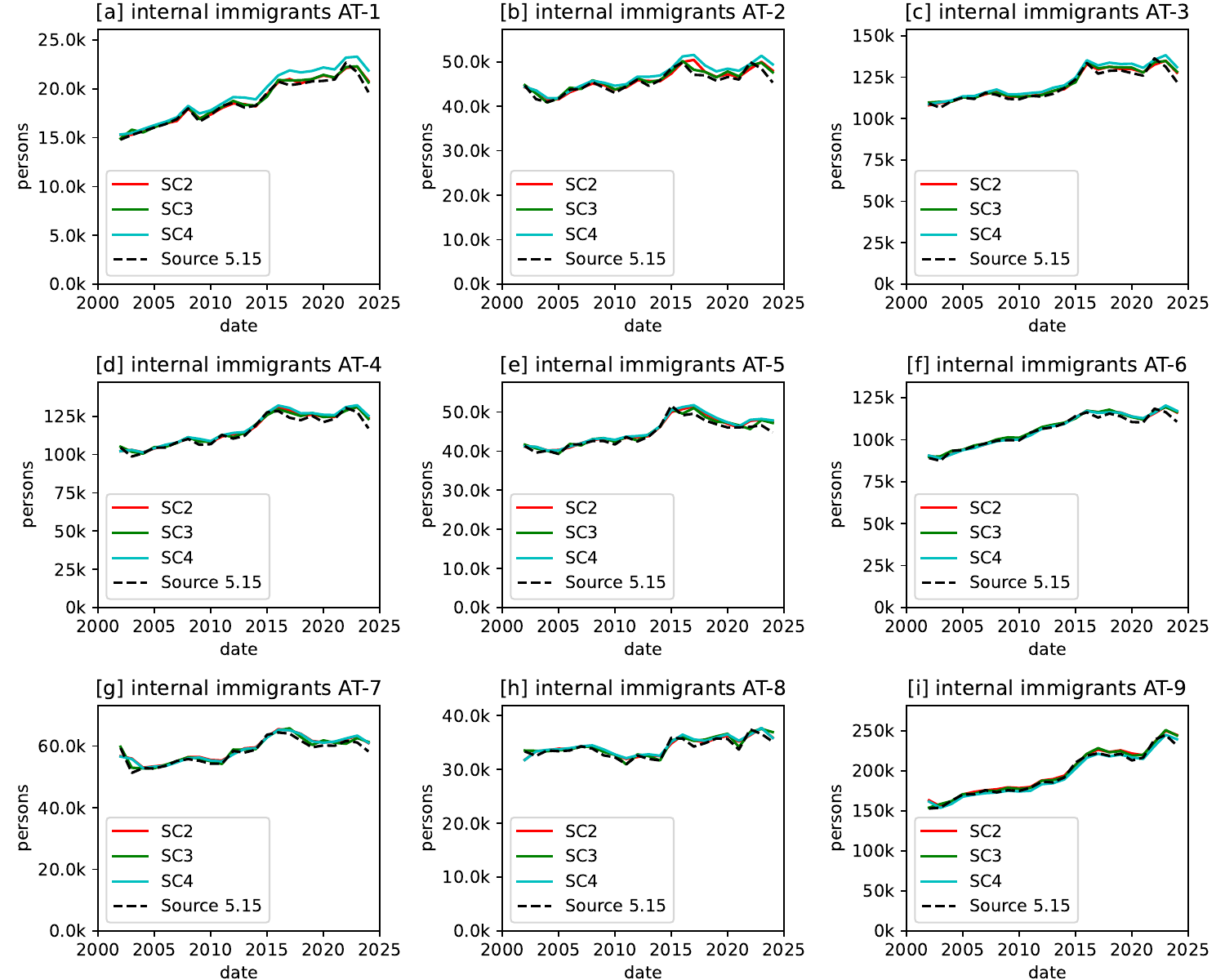}
\caption{Comparison between internal immigrants data from Source \ref{src:internalMigrantsStatCube} and the simulation scenarios SC2-SC4 for the nine federal-states of Austria.}
\label{fig:internal_immigrants_fed}
\end{figure}
\begin{figure}
\centering
\includegraphics[width=\linewidth]{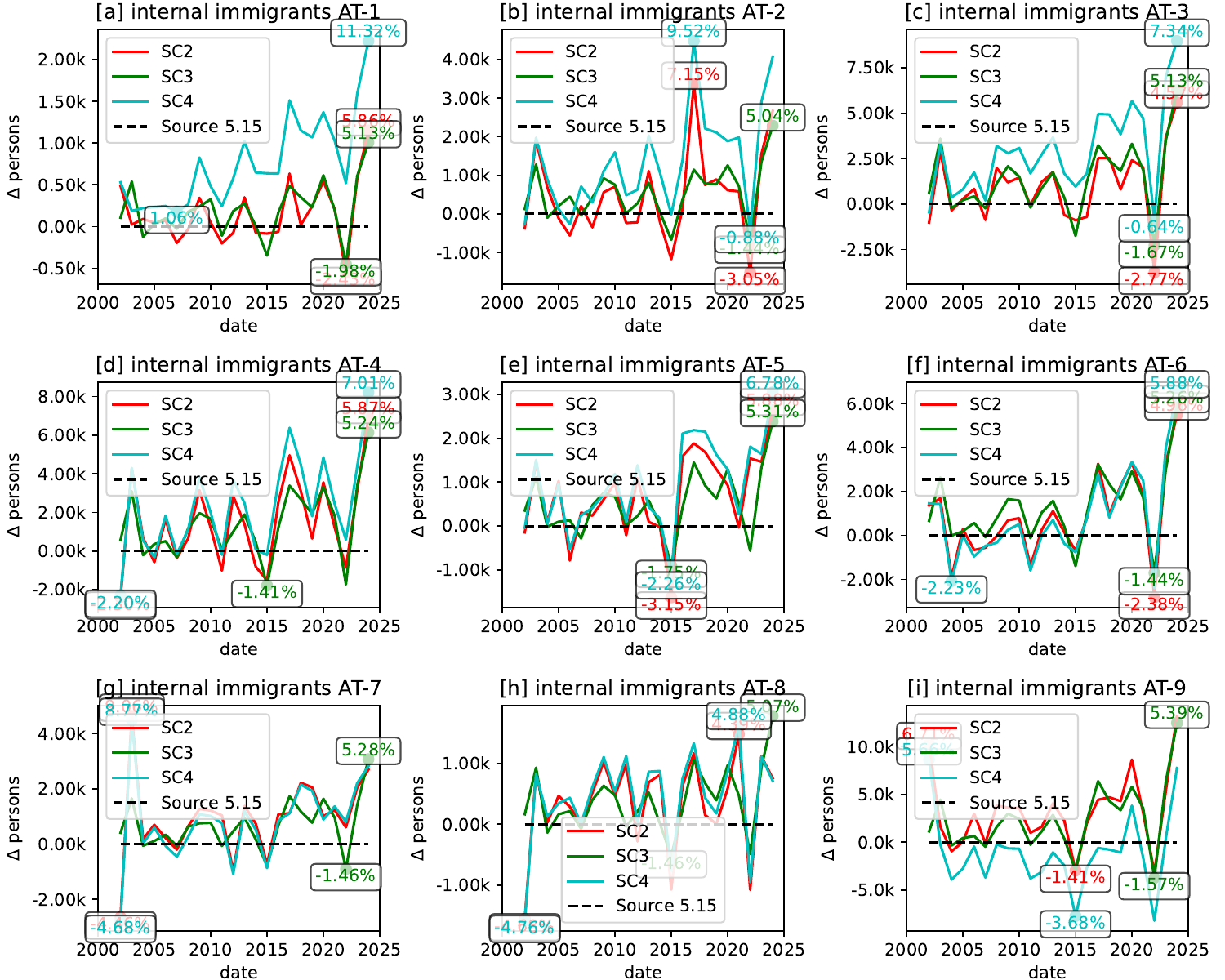}
\caption{Differences between internal immigrants data from Source \ref{src:internalMigrantsStatCube} and the simulation scenarios SC2-SC4 for the nine federal-states of Austria.}
\label{fig:internal_immigrants_fed_diff}
\end{figure}
\end{document}